%% file: main.tex
\documentclass[11pt,a4paper,logo]{googledeepmind}

\setleftlogo[110pt]{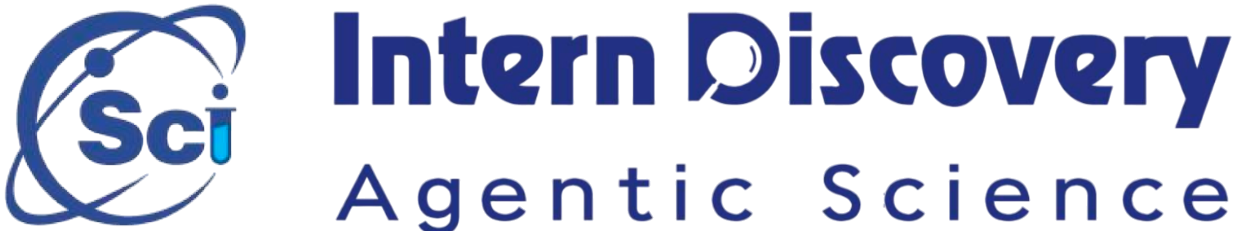}  
\setrightlogo[160pt]{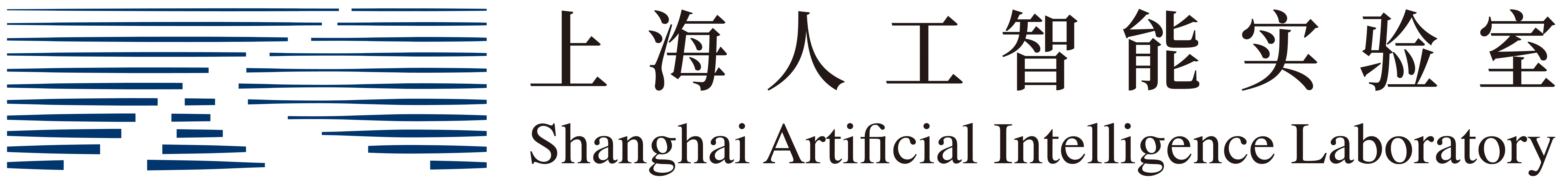}

\usepackage[T1]{fontenc}
\usepackage{tgcursor}
\usepackage{pifont}
\usepackage[
    natbib=true,
    backend=biber,
    style=numeric, 
    sorting=none 
]{biblatex}
\AtEveryBibitem{\clearfield{month}}
\AtEveryBibitem{\clearfield{day}}
\usepackage{csquotes}

\title{Probing Scientific General Intelligence of LLMs with Scientist-Aligned Workflows}

\newcommand{\homepage}{\raisebox{-1.5pt}{\includegraphics[height=1em]{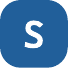}}}
\newcommand{\github}{\raisebox{-1.5pt}{\includegraphics[height=1em]{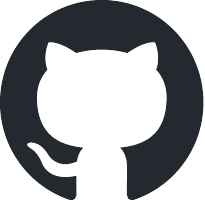}}}
\newcommand{\huggingface}{\raisebox{-1.5pt}{\includegraphics[height=1em]{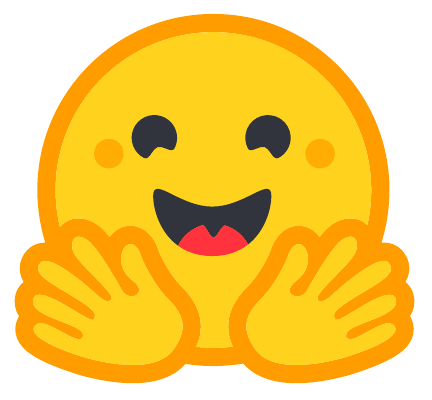}}}

\author{Shanghai Artificial Intelligence Laboratory}

\usepackage{pdflscape}

\usepackage{textcomp}
\usepackage{rotating}

\usepackage{setspace}
\usepackage{microtype} 

\usepackage{graphicx}
\usepackage{subcaption}
\usepackage{caption}

\usepackage{booktabs}
\usepackage[table]{xcolor}
\usepackage{xcolor}

\usepackage{array}
\usepackage{threeparttable}
\usepackage{multirow}

\usepackage{amsmath}
\usepackage{siunitx}

\usepackage{enumitem}
\usepackage{float}
\usepackage{seqsplit}
\usepackage{framed}

\usepackage{tikz}

\usepackage{caption} 
\usepackage[export]{adjustbox}

\usepackage{ragged2e}
\usepackage[most]{tcolorbox}

\usepackage{makecell}
\usepackage{tabularray}

\usepackage{longtable}
\usepackage{listings}
\definecolor{CaseGreen}{HTML}{2E7D32} 
\definecolor{CaseOrange}{HTML}{F57C00} 
\definecolor{CaseGray}{HTML}{F6F7F8}   
\definecolor{CaseInk}{HTML}{212121}    
\definecolor{CaseWhite}{HTML}{F5F5DC}
\definecolor{DeepBlue}{HTML}{003366} 
\definecolor{LightBlue}{HTML}{99CCFF} 
\definecolor{DeepPurple}{HTML}{673AB7} 
\definecolor{MiddlePurple}{HTML}{9C7FD0}
\definecolor{LightPurple}{HTML}{D1C4E9} 
\definecolor{HotPink}{HTML}{FF69B4} 
\definecolor{SoftPink}{HTML}{F8BBD0} 
\definecolor{Crimson}{HTML}{DC143C} 
\definecolor{Teal}{HTML}{008080} 
\definecolor{Cyan}{HTML}{00BCD4} 
\definecolor{SoftGray}{HTML}{EEEEEE}       
\definecolor{LighterGray}{HTML}{FAFAFA} 

\newtcolorbox{stagebox}[1]{
  breakable, enhanced,
  colback=CaseGray, colframe=DeepBlue, coltitle=CaseWhite,
  title=\bfseries #1, fonttitle=\bfseries,
  left=1.2mm,right=1.2mm,top=1.2mm,bottom=1.2mm, boxrule=0.6pt
}

\newtcblisting{verbatimbox}{
  listing only, breakable, enhanced,
  colback=white, colframe=LightBlue, boxrule=0.5pt,
  left=1.2mm,right=1.2mm,top=1.2mm,bottom=1.2mm
}

\newcommand{\gold}{\raisebox{-0.5pt}{\includegraphics[height=5pt]{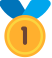}}}
\newcommand{\silver}{\raisebox{-0.5pt}{\includegraphics[height=5pt]{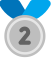}}}
\newcommand{\copper}{\raisebox{-0.5pt}{\includegraphics[height=5pt]{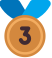}}}

\usepackage[symbol]{footmisc}
\newcolumntype{Y}{>{\RaggedRight\arraybackslash}X}

\usepackage{hyperref}
\hypersetup{
    colorlinks=true,
    linkcolor=blue, 
    citecolor=blue,  
    filecolor=black,
    urlcolor=blue    
}

\usepackage{tocloft}

\usepackage{etoolbox}
\usepackage{arydshln}
\makeatletter
\patchcmd{\@tocline}
    {\hfil}
    {\leaders\hbox{\hfil}\hfil}
    {}{}
\makeatother

\input{sections/0-abstract}

\begin{document}

\sloppy 
\maketitle


\begin{center}
    \centering
    \captionsetup{type=figure}
    \includegraphics[width=0.95\linewidth]{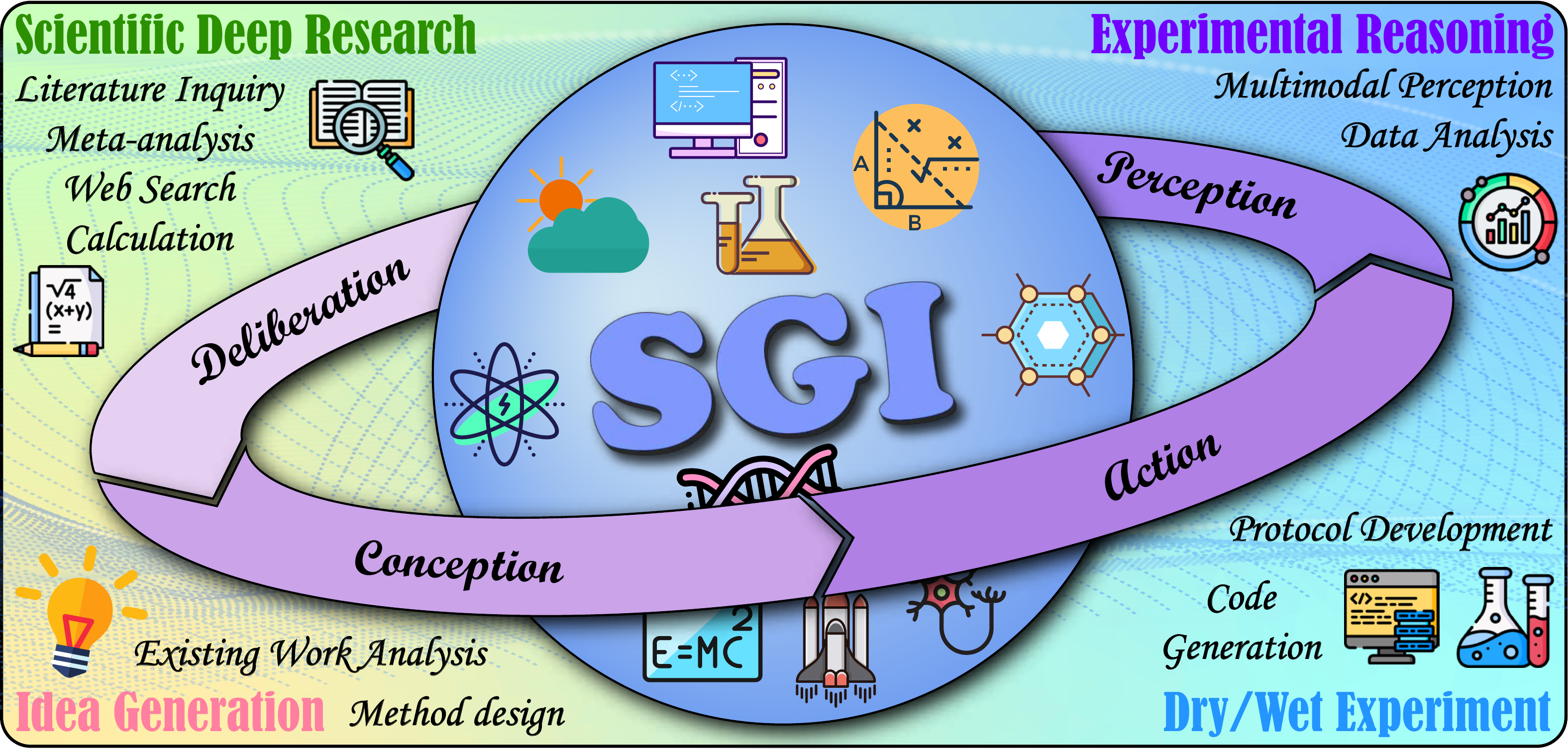}
    \captionof{figure}{\textbf{Scientific General Intelligence (SGI)} We define SGI as an AI that can autonomously navigate the complete, iterative cycle of scientific inquiry with the versatility and proficiency of a human scientist. The teaser illustrates the Practical Inquiry Model's four quadrants—Deliberation (synthesis and critical evaluation of knowledge), Conception (idea generation), Action (experimental execution), and Perception (interpretation)—and how SGI-Bench operationalizes them through four task categories and an agent-based evaluation paradigm, together providing a principle-grounded, measurable framework for assessing scientific intelligence.}
    \label{fig:teaser}
\end{center}%

\clearpage

\tableofcontents
\enlargethispage{1cm} 
\thispagestyle{fancy} 

\clearpage

\input{sections/1-introduction}
\input{sections/2-benchmark}
\input{sections/3-framework}
\input{sections/4-evaluation}
\input{sections/5-discussion}
\input{sections/6-related_work}
\input{sections/7-conclusion}



\begingroup
\sloppy
\clearpage
\printbibliography[heading=bibintoc]
\endgroup

\clearpage
\input{sections/X-appendix}

\end{document}

%% file: sections/0-abstract.tex
\begin{abstract}

\vspace{-0.6cm}

\homepage\ \textbf{Page} \texttt{\url{https://InternScience.github.io/SGI-Page/}}

\github\ \textbf{Code} \texttt{\url{https://github.com/InternScience/SGI-Bench}}

\huggingface\ \textbf{Data} \texttt{\url{https://huggingface.co/collections/InternScience/sgi-bench}}

\vspace{0.2cm}

\textbf{Abstract:}

Despite advances in scientific AI, a coherent framework for Scientific General Intelligence (SGI)—the ability to autonomously conceive, investigate, and reason across scientific domains—remains lacking. 
We present an operational SGI definition grounded in the Practical Inquiry Model (PIM: Deliberation, Conception, Action, Perception) and operationalize it via four scientist-aligned tasks: deep research, idea generation, dry/wet experiments, and experimental reasoning. 
SGI-Bench comprises over 1,000 expert-curated, cross-disciplinary samples inspired by Science's 125 Big Questions, enabling systematic evaluation of state-of-the-art LLMs. Results reveal gaps: low exact match (10--20\%) in deep research despite step-level alignment; ideas lacking feasibility and detail; high code executability but low execution result accuracy in dry experiments; low sequence fidelity in wet protocols; and persistent multimodal comparative-reasoning challenges. 
We further introduce Test-Time Reinforcement Learning (TTRL), which optimizes retrieval-augmented novelty rewards at inference, enhancing hypothesis novelty without reference answer. 
Together, our PIM-grounded definition, workflow-centric benchmark, and empirical insights establish a foundation for AI systems that genuinely participate in scientific discovery.

\end{abstract}

%% file: sections/1-introduction.tex
\section{Introduction}

Large language models (LLMs)~\cite{guo2025deepseek, zhao2023survey, naveed2025comprehensive, hu2025survey, comanici2025gemini} are achieving and even exceeding human-level performance on a diverse array of tasks, spanning multidisciplinary knowledge understanding, mathematical reasoning, and programming. This rapid progress has ignited a vibrant debate: some view these models as early signals of artificial general intelligence (AGI)~\cite{fei2022towards, bubeck2023sparks}, whereas others dismiss them as mere ``stochastic parrots~\cite{bender2021dangers},'' fundamentally constrained by their training data. As these models evolve, the frontier of AGI research is shifting towards the most complex and structured of human endeavors: scientific inquiry~\cite{hu2025surveyscientificlargelanguage}. We argue that demonstrating genuine \textbf{scientific general intelligence (SGI)} represents a critical leap toward AGI, serving as a definitive testbed for advanced reasoning, planning, and knowledge creation capabilities. However, much like AGI, the concept of SGI remains frustratingly nebulous, often acting as a moving goalpost that hinders clear evaluation and progress.

This paper aims to provide a comprehensive, quantifiable framework to cut through this ambiguity, starting with a concrete definition grounded in established theory:
\begin{quote}
\textbf{\emph{"SGI is an AI that can autonomously navigate the complete, iterative cycle of scientific inquiry with the versatility and proficiency of a human scientist"}}
\end{quote}

To operationalize this definition, we ground our approach in the \textbf{Practical Inquiry Model}~\cite{garrison1999critical, garrison2001critical}, a theoretical framework that deconstructs the scientific process into a cycle of four core cognitive activities. This model provides a taxonomic map of scientific cognition through four distinct, interdependent quadrants (Figure~\ref{fig:teaser}): \textbf{Deliberation} (the search, synthesis, and critical evaluation of knowledge), \textbf{Conception} (the generation of ideas), \textbf{Action} (the practical implementation via experiments), and \textbf{Perception} (the awareness and interpretation of results). An AI exhibiting true SGI must possess robust capabilities across this entire spectrum.
This four-quadrant framework provides a conceptual taxonomy of scientific cognition and forms the foundation for an \emph{operational definition} of SGI—one that specifies what kinds of planning, knowledge creation and reasoning an AI must demonstrate to qualify as scientifically intelligent. Translating this operational definition into measurable criteria requires examining how current evaluations of AI intelligence align with, or deviate from, this framework. Identifying these gaps is essential for clarifying what existing assessments capture and what they overlook in defining \textbf{Scientific General Intelligence}.

Grounded in this four-quadrant definition of SGI, we examine how existing benchmarks operationalize scientific reasoning. Most current evaluations capture only fragments of the SGI spectrum. For instance, MMLU~\cite{hendrycks2020measuring} and SuperGPQA~\cite{du2025supergpqa} focus on multidisciplinary knowledge understanding—corresponding mainly to the \textit{Deliberation} quadrant—while GAIA~\cite{mialon2023gaia} emphasizes procedural tool use aligned with \textit{Action}. HLE~\cite{phan2025humanity} further raises difficulty through complex reasoning, yet still isolates inquiry stages without integrating the practical or interpretive cycles that characterize real scientific investigation.
Collectively, these benchmarks present a fragmented view of scientific intelligence. Their disciplinary scope remains narrow, their challenges seldom reach expert-level reasoning, and—most crucially—they frame inquiry as a static, closed-domain question-answering task. This abstraction neglects the creative, procedural, and self-corrective dimensions central to SGI, meaning that what is currently measured as “scientific ability” reflects only a limited slice of true Scientific General Intelligence.

Thus, to concretize the proposed definition of \textbf{Scientific General Intelligence (SGI)}, we develop \textbf{SGI-Bench: A Scientific Intelligence Benchmark for LLMs via Scientist-Aligned Workflows}. Rather than serving as yet another performance benchmark, SGI-Bench functions as an \emph{operational instantiation} of the SGI framework, quantitatively evaluating LLMs across the full spectrum of scientific cognition defined by the \textbf{Practical Inquiry Model}. By design, SGI-Bench is comprehensive in its disciplinary breadth, challenging in its difficulty, and unique in its explicit coverage of all four capabilities central to our definition of SGI. The benchmark structure is therefore organized into four corresponding task categories:
\begin{itemize}
\item \textbf{Scientific Deep Research (Deliberation):} This task evaluates models' ability to perform iterative, multi-step reasoning over complex scientific content.
\item \textbf{Idea Generation (Conception):} This task assesses creativity and methodological planning by asking models to generate novel hypotheses or experimental designs.
\item \textbf{Dry/Wet Experiment (Action):} This task evaluates the ability to plan and execute computational (dry) or laboratory-style (wet) experiments.
\item \textbf{Experimental Reasoning (Perception):} This task requires models to analyze experimental results, interpret data trends, and identify meaningful conclusions.
\end{itemize}

Building upon our theoretical framework, the construction of SGI-Bench operationalizes the proposed definition of \textbf{Scientific General Intelligence (SGI)}. We began with foundational topics drawn from \textit{Science's 125 Big Questions for the 21st Century}~\cite{sanders2021125}, spanning ten major disciplinary areas. Through multi-round collaborations with domain experts, we identified high-impact research problems and curated raw source materials from leading journals such as \textit{Nature}, \textit{Science}, and \textit{Cell}. Together with PhD-level researchers, we implemented a multi-stage quality control pipeline involving human annotation, model-based verification, and rule-based consistency checks. The resulting benchmark comprises over 1,000 expert-curated samples that concretely instantiate the reasoning, creativity, and experimental competencies central to our definition of SGI.

To evaluate performance across these four dimensions, we found that conventional “LLM-as-a-judge”~\cite{li2025generation} paradigms are insufficient to handle the diverse and specialized metrics required by SGI assessment. To address this, we developed an agent-based evaluation framework following an \textbf{Agent-as-a-judge}~\cite{zhuge2024agent} paradigm. Equipped with tools such as a web search interface, Python interpreter, file reader, PDF parser, and discipline-specific metric functions, this framework ensures rigor, scalability, and transparency. It operates through four interdependent stages—\textit{Question Selection}, \textit{Metric Customization}, \textit{Prediction \& Evaluation}, and \textit{Report Generation}—each coordinated by specialized agents aligned with different aspects of scientific inquiry.

Applying SGI-Bench to a wide spectrum of state-of-the-art LLMs reveals a unified picture: while modern models achieve pockets of success, they fall far short of the integrated reasoning required for scientific intelligence.
\begin{itemize}

\item In deep scientific research, models can retrieve relevant knowledge but struggle to perform quantitative reasoning or integrate multi-source evidence; exact-match accuracy remains below 20\% and often collapses on numerical or mechanistic inference.

\item  In idea generation, models show substantial deficits in realization. This manifests in underspecified implementation steps and frequent proposals that lack actionable detail or fail basic feasibility checks.

\item  In dry experiments, even strong models fail on numerical integration, simulation fidelity, and scientific code correctness, revealing a gap between syntactic code fluency and scientific computational reasoning.

\item In wet experiments, workflow planning shows low sequence similarity and error-prone parameter selection, with models frequently omitting steps, misordering actions, or collapsing multi-branch experimental logic.

\item In multimodal experimental reasoning, models perform better on causal and perceptual reasoning but remain weak in comparative reasoning and across domains such as materials science and earth systems.

\item Across tasks, closed-source models demonstrate only a marginal performance advantage over open-source models. Even the best closed-source system achieves an SGI-Score of around 30/100, reflecting that current AI models possess relatively low capability in multi-task scientific research workflows, and remain far from proficient for integrated, real-world scientific inquiry.

\end{itemize}

Collectively, these findings demonstrate that current LLMs instantiate only isolated fragments of scientific cognition. They remain constrained by their linguistic priors, lacking the numerical robustness, procedural discipline, multimodal grounding, and self-corrective reasoning loops essential for scientific discovery.

Because genuine scientific inquiry is inherently open-ended and adaptive, we further explore how SGI may emerge under test-time learning dynamics. Preliminary experiments using test-time scaling~\cite{zhang2025survey} and reinforcement learning~\cite{zuo2025ttrl} suggest that models can enhance hypothesis formation and reasoning through minimal unlabeled feedback. This adaptive improvement provides empirical support for viewing \textbf{Scientific General Intelligence} not as a static property, but as a dynamic capacity that can evolve through iterative, self-reflective reasoning cycles.

In summary, this work provides a principle-grounded definition of \textbf{Scientific General Intelligence (SGI)} and a corresponding framework for its empirical study. By formalizing the cognitive cycle of scientific inquiry and operationalizing it through SGI-Bench, we clarify what it means for an AI to exhibit scientific intelligence in both theory and practice. While not a final answer, this definition establishes a concrete path for future research—linking conceptual understanding with measurable progress toward AI systems capable of genuine scientific reasoning and discovery.

%% file: sections/2-benchmark.tex
\section{Scientific General Intelligence: Concept and Operational Definition}

Scientific General Intelligence (SGI) refers to an AI system capable of engaging in the full cycle of scientific inquiry with autonomy, versatility, and methodological rigor. Unlike systems that excel at isolated reasoning tasks, an SGI-capable model must integrate knowledge retrieval, idea formation, action execution, and evidence-based interpretation into a coherent, iterative workflow.

To formalize this notion, we characterize scientific cognition through four interdependent stages: \textbf{Deliberation} (evidence search, synthesis, and critical assessment), \textbf{Conception} (generation of hypotheses and ideas), \textbf{Action} (implementation of experiments or simulations), and \textbf{Perception} (interpretation of empirical results). 

Grounded in this framework, we provide an operational definition: an AI system exhibits SGI if it can (1) retrieve, synthesize, and critically evaluate knowledge; (2) generate scientifically grounded and novel ideas; (3) plan and execute experimental procedures; (4) interpret empirical outcomes with causal and contextual awareness.

This definition highlights a central limitation in existing benchmarks~\cite{hendrycks2020measuring, du2025supergpqa, mialon2023gaia, phan2025humanity}: most evaluate factual recall or single-step reasoning, but few examine the structured, long-horizon workflows that constitute real scientific inquiry. 

Building on the operational definition of \textbf{SGI} established in the previous section, we introduce \textit{SGI-Bench} (Scientific Intelligence Benchmark for LLMs via Scientist-Aligned Workflows) — a benchmark designed to empirically evaluate the extent to which large language models (LLMs), vision-language models (VLMs), and agent-based systems exhibit the cognitive and procedural abilities required for scientific discovery. 

SGI-Bench systematically measures AI performance across 10 core scientific domains — astronomy, chemistry, earth science, energy, information science, life science, materials science, neuroscience, physics and math — providing a panoramic view of how AI systems engage with scientific reasoning across disciplines. Its task design draws inspiration from the seminal article \textit{125 Questions: Exploration and Discovery}~\cite{sanders2021125} published in \textit{Science}, ensuring both disciplinary breadth and societal relevance.

At the heart of SGI-Bench lies the principle of \textit{scientist alignment}—the commitment to evaluating models under conditions that authentically mirror real scientific workflows.
This concept manifests in several ways:
\begin{itemize}
    \item The task designs closely mirror the real-world research scenarios encountered by scientists in their work, ensuring that each task is intrinsically tied to the scientific discovery process.
    \item The raw materials used in task construction are sourced directly from scientists, ensuring the authenticity and relevance of the content.
    \item Scientists have been closely involved in the process of constructing the benchmark, with a \textit{scientist-in-the-loop} approach, ensuring the tasks reflect the nuances of actual scientific workflows.
    \item The final evaluation scores are aligned with the checklist based on the needs of real scientific research scenarios from scientists, which ensures that the assessments genuinely reflect the scientific utility of the models.
\end{itemize}

SGI-Bench departs from conventional benchmarks that emphasize factual recall or single-turn reasoning. Instead, it operationalizes the long-horizon workflow of scientific discovery into four interdependent stages: literature review(Deliberation), methodology design(conception), experiment implementation(Action), and experimental analysis(Perception). These stages correspond to fundamental capabilities required of AI systems: information integration and understanding(Scientific Deep Research), design and planning(Idea Generation), experimental execution(Dry/Wet Experiment), and reasoning-based interpretation(Experimental Reasoning). Together, they form a unified framework that measures not only what models know but how they think, plan, and adapt in pursuit of new knowledge.

\begin{figure}[ht]
\centerline
{\includegraphics[width=16cm]{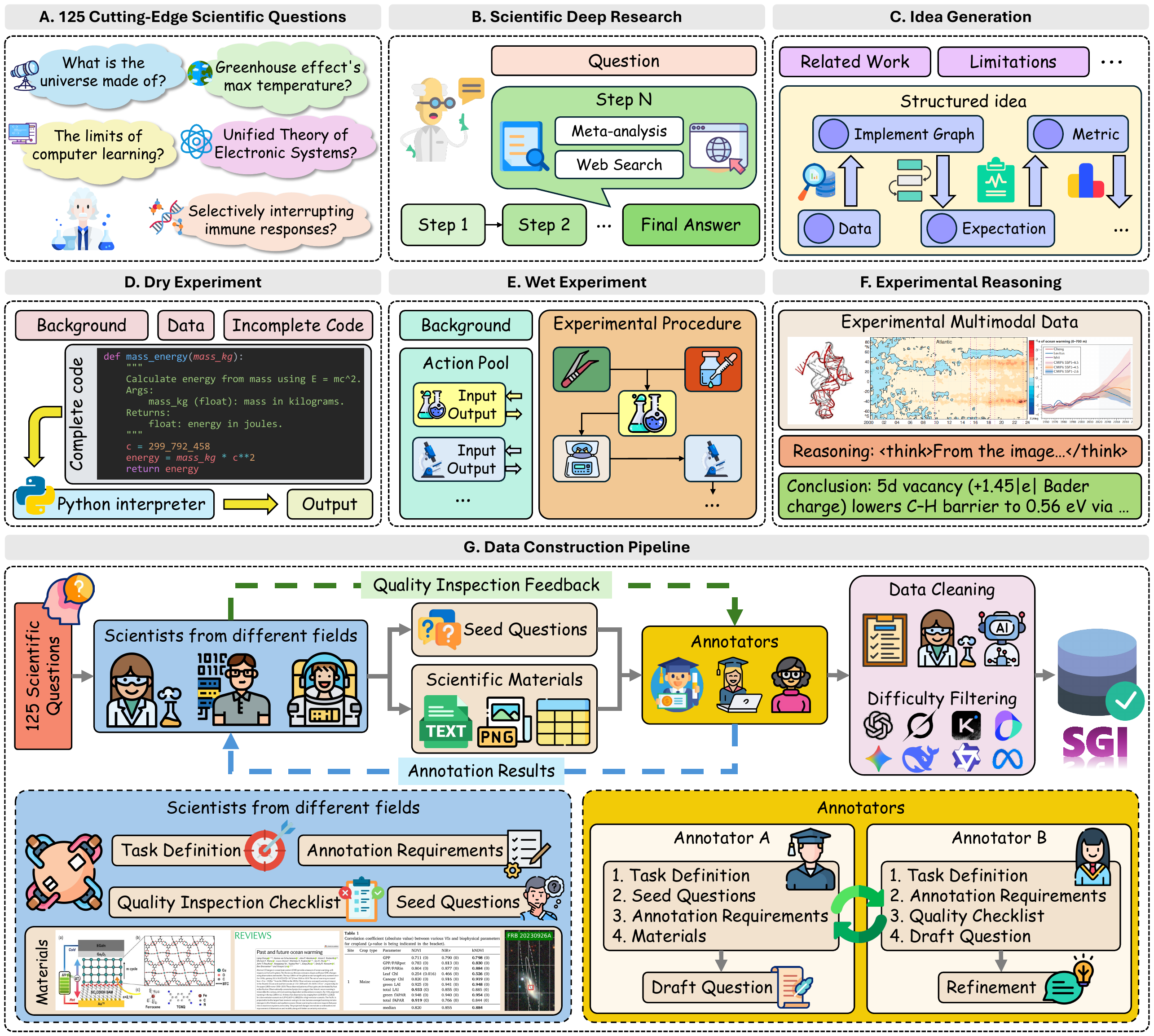}}
\caption{\textbf{SGI-Bench Workflow Pipeline}: The end-to-end four-stage framework (Deliberation, Conception, Action, Perception) that operationalizes scientific discovery, mapping tasks to capabilities and aligning evaluation with scientist practice.}
\label{fig: pipeline}
\end{figure}

\subsection{Task Definition in Scientific Workflow}

\subsubsection{Scientific Deep Research}
Scientific deep research refers to a thorough and comprehensive investigation of a specific scientific topic, combining elements of both AI-driven deep research~\cite{xu2025comprehensive, hu2025flowsearch, shi2025dualresearch} and scientific meta-analysis~\cite{field2010meta, xu2025manalyzer}. This task typically involves multi-step reasoning, web searches, document retrieval, and data analysis~\cite{ac1f09077393404a8bea5141d8710259,trivedi-etal-2023-interleaving,asai2024selfrag}. Drawing inspiration from AI's deep research, which often relies on multi-hop searches to gather diverse information across multiple sources~\cite{shao-etal-2024-assisting}, it also incorporates the methodology of meta-analysis from the scientific community. Meta-analysis, a rigorous form of scientific research, synthesizes existing literature to derive precise, data-driven conclusions and extract quantitative insights from a large body of studies. Unlike general deep research, which may focus on qualitative understanding, meta-analysis centers on aggregating and analyzing data to produce statistically significant results. By combining the multi-hop search nature of AI's deep research with the systematic, evidence-based approach of meta-analysis, this task ensures results that are both scientifically precise and meaningful. The ability to perform scientific deep research is crucial for advancing scientific knowledge, as it enables AI models to replicate the process of reviewing, synthesizing, and analyzing existing research to formulate new, data-driven hypotheses.~\cite{Wang2023,lu2024aiscientistfullyautomated}

Deep Research comprises multiple forms including literature inquiry~\cite{Bosse2025DeepRB}, report-style reasoning~\cite{du2025deepresearchbenchcomprehensivebenchmark} and so on. In this benchmark, we focus on literature-inquiry–centric deep research, where the model identifies and integrates relevant scientific knowledge from provided sources. This process often involves unit verification, quantitative interpretation, and causal assessment—abilities fundamental to scientific reasoning and still challenging for current AI systems. By constraining the task to literature inquiry rather than broader report-generation settings, we ensure greater reproducibility and more reliable evaluation, while still probing a core component of scientific inquiry.

In order to capture the diversity of real-world scientific inquiries, we divide the task of scientific deep research into four representative types: data, properties, micro-experiments, and macro-experiments, as illustrated in Table~\ref{tab:deep_research_types}. This division reflects the major types of questions scientists often confront, ranging from data-centric queries to property characterization, and from small-scale controlled experiments to large-scale natural events. By organizing the task in this way, the benchmark ensures that AI systems are evaluated across the breadth of literature review and data-driven investigation.

\begin{table}[ht]
\centering
\caption{\textbf{Scientific Deep Research Types}: Four representative categories of inquiry targets and their roles in the scientific workflow.}
\label{tab:deep_research_types}
\resizebox{16.0cm}{!}{
\begin{tabular}{p{3.5cm}p{6cm}p{6cm}}
\toprule
\textbf{Type} & \textbf{Core Description} & \textbf{Role in Scientific Workflow} \\
\midrule
\rowcolor{blue!5}Data & Focused on retrieving or analyzing structured datasets, such as event counts, statistical summaries, or dataset-specific attributes. & Supports quantitative literature review and provides a foundation for identifying trends or anomalies. \\
\addlinespace
Property & Concerned with identifying or inferring material, molecular, or system properties, often requiring interpretation of experimental results or theoretical knowledge. & Bridges literature review with methodology design by clarifying key parameters. \\
\addlinespace
\rowcolor{blue!5}Micro-experiment & Small-scale controlled experiments, often involving chemical reactions, physical transformations, or laboratory processes under specific conditions. & Provides simulated reasoning over experimental procedures and outcomes. \\
\addlinespace
Macro-experiment & Large-scale or natural experiments, such as astronomical events, climate observations, or geophysical phenomena. & Extends literature review to global or long-term observations, anchoring hypotheses in real-world contexts. \\
\bottomrule
\end{tabular}
}
\end{table}

In real-world scientific workflows, deep research corresponds to the literature review stage. During this stage, scientists investigate existing studies, gather data, and analyze findings to understand the current state of knowledge and identify knowledge gaps that require further investigation.

\begin{tcolorbox}[
    breakable,
    title=Task Definition of Scientific Deep Research,
    colback=LighterGray,
    colframe=DeepPurple,
    colbacktitle=DeepPurple,
    coltitle=White
]
\textbf{\emph{\textcolor{DeepPurple}{Task Input}}}
\begin{itemize}
    \item \texttt{\textbf{Background (B)}}: A detailed background of the research topic, including the scientific field and subfields, to avoid ambiguities in terminology.
    \item \texttt{\textbf{Constraints (C)}}: Constraints such as experimental settings, scientific assumptions, and data sources that frame the problem appropriately.
    \item \texttt{\textbf{Data (D)}}: Any experimental or empirical data directly mentioned in the task, which might be either explicitly provided or inferred.
    \item \texttt{\textbf{Question (Q)}}: A specific, focused question that the task aims to address, such as determining a particular quantity or its variation over time.
    \item \texttt{\textbf{Response Requirements (R)}}: Specifications for the answer, including the required units and whether the answer should be an integer or a decimal with a specified number of decimal places.
\end{itemize}

\textbf{\emph{\textcolor{DeepPurple}{Task Output}}}
\begin{itemize}
    \item \texttt{\textbf{Steps (S)}}: A detailed, step-by-step approach that the system uses to retrieve and process data or perform reasoning.
    \item \texttt{\textbf{Answer (A)}}: A precise numerical or string-based response, such as a specific value or a phrase.
\end{itemize}

\textbf{\emph{\textcolor{DeepPurple}{Task Formulation}}}
\[
\texttt{S, A} = \texttt{LLM/Agent}(\texttt{B, C, D, Q, R})
\]
\end{tcolorbox}

\begin{figure}[ht]
    \vspace{0.01cm}
    \caption{\textbf{Scientific Deep Research Task}: Inputs, outputs, and formulation for literature-driven quantitative inquiry combining multi-step reasoning and meta-analysis.}
    \label{fig:Task Definition of Scientific Deep Research}
\end{figure}

\subsubsection{Idea Generation}
\label{sec: Idea gen}
Idea generation is a critical component of the scientific process, corresponding to the stage of research methodology design. At this stage, researchers synthesize existing knowledge, engage in associative and creative thinking, and propose new approaches to address current challenges. It embodies the creative essence of scientific inquiry and shapes the direction and potential impact of subsequent research.

In real-world scientific workflows, idea generation typically occurs after researchers have completed a thorough literature review. They integrate prior findings, identify limitations or knowledge gaps, and use creative reasoning to formulate new hypotheses, methods, or frameworks aimed at overcoming these shortcomings. In this sense, idea generation serves as the crucial link between literature understanding and methodological innovation. 

However, because idea generation is an open-ended and highly creative task, its evaluation is inherently challenging. In principle, scientific ideas span a wide spectrum from high-level hypotheses to fully specified methodological plans~\cite{Wan2025DeepResearchAT, popper2005logic, yang2024moose}. Evaluating the quality of open-ended hypotheses—those with substantial conceptual freedom and without explicit implementation structure—requires extensive human expert review to achieve even a modest degree of inter-rater reliability and public defensibility. Such large-scale expert adjudication is beyond the practical scope of this version of the benchmark.

Consequently, our current Idea Generation evaluation focuses on the methodological-design component of an idea—i.e., how a proposed approach is operationalized through data usage, step-by-step procedures, evaluation protocols, and expected outcomes. This component offers a more constrained structure that enables measurable, partially automatable assessment while still reflecting an essential aspect of scientific ideation. We view this as a pragmatic starting point, and future versions of the benchmark may incorporate broader hypothesis-level evaluation once sufficiently robust expert-sourced ground truth becomes feasible.

To make the assessment more systematic and tractable, we decompose an originally holistic idea into several interrelated components, forming a structured representation of the idea. This decomposition enables more fine-grained evaluation along dimensions such as effectiveness, novelty, level of detail, and feasibility~\cite{moose}.

\begin{tcolorbox}[
    breakable,
    title=Task Definition of Idea Generation,
    colback=LighterGray,
    colframe=DeepPurple,
    colbacktitle=DeepPurple,
    coltitle=White
]
\textbf{\emph{\textcolor{DeepPurple}{Task Input}}}
\begin{itemize}
    \item \texttt{\textbf{Related Work (RW)}}: A summary of existing research relevant to a certain research direction, providing context for new ideas.
    \item \texttt{\textbf{Challenge (C)}}: The current challenges in the field and the limitations of existing solutions.
    \item \texttt{\textbf{Limitation (L)}}: Specific shortcomings or constraints of current research that new ideas need to address.
    \item \texttt{\textbf{Motivation (M)}}: The perspective and motivation of addressing the limitations in this research direction.
    \item \texttt{\textbf{Task Objective (TO)}}: The primary goal of the task, such as generating ideas that solve identified challenges or improve existing solutions.
    \item \texttt{\textbf{Existing Solutions (ES)}}: A description of the current approaches or solutions available in the field.
\end{itemize}

\textbf{\emph{\textcolor{DeepPurple}{Task Output}}}
\begin{itemize}
    \item \texttt{\textbf{Core Idea (CI)}}: The central novel idea or concept generated to address the research challenge.
    \item \texttt{\textbf{Implementation Steps (IS)}}: The steps or procedures required to implement the core idea.
    \item \texttt{\textbf{Implementation Order (IO)}}: The sequence in which the implementation steps should be executed.
    \item \texttt{\textbf{Data (D)}}: The data that will be used to implement the idea or evaluate its effectiveness.
    \item \texttt{\textbf{Evaluation Metrics (EM)}}: The criteria for assessing the success or relevance of the generated idea.
    \item \texttt{\textbf{Expected Outcome (EO)}}: The anticipated result or contribution the idea is expected to achieve.
\end{itemize}

\textbf{\emph{\textcolor{DeepPurple}{Task Formulation}}}
\[
\texttt{CI, IS, IO, D, EM, EO} = \texttt{LLM/Agent}(\texttt{RW, C, L, M, TO, ES})
\]
\end{tcolorbox}

\begin{figure}[ht]
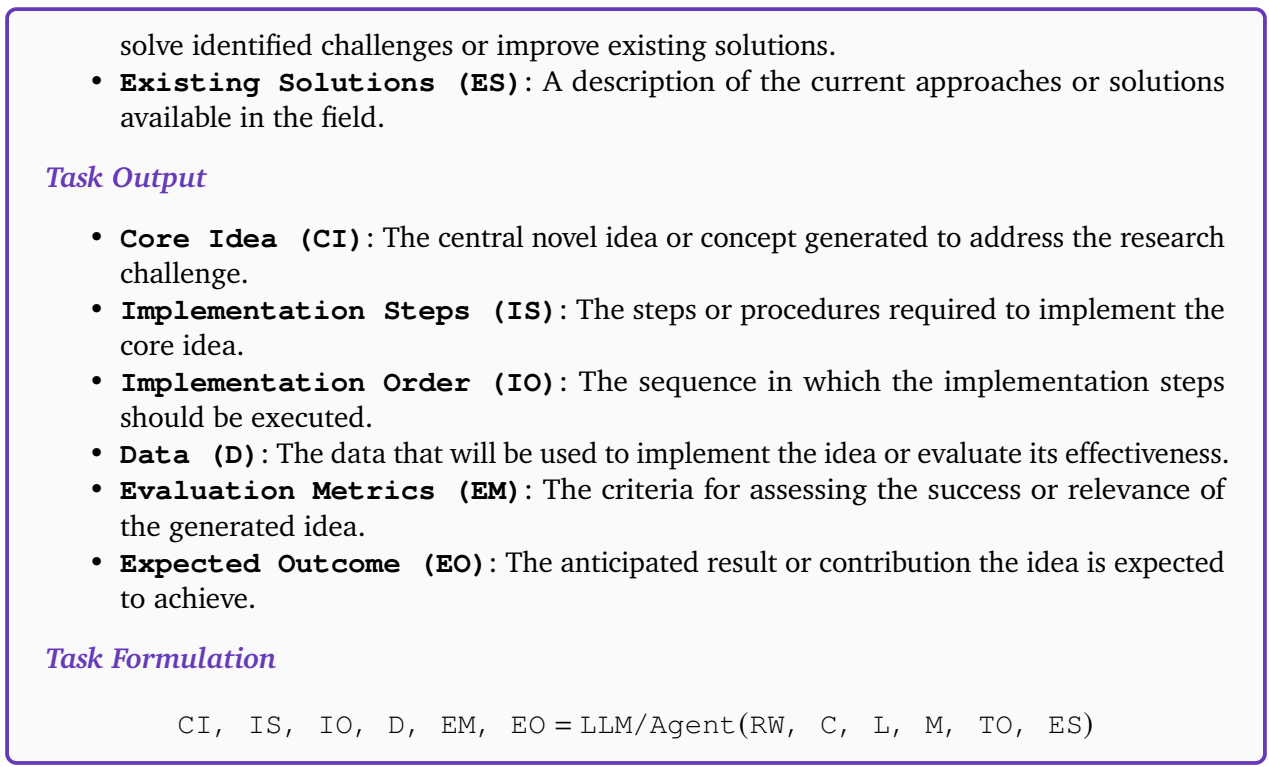

    \vspace{0.01cm}
    \caption{\textbf{Idea Generation Task}: Inputs, outputs, and formulation for methodology design, integrating evaluation metrics and structured implementation planning.}
    \label{fig:Task Definition of Idea Generation}
\end{figure}

\subsubsection{Dry/Wet Experiment}
\label{sec: Task Definition of Experiment}
Scientific experimentation represents the core of the discovery process, bridging theoretical formulation and empirical validation~\cite{Wang2023}. Within SGI-Bench, we formalize this process into two complementary categories: \textit{dry} and \textit{wet} experiments. Dry experiments capture computational and simulation-based studies—where AI assists in generating, refining, or executing scientific code that models physical phenomena.~\cite{RomeraParedes2023,ma2024eureka} Wet experiments, by contrast, simulate laboratory-based workflows, requiring the model to plan and reason about sequences of actions involving physical instruments, reagents, and procedural parameters~\cite{Boiko2023,bran2023augmenting}. Together, these two categories span the continuum from theoretical abstraction to empirical realization, offering a holistic evaluation of how AI can assist scientists in both virtual and physical experimentation.

Computational and laboratory experiments take many forms in real scientific practice. For dry experiments, possible tasks range from full pipeline construction to simulation design and multi-module scientific computing; in this benchmark, we adopt a code-completion–based formulation~\cite{tian2024scicoderesearchcodingbenchmark}, where the model fills in missing components of an existing scientific script rather than generating an entire project from scratch. For wet experiments, laboratory workflows span diverse operational activities, yet we focus on the protocol-design aspect~\cite{Liu2025BioProBenchCD}, where the model composes a sequence of experimental actions and parameters from a predefined action space.

By constraining dry and wet experiments to code completion and protocol design respectively, we retain core aspects of computational and laboratory reasoning while ensuring reproducibility, controlled variability, and reliable evaluation across models.

\paragraph{Dry Experiment}
Dry experiments emphasize computational problem-solving, reflecting the growing role of AI in automating simulation-driven science. Each task presents the model with incomplete or masked scientific code that encapsulates domain-specific computations, such as molecular dynamics, climate modeling, or numerical solvers in physics~\cite{cox2025mdcrow}. The model must infer the missing logic, reconstruct executable code, and ensure that the resulting program produces correct and efficient outcomes. This task thus evaluates a model's ability to integrate scientific understanding with code synthesis—testing not only syntactic correctness but also conceptual fidelity to the underlying scientific problem~\cite{tian2024scicoderesearchcodingbenchmark}.

To better characterize the scope of dry experiments, we categorize representative computational functions commonly encountered across disciplines, including numerical calculation, statistical analysis, simulation, metric calculation, data processing, and predictive modeling, as shown in Table~\ref{tab:dry_experiment_functions}. The completion or generation of these functions offers a rigorous measure of how well AI systems can operationalize scientific intent into executable form.

\begin{table}[ht]
\centering
\caption{\textbf{Dry Experiment Function Types}: Representative computational functions and their roles across scientific code-completion tasks.}
\label{tab:dry_experiment_functions}
\resizebox{15.5cm}{!}{
\begin{tabular}{p{5cm}p{10cm}}
\toprule
\textbf{Function Category} & \textbf{Core Role in Scientific Experiments} \\
\midrule
\rowcolor{blue!5}Numerical Calculation & Basic mathematical computations required to support physical or chemical modeling. \\
\addlinespace
Statistical Analysis & Processing experimental data using descriptive or inferential statistics to identify trends and distributions. \\
\addlinespace
\rowcolor{blue!5}Simulation & Running computational simulations (e.g., molecular dynamics, finite element analysis) and filtering results for relevant conditions. \\
\addlinespace
Metric Calculation & Computing evaluation metrics such as accuracy, error, or performance indicators for validating experiments. \\
\addlinespace
\rowcolor{blue!5}Data Processing & Handling raw data before and after experiments, including normalization, cleaning, and feature extraction. \\
\addlinespace
Predictive Modeling & Applying machine learning methods to categorize, predict, or group experimental results. \\
\bottomrule
\end{tabular}
}
\end{table}

In real scientific workflows, dry experiments correspond to the stage of experimental design in computational and simulation-based studies. Following hypothesis formulation, researchers employ virtual experiments to anticipate and evaluate potential outcomes prior to empirical validation, enabling a cost-efficient and theoretically grounded pre-assessment of experimental feasibility.

\begin{tcolorbox}[
    breakable,
    title=Task Definition of Dry Experiment,
    colback=LighterGray,
    colframe=DeepPurple,
    colbacktitle=DeepPurple,
    coltitle=White
]
\textbf{\emph{\textcolor{DeepPurple}{Task Input}}}
\begin{itemize}
    \item \texttt{\textbf{Background (B)}}: Information from relevant scientific code, providing context for the dry experiment.
    \item \texttt{\textbf{Data Code (D)}}: The data used in the experiment, including any code snippets or predefined inputs.
    \item \texttt{\textbf{Main Code (M)}}: The core experimental code where some functions may be masked or missing.
\end{itemize}

\textbf{\emph{\textcolor{DeepPurple}{Task Output}}}
\begin{itemize}
    \item \texttt{\textbf{Functions (F)}}: The missing functions in the main code \( M \), which the system is tasked with generating or completing.
\end{itemize}

\textbf{\emph{\textcolor{DeepPurple}{Task Formulation}}}
\[
\texttt{F} = \texttt{LLM/Agent}(\texttt{B, D, M})
\]
\end{tcolorbox}

\begin{figure}[ht]
    \vspace{0.01cm}
    \caption{\textbf{Dry Experiment Task}: Inputs, outputs, and formulation for code-completion based computational studies with masked functions.}
    \label{fig:Task Definition of Dry Experiment}
\end{figure}

\paragraph{Wet Experiment}
Wet experiments represent the physical realization of scientific inquiry, encompassing laboratory and field-based procedures that transform theoretical designs into empirical evidence. These tasks simulate the execution phase of real-world experiments, where models are required to plan, organize, and reason through sequences of atomic actions involving materials, instruments, and procedural parameters. Given inputs describing experimental objectives, configurations, and available tools, the model must generate structured, executable protocols that are both accurate and practically feasible. Evaluation considers not only the correctness of individual steps but also their procedural coherence and alignment with established laboratory conventions.

In real scientific workflows, wet experiments correspond to the execution and validation stages of discovery. This is where hypotheses are tested against the physical world, data are collected, and evidence is generated to confirm, refine, or refute prior assumptions. By assessing how effectively AI systems can design and reason through these embodied experimental processes, this task provides a window into their capacity to bridge symbolic understanding with real-world scientific practice.

\begin{tcolorbox}[
    breakable,
    title=Task Definition of Wet Experiment,
    colback=LighterGray,
    colframe=DeepPurple,
    colbacktitle=DeepPurple,
    coltitle=White
]
\textbf{\emph{\textcolor{DeepPurple}{Task Input}}}
\begin{itemize}
    \item \texttt{\textbf{Background (B)}}: Information from relevant experimental procedure.
    \item \texttt{\textbf{Action Pool (AP)}}: A predefined set of atomic actions that can be used in the experiment, along with explanations and corresponding input/output definitions.
\end{itemize}

\textbf{\emph{\textcolor{DeepPurple}{Task Output}}}
\begin{itemize}
    \item \texttt{\textbf{Atomic Action Order (AAO)}}: The order in which atomic actions should be executed.
    \item \texttt{\textbf{Atomic Action Parameters (AAP)}}: The parameters associated with each atomic action (e.g., reagents, temperature).
\end{itemize}

\textbf{\emph{\textcolor{DeepPurple}{Task Formulation}}}
\[
\texttt{AAO, AAP}= \texttt{LLM/Agent}(\texttt{B, AP})
\]
\end{tcolorbox}

\begin{figure}[ht]
    \vspace{0.01cm}
    \caption{\textbf{Wet Experiment Task}: Inputs, outputs, and formulation for laboratory protocol planning via atomic actions and parameters.}
    \label{fig:Task Definition of Wet Experiment}
\end{figure}

\subsubsection{Experimental Reasoning}

Experimental reasoning refers to the process of interpreting scientific observations and data to reach justified conclusions. In this benchmark, we focus on data-analysis–oriented reasoning~\cite{zhou2025scientistsexamprobingcognitive}, where the model must extract relevant visual or numerical cues from multi-modal sources~\cite{zhang2024cmmmuchinesemassivemultidiscipline}, compare conditions, and identify causal or descriptive patterns. This formulation emphasizes analytical interpretation rather than open-form scientific narrative, enabling reliable assessment while capturing an essential part of empirical scientific reasoning.

We consider five representative modalities as shown in Table~\ref{tab:modalities}: a) process images that integrate symbolic and textual information to depict workflows or variable relationships; b) observation images representing raw data captured by instruments such as telescopes, satellites, or microscopes; c) experiment images documenting laboratory setups and procedures; d) simulation images generated by computational models to visualize physical or chemical processes; and e) visualization images such as plots or charts that reveal patterns within structured datasets. Collectively, these modalities reflect the multi-faceted and evidence-driven nature of scientific inquiry.

\begin{table}[ht]
\centering
\caption{\textbf{Experimental Reasoning Modalities}: Five visual modalities used for multi-modal evidence and analysis.}
\label{tab:modalities}
\resizebox{16.0cm}{!}{
\begin{tabular}{p{4cm}p{6cm}p{6cm}}
\toprule
\textbf{Modality} & \textbf{Core Description} & \textbf{Scientific Role} \\
\midrule
\rowcolor{blue!5}Process Images & Graphical symbols + text describing workflows or variable relations. & Capture the logical flow of experiments and research design. \\
\addlinespace
Observation Images & Raw data from instruments (e.g., telescope, satellite, microscope). & Provide direct evidence of natural or physical phenomena. \\
\addlinespace
\rowcolor{blue!5}Experiment Images & Photos of instruments, setups, or lab operations. & Document experimental configurations and operational details. \\
\addlinespace
Simulation Images & Generated from computational models/software. & Visualize theoretical predictions of physical or chemical processes. \\
\addlinespace
\rowcolor{blue!5}Visualization Images & Processed structured data into charts/plots. & Reveal patterns, comparisons, or correlations from datasets. \\
\bottomrule
\end{tabular}
}
\end{table}

To reason effectively over such diverse inputs, we define four complementary reasoning paradigms as shown in Table~\ref{tab:reasoning_paradigms}: a) signal perception, focusing on the extraction of direct patterns from visual signals; b) attribute understanding, which demands domain knowledge to interpret key visual or contextual features; c) comparative reasoning, involving integration and comparison across multiple sources to ensure consistency and rigor; and d) causal reasoning, aimed at uncovering underlying mechanisms and scientific principles. These paradigms collectively span the hierarchy from low-level perception to high-level scientific inference.

\begin{table}[ht]
\centering
\caption{\textbf{Experimental Reasoning Paradigms}: Four reasoning paradigms spanning perception to causality with examples and requirements.}
\label{tab:reasoning_paradigms}
\resizebox{16.0cm}{!}{
\begin{tabular}{p{4cm}p{6cm}p{6cm}}
\toprule
\textbf{Reasoning Paradigm} & \textbf{Core Requirement} & \textbf{Typical Example} \\
\midrule
\rowcolor{blue!5}Signal Perception & Direct extraction of information from visual signals without heavy prior knowledge. & Identifying patterns in telescope images or microscope slides. \\
\addlinespace
Attribute Understanding & Requires disciplinary background to interpret key features and scientific attributes. & Recognizing crystalline structures in materials science images. \\
\addlinespace
\rowcolor{blue!5}Comparative Reasoning & Integrates and contrasts information across multiple images, often cross-domain. & Comparing climate model simulations with satellite observations. \\
\addlinespace
Causal Reasoning & Goes beyond correlation to infer mechanisms or propose hypotheses. & Inferring causal pathways in gene expression from multi-modal experimental data. \\
\bottomrule
\end{tabular}
}
\end{table}

In real-world scientific workflows, experimental reasoning corresponds to the data analysis stage, during which scientists interpret experimental and simulated data, perform comparative analyses, and refine hypotheses based on empirical evidence.

\begin{tcolorbox}[
    breakable,
    title=Task Definition of Experimental Reasoning,
    colback=LighterGray,
    colframe=DeepPurple,
    colbacktitle=DeepPurple,
    coltitle=White
]
\textbf{\emph{\textcolor{DeepPurple}{Task Input}}}
\begin{itemize}
    \item \texttt{\textbf{Multiple Experimental Images (MEI)}}: A set of images representing various experimental outcomes or data collected from instruments.
    \item \texttt{\textbf{Question (Q)}}: A specific question or hypothesis related to the experimental data that requires reasoning or analysis.
\end{itemize}

\textbf{\emph{\textcolor{DeepPurple}{Task Output}}}
\begin{itemize}
    \item \texttt{\textbf{Reasoning (R)}}: The specific steps in the reasoning process, including calculation, thinking, analysis, etc..
    \item \texttt{\textbf{Answer (A)}}: The conclusion drawn from analyzing the experimental data, answering the specified question or hypothesis.
\end{itemize}

\textbf{\emph{\textcolor{DeepPurple}{Task Formulation}}}
\[
\texttt{R, A} = \texttt{LLM/Agent}(\texttt{MEI, Q})
\]
\end{tcolorbox}

\begin{figure}[ht]
    \vspace{0.01cm}
    \caption{\textbf{Experimental Reasoning Task}: Inputs, outputs, and formulation for multi-modal analysis with step-by-step reasoning and final answers.}
    \label{fig:Task Definition of Experimental Reasoning}
\end{figure}

\subsection{Multi-Dimensional Metrics}
\label{sec:Metrics}

To align with the scientific characteristics of each task, we have designed multi-dimensional evaluation metrics for every task. This approach avoids a one-size-fits-all binary judgment and instead provides a more fine-grained assessment.

\subsubsection{Metrics of Scientific Deep Research}

The Scientific Deep Research task draws inspiration from AI's deep research paradigms~\cite{wenxiaobai,DeepResearch_OpenAI2025,Perplexity_DeepResearch_2025,KimiResearcher_2025,GrokDeepSearch_2025} while incorporating methodologies from meta-analysis in the scientific domain. The former emphasizes multi-step reasoning, where solving a problem often requires iterative searches, calculations, and inferences; the correctness of each step directly impacts the accuracy of the final answer. The latter focuses on systematically extracting and synthesizing data from literature, requiring highly precise results. Accordingly, our metrics capture both step-by-step reasoning fidelity and final answer accuracy.

\begin{tcolorbox}[
    breakable,
    title=Metric Definition of Exact Match,
    colback=LighterGray,
    colframe=DeepPurple,
    colbacktitle=DeepPurple,
    coltitle=White
]
\textbf{Exact Match (EM)}: Since the Scientific Deep Research tasks are designed to have short, unique, and easily verifiable answers, we use exact match as a hard metric to assess whether the model's final answer is correct. The model receives a score of 1 if the output exactly matches the reference answer, and 0 otherwise.
\end{tcolorbox}

\begin{tcolorbox}[
    breakable,
    title=Metric Definition of Step-Level Accuracy,
    colback=LighterGray,
    colframe=DeepPurple,
    colbacktitle=DeepPurple,
    coltitle=White
]
\textbf{Step-Level Accuracy (SLA)}: Models are required to produce step-by-step solutions. We employ an LLM-based judge to compare each model-generated step against the reference solution steps. For each step, the judge determines whether it is correct and provides reasoning. This fine-grained evaluation avoids binary correctness judgments for the entire solution, allowing precise assessment of reasoning accuracy at each inference step. The metric is computed as the proportion of steps correctly solved relative to the total number of steps. The score is calculated as
\[
\text{SLA} = \frac{\text{Number of correct reasoning steps}}{\text{Total number of reasoning steps}}.
\]

\end{tcolorbox}

\subsubsection{Metrics of Idea Generation}
\label{Metrics: Idea Gen}

To evaluate the open-ended nature of idea generation, we adopt a hybrid framework that integrates both subjective and objective metrics. We assess each idea along four dimensions---\textbf{effectiveness}, \textbf{novelty}, \textbf{detailedness}, and \textbf{feasibility}---which together characterize an idea's scientific quality, creativity, and executability~\cite{moose,ruan2025liveideabenchevaluatingllmsdivergent}.

\textbf{Subjective Evaluation via LLM Judges.}
For subjective scoring, we perform pairwise comparisons between model-generated ideas and expert-written reference ideas. For each of the four dimensions, an LLM judge selects which idea is superior. To ensure fairness and robustness, we employ three different LLM judges, each casting two independent votes, resulting in a total of six votes per dimension. The pairwise win rate against the reference idea is then used as the subjective component of the score for each dimension.

\textbf{Objective Evaluation via Computable Metrics.}
In addition to subjective judgments, we design dimension-specific computational metrics that capture structured properties of the ideas.

\begin{tcolorbox}[
    breakable,
    title=Metric Definition of Effectiveness,
    colback=LighterGray,
    colframe=DeepPurple,
    colbacktitle=DeepPurple,
    coltitle=White
]

For each reference idea, human experts extract its 3--5 most essential keywords. We compute the hit rate of these keywords in the model-generated idea, allowing semantic matches to avoid underestimating effectiveness. The final effectiveness score is the average of the keyword hit rate and the LLM-judge win rate:
\[
\text{Effectiveness} = \frac{\text{Keyword Hit Rate} + \text{LLM Win Rate}}{2}.
\]

\end{tcolorbox}

\begin{tcolorbox}[
    breakable,
    title=Metric Definition of Novelty,
    colback=LighterGray,
    colframe=DeepPurple,
    colbacktitle=DeepPurple,
    coltitle=White
]
We measure novelty by computing the dissimilarity between the model-generated idea and prior related work. Lower similarity indicates that the model proposes ideas not present in existing literature and therefore exhibits higher creativity. The final novelty score averages the dissimilarity score and the subjective win rate:
\[
\text{Novelty} = \frac{\text{Dissimilarity Score} + \text{LLM Win Rate}}{2}.
\]

\end{tcolorbox}

\begin{tcolorbox}[
    breakable,
    title=Metric Definition of Detailedness,
    colback=LighterGray,
    colframe=DeepPurple,
    colbacktitle=DeepPurple,
    coltitle=White
]

We evaluate detailedness from two angles: a) content completeness, which checks whether the idea contains required components (Core Idea, Implementation Steps, Implementation Order, Dataset, Evaluation Metrics, Expected Outcome), and b) redundancy penalty, computed via sentence-level semantic similarity. Ideas with many repetitive sentences are penalized, as verbosity without substance does not constitute genuine detail. The final detailedness score is:
\[
\text{Detailedness} = \frac{\text{Completeness Score (with Penalty)} + \text{LLM Win Rate}}{2}.
\]

\end{tcolorbox}

\begin{tcolorbox}[
    breakable,
    title=Metric Definition of Feasibility,
    colback=LighterGray,
    colframe=DeepPurple,
    colbacktitle=DeepPurple,
    coltitle=White
]

For each research direction, domain experts provide a standardized \textit{implementation graph} containing the essential nodes and their execution order. We extract an implementation graph from each model-generated idea and compute its similarity to the expert template. A low similarity indicates that the proposed idea does not align with accepted solution workflows and is therefore infeasible. The final feasibility score is:
\[
\text{Feasibility} = \frac{\text{Graph Similarity} + \text{LLM Win Rate}}{2}.
\]

\end{tcolorbox}

Taken together, the hybrid subjective--objective design provides a robust, interpretable, and comprehensive assessment of LLMs' scientific idea generation capabilities across creativity, structural clarity, and practical executability.

\subsubsection{Metrics of Dry/Wet Experiment}

\paragraph{Dry Experiment}
\label{sec: Metric of Dry Experiment}
Dry experiments focus on code generation task. Specifically, each problem includes background information, data code, and main code with certain functions masked. The model is tasked with completing the missing functions. Each problem contains 5 unit tests. Our metrics capture both correctness and execution behavior of the generated code~\cite{jain2024livecodebenchholisticcontaminationfree}.

\begin{tcolorbox}[
    breakable,
    title=Metric Definition of Pass All k Unit Tests,
    colback=LighterGray,
    colframe=DeepPurple,
    colbacktitle=DeepPurple,
    coltitle=White
]
\textbf{Pass all k Unit Tests(PassAll@k)}: This metric measures the proportion of problems with k or more unit tests passed successfully. It's important to distinguish this from Pass@k. While Pass@k requires only one successful attempt out of k trials, PassAll@k demands that at least k attempts pass the unit tests. Consequently, PassAll@5 represents the most challenging criterion. The score is calculated as
\[
\text{PassAll@k} = \frac{\text{Number of problems with k or more unit tests passed}}{\text{Total number of problems}}.
\]
\end{tcolorbox}

\begin{tcolorbox}[
    breakable,
    title=Metric Definition of Average Execution Time,
    colback=LighterGray,
    colframe=DeepPurple,
    colbacktitle=DeepPurple,
    coltitle=White
]
\textbf{Average Execution Time (AET)}: This metric captures the efficiency of the generated code by measuring the average runtime across all test cases:
\[
\text{AET} = \frac{1}{N} \sum_{i=1}^{N} t_i,
\]
where \(t_i\) is the execution time of the \(i\)-th test case and \(N\) is the total number of test cases.
\end{tcolorbox}

\begin{tcolorbox}[
    breakable,
    title=Metric Definition of Smooth Execution Rate,
    colback=LighterGray,
    colframe=DeepPurple,
    colbacktitle=DeepPurple,
    coltitle=White
]
\textbf{Smooth Execution Rate (SER)}: This metric measures the proportion of generated code that runs without any runtime errors, regardless of correctness of output. It reflects adherence to basic coding standards and robustness:
\[
\text{SER} = \frac{\text{Number of code executions without errors}}{\text{Total number of code executions}}.
\]
\end{tcolorbox}

\paragraph{Wet Experiment}

Wet experiments involve procedural steps using laboratory instruments. Correct execution requires both the correct sequence of actions and proper parameter settings. Accordingly, we propose the following metrics:

\begin{tcolorbox}[
    breakable,
    title=Metric Definition of Sequence Similarity,
    colback=LighterGray,
    colframe=DeepPurple,
    colbacktitle=DeepPurple,
    coltitle=White
]
\textbf{Sequence Similarity (SS)}: This metric evaluates the similarity between the order of atomic actions provided by the model and the reference sequence. Let \(\text{seq}_\text{model}\) and \(\text{seq}_\text{ref}\) be the sequences of atomic actions from the model and the reference, respectively. Denote by \(\text{Inv}(\text{seq}_\text{model}, \text{seq}_\text{ref})\) the number of discordant pairs between the sequences. For sequences of length \(n\), the score is computed as:
\[
\text{SS} = 1 - \frac{\text{Inv}(\text{seq}_\text{model}, \text{seq}_\text{ref})}{\frac{n(n-1)}{2}},
\]
where \(\frac{n(n-1)}{2}\) is the maximum possible number of inversions. By definition, \(\text{SS} = 1\) indicates that the sequences are identical, while \(\text{SS} = 0\) indicates maximal disorder relative to the reference sequence.
\end{tcolorbox}

\begin{tcolorbox}[
    breakable,
    title=Metric Definition of Parameter Accuracy,
    colback=LighterGray,
    colframe=DeepPurple,
    colbacktitle=DeepPurple,
    coltitle=White
]
\textbf{Parameter Accuracy (PA)}: This metric measures the correctness of input parameters for each atomic action compared to the reference, including reagent types, concentrations, volumes, or other domain-specific parameters. The score is calculated as the proportion of correctly specified parameters across all actions:
\[
\text{PA} = \frac{\text{Number of correctly specified parameters}}{\text{Total number of parameters}}.
\]
\end{tcolorbox}

\subsubsection{Metrics of Experimental Reasoning}
The Experimental Reasoning task assesses the multi-modal scientific reasoning capabilities of LLMs and agents. 
Specifically, given several images and a corresponding question, the model is required to select the correct option from no fewer than 10 candidates. 
For evaluation, the correctness of the final answer and the validity of intermediate reasoning are equally critical. 
Therefore, two evaluation metrics are adopted, as detailed below.

\begin{tcolorbox}[
    breakable,
    title=Metric Definition of MCA,
    colback=LighterGray,
    colframe=DeepPurple,
    colbacktitle=DeepPurple,
    coltitle=White
]
\textbf{Multi-choice Accuracy (MCA)}: Given several options, the model receives a score of 1 if the selected option exactly matches the reference answer, and 0 otherwise.
The final score of MCA is the average of all individual scores across all test samples. 
This metric directly quantifies the model's ability to pinpoint the correct solution from a large candidate pool, serving as a foundational measure of its end-to-end scientific reasoning accuracy in the multi-modal task.
\end{tcolorbox}

\begin{tcolorbox}[
    breakable,
    title=Metric Definition of Reasoning Validity,
    colback=LighterGray,
    colframe=DeepPurple,
    colbacktitle=DeepPurple,
    coltitle=White
]
\textbf{Reasoning Validity (RV)}: 
Models are required to generate step-by-step logical reasoning to justify their selected answers. 
An LLM-based judge is utilized to assess the model-generated reasoning against a reference reasoning. 
For each test sample, the LLM judge assigns a validity score ranging from 0 (completely invalid, contradictory, or irrelevant) to 10 (fully rigorous, logically coherent, and perfectly aligned with the reference reasoning), accompanied by justifications for the assigned score. 
This fine-grained scoring paradigm circumvents the limitations of binary correctness assessments, enabling precise quantification of reasoning quality, including the validity of premises, logical transitions, and alignment with scientific principles. 
The final RV score is computed as the mean of individual sample scores across the entire test set, reflecting the model's overall capability to perform interpretable and reliable scientific reasoning.
\end{tcolorbox}

\subsection{Scientist-Aligned Data Construction}

\begin{figure}[ht]
\centerline
{\includegraphics[width=16cm]{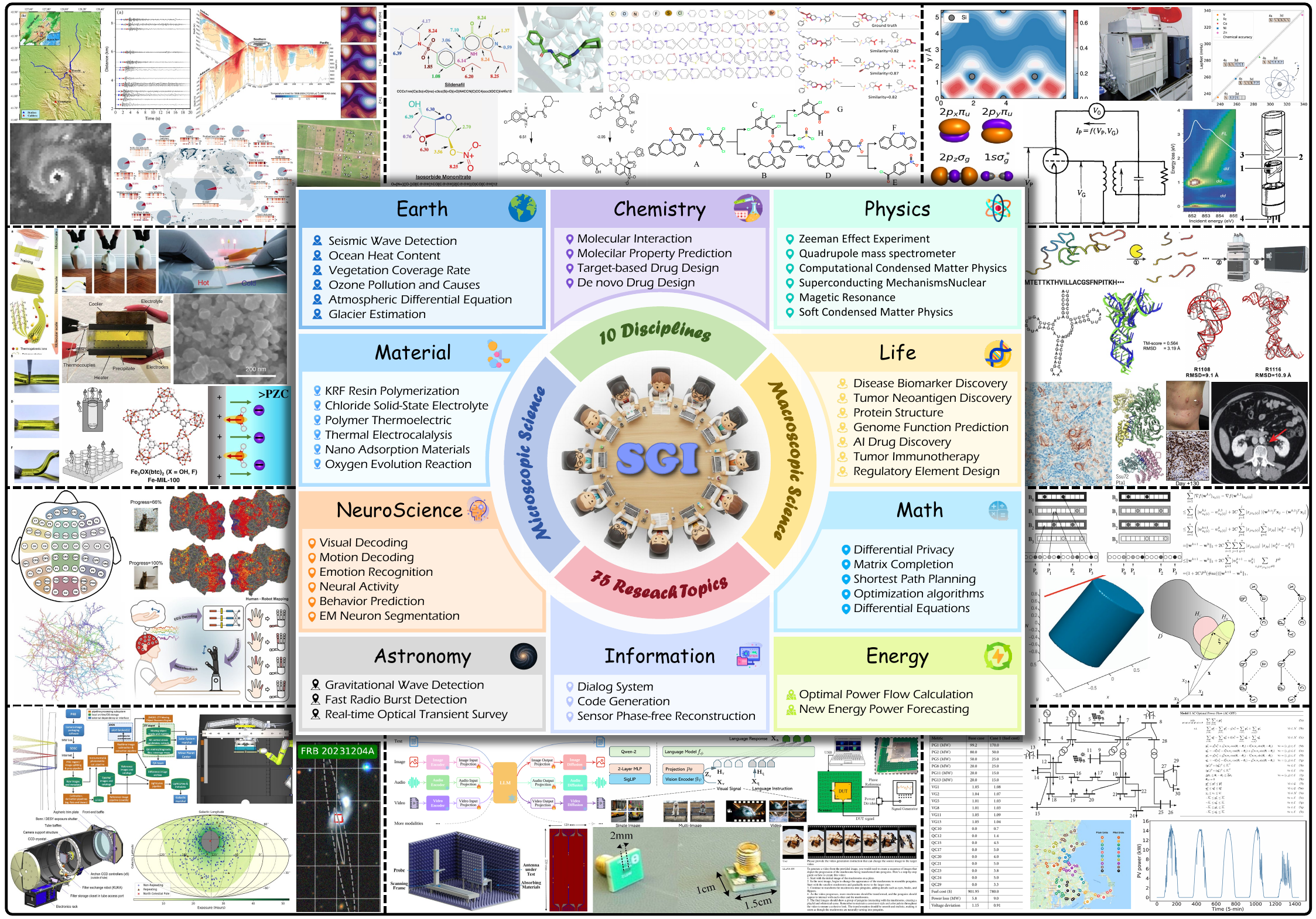}}
\caption{\textbf{Benchmark Subjects}: Overview of 10 scientific domains covered by SGI-Bench.}
\label{fig: subjects}
\end{figure}

\paragraph{Raw Corpus Collection}  
In this stage, we conducted multiple discussions with experts from diverse scientific disciplines, drawing from both the 125 important scientific questions published in \textit{Science}, and the prominent research directions in various disciplines with significant scientific impact. Ultimately, we curated 75 research directions spanning ten scientific domains, as shown in Figure~\ref{fig: subjects}. Please refer to Appendix~\ref{sec: all disciplines} for a complete list of research directions.

Subsequently, we collected raw data provided by experts and researchers, primarily consisting of scientific texts and images across the various disciplines. The texts mainly cover knowledge introduction, methodological design, experimental procedures, and data analysis. The images include experiment figures, data visualizations, and observational images, each accompanied by detailed descriptions.

In addition, these experts and researchers will provide seed questions and annotation requirements for annotation, which provide initial examples for the subsequent annotation process, as illustrated in Figure~\ref{fig: pipeline} (G).

\paragraph{Question Construction}  
After gathering the raw data, we recruited over 100 Master's and PhD holders from different disciplines to construct benchmark questions according to the task definitions. Annotators first analyzed the collected texts and images, and then created questions according to annotation requirements and seed questions. Several rules were applied to ensure scientific validity and authenticity. Specifically, annotators were required to reference the original data source and paragraph for each question, ensuring traceability to scientist-provided data. Furthermore, all questions are constructed by at least two annotators, one of whom is responsible for generating complex draft questions, and the other is responsible for refining them, as shown in Figure~\ref{fig: pipeline} (G).

During question construction, experts continuously reviewed the generated questions. Each question was immediately submitted to the relevant expert for evaluation, who assessed its scientific value. For instance, a question with an experiment configuration that lacks general applicability would be deemed scientifically invalid. Experts provided feedback to annotators, who then revised the questions accordingly, ensuring that the constructed questions remain aligned with the perspectives and standards of domain scientists.

\paragraph{Data Cleaning}  
Once all questions were constructed, we applied three layers of data cleaning:  
1. \textit{Rule-based cleaning}: Questions that did not meet task-specific criteria were removed. For example, for Scientific Deep Research, steps must be short sentences forming a list, each representing one step; for Wet Experiments, each action must exist in the predefined action pool.  
2. \textit{Model-based cleaning}: Large language models were used to detect and remove questions with semantic errors or potential logical inconsistencies.  
3. \textit{Expert quality check}: All questions were reviewed by the original data-providing scientists, removing incomplete questions, questions with non-unique answers, or questions whose research direction did not align with the source data. For Dry Experiments, Python environments were used to test all code snippets to ensure executability.

\paragraph{Difficulty Filtering}  
After data cleaning, we filtered questions based on difficulty using mainstream LLMs. We evaluated each question with six high-performance models (e.g., GPT-5~\cite{GPT5_SystemCard2025}, Gemini-2.5-Pro~\cite{comanici2025gemini}, DeepSeek-R1~\cite{Guo2025}, Kimi-k2~\cite{team2025kimi}) under a setup allowing web search and deep-reasoning modes. Questions that more than half of the models could correctly answer were removed. This process ensures that the benchmark remains highly challenging.

Through these four steps, we guarantee that all benchmark questions are derived from authentic scientific data, aligned with domain scientists' judgment of scientific value, and maintain both high quality and high challenge.

\subsection{Data Distribution}

\begin{figure}[ht]
\centerline
{\includegraphics[width=15cm]{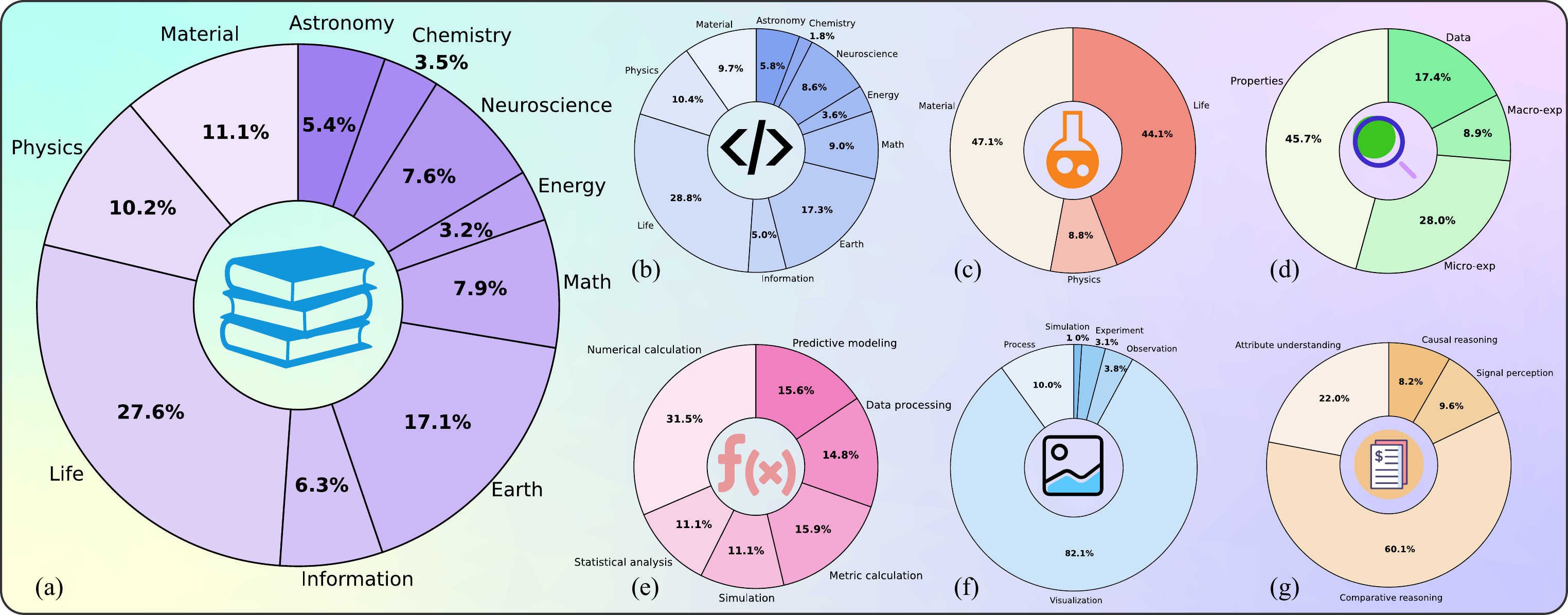}}
\caption{\textbf{Benchmark Data Distribution}: (a) Overall discipline distribution; (b) Dry experiment discipline distribution; (c) Wet experiment discipline distribution; (d) Scientific Deep Research question types; (e) Dry Experiment function types; (f) Experimental Reasoning image modalities; (g) Experimental Reasoning reasoning paradigms.}
\label{fig: data_distribution}
\end{figure}

After the data construction process, we obtained the complete SGI-Bench benchmark, which contains 318 Scientific Deep Research questions, 315 Idea Generation questions, 271 Dry Experiment questions, 68 Wet Experiment questions, and 291 Experimental Reasoning questions. The discipline distributions for Scientific Deep Research, Idea Generation, and Experimental Reasoning are identical, as shown in Figure~\ref{fig: data_distribution} (a). The discipline distributions for Dry and Wet Experiments are presented in Figure~\ref{fig: data_distribution} (b) and Figure~\ref{fig: data_distribution} (c), respectively, with Wet Experiments covering only a subset of disciplines, such as Biology and Chemistry.

In addition to discipline-level distributions, we further categorized the tasks at a finer granularity. For Scientific Deep Research, questions are grouped based on the type of target being investigated into four categories: Data, Properties, Micro-Experiments, and Macro-Experiments, as detailed in Table~\ref{tab:deep_research_types}. The distribution of these types is illustrated in Figure~\ref{fig: data_distribution} (d). For Dry Experiments, questions are classified into six types according to the masked function type, as shown in Table~\ref{tab:dry_experiment_functions}, with the corresponding distribution displayed in Figure~\ref{fig: data_distribution} (e). In Experimental Reasoning, the task inputs include images spanning multiple modalities, including Process Images, Observation Images, Experiment Images, Simulation Images, and Visualization Images, summarized in Table~\ref{tab:modalities} and visualized in Figure~\ref{fig: data_distribution} (f). Moreover, based on the type of reasoning required, questions are further categorized into Signal Perception, Attribute Understanding, Comparative Reasoning, and Causal Reasoning, as detailed in Table~\ref{tab:reasoning_paradigms}, with distributions shown in Figure~\ref{fig: data_distribution} (g).

These fine-grained categorizations by discipline and task type facilitate a detailed analysis of the limitations of evaluated LLMs and agents across scientific domains and research tasks. Such insights provide clear directions for advancing AI-assisted scientific discovery.

%% file: sections/3-framework.tex
\section{SGIEvalAgent: Agentic Evaluation Framework}

\begin{figure}[t]
\centerline
{\includegraphics[width=16cm]{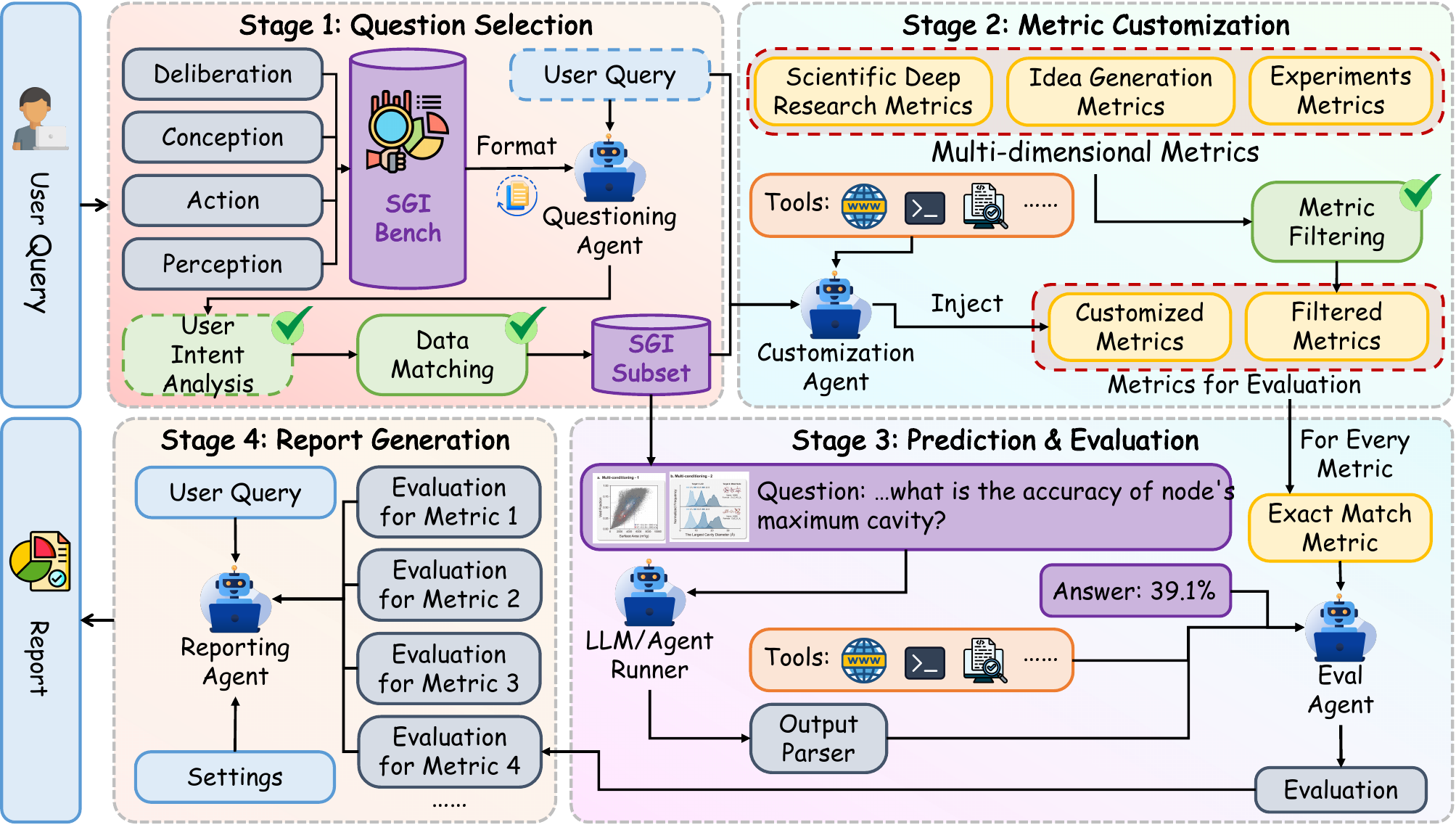}}
\caption{\textbf{Evaluation Framework.}}
\label{fig:evaluation-framework}
\end{figure}

Given the inherent complexity of scientific discovery, evaluating the performance of LLMs and agents in this domain presents formidable challenges. 
Rather than merely employing LLMs as evaluators, we develope a comprehensive, agent-based evaluation framework augmented with diverse capabilities (\textit{e.g.}, web search, Python interpreter, file reader, PDF parser, metric-specific Python functions~\cite{SmolAgents_2025}) to ensure rigorous, accurate, and scalable evaluations. 
As illustrated in Figure~\ref{fig:evaluation-framework}, this framework is structured into four interconnected stages: Question Selection, Metric Customization, Predict \& Eval, and Report Generation, each orchestrated by specialized agents to address distinct facets of the evaluation workflow.

\subsection{Question Selection}

The Question Selection stage is managed by a dedicated \emph{questioning agent}, which interprets user queries to retrieve relevant questions from the SGI-Bench question bank. 
The agent filters questions according to multiple criteria, including disciplinary domain, task category, and evaluation intent specified in the input query. 
In scenarios where no user query is provided, the agent defaults to systematically selecting all questions from the SGI-Bench, thereby ensuring comprehensive coverage across all scientific tasks. 
This stage effectively defines the evaluation scope by specifying the precise set of problems that subsequent stages will assess.

\begin{tcolorbox}[
    breakable,
    title=Question Agent Definition,
    colback=LighterGray,
    colframe=DeepPurple,
    colbacktitle=DeepPurple,
    coltitle=White
]
\textbf{\emph{\textcolor{DeepPurple}{Agent Input}}}
\begin{itemize}
    \item \texttt{\textbf{User Query (Q)}}: Any content input by users for obtaining relevant information, which can be in various forms such as text, keywords, or questions.
    \item \texttt{\textbf{SGI-Bench Data (D)}}: All constructed datasets in SGI-Bench, each of which is associated with a specific discipline and corresponding research area.
    \item \texttt{\textbf{K-value (K)}}: A positive integer indicating the number of most relevant items to select from the SGI-Bench Data based on the User Query.
\end{itemize}

\textbf{\emph{\textcolor{DeepPurple}{Agent Output}}}
\begin{itemize}
    \item \texttt{\textbf{Selected Indices (SI)}}: The selected indices for locating and retrieving the target data.
\end{itemize}

\end{tcolorbox}

\subsection{Metric Customization}
In the metric customization stage, a metric customization agent first dynamically generates novel evaluation metrics based on user queries and selected questions.
The agent parses the evaluation intent from user input to formalize customized metric instructions with advanced tools like web search and PDF parser, enabling flexible prioritization of metrics or integration of novel evaluation dimensions.
Then, the customized metrics will be aggregated with predefined scientist-aligned metrics given different question types, as described in Section~\ref{sec:Metrics}, to form the final metrics for evaluation.
By synergizing pre-defined and user-customized metrics, this stage ensures the framework aligns with both standardized benchmarks and domain-specific demands.

\begin{tcolorbox}[
    breakable,
    title=Customization Agent Definition,
    colback=LighterGray,
    colframe=DeepPurple,
    colbacktitle=DeepPurple,
    coltitle=White
]
\textbf{\emph{\textcolor{DeepPurple}{Agent Input}}}
\begin{itemize}
    \item \texttt{\textbf{User Query (UQ)}}: Any content input by users for obtaining relevant information, which can be in various forms such as text, keywords, or questions.
    \item \texttt{\textbf{SGI-Bench Data (D)}}: All constructed datasets in SGI-Bench, each of which is associated with a specific discipline and corresponding research area.
    \item \texttt{\textbf{Selected Indices (SI)}}: The selected indices for locating and retrieving the target data.
    \item \texttt{\textbf{Tool Pool(T)}}: A set of pre-configured tools for agents to call, including web search, PDF parser, Python Interpreter, etc.
    \item \texttt{\textbf{Metric Pool(M)}}: A set of pre-defined task-specific metrics presented in Section~\ref{sec:Metrics}. 
\end{itemize}

\textbf{\emph{\textcolor{DeepPurple}{Agent Output}}}
\begin{itemize}
    \item \texttt{\textbf{Metrics for Evaluation (ME)}}: Generated novel metrics based on the user query.
\end{itemize}

\end{tcolorbox}

\subsection{Inference and Evaluation}
The predict \& eval stage leverages a tool pool that includes utilities like web search, PDF parser, and Python interpreter to first execute inference for target LLMs or agents on the questions selected in the first stage. 
Subsequently, a dedicated Science Eval Agent (SGI-Bench Agent) applies the metrics finalized in the second stage to score the inference results. 
For each score, the agent generates a rationale grounded in reference answers, question context, and supplementary information retrieved via tools if necessary, thereby ensuring transparency and reproducibility. 
By integrating tool-augmented inference with systematic, metric-driven scoring, this stage effectively addresses the multi-dimensional and complex nature of scientific reasoning assessment.

\begin{tcolorbox}[
    breakable,
    title=Evaluation Agent Definition,
    colback=LighterGray,
    colframe=DeepPurple,
    colbacktitle=DeepPurple,
    coltitle=White
]
\textbf{\emph{\textcolor{DeepPurple}{Agent Input}}}
\begin{itemize}
    \item \texttt{\textbf{SGI-Bench Data (D)}}: All constructed datasets in SGI-Bench, each of which is associated with a specific discipline and corresponding research area.
    \item \texttt{\textbf{Selected Indices (SI)}}: The selected indices for locating and retrieving the target data.
    \item \texttt{\textbf{Responses (R)}}: Generated responses by the evaluation target in the Testbed.
    \item \texttt{\textbf{Tool Pool(T)}}: A set of pre-configured tools for agents to call, including web search, PDF parser, Python Interpreter, etc.
    \item \texttt{\textbf{Metrics for Evaluation (ME)}}: Generated novel metrics based on the user query.
\end{itemize}

\textbf{\emph{\textcolor{DeepPurple}{Agent Output}}}
\begin{itemize}
    \item \texttt{\textbf{Score (S)}}: A single integer score from 0–10, where 10 means the response is fully correct compared to the answer. Higher scores indicate the Prediction is better, and lower scores indicate it is worse.  
    \item \texttt{\textbf{Rationale (RN)}}: A brief explanation of why the response is correct or incorrect with respect to accuracy, completeness, clarity, and supporting evidence.
\end{itemize}

\end{tcolorbox}

\subsection{Report Generation}

The report generation stage is orchestrated by a dedicated reporting agent, which aggregates the user evaluation intents, finalized metric specifications, and the results produced during the Predict \& Eval stage. 
The agent then compiles a comprehensive report that both visualizes and quantifies the performance of different LLMs and agents across the selected questions and metrics. 
Beyond summarizing raw results, the report contextualizes the findings within the broader landscape of scientific discovery capabilities, thereby enabling users to extract actionable insights and make informed decisions efficiently.

\begin{tcolorbox}[
    breakable,
    title=Reporting Agent Definition,
    colback=LighterGray,
    colframe=DeepPurple,
    colbacktitle=DeepPurple,
    coltitle=White
]
\textbf{\emph{\textcolor{DeepPurple}{Agent Input}}}
\begin{itemize}
    \item \texttt{\textbf{Score List(SL)}}: A list of integers score from 0–10, where 10 means the response is fully correct compared to the answer. Higher scores indicate the Prediction is better, and lower scores indicate it is worse.  
    \item \texttt{\textbf{Rationale List(RNL)}}: A list of explanations of why the response is correct or incorrect with respect to accuracy, completeness, clarity, and supporting evidence.
    \item \texttt{\textbf{User-customized Metric (UM)}}: Generated novel metrics based on the user query. 
\end{itemize}

\textbf{\emph{\textcolor{DeepPurple}{Agent Output}}}
\begin{itemize}
    \item \texttt{\textbf{Report (R)}}: A comprehensive final evaluation report that demonstrates the scientific discovery capabilities of different LLMs and agents.
\end{itemize}

\end{tcolorbox}

%% file: sections/4-evaluation.tex
\section{Evaluation Results}

\subsection{Evaluation Setup}
To comprehensively evaluate different models throughout the scientific discovery workflow, we performed quantitative assessments across diverse LLMs and agents using scientist-aligned metrics.
\begin{itemize}
    \item For open-weight LLMs, we evaluated DeepSeek-V3.2~\cite{DeepSeekV3.2_2025}, DeepSeek-R1~\cite{Guo2025}, Intern-S1 and Intern-S1-mini~\cite{bai2025intern}, Kimi-k2~\cite{team2025kimi}, Qwen3-VL-235B-A22B~\cite{bai2025qwen3}, Qwen3-235B-A22B, Qwen3-Max, and Qwen3-8B~\cite{yang2025qwen3}, and Llama-4-Scout~\cite{Llama4_Release2025}.
    \item For closed-weight LLMs, we assessed GPT-4o~\cite{GPT4o_SystemCard2024}, GPT-4.1~\cite{GPT4.1_Docs2025}, GPT-5~\cite{GPT5_SystemCard2025}, GPT-5.1~\cite{GPT5.1_Addendum2025}, GPT-5.2-Pro~\cite{GPT5.2}, o3 and o4-mini~\cite{OpenAI_o3_2025}, Gemini-2.5-Flash and Gemini-2.5-Pro~\cite{comanici2025gemini}, Gemini-3-Pro~\cite{Gemini3_DeepMind2025}, Claude-Opus-4.1~\cite{ClaudeOpus4.1_2025}, Claude-Sonnet-4.5~\cite{ClaudeSonnet4.5_2025}, Grok-3~\cite{Grok3_2025}, and Grok-4~\cite{Grok4_ModelCard2025}.
    \item For open-source agents, we tested SmolAgents(GPT-4.1) and SmolAgents(Gemini-2.5-Flash)~\cite{SmolAgents_2025}, Owl(GPT-4.1) and Owl(Gemini-2.5-Flash)~\cite{hu2025owl}, WebThinker~\cite{WebThinker_NeurIPS2025}, XMaster~\cite{XMaster_Arxiv2025}, and InternAgent~\cite{team2025novelseek}.
    \item For closed-source agents, we evaluated OpenAI DeepResearch(o3) and OpenAI DeepResearch(o4-mini)~\cite{DeepResearch_OpenAI2025}, Kimi-Search(Kimi-k2)~\cite{KimiResearcher_2025}, Doubao-Search(Seed-1-6), Grok-Search(Grok-4)~\cite{GrokDeepSearch_2025}, and Perplexity(Sonar-Pro)~\cite{Perplexity_DeepResearch_2025}.
\end{itemize}
For benchmarking consistency, we set the temperature of all configurable models to 0 to minimize randomness and used a standard zero-shot, task-specific prompt template across all tasks.

\subsection{Overview}

Table~\ref{tab:llm_comparison} provides a cross‑task snapshot of current capabilities. Overall, SGI‑Score remains low across families (typically ~30±5), with the best aggregate result at 33.83 (Gemini‑3‑Pro). Closed‑source models show only a marginal edge over leading open‑source systems (e.g., Claude‑Sonnet‑4.5 at 32.16 vs. Qwen3‑Max at 31.97), indicating that scale and access alone do not translate into robust scientific cognition. At the task level, Deep Research is the most brittle under the strict Exact‑Match metric (best 18.48; many models around 8–16), revealing the difficulty of end‑to‑end, multi‑source evidence integration and numerically faithful inference. Idea Generation exhibits the opposite pattern—strong surface performance but weak realizability: while GPT‑5 attains the highest average (55.40), feasibility remains uniformly low across models, reflecting underspecified implementation details and missing resource/parameter assumptions. In Dry Experiments, high executability does not imply correctness: even the best PassAll@5 peaks at 36.64 (Gemini‑3‑Pro), underscoring persistent gaps in numerical stability and scientific algorithm selection. Wet Experiments remain challenging, with uniformly low action‑sequence similarity and only moderate parameter accuracy, driven by errors in step ordering, temporal coordination, and branch/sample bookkeeping. Multimodal Experimental Reasoning shows relatively stronger results (best MCA 41.92), yet remains far from reliable scientific discrimination. Taken together, these patterns validate our SGI framing: contemporary models possess fragments of the Deliberation–Conception–Action–Perception cycle but fail to integrate them into a coherent, workflow‑faithful intelligence—pointing to the need for meta‑analytic retrieval with numerical rigor, planning‑aware conception, and procedure‑level consistency constraints.

\begin{table}[t]
\centering
\renewcommand{\arraystretch}{0.9}
\setlength{\tabcolsep}{2pt}
\tiny
\resizebox{15cm}{!}{
\begin{tabular}{lccccc c}
\toprule
\rowcolor{blue!15}\textbf{Model} & \textbf{DeepResearch} & \textbf{IdeaGen} & \textbf{DryExp} & \textbf{WetExp} & \textbf{ExpReasoning} & \textbf{SGI-Score} \\
\midrule

\rowcolor{blue!5}\multicolumn{7}{c}{\emph{Open-source LLM}} \\ \arrayrulecolor{black!30}\midrule
DeepSeek-V3.2 & 12.70 & 37.45 & 23.62 & 20.95 & - & \cellcolor{red!10}- \\
DeepSeek-R1 & 15.03 & 40.16 & 33.33 & 21.12 & - & \cellcolor{red!10}- \\
Intern-S1 & 15.74 & 38.09 & 28.79 & 29.02 & 28.87 & \cellcolor{red!10}28.10 \\
Intern-S1-mini & 11.06 & 36.04 & 16.97 & 12.42 & 16.84 & \cellcolor{red!10}18.67 \\
Kimi-k2 & 13.11 & 43.17 & 29.52 & 25.76 & - & \cellcolor{red!10}- \\
Qwen3-VL-235B-A22B & 11.97 & 39.28 & 28.41 & 30.30 & 31.62 & \cellcolor{red!10}28.32 \\
Qwen3-235B-A22B & 14.19 & 39.45 & 28.89 & 26.40 & - & \cellcolor{red!10}- \\
Qwen3-Max & 15.38 & 39.83 & 33.21 & 33.62 & 37.80$^{*}$ & \cellcolor{red!10}31.97$^{*}$\copper \\
Qwen3-8B & 8.18 & 35.78 & 18.45 & 9.96 & 23.37$^{*}$ & \cellcolor{red!10}19.15$^{*}$ \\
Llama-4-Scout & 7.86 & 29.72 & 20.37 & 21.66 & 25.77 & \cellcolor{red!10}21.08 \\
\arrayrulecolor{black!100}\midrule

\rowcolor{blue!5}\multicolumn{7}{c}{\emph{Closed-source LLM}} \\ \arrayrulecolor{black!30}\midrule
GPT-4o & 7.86 & 35.95 & 26.94 & 31.31 & 32.30 & \cellcolor{red!10}26.87 \\
GPT-4.1 & 11.32 & 36.49 & 34.32 & 36.63 & 38.49 & \cellcolor{red!10}31.45 \\
GPT-5 & 14.47 & \cellcolor{blue!10}\textbf{55.40} & 29.89 & 16.31 & 38.14 & \cellcolor{red!10}30.84 \\
GPT-5.1 & 11.64 & 47.12 & 31.00 & 22.77 & 34.02 & \cellcolor{red!10}29.31 \\
GPT-5.2-Pro & 15.72 & 55.03 & 28.04 & 17.50 & 39.18 & \cellcolor{red!10}31.09 \\
o3 & 12.89 & 46.07 & 31.73 & 30.04 & 32.65 & \cellcolor{red!10}30.68 \\
o4-mini & 11.95 & 40.78 & 35.79 & 28.86 & 33.33 & \cellcolor{red!10}30.14 \\
Gemini-2.5-Flash & 10.69 & 39.13 & 21.03 & 18.55 & 34.36 & \cellcolor{red!10}24.75 \\
Gemini-2.5-Pro & 15.09 & 39.95 & 22.51 & 22.05 & 41.24 & \cellcolor{red!10}28.17 \\
Gemini-3-Pro & \cellcolor{blue!10}\textbf{18.48} & 39.68 & \cellcolor{blue!10}\textbf{36.64} & 32.45 & \cellcolor{blue!10}\textbf{41.92} & \cellcolor{red!10}\textbf{33.83}\gold  \\
Claude-Opus-4.1 & 12.93 & 40.29 & 34.69 & 25.38 & 38.83 & \cellcolor{red!10}30.42 \\
Claude-Sonnet-4.5 & 13.84 & 43.20 & 35.79 & 30.15 & 37.80 & \cellcolor{red!10}32.16\silver \\
Grok-3 & 13.52 & 35.98 & 27.31 & \cellcolor{blue!10}\textbf{37.92} & - & \cellcolor{red!10}- \\
Grok-4 & 13.31 & 37.12 & 33.71 & 29.01 & 30.24 & \cellcolor{red!10}28.68 \\
\arrayrulecolor{black!100}\bottomrule
\end{tabular}
}
\caption{\textbf{Overview Results Across SGI-Bench Tasks}: Aggregated performance across Deep Research, Idea Generation, Dry/Wet Experiment, and Experimental Reasoning. The scores for Deep Research are based on the exact match metric (the strictest metric). Idea Generation scores are the average of four metrics evaluating ideas. Dry Experiment scores are based on PassAll@5 (the strictest metric). Wet Experiment scores are the average of action sequence similarity and parameter accuracy. Experimental Reasoning scores are based on the multi-choice accuracy metric (the strictest metric). The SGI-Score is the average across these tasks, reflecting the overall capability of an AI model in various scientific research scenarios. An asterisk $^{*}$ indicates results from different versions of the same series of multimodal models.}
\label{tab:llm_comparison}
\end{table}

\subsection{Scientific Deep Research}

\begin{figure}[ht]
\centerline
{\includegraphics[width=16cm]{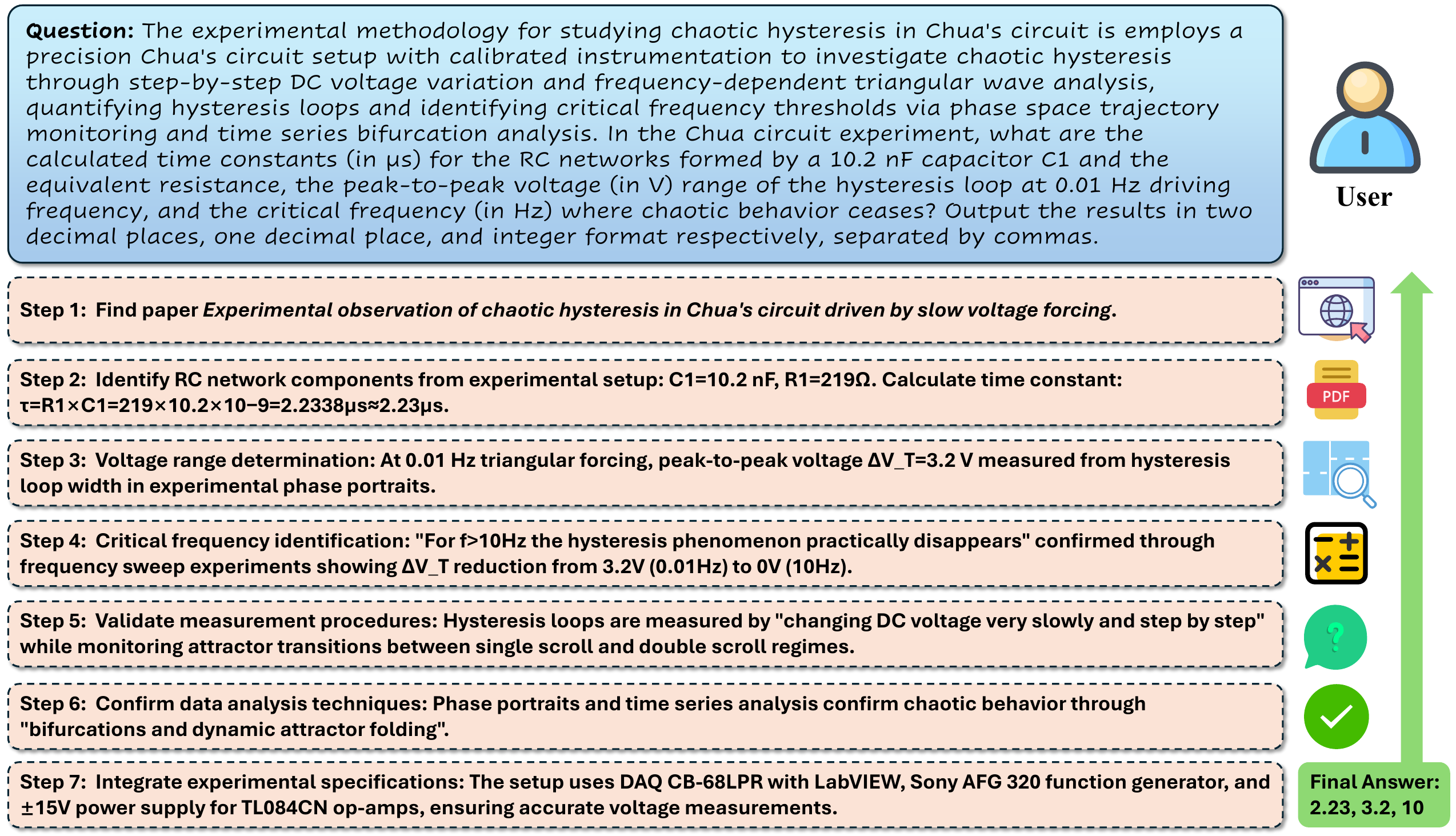}}
\caption{\textbf{Scientific Deep Research Case}: Example multi-hop workflow illustrating data retrieval, evidence synthesis, and quantitative analysis.}
\label{fig: search_case1}
\end{figure}

\begin{figure}[ht]
\centerline
{\includegraphics[width=17cm]{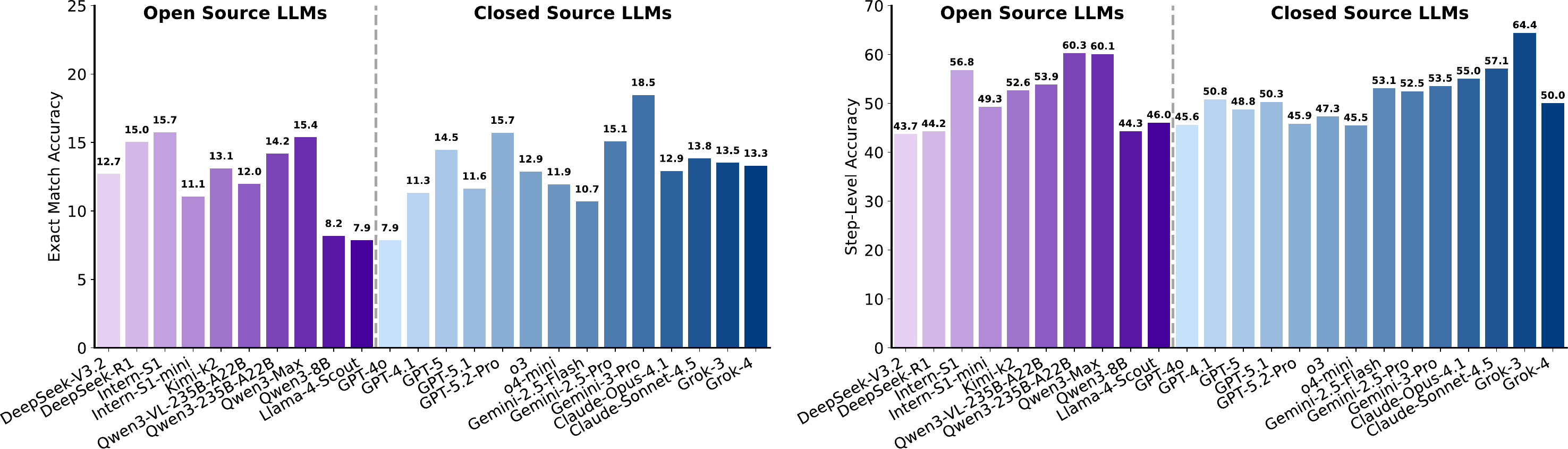}}
\caption{\textbf{Scientific Deep Research Evaluation of LLMs}: Exact Match (EM) and Step-Level Accuracy (SLA) across models using scientist-aligned metrics.}
\label{fig: llms deep research}
\end{figure}

\begin{figure}[ht]
\centerline
{\includegraphics[width=17cm]{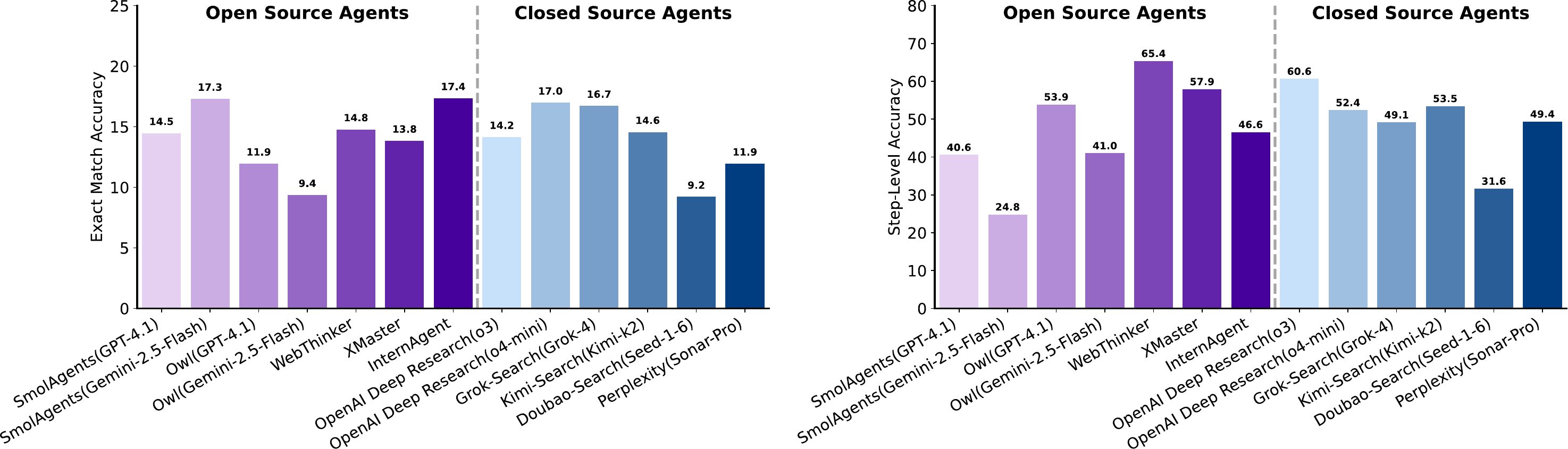}}
\caption{\textbf{Scientific Deep Research Evaluation of Multi-Agent Systems}: EM and SLA for tool-augmented agent systems.}
\label{fig: agents deep research}
\end{figure}

The results for LLMs and agents are presented in Figs.~\ref{fig: llms deep research} and~\ref{fig: agents deep research}. Exact Match (EM) evaluates the correctness of the final answer, while Step-Level Accuracy (SLA) measures alignment with the reference reasoning trajectory. EM remains low across all evaluated systems, typically around 10\% and seldom above 20\%, indicating that current models capture only a narrow fraction of the analytical depth required for scientific deep research. While top-performing tool-augmented agents slightly outperform the best offline LLMs on SLA, the overall distributions overlap substantially; several agent systems underperform many LLMs, and EM differences are marginal with the best LLMs matching or exceeding the best agents.

\textbf{SLA substantially exceeds EM across nearly all systems.}
Multiple systems, including several agents—achieve SLA above 50\%, with the best around 65\%. This disparity suggests that models frequently produce partially correct or locally consistent reasoning steps but struggle to maintain coherence and correctness across the full reasoning chain. Such behavior underscores the intrinsic difficulty of end-to-end scientific reasoning and the importance of step-wise decomposition for improving task success.

\textbf{Newer large-scale LLMs do not universally outperform predecessor models. }
For example, Grok-4 exhibits lower EM and SLA than Grok-3 on this benchmark, suggesting that large-scale training may introduce regressions or reduce retention of specialized scientific knowledge. These results collectively highlight the current limitations of frontier AI systems in executing the multi-faceted and rigorously structured reasoning processes required for Scientific Deep Research.

\begin{figure}[ht]
\centerline
{\includegraphics[width=17cm]{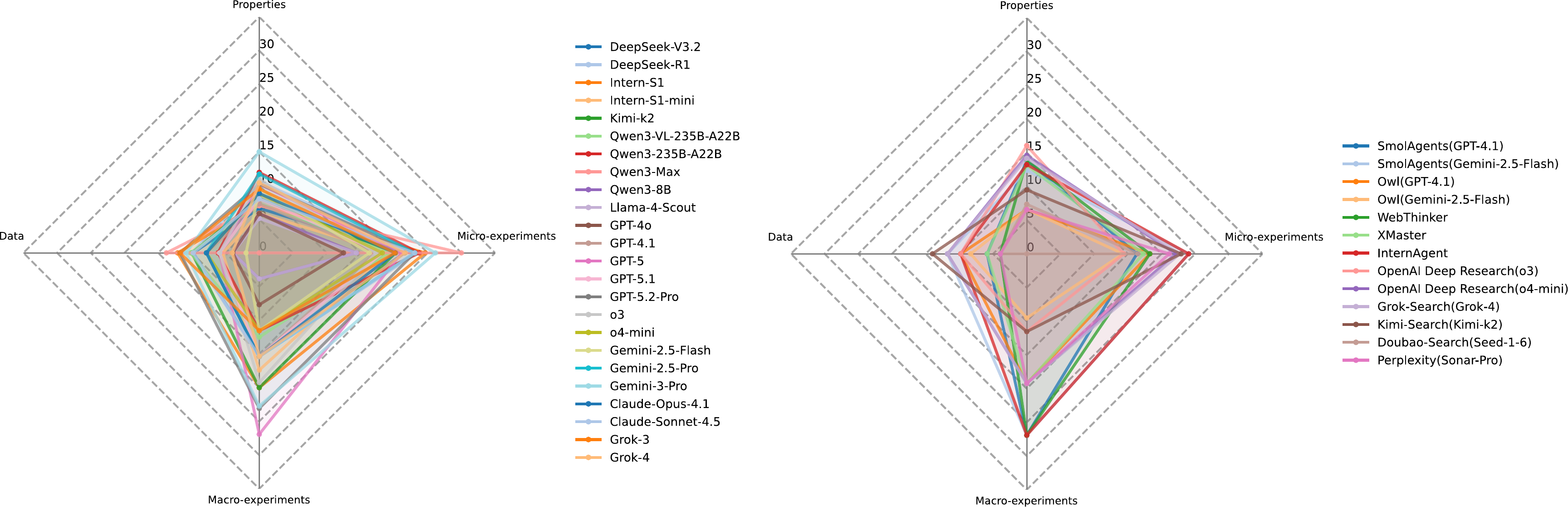}}
\caption{\textbf{Scientific Deep Research Performance by Type}: Comparison across Data, Properties, Micro-Experiments, and Macro-Experiments categories.}
\label{fig: deep research on different task}
\end{figure}

\textbf{Most models exhibit substantially lower performance on the Data and Properties tasks, but somewhat better—though still modestly—on Micro- and Macro-experiment tasks.}
Based on the focus of each question, we categorize the tasks into four types: Data, Properties, Micro-experiments, and Macro-experiments (Table~\ref{tab:deep_research_types}). Figure~\ref{fig: deep research on different task} summarizes the performance of LLMs and agents across these categories. Notably, performance across all four categories rarely exceeds 30\% (with only a few Macro cases slightly above), underscoring the intrinsic difficulty of scientific deep research. This disparity can be attributed to the nature of the information required. Data- and property-related questions often rely on detailed numerical specifications or contextual descriptions scattered across disparate sources in the literature, demanding precise retrieval, cross-referencing, and aggregation. In contrast, Micro- and Macro-experiment tasks tend to provide more structured protocols or clearer experimental outcomes, enabling LLMs and agents to reason with fewer retrieval uncertainties.

In summary, the relatively stronger model performance on experiment-oriented tasks suggests that recent advances in LLM pretraining and instruction tuning have enhanced models' abilities to process structured procedures and numerical patterns. Nevertheless, the consistently low scores across all categories indicate that contemporary LLMs, even when augmented with tool-based agents, remain far from mastering the breadth and depth of reasoning required for robust scientific deep research.

\subsection{Idea Generation}

\begin{figure}[ht]
\centerline
{\includegraphics[width=17cm]{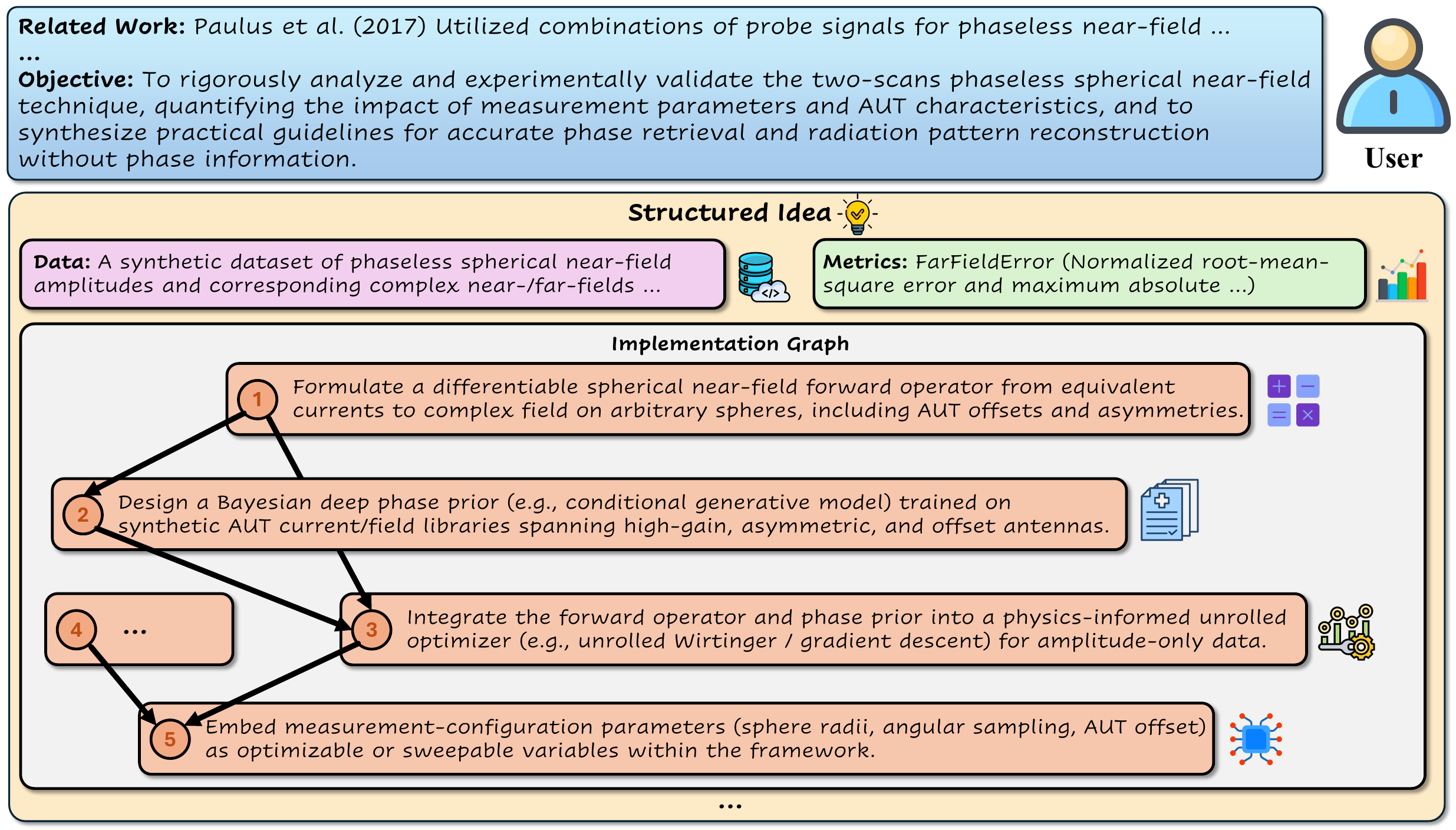}}
\caption{\textbf{Idea Generation Case}: Input information such as related work and objective, and output a structured idea, including a graph consisting of specific implementation steps.}
\label{fig: idea_case}
\end{figure}

\begin{table}[t]
\centering
\renewcommand{\arraystretch}{0.9}
\setlength{\tabcolsep}{2pt}
\tiny
\resizebox{13cm}{!}{
\begin{tabular}{lccccc}
\toprule
\textbf{Model} & \textbf{Effectiveness} & \textbf{Novelty} & \textbf{Detailedness} & \textbf{Feasibility}  & \textbf{Average} \\
\midrule

\rowcolor{blue!5}\multicolumn{6}{c}{\emph{Open-source LLM}} \\ \arrayrulecolor{black!30}\midrule
DeepSeek-V3.2 & 28.09 & 54.09 & 47.34 & 20.28 & 37.45 \\
DeepSeek-R1 & 27.73 & 63.64 & 50.06 & 19.20 & 40.16 \\
Intern-S1 & 26.38 & 56.47 & 49.10 & 20.42 & 38.09 \\
Intern-S1-mini & 24.95 & 55.71 & 48.07 & 15.44 & 36.04 \\
Kimi-k2 & 25.24 & 69.49 & 59.20 & 18.74 & 43.17 \\
Qwen3-VL-235B-A22B & 27.24 & 59.53 & 50.23 & 20.14 & 39.28 \\
Qwen3-235B-A22B & 26.63 & 62.05 & 49.73 & 19.40 & 39.45 \\
Qwen3-Max & 28.74 & 59.01 & 50.61 & 20.98 & 39.83 \\
Qwen3-8B & 26.12 & 49.36 & 47.09 & 20.58 & 35.78 \\
Llama-4-Scout & 28.50 & 33.25 & 43.08 & 14.06 & 29.72 \\
\arrayrulecolor{black!100}\midrule

\rowcolor{blue!5}\multicolumn{6}{c}{\emph{Closed-source LLM}} \\ \arrayrulecolor{black!30}\midrule
GPT-4o & 27.28 & 48.19 & 47.85 & 20.51 & 35.95 \\
GPT-4.1 & 27.49 & 48.72 & 47.88 & 21.87 & 36.49 \\
GPT-5 & 40.92 & \cellcolor{blue!10}\textbf{76.08} & \cellcolor{blue!10}\textbf{85.72} & 18.87 & \cellcolor{blue!10}\textbf{55.40} \\
GPT-5.1 & 36.07 & 66.98 & 66.62 & 18.83 & 47.12 \\
GPT-5.2-Pro & \cellcolor{blue!10}\textbf{51.36} & 71.19 & 78.03 & 19.53 & 55.03 \\
o3 & 29.42 & 73.74 & 58.22 & \cellcolor{blue!10}\textbf{22.90} & 46.07 \\
o4-mini & 27.26 & 63.33 & 50.53 & 22.01 & 40.78 \\
Gemini-2.5-Flash & 28.45 & 56.91 & 50.49 & 20.69 & 39.13 \\
Gemini-2.5-Pro & 30.98 & 57.54 & 52.21 & 19.06 & 39.95 \\
Gemini-3-Pro & 28.38 & 59.41 & 51.07 & 19.87 & 39.68 \\
Claude-Opus-4.1 & 26.52 & 64.40 & 50.16 & 20.07 & 40.29 \\
Claude-Sonnet-4.5 & 32.01 & 58.00 & 61.75 & 21.03 & 43.20 \\
Grok-3 & 28.37 & 46.27 & 48.35 & 20.93 & 35.98 \\
Grok-4 & 28.46 & 50.93 & 49.48 & 19.60 & 37.12 \\
\arrayrulecolor{black!100}\bottomrule
\end{tabular}
}
\caption{\textbf{Idea Generation Results}: The ideas generated by the model outperformed the average proportion of the original papers in the four dimensions of Effectiveness, Novelty, Detailedness, and Feasibility.}
\label{tab:idea_gen_res}
\end{table}

Figure~\ref{fig: idea_case} illustrates the evaluation pipeline for Idea Generation in SGI-Bench, and more experimental details can be found in the section~\ref{Metrics: Idea Gen}. Table~\ref{tab:idea_gen_res} shows the quantitative experimental results of idea generation, including effectiveness, novelty, detailedness, and feasibility. We could see that GPT-5 achieves the best average performance, and achieves the best performance in three aspects only excluding the feasibility. Moreover, across models, a clear pattern emerges: Novelty is generally high, especially among closed-source systems (e.g., o3 73.74, GPT-5 76.08). This indicates that modern LLMs possess a robust capacity for generating conceptually novel scientific ideas. Such behavior aligns with the growing empirical use of LLMs as inspiration engines for scientific hypothesis generation and exploratory research.

Mechanistically, this strength likely stems from their broad pretraining over heterogeneous scientific corpora, which enables them to recombine distant concepts across domains, as well as their ability to internalize high-level research patterns (problem--method--evaluation triples). 
As a result, LLMs are particularly effective at proposing \emph{plausible and novel conceptual directions}, often exceeding what a single human researcher can enumerate in a short time window.

\textbf{Novelty is relatively high while feasibility lags.}
In contrast, Effectiveness is modest for most models and Feasibility consistently lags behind the other dimensions. Even the best-performing GPT-5, which achieves high Detailedness (85.72) and the highest Average (55.40), attains only scores 18.87 in Feasibility, confirming that conceptual richness does not reliably translate into implementation-ready plans. The top Feasibility model by our metric is o3 (22.90), while open-source feasibility peaks at Qwen3-8B (20.58); other models cluster in the 14–20 range. Open-source models exhibit the same trend: Kimi-k2 reaches higher Detailedness (59.20) but remains limited in Feasibility (18.74); similarly, Qwen3-VL-235B-A22B reaches only 20.14 in Feasibility despite substantially higher conceptual elaboration (50.23).

\textbf{Execution details are often underspecified.}
These outcomes reveal a realization bottleneck in current idea generation: While models can articulate sophisticated pipelines at a high level, they frequently omit or under-specify key executable details. Typical failure issues include: (i) data references without acquisition or preprocessing plans; (ii) training and optimization loops that omit concrete hyperparameters or resource assumptions; (iii) algorithmic modules named but not grounded in precise choices (e.g., solver type, training objective, evaluation protocol); (iv) integration steps that fail to specify interfaces, ordering, or data flow. Consequently, many proposals fail feasibility checks not because they are conceptually unsound, but because they rely on implicit, unparameterized execution assumptions that cannot be validated under realistic experimental conditions. 
This gap highlights a fundamental limitation of current LLMs: they excel at linguistic and conceptual abstraction, yet struggle with the procedural, resource-aware, and constraint-grounded planning required for real scientific implementation. 

Overall, the Idea Generation results indicate that contemporary LLMs are adept at proposing novel directions but struggle to turn them into fully executable plans. Bridging this gap will require constraint-aware planning, stronger priors over experimental and engineering practice, tool-augmented verification (e.g., property simulators, data/API discovery, and reproducibility scaffolds), and training signals that reward concrete, parameterized, and testable implementation steps rather than stylistic innovation.

\subsection{Dry/Wet Experiment}

Experiments form the critical bridge between idea generation and scientific reasoning, providing the most direct avenue for validating hypotheses and uncovering new phenomena. Within SGI-Bench, we evaluate two complementary forms of experiments: \emph{dry experiments}, which involve computational analyses or simulations, and \emph{wet experiments}, which require laboratory procedures and operational planning. Across both categories, current AI models exhibit substantial limitations, revealing a persistent gap between linguistic fluency and experimentally actionable competence.

\subsubsection{Dry Experiment}

\begin{figure}[ht]
\centerline
{\includegraphics[width=16cm]{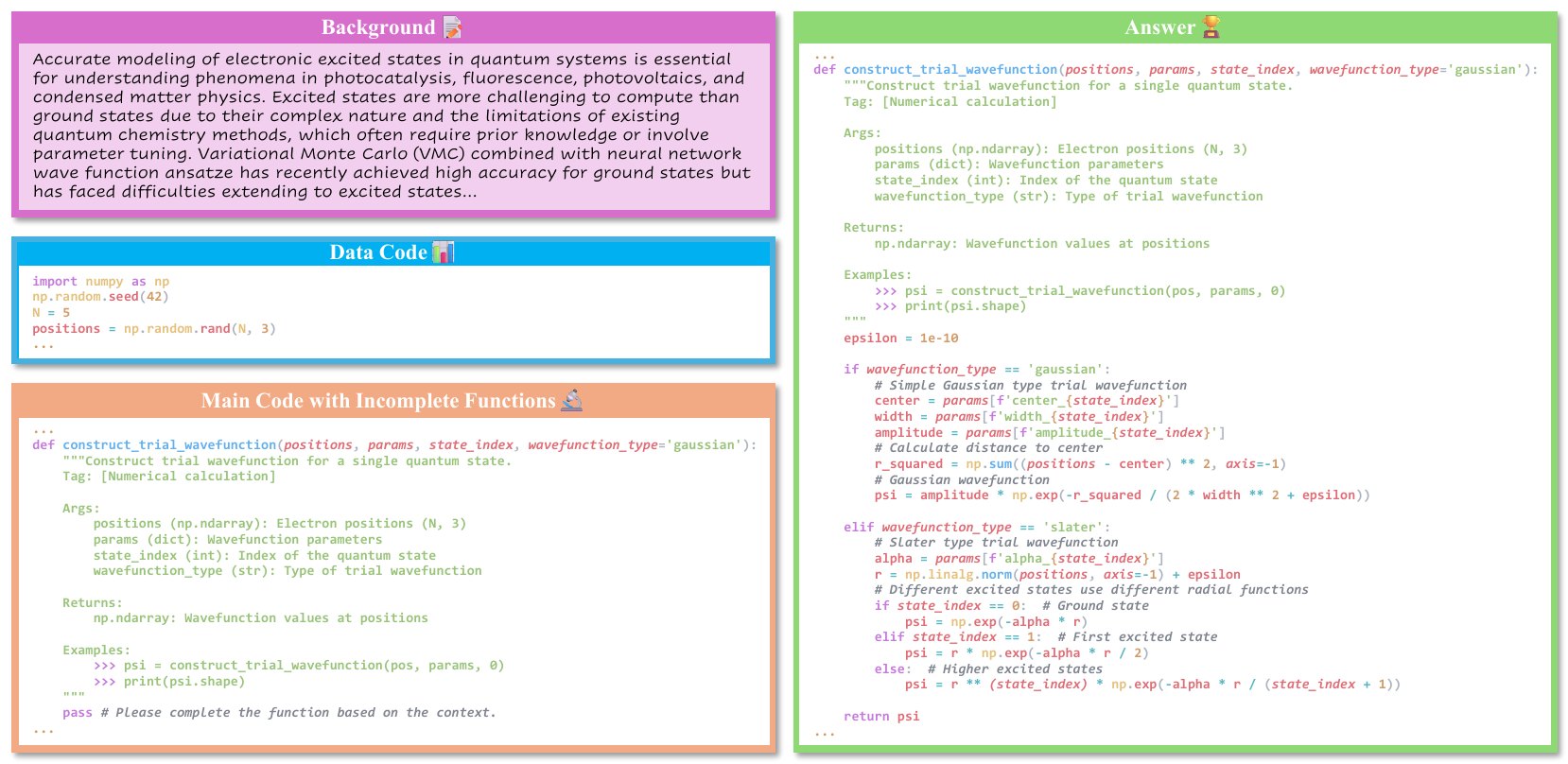}}
\caption{\textbf{Dry Experiment Code Examples}: Masked-function completion setup with I/O formats, and functional descriptions.}
\label{fig: code_case1}
\end{figure}

As introduced in Section~\ref{sec: Task Definition of Experiment}, each dry experiment contains three components: a description of scientific background, a complete data-construction script, and an analysis script with masked functions. The model must infer and complete these missing functions using contextual understanding. For fairness and structural clarity, function headers, including names, signatures, and functional descriptions, are preserved, as shown in Figure~\ref{fig: code_case1}. This setup isolates the model's ability to infer algorithmic logic rather than boilerplate structure.

\begin{table}[t]
\centering
\renewcommand{\arraystretch}{0.9}
\setlength{\tabcolsep}{2pt}
\tiny
\resizebox{14cm}{!}{
\begin{tabular}{lccccc}
\toprule
\textbf{Model} & \textbf{PassAll@5(\%)$\uparrow$} & \textbf{PassAll@3(\%)$\uparrow$} & \textbf{PassAll@1(\%)$\uparrow$} & \textbf{AET(s)$\downarrow$} & \textbf{SER(\%)$\uparrow$} \\
\midrule

\rowcolor{blue!5}\multicolumn{6}{c}{\emph{Open-source LLM}} \\ \arrayrulecolor{black!30}\midrule
DeepSeek-V3.2 & 23.62 & 26.94 & 29.52 & 29.96 & 68.27 \\
DeepSeek-R1 & 33.33 & 35.56 & 37.41 & 28.09 & 91.70 \\
Intern-S1 & 28.79 & 31.44 & 34.09 & 31.04 & 87.58 \\
Intern-S1-mini & 16.97 & 17.34 & 18.08 & 14.55 & 79.83 \\
Kimi-k2 & 29.52 & 32.10 & 36.16 & 33.42 & 90.26 \\
Qwen3-VL-235B-A22B & 28.41 & 31.37 & 33.58 & 32.74 & 91.22 \\
Qwen3-235B-A22B & 28.89 & 31.48 & 34.44 & 30.68 & 90.81 \\
Qwen3-Max & 33.21 & 35.42 & 37.27 & 35.25 & 90.33 \\
Qwen3-8B & 18.45 & 20.30 & 21.03 & 21.13 & 71.51 \\
Llama-4-Scout & 20.37 & 21.48 & 22.59 & 24.24 & 68.52 \\
\arrayrulecolor{black!100}\midrule

\rowcolor{blue!5}\multicolumn{6}{c}{\emph{Closed-source LLM}} \\ \arrayrulecolor{black!30}\midrule
GPT-4o & 26.94 & 29.89 & 32.10 & 37.90 & 79.78 \\
GPT-4.1 & 34.32 & 37.64 & 40.22 & 40.54 & 94.10 \\
GPT-5 & 29.89 & 32.84 & 34.69 & 34.54 & 75.50 \\
GPT-5.1 & 31.00 & 35.42 & 38.01 & 23.87 & 96.53 \\
GPT-5.2-Pro & 28.04 & 33.21 & 39.48 & 23.73 & 96.60 \\
o3 & 31.73 & 34.32 & 37.64 & 34.06 & 85.17 \\
o4-mini & 35.79 & 39.11 & 41.70 & 31.34 & 87.60 \\
Gemini-2.5-Flash & 21.03 & 22.51 & 24.72 & 15.09 & 44.65 \\
Gemini-2.5-Pro & 22.51 & 23.99 & 24.72 & \cellcolor{blue!10}\textbf{13.94} & 44.65 \\
Gemini-3-Pro & \cellcolor{blue!10}\textbf{36.64} & \cellcolor{blue!10}\textbf{40.46} & 41.98 & 21.16 & \cellcolor{blue!10}\textbf{98.85} \\
Claude-Opus-4.1 & 34.69 & 37.27 & 40.59 & 31.67 & 94.32 \\
Claude-Sonnet-4.5 & 35.79 & 38.75 & \cellcolor{blue!10}\textbf{42.07} & 31.59 & 94.83 \\
Grok-3 & 27.31 & 29.15 & 32.10 & 35.30 & 91.22 \\
Grok-4 & 33.71 & 37.12 & 40.53 & 33.74 & 94.09 \\

\arrayrulecolor{black!100}\bottomrule
\end{tabular}
}
\caption{\textbf{Dry Experiment Metrics Across Models}: PassAll@k, Average Execution Time (AET), and Smooth Execution Rate (SER) under five unit tests per problem.}
\label{tab:code metrics}
\end{table}

Table~\ref{tab:code metrics} summarizes three metrics defined in Section~\ref{sec: Metric of Dry Experiment}: \textit{PassAll@k}, \textit{Average Execution Time (AET)}, and \textit{Smooth Execution Rate (SER)}. Here, PassAll@k denotes passing at least $k$ out of five unit tests per problem. Under the lenient criterion ($k{=}1$), the best models achieve a \textit{PassAll@1} score of 42.07\%, whereas the strictest requirement ($k{=}5$) reduces performance to 36.64\%. These results underscore that scientific code completion remains a significant bottleneck, even for frontier LLMs. Notably, closed-source models generally achieve higher PassAll@k than leading open-source models, though the advantage is modest and distributions overlap, suggesting that scientific code synthesis in dry experiments remains underdeveloped across architectures.

\textbf{High execution rates do not guarantee correctness.}
The \textit{SER} metric captures whether the generated code executes without error, independent of correctness. While many top models achieve high SER values (>90\%), performance varies widely across systems; several models are substantially below this threshold (e.g., Gemini-2.5-Flash/Pro, Qwen3-8B, Llama-4-Scout, GPT-5, GPT-4o), indicating nontrivial robustness gaps. This suggests that basic structural and API-level reasoning has matured for some models; however, the persistent gap between \textit{SER} and accuracy metrics highlights that structural validity is far easier than algorithmic correctness in scientific contexts.

\begin{figure}[ht]
\centerline
{\includegraphics[width=16cm]{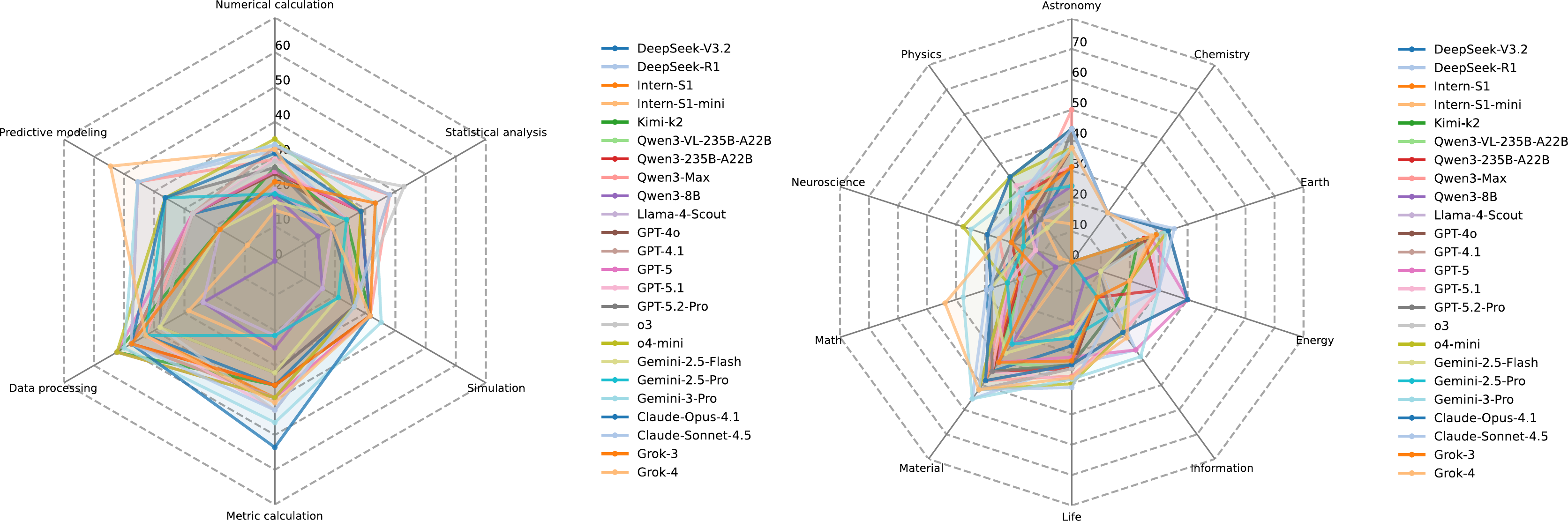}}
\caption{\textbf{PassAll@5 by Function Category}: Completion accuracy across numerical calculation, statistical analysis, simulation, metric calculation, data processing, and predictive modeling.}
\label{fig: dry_task_metric}
\end{figure}

\textbf{Numerical and simulation functions are the most challenging.}
Figure~\ref{fig: dry_task_metric} breaks down \textit{PassAll@5} across functional types. Models perform relatively well on \textit{Data Processing} and \textit{Predictive Modeling}, where multiple valid implementations exist and errors are less amplified. In contrast, \textit{Numerical Calculation} and simulation-oriented functions prove substantially more difficult. These tasks typically require precise numerical stability, accurate discretization, or careful handling of domain-specific constraints, all of which amplify small reasoning inconsistencies. This pattern reveals a striking asymmetry: models exhibit reasonable flexibility in tasks with diverse valid outputs but struggle with tasks requiring exact numerical fidelity.

\begin{figure}[ht]
\centerline
{\includegraphics[width=16cm]{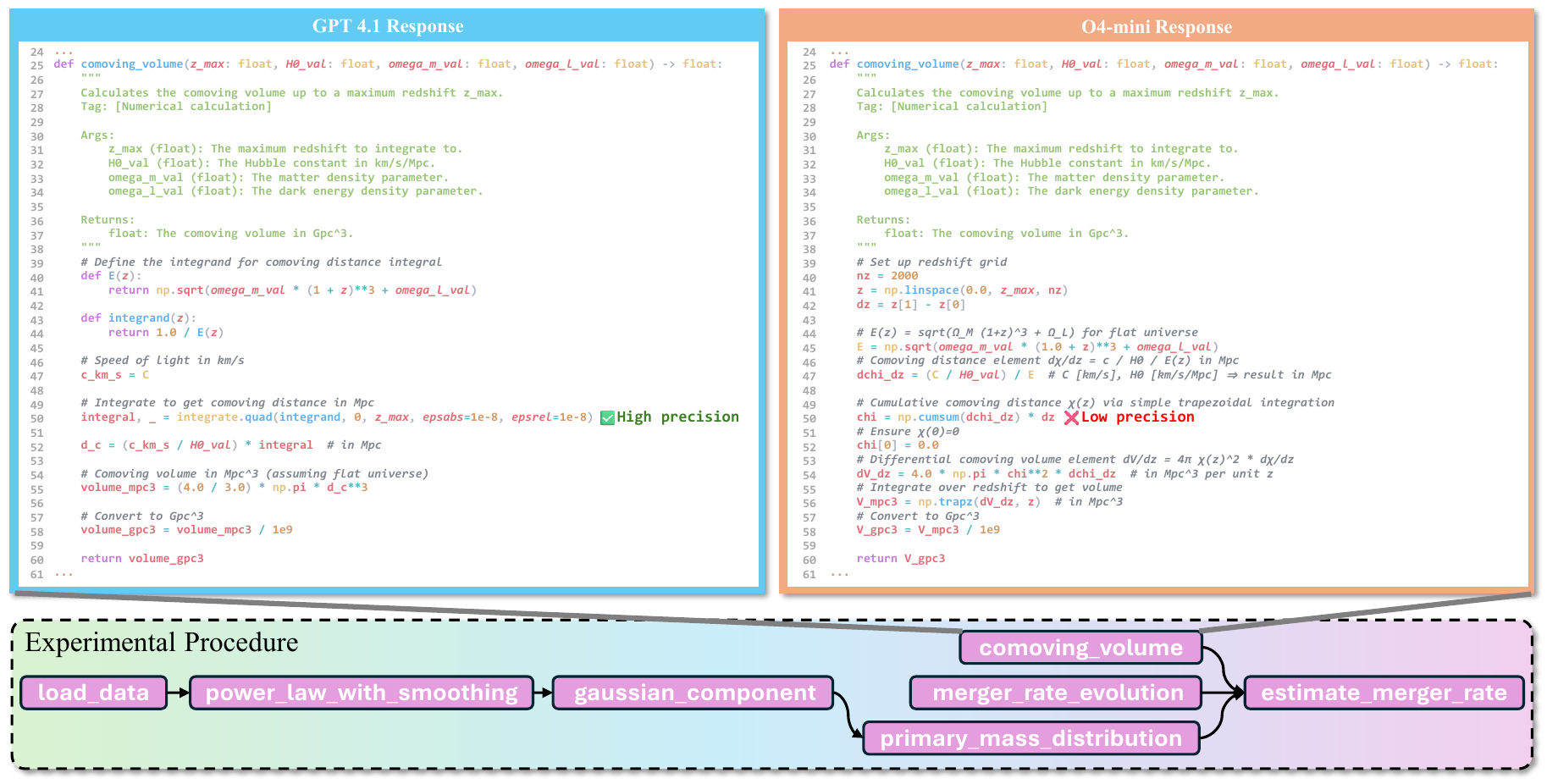}}
\caption{\textbf{Dry Experiment Case Study}: Gravitational-wave computation highlighting the impact of numerical integration strategy on scientific outcomes.}
\label{fig: code_case2}
\end{figure}

\textbf{Methodological choices critically affect outcomes.}
The case shown in Figure~\ref{fig: code_case2} illustrates this issue in an astronomical dry experiment involving the computation of gravitational-wave observables from LIGO/Virgo–like detectors. The o4-mini model employs a naïve numerical integration via \texttt{np.cumsum}, effectively using a forward Euler approximation for 
\[
\chi(z) = \int_0^z \frac{d\chi}{dz} dz,
\]
which introduces substantial cumulative error when the discretization is coarse. In contrast, GPT-4.1 correctly adopts \texttt{scipy.integrate.quad}, leveraging adaptive integration schemes that preserve numerical precision. Because errors in $\chi(z)$ propagate directly to the comoving volume element 
\[
\frac{dV}{dz} = 4\pi \chi(z)^2 \frac{d\chi}{dz},
\]
the flawed integration strategy in o4-mini leads to a significant deviation in the final volume estimate $V_{\mathrm{Gpc}^3}$. This example highlights a broader challenge: LLMs often fail to capture the numerical sensitivity and methodological nuance essential for scientific computation.

Overall, these findings reveal that while current models can generate syntactically valid code with high reliability, their deeper limitations stem from (i) incomplete numerical reasoning, (ii) superficial understanding of scientific algorithms, and (iii) the inability to select appropriate computational strategies under domain constraints. AI-assisted scientific experimentation thus remains a demanding frontier, requiring future models to incorporate domain-aware numerical reasoning, fine-grained algorithmic priors, and training signals beyond natural-language supervision.

\subsubsection{Wet Experiment}

\begin{figure}[ht]
\centerline
{\includegraphics[width=16cm]{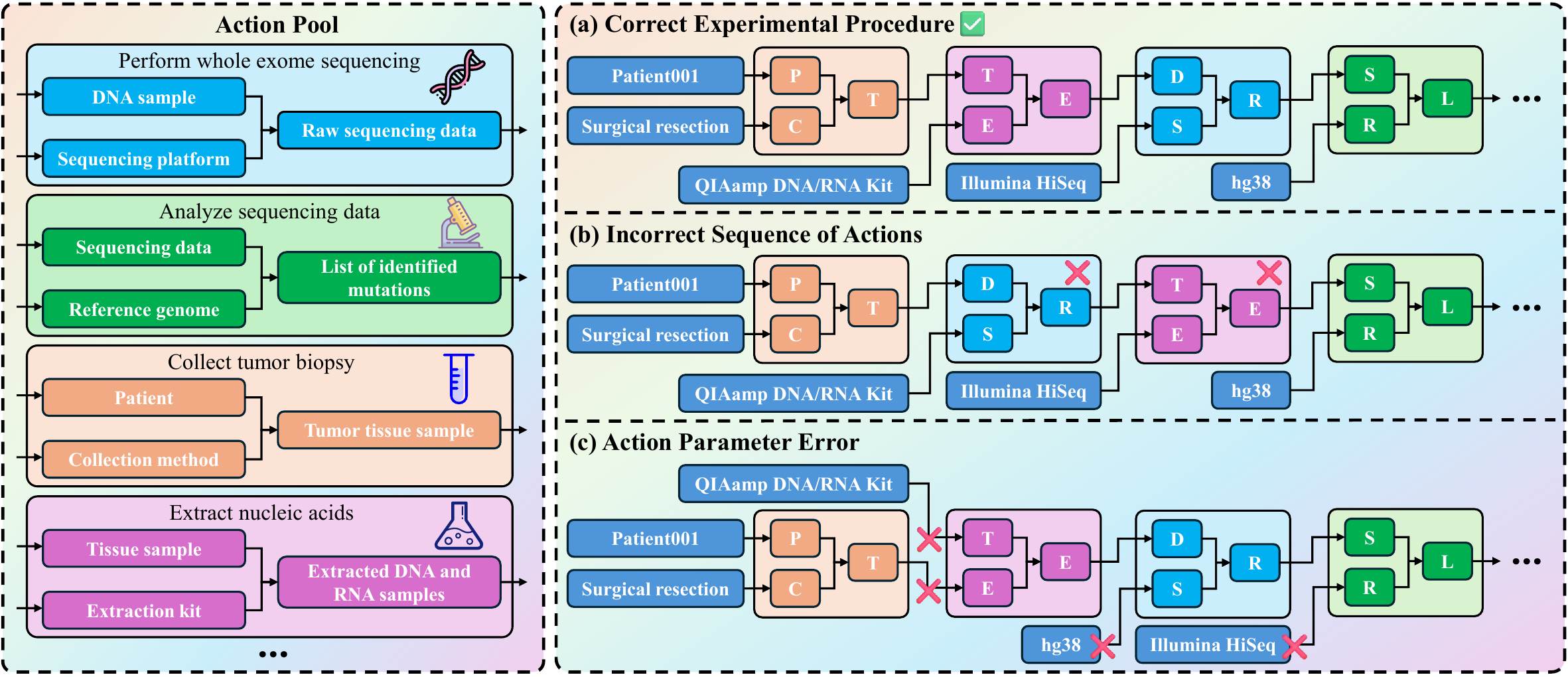}}
\caption{\textbf{Wet Experiment Workflow}: Action-pool based protocol construction with typical errors in step sequencing and parameter specification.}
\label{fig: wet_case1}
\end{figure}

For wet experiments, we provide models with an action pool containing standardized experimental operations and detailed descriptions. Given the experimental context, the model is required to synthesize a complete workflow, including both the selection and ordering of actions as well as all associated parameters (Figure~\ref{fig: wet_case1}). As illustrated in the figure, the model outputs typically exhibit two major categories of errors: (i) incorrect ordering of experimental steps and (ii) inaccurate or inconsistent parameter specification.

\begin{figure}[ht]
\centerline
{\includegraphics[width=17cm]{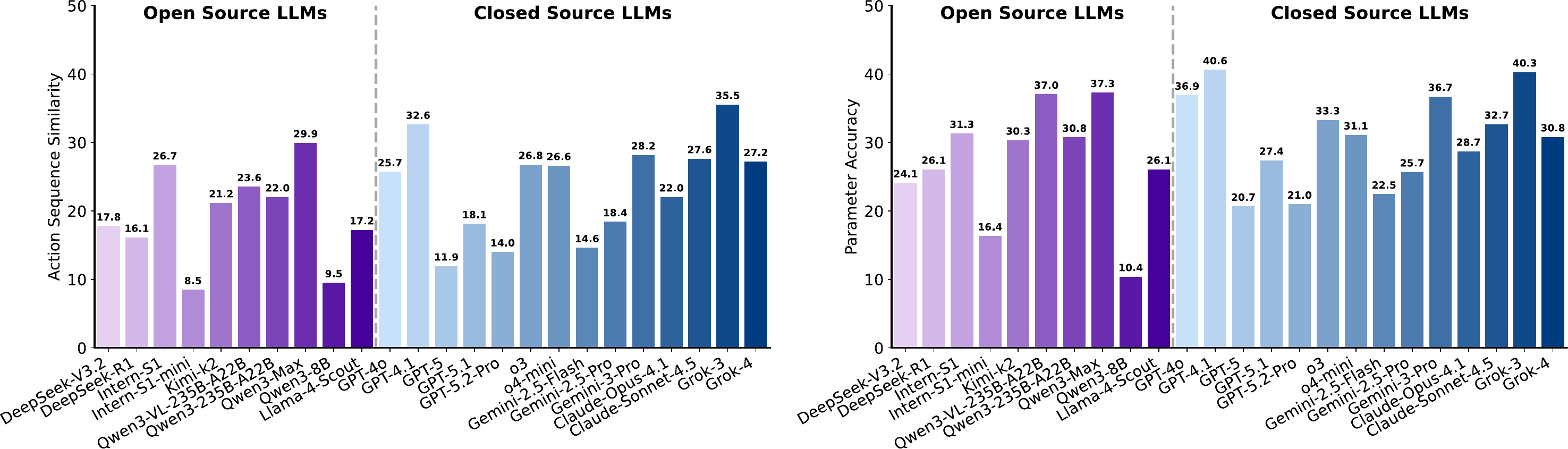}}
\caption{\textbf{Wet Experiment Evaluation}: Sequence Similarity (SS) and Parameter Accuracy (PA) across models for laboratory protocol planning.}
\label{fig: wet_metrics}
\end{figure}

\textbf{Wet experiments reasoning remains brittle.}
Figure~\ref{fig: wet_metrics} summarizes performance in terms of sequence similarity (SS) and parameter accuracy (PA). For SS, closed-source models in general achieve higher scores than open-source ones (with the best closed-source model around 35.5 versus the best open-source below 30), yet SS remains uniformly low across all systems. In contrast, PA exhibits a mixed pattern: although the top result is obtained by a closed-source model (around 40.6), several open-source models are competitive, and some closed-source models drop markedly (e.g., near 20.7). PA appears slightly more optimistic also since permutation-equivalent parameter groups are treated as identical (e.g., $\langle \text{action 1} \rangle(B, C)$ and $\langle \text{action 1} \rangle(X, Y)$ are identical when $B{=}X$ and $C{=}Y$), but both families still achieve only modest scores. Across outputs, errors recur in three patterns: insertion of unnecessary steps, omission of essential steps, and incorrect ordering of valid steps.

\begin{figure}[ht]
\centerline
{\includegraphics[width=16cm]{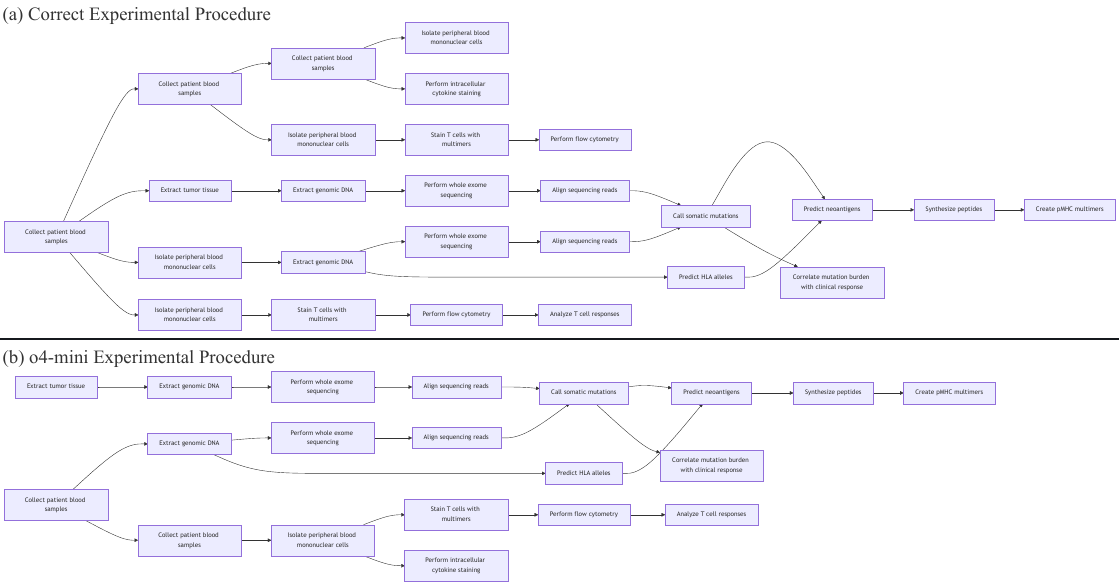}}
\caption{\textbf{Wet Experiment Case Study}: NSCLC anti–PD-1 immunotherapy workflow—ground-truth protocol versus model-generated design.}
\label{fig: wet_case2}
\end{figure}

\textbf{Temporal and branch-aware planning is often broken.}
Figure~\ref{fig: wet_case2} presents an experiment examining how tumor mutational burden and neoantigen load influence the efficacy of anti–PD-1 immunotherapy in non–small cell lung cancer. The ground-truth workflow (Figure~\ref{fig: wet_case2}~a) features a deeply branched structure with precisely coordinated timing and sample-handling procedures. In contrast, the workflow generated by o4-mini is substantially simplified and deviates from several core principles of experimental design.

First, the model collapses longitudinal sampling into a single blood draw and does not distinguish time windows, precluding any meaningful reconstruction of T‑cell dynamics. Second, PBMC isolation is executed only once rather than per time point, causing misalignment with downstream staining and flow cytometry. Functional assays (e.g., intracellular cytokine staining) are performed on a single PBMC aliquot without branching by time point or antigenic stimulation, and flow cytometry is likewise conducted only once, failing to capture temporal variation. Finally, the blood-sample branch conflates genomic and immunophenotyping workflows: “Extract genomic DNA” is executed in parallel with PBMC isolation and downstream immunology, leading to duplicated and cross‑purpose use of peripheral blood. These design flaws mirror the low sequence similarity and only moderate parameter accuracy observed in Figure~\ref{fig: wet_metrics}, underscoring failures in temporal coordination, branch-aware planning, and sample bookkeeping.

Overall, the deviations highlight a critical limitation of current AI models: while they can enumerate plausible wet experiment actions, they struggle to construct experimentally valid, temporally consistent, and branch-aware protocols. These limitations point to fundamental gaps in reasoning about experimental constraints, biological timing, and multi-sample coordination—elements essential for real-world scientific experimentation.

\subsection{Experimental Reasoning}

\begin{figure}[ht]
\centerline
{\includegraphics[width=16cm]{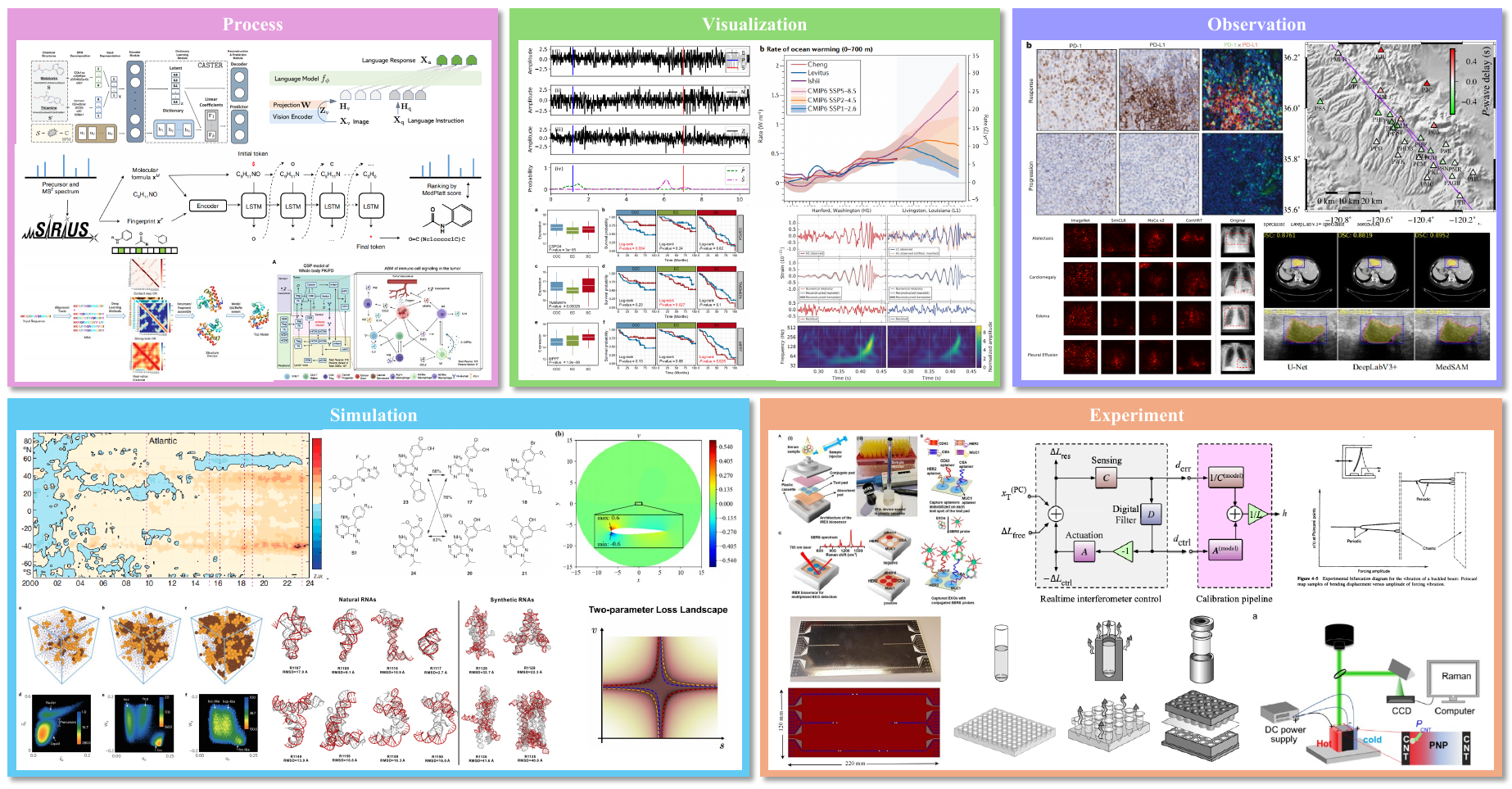}}
\caption{\textbf{Experimental Reasoning Modalities}: Examples of process, visualization, observation, simulation, and experiment images used as multi-modal evidence.}
\label{fig: image_case1}
\end{figure}

Experimental Reasoning evaluates the ability of multimodal LLMs to interpret experimental observations, integrate heterogeneous scientific evidence, and refine testable hypotheses. As illustrated in Figure~\ref{fig: image_case1}, the visual inputs span five representative modalities in scientific practice—process diagrams, data visualizations, natural observations, numerical simulations, and laboratory experiments—reflecting the diversity of multimodal information that underpins real-world scientific inquiry.

\begin{figure}[ht]
\centerline
{\includegraphics[width=16cm]{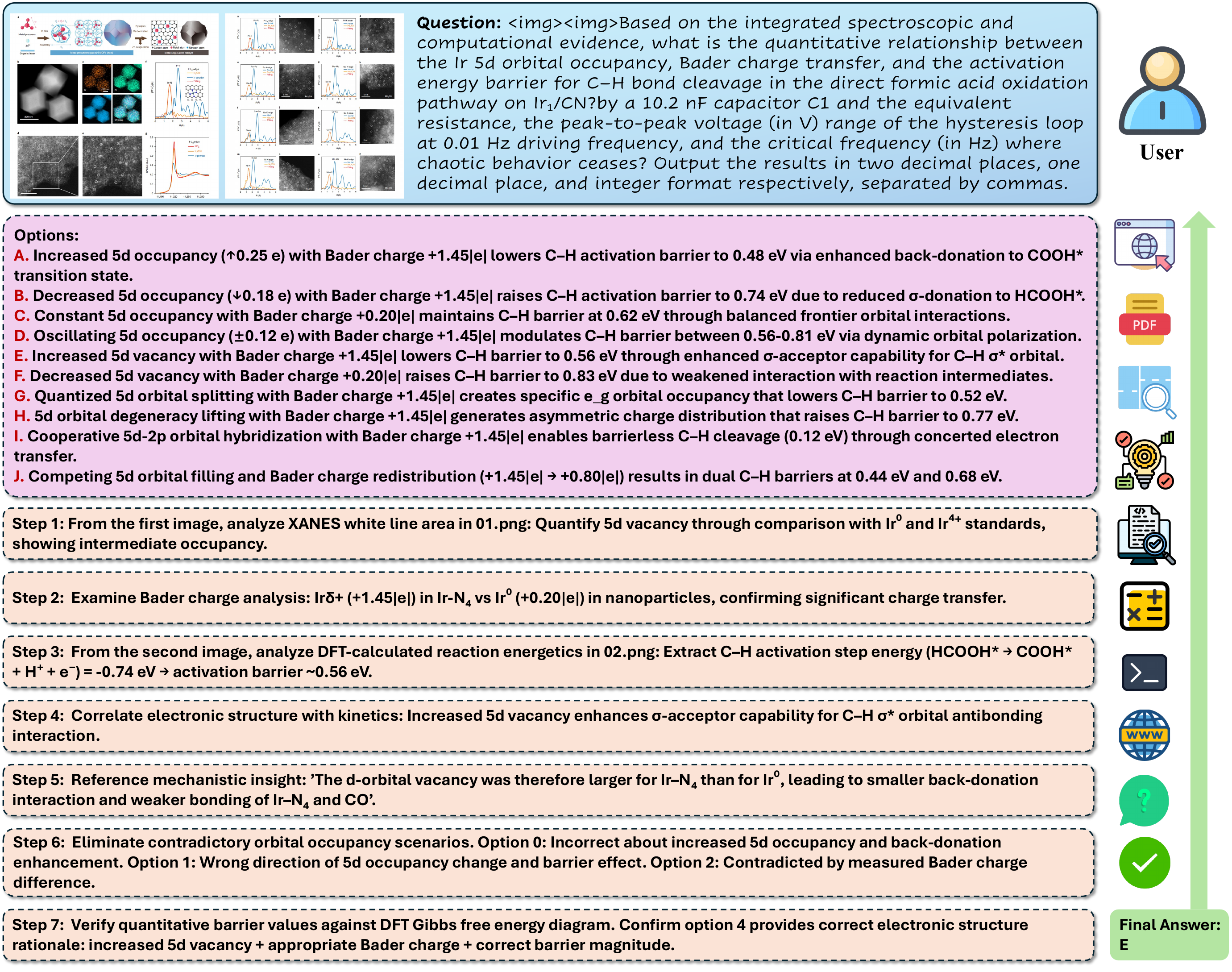}}
\caption{\textbf{Experimental Reasoning Case}: Multi-image question requiring cross-modal synthesis and step-wise reasoning.}
\label{fig: mm_case1}
\end{figure}

In this task, models are provided with several images accompanied by a question and must select the correct answer from at least ten candidates (Figure~\ref{fig: mm_case1}). Solving these problems requires multi-step inferential reasoning: identifying relevant variables, synthesizing multimodal cues, evaluating competing hypotheses, and ultimately validating consistency across the provided evidence. We therefore evaluate model performance using both Multi-choice Accuracy and Reasoning Validity, the latter assessing whether the model's explanation follows logically from the scientific evidence.

\begin{figure}[ht]
\centerline
{\includegraphics[width=16cm]{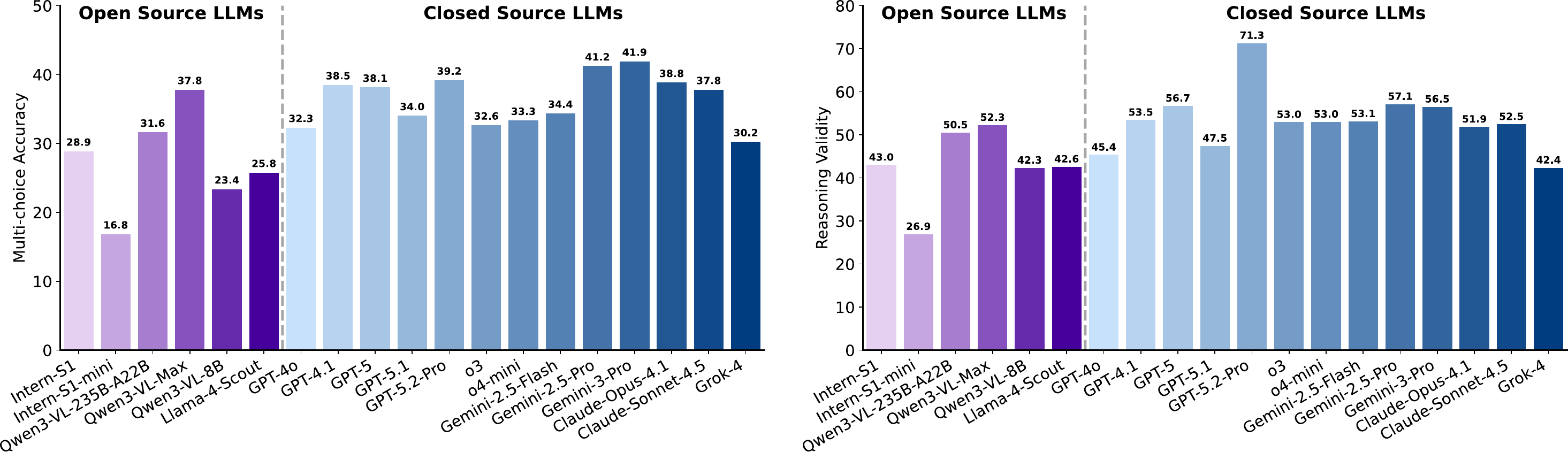}}
\caption{\textbf{Experimental Reasoning Evaluation}: Multi-Choice Accuracy (MCA) and Reasoning Validity (RV) across models on multimodal tasks.}
\label{fig: mm_results}
\end{figure}

\textbf{Reasoning validity often exceeds answer accuracy.}
As shown in Figure~\ref{fig: mm_results}, closed-source LLMs generally outperform open-source counterparts on both metrics, with the best closed-source models achieving higher MCA (e.g., up to 41.9) and RV (e.g., up to 71.3) than the best open-source models (MCA 37.8, RV 52.3). However, several open-source models remain competitive with or exceed some closed-source systems in specific metrics (e.g., Qwen3-VL-235B-A22B RV 50.5 > GPT-4o RV 45.4), indicating nontrivial overlap. Most models score higher in Reasoning Validity than in Multi-choice Accuracy, suggesting that even when the final choice is incorrect, explanations often preserve partial logical coherence. Variance is moderate—particularly among closed-source models—while only a few models (e.g., Intern-S1-mini) show noticeably lower performance, pointing to the importance of scale for robust multimodal scientific reasoning.

\begin{figure}[ht]
\centerline
{\includegraphics[width=16cm]{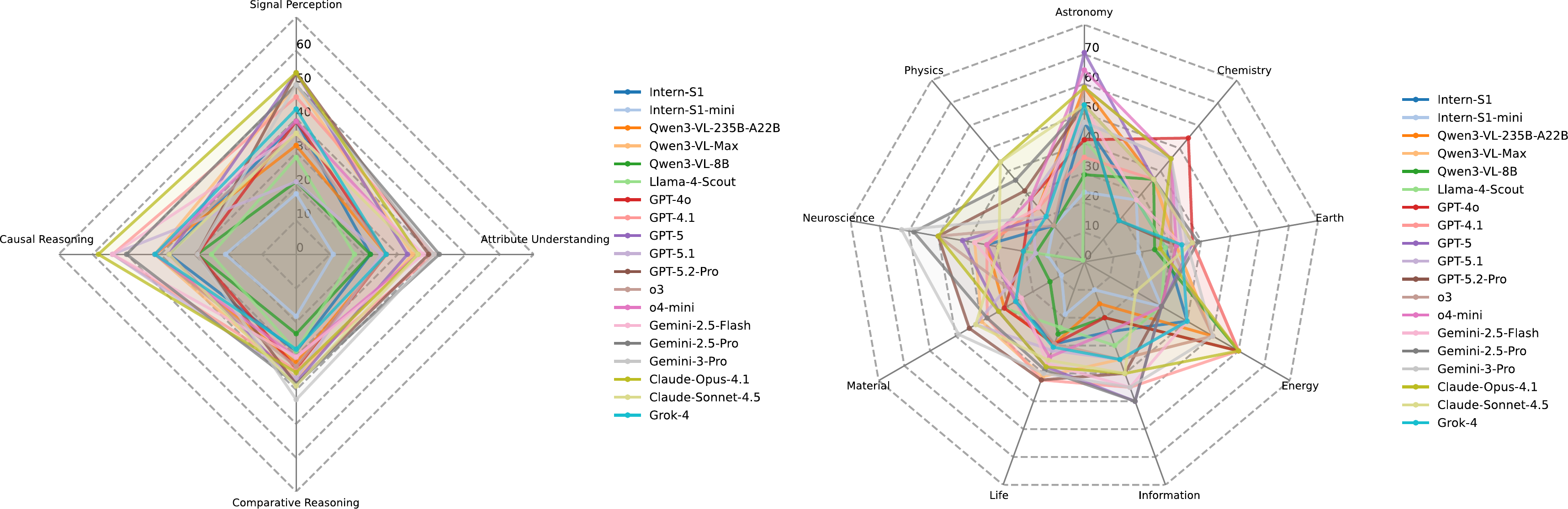}}
\caption{\textbf{Experimental Reasoning Performance by Type and Discipline}: Breakdown across reasoning paradigms (signal, attribute, comparative, causal) and 10 scientific domains.}
\label{fig: mm_tasks_subjects}
\end{figure}

\textbf{Comparative reasoning is the most challenging across domains.}
To further dissect these capabilities, we analyze performance across reasoning types and disciplinary domains (Figure~\ref{fig: mm_tasks_subjects}).
From the perspective of reasoning categories, including signal perception, attribute understanding, comparative reasoning, and causal reasoning, LLMs perform consistently well in causal reasoning and perceptual recognition. In contrast, comparative reasoning emerges as a persistent weakness. This indicates that models struggle when required to contrast subtle quantitative or qualitative differences, a cognitive operation fundamental to scientific evaluation and hypothesis discrimination.
When examining performance across 10 scientific disciplines, an intriguing pattern emerges. Models achieve their highest accuracy in astronomy, followed by chemistry, energy science, and neuroscience. These domains often feature structured visual patterns or canonical experimental setups, which may align well with LLMs' prior training data. Conversely, performance declines substantially in materials science, life sciences, and Earth sciences, where visual cues are more heterogeneous, context-dependent, or experimentally nuanced. This divergence suggests that domain-specific complexity and representation diversity strongly influence multimodal reasoning performance.

Overall, these findings reveal that while current LLMs demonstrate encouraging abilities in integrating scientific evidence and conducting basic causal analyses, they still fall short in tasks requiring precise discrimination, cross-sample comparison, and nuanced interpretation of domain-specific observations. The relatively narrow performance gap among leading models underscores that scale alone is insufficient; advancing experimental reasoning will require improved multimodal grounding, finer-grained visual understanding, and training paradigms explicitly aligned with scientific inquiry.

%% file: sections/5-discussion.tex
\section{Analysis}

\subsection{Test Time Reinforcement Learning}

Large Language Models (LLMs) have demonstrated remarkable capabilities in reasoning and problem-solving, primarily driven by supervised fine-tuning and reinforcement learning on extensive labeled datasets. However, applying these models to the frontier of scientific discovery, particularly in the task of scientific idea generation, presents a fundamental challenge: the inherent absence of ground truth. Unlike closed-domain tasks such as mathematical reasoning or code generation, where solutions can be verified against a correct answer, the generation of novel research ideas is an open-ended problem with no pre-existing ``gold standard'' labels. This limitation renders traditional offline training pipelines insufficient for adapting to dynamic and unexplored scientific territories. 

Consequently, the critical research question becomes: \textit{How can we enhance a model's capability during the inference phase in the absence of ground-truth supervision?} To address this, we adopt the paradigm of \textbf{Test-Time Reinforcement Learning (TTRL)}~\cite{zuo2025ttrl}. This framework enables models to self-evolve on unlabeled test data by optimizing policies against rule-based rewards derived from the model's own outputs or environmental feedback. Distinct from the original implementation~\cite{zuo2025ttrl}, which primarily leveraged consensus-based consistency as a reward mechanism for logical reasoning tasks, we establish \textbf{novelty} as our core optimization objective in the current context. Consequently, we introduce a TTRL framework where the reward signal is constructed based on the dissimilarity between generated ideas and retrieved related works, guiding the model to actively explore the solution space and maximize innovation at test time.

\subsubsection{Methodology}

\begin{figure}[htbp]
    \centering
    \includegraphics[width=0.95\textwidth]{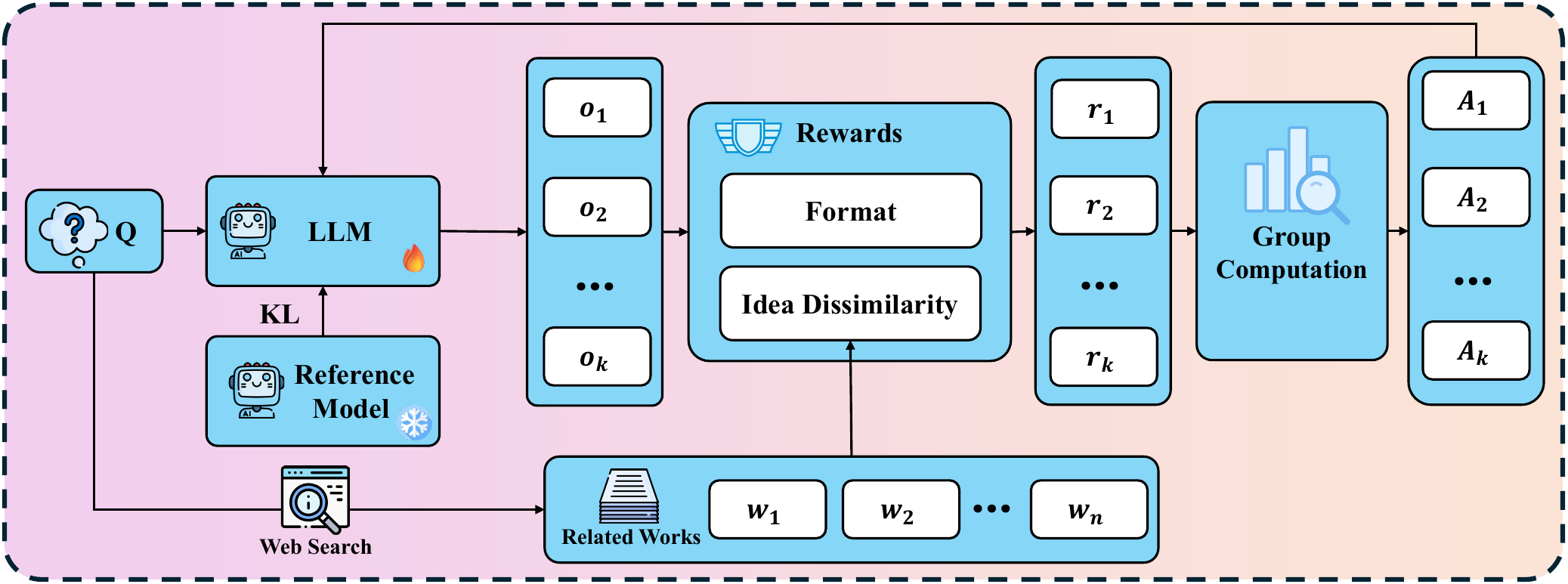}
    \caption{\textbf{TTRL Training Framework}: The model generates candidate ideas evaluated against online retrieved related works to calculate novelty rewards, guiding GRPO updates.}
    \label{fig:train_process}
\end{figure}

To address the absence of ground truth in scientific idea generation, we propose a generalizable reward mechanism based on \textbf{online retrieval}. Instead of relying on static labels, we utilize real-time search to fetch existing related works, serving as a dynamic baseline for comparison. This approach enables us to quantify novelty as the semantic dissimilarity between the model's output and the retrieved context, effectively converting an open-ended exploration problem into a measurable optimization task. The overall training framework is illustrated in Figure \ref{fig:train_process}.

We employ Group Relative Policy Optimization (GRPO)~\cite{guo2025deepseek} as our training backbone. For a given query $Q$, the policy model $\pi_\theta$ generates a group of $k$ outputs $\{o_1, \dots, o_k\}$. The optimization is guided by a composite reward function, defined as the unweighted sum of a format constraint and a novelty metric (labeled as \textbf{Idea Dissimilarity} in Figure \ref{fig:train_process}):
\begin{equation}
    R(o) = R_{\text{format}}(o) + R_{\text{novelty}}(o, \mathcal{W})
\end{equation}
where $\mathcal{W} = \{w_1, \dots, w_n\}$ denotes the set of related works obtained via online search.

\paragraph{Format Reward ($R_{\text{format}}$).} 
To guarantee interpretable reasoning, we enforce a strict XML structure. The model must encapsulate its chain of thought within \texttt{<think>...</think>} and the final proposal within \texttt{<answer>...</answer>}. The format reward is binary:
\begin{equation}
    R_{\text{format}}(o) = \mathbb{I}\left(o \text{ follows the specified XML structure}\right)
\end{equation}

\paragraph{Novelty Reward ($R_{\text{novelty}}$).} 
We quantify novelty by measuring the vector space dissimilarity between the generated idea and the retrieved literature. Let $\mathbf{e}_{\text{idea}}$ be the embedding of the generated answer, and $\{\mathbf{e}_{w_j}\}_{j=1}^n$ be the embeddings of $n$ retrieved papers (denoted as $w_1, \dots, w_n$ in the figure). We compute the average cosine similarity $S_{\text{avg}}$:
\begin{equation}
    S_{\text{avg}} = \frac{1}{n} \sum_{j=1}^n \frac{\mathbf{e}_{\text{idea}} \cdot \mathbf{e}_{w_j}}{\|\mathbf{e}_{\text{idea}}\| \|\mathbf{e}_{w_j}\|}
\end{equation}
An innovation score $S_{\text{inn}} \in [0, 10]$ is then derived to reward divergence:
\begin{equation}
    S_{\text{inn}} = \text{clip}\left( (1 - S_{\text{avg}}) \times 10, \ 0, \ 10 \right)
\end{equation}
Using a gating threshold $\tau=5$, the final novelty reward is defined as:
\begin{equation}
    R_{\text{novelty}}(o, \mathcal{W}) = \mathbb{I}(S_{\text{inn}} > \tau)
\end{equation}
This mechanism incentivizes the model to produce ideas that are semantically distinct from existing work.

\subsubsection{Experimental Setup}

\begin{figure}[htbp]
    \centering
    \begin{minipage}[c]{0.40\textwidth} 
        \centering
        \captionof{table}{\textbf{TTRL Hyperparameters}: Key training configuration for GRPO-based test-time reinforcement learning.}
        \label{tab:hyperparameters}
        \resizebox{0.9\linewidth}{!}{
            \begin{tabular}{lc}
                \toprule
                \textbf{Hyperparameter} & \textbf{Value} \\
                \midrule
                Base Model & Qwen3-8B \\
                RL Algorithm & GRPO \\
                Precision & bfloat16 \\
                Learning Rate & $5 \times 10^{-7}$ \\
                Max Length & 2048 \\
                Generations ($G$) & 8 \\
                Temperature & 1.0 \\
                Batch Size & 4 \\
                Related Works ($n$) & 4 \\
                Weights & 1:1 \\
                \bottomrule
            \end{tabular}
        }
    \end{minipage}
    \hfill
    \begin{minipage}[c]{0.55\textwidth} 
        \centering
        \includegraphics[width=0.95\linewidth, valign=c]{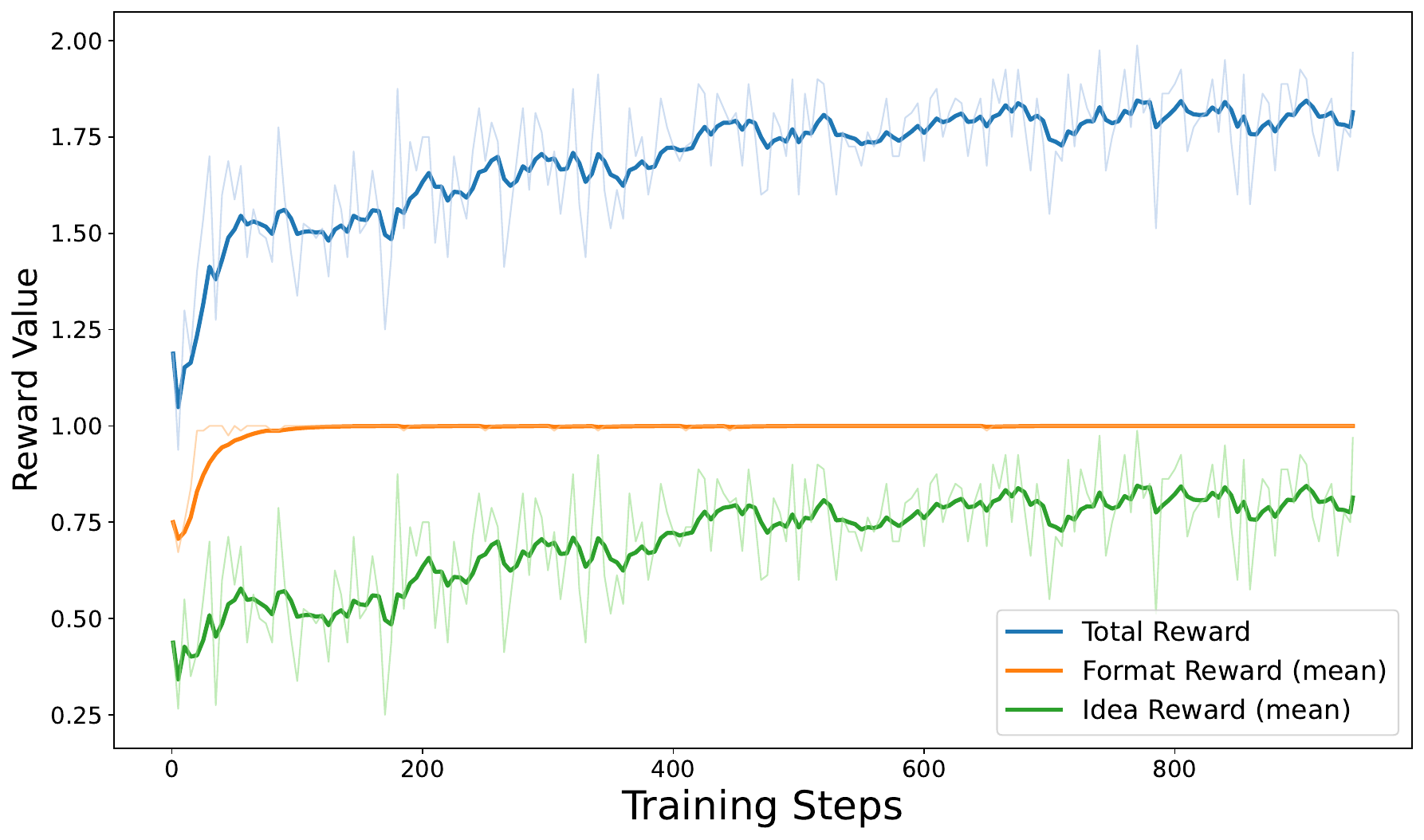}
        \caption{\textbf{TTRL Training Dynamics}: Format reward saturates quickly, followed by steady growth in idea novelty.}
        \label{fig:reward_curves}
    \end{minipage}
\end{figure}

We employ Qwen3-8B as the base model, trained using the GRPO algorithm within the ms-swift~\cite{zhao2025swiftascalablelightweightinfrastructure} framework. To facilitate diverse exploration, we utilize a high sampling temperature. Key hyperparameters are detailed in Table \ref{tab:hyperparameters}.

\subsubsection{Experimental Results}

The training dynamics of our TTRL framework are illustrated in Figure \ref{fig:reward_curves}. The curves demonstrate a clear two-phase optimization process. Initially, the \textbf{Format Reward} (orange line) rises rapidly and saturates near 1.0 within the first few steps, indicating that the model quickly adapts to the rigid XML structural constraints (\texttt{<think>} and \texttt{<answer>} tags). Once the format is stabilized, the \textbf{Idea Reward} also starts to rise (green line). Despite the inherent difficulty of the task, the Idea Reward exhibits a consistent upward trend throughout the training steps, driving the total reward (blue line) to converge at a higher value.

Quantitatively, this self-evolution process yields a significant improvement in the quality of generated ideas. The average novelty score of the model's outputs increased from a baseline of \textbf{49.36} to \textbf{62.06}. It is important to emphasize that this performance gain was achieved \textit{entirely without ground-truth labels}. The model improved solely by leveraging the online retrieval feedback loop, validating the hypothesis that LLMs can self-improve on open-ended scientific discovery tasks through test-time reinforcement learning.

\subsubsection{Case Study of TTRL}

\begin{figure}[htbp]
    \centering
    \includegraphics[width=0.95\textwidth]{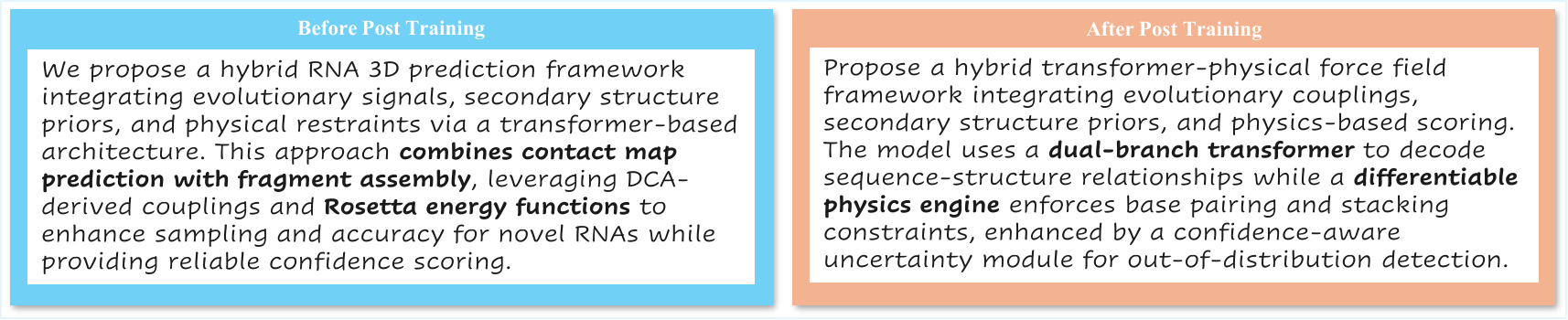}
    \caption{\textbf{TTRL Case Study}: Comparison of generated research ideas before and after TTRL, highlighting structural innovation (dual-branch transformer, differentiable physics engine) versus generic pre-training assembly.}
    \label{fig:case_study}
\end{figure}

To visually demonstrate the impact of TTRL on scientific idea generation, we present a comparative case study in Figure \ref{fig:case_study}. The task requires the model to propose a novel framework for RNA 3D structure prediction.

Comparing the responses before and after training reveals a noticeable improvement in both specificity and novelty. 
The \textbf{Pre-Training Response} suggests a standard combination of existing components, essentially assembling "contact map prediction" with "Rosetta energy functions." While logical, it represents a conventional approach without distinct architectural details.

In contrast, the \textbf{Post-Training Response} introduces more structurally specific and technically distinct concepts. It explicitly proposes a \textbf{"dual-branch transformer"} and replaces static energy functions with a \textbf{"differentiable physics engine."} Additionally, it incorporates a \textbf{"confidence-aware uncertainty module"} to address the reliability challenge. This shift indicates that the model has moved beyond generic component assembly toward generating more detailed and differentiated technical proposals.

\paragraph{Summary.}
In conclusion, our experiments demonstrate that Test-Time Reinforcement Learning (TTRL), driven by retrieval-based novelty rewards, effectively enhances model capabilities in the absence of ground-truth supervision. The observed improvements in both quantitative novelty metrics and qualitative technical specificity indicate that the model can successfully self-evolve beyond conventional patterns. These findings suggest that TTRL is a promising paradigm for adapting Large Language Models to the open-ended and unexplored frontiers of real-world scientific discovery.

\subsection{Agent Tool Integrated Reasoning}

\subsubsection{Retrieve–Browse Loop Analysis}
Tool-Integrated Reasoning (TIR) in real tasks unfolds as a dynamic, opportunistic process rather than a fixed linear chain\cite{paranjape2023artautomaticmultistepreasoning}. As shown in \autoref{fig:agent_tool_compare} (left), the model-to-tool flow concentrates heavily on retrieval actions: \texttt{web\_search} is the most frequently invoked tool with 539 calls (33.98\% of all), followed by \texttt{visit\_webpage} (385, 24.27\%), \texttt{final\_answer} (358, 22.57\%), python\_interpreter (200, 12.61\%), and \texttt{wikipedia\_search} (104, 6.56\%). This distribution indicates that an external “retrieve-then-browse” loop remains the dominant path for contemporary agentic systems, reflecting persistent limits in time-sensitive and domain-specific knowledge available to base LLMs. Importantly, models differ in how efficiently they traverse this loop: for example, GPT-4.1 issues large volumes of \texttt{web\_search} (168) and \texttt{visit\_webpage} (110) that frequently land in slow tiers, whereas Qwen3-Max completes comparable coverage with far fewer retrieval and browsing steps (61 and 59, respectively). Practically, this pattern implies that reducing redundant retrieval iterations—via better query formulation and higher-quality extraction on the first pass—has immediate leverage on end-to-end latency, often exceeding gains from marginal improvements to raw model inference.

\subsubsection{Tool Efficiency Analysis}
\begin{figure}[ht]
\centerline
{\includegraphics[width=16cm]{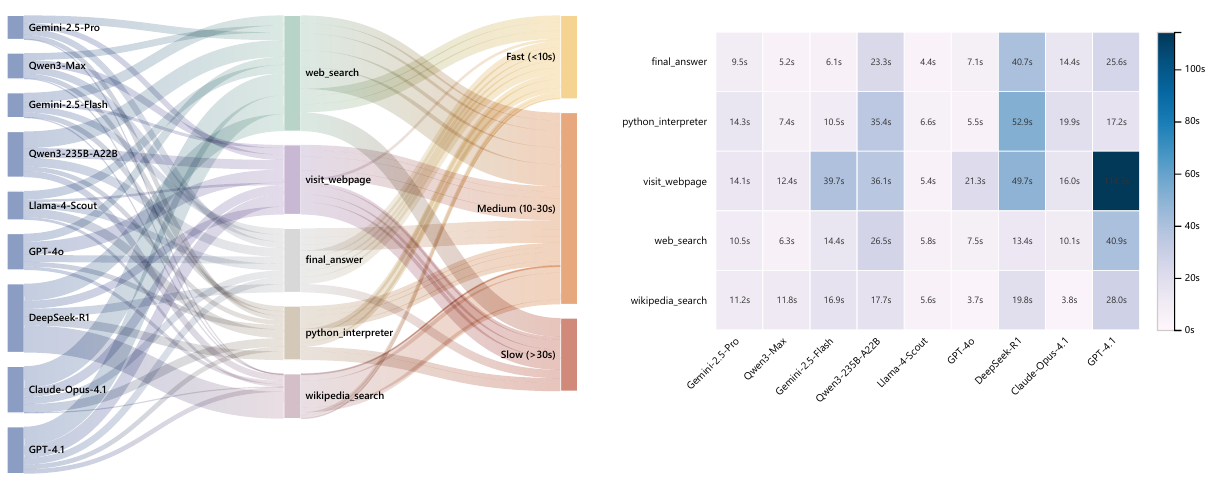}}
\caption{\textbf{Agent Tool Calls}: Frequency (left) and efficiency (right) across leading models.}
\label{fig:agent_tool_compare}
\end{figure}

Latency variation is predominantly tool-dependent, as visualized in \autoref{fig:agent_tool_compare} (right). The primary bottleneck is \texttt{visit\_webpage}, whose cross-model latency spans from 5.37s (Llama-4-Scout) to 114.29s (GPT-4.1), a 21.28× spread. This reflects the intrinsic cost of browser-level execution—network I/O, DOM parsing, and event replay—rather than LLM reasoning alone. In contrast, more atomic operations such as \texttt{wikipedia\_search} still exhibit a substantial 7.59× spread (3.69–28.03s), underscoring that I/O pathways and parsing routines meaningfully shape end-to-end time even for ostensibly simple tools. These observations suggest a design priority: engineering optimizations in the retrieval-and-browsing pipeline (e.g., smarter caching, incremental browsing, selective content extraction) will reduce both long-tail latencies and overall wall-clock time more reliably than tuning model-only parameters.

\subsubsection{Reasoning Cost Analysis}
The \texttt{python\_interpreter} tool exhibits a 9.65× cross-model range (5.48–52.94s), indicating that measurements capture the full “reason–execute–debug–repair” loop rather than a single code run. The slowest average arises for DeepSeek-R1 (52.94s), consistent with more frequent multi-step error analysis and correction; the fastest is GPT-4o (5.48s), reflecting a low-latency, near single-shot execution path. This divergence reveals a strategic trade-off: systems optimized for first-attempt correctness minimize tool time but may forgo deeper self-correction, whereas systems favoring iterative refinement accrue longer tool-side latency while potentially achieving more robust final solutions. In practice, aligning tool routing, retry policy, and verification depth with a model's characteristic behavior can reduce wasted cycles and sharpen the latency–quality frontier.

\subsection{SGIEvalAgent}

\begin{figure}[ht]
\centerline
{\includegraphics[width=16cm]{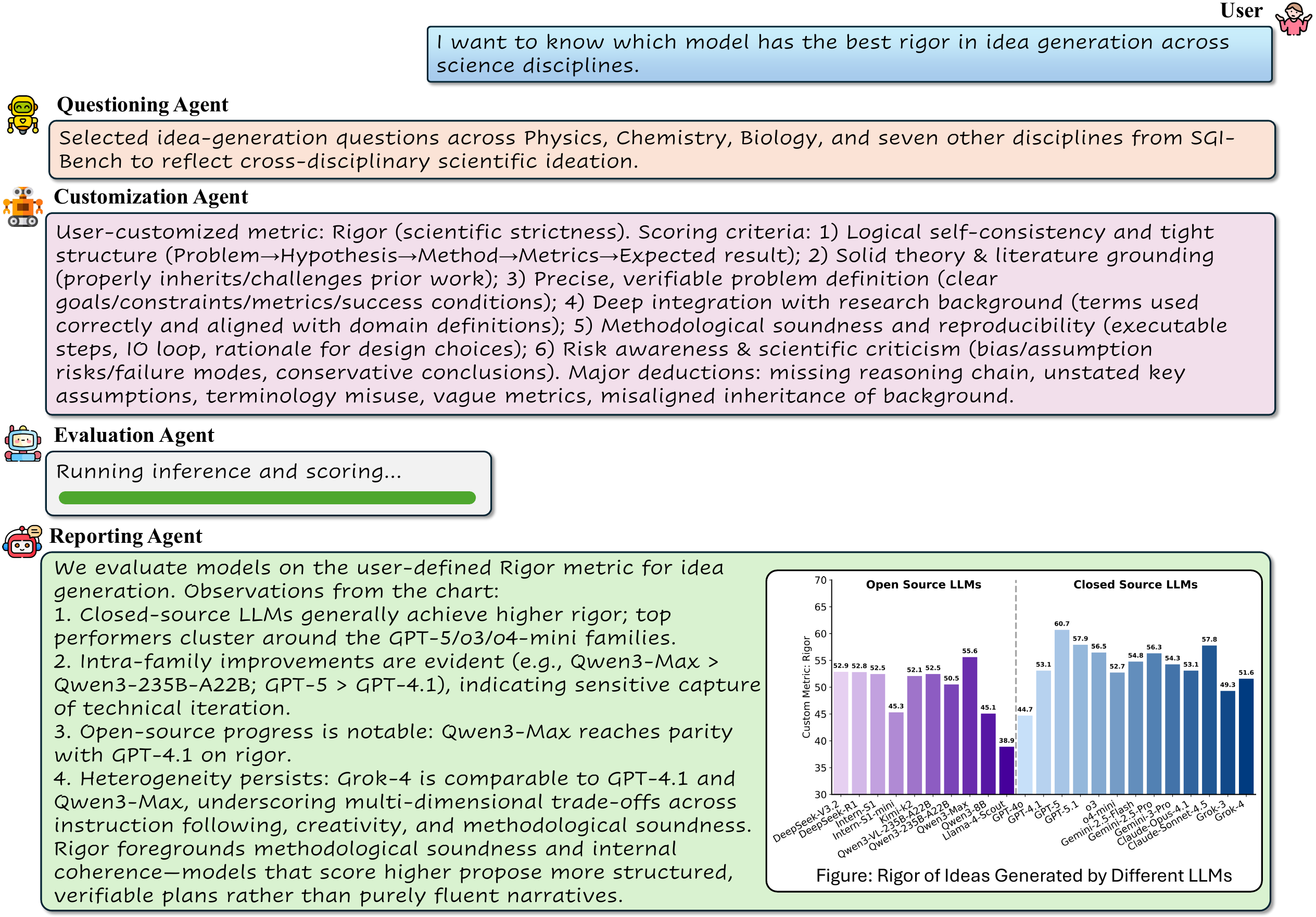}}
\caption{\textbf{SGIEvalAgent Case}: Model Users describe their evaluation needs, SGIEvalAgent customizes the evaluation plan and metrics based on these needs, and finally provides an evaluation report..}
\label{fig: idea_custom_metric}
\end{figure}

\subsubsection{User-customized Metric}
SGIEvalAgent interprets the user's evaluation intent and turns it into a rubric that can be applied consistently across the selected idea-generation questions. In the case shown in Figure~\ref{fig: idea_custom_metric}, the user asks to compare models on “rigor” in cross-disciplinary idea generation. The system formalizes Rigor (scientific strictness) for idea proposals so that it reflects how scientists judge whether a plan is internally coherent, well grounded, and practically verifiable.

The rubric expresses six aspects in prose rather than checklists. First, it checks logical self-consistency and completeness of the pipeline from problem to hypothesis, method, metrics, and expected results. Second, it requires theory and literature grounding that either correctly inherits prior work or responsibly challenges it with evidence. Third, it demands precise and verifiable problem definitions that state goals, constraints, evaluation metrics, and success conditions. Fourth, it looks for deep fit with the research background and correct, discipline-aligned terminology. Fifth, it evaluates methodological soundness and reproducibility through executable steps, a clear I/O loop, and explicit rationale for key design choices. Sixth, it considers risk awareness and scientific criticism by articulating assumptions, potential failure modes, bias sources, and avoiding over-confident conclusions. Major deductions apply when the reasoning chain is missing, key assumptions are unstated, terminology is misused, metrics are vague or non-verifiable, or inheritance from background knowledge is misaligned.

Scores are produced on a 0–10 scale for each aspect and aggregated with default equal weights into a single rigor score; the result is linearly mapped to a 0–100 axis for visualization without changing rank order. The evaluation agent generates textual rationales that cite reference answers and problem context so that decisions are transparent and reproducible. Customized metrics are reported alongside SGI-Bench's predefined task metrics rather than replacing them, preserving standardized comparability while highlighting the user's domain-specific focus.

\subsubsection{Automated Evaluation Report}
The reporting agent summarizes the customized metric and the evaluation outputs into a concise narrative with figures. In Figure~\ref{fig: idea_custom_metric}, the report contrasts open-source and closed-source systems on the user-defined rigor metric for idea generation and highlights what the scores mean in practice.

The core takeaway is straightforward: closed-source models generally exhibit higher rigor under this rubric, intra-family iterations capture measurable gains, and leading open-source models show notable progress that narrows the gap. Higher rigor reflects more structured, well-grounded, and verifiable research plans rather than merely fluent narratives. The report therefore gives users a clear, scientist-aligned comparison they can directly use for model selection and iteration in research workflows.

\section{Challenges and Future Directions}
Grounded in our operational definition of SGI and instantiated through SGI-Bench, the evaluation results reveal a consistent message: contemporary LLMs and agentic systems exhibit \emph{localized scientific cognition and segmented scientific reasoning}
They may solve isolated sub-problems, but fail to robustly close the iterative loop spanning \textit{Deliberation, Conception, Action, and Perception}. Below we summarize the main limitations across tasks and disciplines, connect them with our TTRL and tool-integrated reasoning analyses, and outline concrete future directions.

\subsection{Fragmentation Across the Four Quadrants of SGI}

\paragraph{Deliberation: Scientific Deep Research remains brittle end-to-end.}
Scientific Deep Research operationalizes the literature-review/meta-analysis stage and is evaluated by Exact Match (EM) and Step-Level Accuracy (SLA). Across both standalone LLMs and tool-augmented agents, EM is consistently low: most systems achieve only $\sim$10\% accuracy, and even the best models rarely exceed 20\% EM (Figure~\ref{fig: llms deep research}, Figure~\ref{fig: agents deep research}). This indicates that current models still fail to produce \emph{verifiable final scientific claims} under multi-source evidence integration.

A notable gap exists between SLA and EM. SLA is substantially higher for nearly all systems, with several agentic systems reaching $\approx$50\% SLA (Figure~\ref{fig: agents deep research}), while EM remains low. This disparity shows that models often produce \emph{locally correct steps} but cannot maintain global coherence across long reasoning chains. The failure mode is therefore not mere knowledge absence, but \emph{reasoning trajectory collapse} under long-horizon scientific inference.

At a finer granularity, Deep Research tasks involving \textbf{Data} and \textbf{Properties} are the weakest: performance on these categories is substantially below that of \textbf{Micro-} and \textbf{Macro-experiment} questions, with \emph{all four categories rarely exceeding 30\%} accuracy (Figure~\ref{fig: deep research on different task}). This aligns with the task design: data/property questions require retrieving dispersed numerical details across heterogeneous papers, while experiment-oriented questions provide more structured evidence. The results thus expose a core SGI bottleneck: \emph{meta-analytic retrieval + numerical aggregation over scattered literature}.

\paragraph{Conception: Ideas lack implementability.}
Idea Generation in SGI-Bench is assessed using \textbf{Effectiveness}, \textbf{Detailedness}, and \textbf{Feasibility} (Table~\ref{tab:idea_gen_res}). \textbf{Feasibility is low across models}: many systems score in the 14–20 range, and the best result reaches 22.90 (\texttt{o3}), indicating that feasibility consistently lags behind novelty and detailedness. \textbf{Detailedness remains insufficient for several models}, with implementation steps frequently missing concrete parameters, resource assumptions, or step ordering; \textbf{Effectiveness is moderate for most systems}, with the highest result of 51.36 (GPT-5.2-Pro) and open-source models clustering around 24.95–28.74 (e.g., DeepSeek-V3.2, Llama-4-Scout).

Recurring issues include: (i) underspecified implementation steps—absent data acquisition or preprocessing plans, missing hyperparameters or compute assumptions, vague module choices (e.g., solver type, training objective, evaluation protocol), and unclear interfaces, ordering, or data flow; and (ii) infeasible procedures—reliance on unavailable instruments or data, uncoordinated pipelines that cannot be executed, and designs lacking reproducibility.

In SGI terms, current systems exhibit \emph{fluent linguistic ideation without sufficient methodological execution grounding}: they articulate concepts clearly but struggle to translate them into \emph{concrete, parameterized, and testable} workflows. The \textbf{feasibility gap} observed in Table~\ref{tab:idea_gen_res} is therefore a persistent bottleneck in realization, including within the Conception quadrant, where ideation quality does not reliably imply executable planning competence.

\paragraph{Action: Experimental execution is limited by numerical and procedural rigor.}
For \textbf{Dry Experiments}, accuracy is measured by PassAll@k. Even under the most lenient setting, the best PassAll@1 is only \textbf{42.07\%} (Claude-Sonnet-4.5), and under the strictest criterion, the best PassAll@5 rises to merely \textbf{36.64\%} (Gemini-3-Pro) (Table~\ref{tab:code metrics}). The spread between PassAll@1 and PassAll@5 (e.g., 42.07$\to$35.79 for Claude-Sonnet-4.5, 41.98$\to$36.64 for Gemini-3-Pro) indicates that models often nail partial logic but fail full scientific correctness.

Importantly, code executability is not the bottleneck: most frontier models achieve \textbf{SER $>$ 90\%} (e.g., GPT-5.1 96.53, Gemini-3-Pro 98.85), while accuracy remains low. This gap confirms a central limitation: \emph{syntactic fluency $\neq$ scientific computational reasoning}. The per-function analysis further shows numerical-calculation and simulation functions as the major failure mode (Figure~\ref{fig: dry_task_metric}), consistent with the case study (Figure~\ref{fig: code_case2}) where naive integration choices lead to cascading scientific errors.

For \textbf{Wet Experiments}, although Parameter Accuracy improves slightly under permutation-equivalence evaluation, \textbf{Sequence Similarity remains uniformly low} across both open and closed models (Figure~\ref{fig: wet_metrics}). Models frequently insert redundant steps, omit critical actions, or misorder multi-branch protocols. The complex oncology workflow case (Figure~\ref{fig: wet_case2}) illustrates that models cannot reliably manage temporal design, branching logic, or multi-sample coordination. Thus, wet-lab action planning remains a profound gap toward embodied SGI.

\paragraph{Perception: Multimodal reasoning is improving, but comparison is a hard frontier.}
In Experimental Reasoning, closed-source models consistently outperform open-source ones (Figure~\ref{fig: mm_results}). Across nearly all models, \textbf{Reasoning Validity (RV) exceeds Multi-choice Accuracy (MCA)}, showing that models can often produce partially coherent narratives even when selecting the wrong option. This echoes the SLA--EM gap in Deep Research and suggests a general pattern: models are better at producing plausible \emph{local reasoning} than globally correct scientific decisions.

Reasoning-type breakdown reveals that models perform relatively well on \textbf{Signal Perception} and \textbf{Causal Reasoning}, but \textbf{Comparative Reasoning is persistently weakest} (Figure~\ref{fig: mm_tasks_subjects}). Scientific comparison requires subtle cross-sample discrimination and quantitative contrast---a cognitive operation central to scientist judgment but not yet robustly captured by current MLLMs. Discipline-wise, astronomy and chemistry are easier, while materials science, life science, and Earth science remain hardest (Figure~\ref{fig: mm_tasks_subjects}), reflecting the mismatch between real scientific visual heterogeneity and training priors.

\subsection{Implications from Test-Time RL and Tool-Integrated Reasoning}
\paragraph{SGI as a dynamic, learnable capacity.}
Our TTRL experiments demonstrate that open-ended scientific ideation can improve \emph{without labeled supervision}. With retrieval-based novelty rewards, Qwen3-8B increases its novelty score from \textbf{49.36} to \textbf{62.06} (Figure~\ref{fig:reward_curves}) and qualitatively progresses from generic component assembly to structured innovation (Figure~\ref{fig:case_study}). These results suggest that SGI should be interpreted not merely as a static benchmark score, but as a \emph{capability that can evolve through test-time learning}. Nevertheless, optimizing for novelty in isolation risks ungrounded or implausible ideas; combining novelty with rigor- or feasibility-based rewards is a crucial next step for reliable scientific ideation.

\paragraph{The retrieval pipeline is the true bottleneck for agentic SGI.}
Tool-Integrated Reasoning (TIR) analysis reveals that agent workflows are heavily dominated by retrieval operations: \texttt{web\_search} accounts for \textbf{539 calls (33.98\%)}, and \texttt{visit\_webpage} for \textbf{385 calls (24.27\%)} (Figure~\ref{fig:agent_tool_compare}). Latency is primarily tool-driven rather than model-driven; \texttt{visit\_webpage} exhibits a \textbf{5.37s--114.29s} range across models (a \textbf{21.28$\times$} spread). This indicates that many gains in SGI performance may stem from \emph{smarter tool routing, reduction of redundant retrievals, and higher-quality first-pass extraction}, rather than simply scaling base LLMs. Analysis of the Python tool further highlights a trade-off between first-shot correctness and iterative self-repair, with a \textbf{9.65$\times$} cross-model latency range, underscoring the need for \emph{model-aware verification and retry policies} in practical agentic workflows.

\subsection{Future Directions Toward Scientific General Intelligence}

Our findings point to several high-leverage research directions:

\paragraph{(1) Meta-analytic reasoning with numerical robustness.}
Deep Research failures on Data/Properties and low EM despite high SLA call for methods that explicitly train \emph{evidence aggregation and numerical synthesis}. Promising routes include retrieval-conditioned quantitative reasoning, uncertainty-calibrated aggregation over multiple sources, and verification-aware step planning that penalizes reasoning-chain drift.

\paragraph{(2) Planning-aware conception and structured supervision.}
To address uniformly low feasibility and sparse implementation detail in Idea Generation, adopt planning-aware constraints with structured supervision: require parameter-complete, dependency-consistent steps, prioritize feasibility-focused rewards (availability checks, resource/cost estimates, reproducibility), and use lightweight tool checks during decoding to block or repair incomplete plans. This shifts fluent proposals into executable, testable designs under realistic scientific constraints.

\paragraph{(3) Scientific code training beyond syntax.}
Dry experiments show high SER but low PassAll@5 (Table~\ref{tab:code metrics}), especially on numerical and simulation functions (Figure~\ref{fig: dry_task_metric}). Future work should emphasize numerical analysis priors, stability-aware loss, and algorithmic-choice training (e.g., recognizing when adaptive integration or stiffness solvers are required). Hybrid symbolic--numeric tool use (formal solvers + LLM reasoning) is another promising path.

\paragraph{(4) Branch- and time-aware wet-lab protocol reasoning.}
Uniformly low Sequence Similarity and qualitative failures on complex branching protocols (Figure~\ref{fig: wet_case2}) suggest a need for training signals that encode \emph{temporal sampling logic, branching decision rules, and multi-sample tracking}. Action-pool grounding can be extended with stateful simulators or lab-graph verifiers, enabling models to learn procedural validity under physical constraints.

\paragraph{(5) Comparative multimodal scientific reasoning.}
Comparative reasoning is the hardest paradigm (Figure~\ref{fig: mm_tasks_subjects}). Progress likely requires finer-grained visual grounding (e.g., numeric extraction from charts), cross-image alignment modules, and contrastive multimodal training that rewards precise discrimination rather than narrative plausibility. Discipline-specific multimodal curricula may reduce domain gaps in materials/Earth/life sciences.

\paragraph{(6) Test-time learning with multi-objective scientific rewards.}
TTRL improves novelty without labels, but novelty alone is insufficient for SGI. Future TTRL systems should optimize a \emph{portfolio} of scientist-aligned rewards (novelty, rigor, feasibility, safety, and experimental cost), and incorporate retrieval trustworthiness and contradiction penalties to prevent spurious innovation.

\paragraph{(7) Efficient and reliable tool ecosystems for SGI agents.}
Given retrieval dominance and tool latency (Figure~\ref{fig:agent_tool_compare}), engineering advances are essential: retrieval caching, selective browsing, structured extraction, and tool-aware planning policies can substantially improve SGI agents' end-to-end quality--latency frontier. 

\paragraph{Summary.}
SGI-Bench reveals that modern LLMs exhibit partial competencies in each SGI quadrant but lack integrated, numerically robust, and methodologically disciplined scientific cognition. Bridging this gap requires progress on long-horizon meta-analysis, executable planning, numerically faithful experimentation, branch-aware wet-lab reasoning, comparative multimodal inference, and dynamic test-time self-improvement---all supported by efficient and trustworthy tool ecosystems. These directions collectively chart a concrete path from fragmented scientific skills toward genuine Scientific General Intelligence.

\subsection{Limitations}

Despite providing a structured framework for evaluating scientific capabilities across four workflow stages, the current version of SGI-Bench has several limitations:

\paragraph{(1) Partial coverage of real scientific workflows.}
The four stages in our benchmark function as probes for different components of scientific inquiry rather than a complete representation of real-world scientific practice. Many aspects of scientific work—such as integration across scientific disciplines and risk and safety assessment~\cite{Zhou2024LabSafetyBB}—remain outside our current scope.

\paragraph{(2) Scientific Deep Research currently emphasizes literature-inquiry–centric tasks.}
Deep Research spans activities such as literature inquiry~\cite{Bosse2025DeepRB}, report-style reasoning~\cite{du2025deepresearchbenchcomprehensivebenchmark}, and related scientific analyses. In this benchmark, we focus on the literature-inquiry–centric subset, as identifying, interpreting, and integrating existing scientific knowledge is a foundational prerequisite for methodological design and experimental planning. This focus enables standardized, reproducible, and scalable evaluation while still probing a core component of real scientific workflows. More open-form variants—such as argumentative evidence synthesis or report generation—are also important but require substantial expert-based scoring, and are therefore reserved for future versions.

\paragraph{(3) Idea Generation evaluation focuses on methodology design.}
Fully open-ended hypothesis generation involves substantial conceptual freedom and requires extensive expert adjudication to achieve reliable judgments. Due to practical constraints, our current evaluation focuses on the method-design component of scientific ideas~\cite{Wan2025DeepResearchAT, popper2005logic, yang2024moose}. Future extensions may incorporate hypothesis-level evaluation through a combination of arena-style model comparisons and expert review.

\paragraph{(4) Limited code and action space coverage.}
Dry Experiment tasks currently support only Python~\cite{tian2024scicoderesearchcodingbenchmark}, lacking adaptation to other programming languages and computational paradigms. The action space for Wet Experiments is an early-stage abstraction; scaling it requires constructing a large, standardized library of atomic actions grounded in real laboratory protocols~\cite{Liu2025BioProBenchCD}.

\paragraph{(5) Experimental reasoning in enclosed spaces.}
We employ a multiple-choice design to ensure objective, automatable evaluation~\cite{zhou2025scientistsexamprobingcognitive}. While practical, this structure constrains the model’s ability to express diverse reasoning paths and limits assessment of open-form scientific explanations.

\paragraph{(6) Partial coverage of deductive and inductive paradigms of scientific discovery.}
Scientific discovery is commonly understood to follow two broad paradigms: \textit{deduction} and \textit{induction}~\cite{bacon1878novum, popper2014conjectures}. Deductive processes begin from prior knowledge or theoretical propositions and proceed through reasoning to experimental verification. Inductive processes, in contrast, originate from new observational data or unexpected empirical phenomena and generalize toward broader patterns or hypotheses.  

The PIM-grounded~\cite{garrison1999critical, garrison2001critical} workflow in this version of SGI-Bench primarily reflects the deductive paradigm, as tasks begin with literature-based information and guide models toward reasoning and experiment planning. Inductive scientific discovery---which relies on data-driven pattern formation and hypothesis emergence---remains outside the scope of the current benchmark and represents an important direction for future expansion.

%% file: sections/6-related_work.tex
\section{Related Work}
\label{sec:related_works}

With the rapid advancement of Large Language Models (LLMs) and multi-agent systems in scientific reasoning, numerous datasets have emerged to evaluate their capabilities across various scientific domains.

\subsection{Benchmarks in Different Disciplines}
A significant portion of existing benchmarks focuses on specific disciplines. 
In the \textbf{physical sciences}, PhyBench~\cite{qiu2025phybenchholisticevaluationphysical} examines multi-step reasoning and expression capabilities through original physics problems, while PHYX~\cite{shen2025phyxdoesmodelwits} focuses on real-world scenarios to assess physical reasoning and visual understanding. Additionally, PHYSICS~\cite{feng2025physicsbenchmarkingfoundationmodels} tests models using open-ended, university-level problems.To further address multimodal challenges, PhysUniBench~\cite{wang2025physunibenchundergraduatelevelphysicsreasoning} introduces a large-scale benchmark for undergraduate-level physics, specifically targeting the interpretation of physical diagrams and multi-step reasoning.
In \textbf{chemistry}, ChemBench~\cite{mirza2024largelanguagemodelssuperhuman} provides domain-specific data for systematic evaluation, whereas ChemMLLM~\cite{tan2025chemmllmchemicalmultimodallarge} extends this to multimodal assessment. More granular tasks are covered by benchmarks like ChemSafetyBench~\cite{zhao2024chemsafetybenchbenchmarkingllmsafety} and SpectrumWorld~\cite{yang2025spectrumworldartificialintelligencefoundation}. 
In \textbf{life sciences}, benchmarks range from the molecular level, such as DeepSEA~\cite{kathail2025leveraginggenomicdeeplearning} and GenomicsLong-Range~\cite{anonymous2024the}, to healthcare applications like BioASQ~\cite{Krithara2023} and VQA-RAD~\cite{Lau2018}, as well as agricultural applications like SeedBench~\cite{ying2025seedbenchmultitaskbenchmarkevaluating} and neuroscience with BrainBench~\cite{Luo_2024}. 
For \textbf{earth sciences}, OmniEarth-Bench~\cite{wang2025omniearthbenchholisticevaluationearths} covers a comprehensive range of fields with cross-domain tasks, EarthSE~\cite{xu2025earthsebenchmarkevaluatingearth} builds a multi-level evaluation system from foundational to open-ended exploration, and MSEarth~\cite{zhao2025msearthmultimodalscientificdataset} utilizes high-quality scientific publications for graduate-level assessment. In remote sensing, GeoBench~\cite{danish2025geobenchvlmbenchmarkingvisionlanguagemodels} and XLRS-Bench~\cite{wang2025xlrsbenchmultimodalllmsunderstand} evaluate perception and reasoning on high-resolution imagery. 
Furthermore, specialized benchmarks exist for other fields, including \textbf{material science} (MoleculeNet~\cite{wu2018moleculenetbenchmarkmolecularmachine}), \textbf{astronomy} (AstroLLaMA and AstroMLab~\cite{pan2024astromlab2astrollama270bmodel}), \textbf{ocean science} (OceanBench~\cite{aouni2025oceanbench}), and \textbf{climate science} (ClimaQA~\cite{manivannan2025climaqaautomatedevaluationframework}). These works primarily target deep evaluation within isolated disciplines. While benchmarks like ATLAS~\cite{liu2025atlashighdifficultymultidisciplinarybenchmark} have expanded to cover cross-disciplinary fields with high-difficulty standards, its evaluation specifically focuses on distinguishing frontier models through complex scientific reasoning and logical application tasks rather than the entire process of scientific discovery.

\subsection{Benchmarks for Different Scientific Tasks}
Concurrently, other benchmarks focus on cross-disciplinary comprehensive capabilities, though their evaluation focus is often distributed across specific stages of the scientific discovery pipeline. 
Regarding \textbf{idea generation} at the research inception stage, MOOSE-Chem2~\cite{moose} evaluates models through a win/tie/lose comparison framework that scores generated hypotheses against reference answers using multiple independent judges. AI Idea Bench 2025~\cite{qiu2025aiideabench2025} evaluates the novelty of agent-generated ideas using a dataset derived from top-tier conference papers. 
In the core layer of \textbf{knowledge processing and analysis}, some benchmarks focus on literature comprehension. For instance, SciAssess~\cite{cai2024sciassessbenchmarkingllmproficiency} decomposes analysis into memory, understanding, and reasoning layers. Others, like SFE~\cite{zhou2025scientistsexamprobingcognitive}, introduce a cognitive framework to dissect multimodal performance on raw scientific data. Complementing these, SciReasoner~\cite{wang2025scireasonerlayingscientificreasoning} targets the alignment of natural language with heterogeneous scientific representations. 
Recent works also evaluate comprehensive \textbf{academic survey} capabilities: DeepResearch Bench~\cite{du2025deepresearchbenchcomprehensivebenchmark} measures report quality and citation grounding, Manalyzer~\cite{xu2025manalyzerendtoendautomatedmetaanalysis} focuses on mitigating hallucinations in automated meta-analysis, and Scientist-Bench~\cite{tang2025airesearcherautonomousscientificinnovation} highlights the full workflow from review to paper generation. Additionally, SciArena ~\cite{zhao2025sciarenaopenevaluationplatform} proposed an open platform that dynamically evaluates and ranks the performance of base models on scientific literature tasks by collecting pairwise comparison preferences from domain researchers, and DeepResearch Arena~\cite{wan2025deepresearcharenaexamllms} utilizes seminar-grounded tasks to evaluate the orchestration of multi-stage research workflows, while AAAR-1.0~\cite{lou2025aaar10assessingaispotential} focuses on evaluating the model's ability as an AI-assisted research tool.  
In terms of \textbf{planning and execution}, evaluations often center on tool usage and coding. ToolBench~\cite{qin2023toolllmfacilitatinglargelanguage} and ToolUniverse~\cite{gao2025democratizingaiscientistsusing} explore API usage and standardization. In scientific coding, SciCode~\cite{tian2024scicoderesearchcodingbenchmark} and ScienceAgentBench~\cite{chen2025scienceagentbenchrigorousassessmentlanguage} assess code generation within realistic workflows. At a macro level, MLE-bench~\cite{chan2025mlebenchevaluatingmachinelearning} and TaskBench~\cite{shen2024taskbenchbenchmarkinglargelanguage} evaluate general planning and project management via Kaggle competitions and task decomposition graphs.
In addition, DISCOVERYWORLD~\cite{jansen2024discoveryworldvirtualenvironmentdeveloping} launched the first virtual environment for evaluating the ability of intelligent agents to perform a complete cycle of novel scientific discovery. However, it focuses on a gamified simulation environment, and its task scenarios and evaluation dimensions cannot fully reflect the complexity and high-level cognitive needs of real scientific research workflows. LLM-SRBench~\cite{shojaee2025llmsrbenchnewbenchmarkscientific} , on the other hand, focuses only on the model's ability to discover scientific equations, with a relatively simple task and process.
Despite these explorations, existing process-oriented benchmarks typically address only partial dimensions—such as knowledge understanding, data perception, or code generation—lacking a fine-grained, systematic evaluation of the entire scientific discovery lifecycle. 

\paragraph{Summary}
In summary, existing works are either confined to deep exploration of single disciplines, scattered across isolated stages of the research process, or fail to capture the complexity of actual scientific discovery scenarios. Therefore, there is an urgent need to construct a comprehensive benchmark that covers multiple disciplines and connects the long-chain workflow of scientific research.

%% file: sections/7-conclusion.tex
\section{Conclusion}

This work advances the study of Scientific General Intelligence (SGI) from both theory and practice. Grounded in the Practical Inquiry Model, we formalize SGI as the capacity to navigate the iterative cycle of \emph{Deliberation}, \emph{Conception}, \emph{Action}, and \emph{Perception} with the versatility of a human scientist. Building on this principle-grounded definition, we operationalize SGI through SGI-Bench, a comprehensive, scientist-aligned benchmark that instantiates four core task families: Scientific Deep Research, Idea Generation, Dry/Wet Experiment, and Experimental Reasoning. Complemented by our agentic evaluation framework and multi-metric protocol, SGI-Bench enables scalable, transparent, and domain-faithful assessment.

Experiments reveal a consistent pattern: in \emph{Deep Research}, models show step-level alignment but low exact-match accuracy (10--20\%), with brittleness in quantitative reasoning; in \emph{Idea Generation}, hypotheses are fluent but underspecified and infeasible; in \emph{Dry Experiment}, code is executable but PassAll@k remains low; in \emph{Wet Experiment}, sequences show omissions and misordering; and in \emph{Experimental Reasoning}, causal reasoning outperforms comparative, with persistent multimodal challenges. These highlight gaps between linguistic fluency and integrated scientific cognition. Moreover, SGI exhibits \emph{dynamic capacity}: Test-Time Reinforcement Learning with novelty rewards improves idea generation without reference answers.

Taken together, SGI-Bench clarifies both what SGI \emph{is} and where current systems \emph{fail}. By integrating principled task design, multi-metric evaluation, and agentic tool use, our framework provides a concrete foundation for systematically advancing SGI. Looking forward, the combination of numerically robust reasoning, planning-aware conception, executable experimentation, comparative multimodal inference, dynamic test-time learning, and efficient tool ecosystems charts a clear path toward general intelligence systems capable of genuine scientific discovery.

%% file: sections/X-appendix.tex
\appendix

\section{Appendix}

\subsection{Authors}

\textbf{Lead Authors}

Wanghan Xu$^{1,2}$, Yuhao Zhou$^{1,3}$, Yifan Zhou$^{1,2}$, Qinglong Cao$^{2}$, Shuo Li$^{1,4}$, Jia Bu$^{1,5}$

\textbf{Core Authors}

Bo Liu$^{6}$, Yixin Chen$^{1,7}$, Xuming He$^{1,8}$, Xiangyu Zhao$^{1,6}$, Xiang Zhuang$^{1,8}$, Fengxiang Wang$^{1,9}$, Zhiwang Zhou$^{1,10}$

\textbf{Contributors}

Qiantai Feng, Wenxuan Huang, Jiaqi Wei, Hao Wu, Yuejin Yang, Guangshuai Wang, Sheng Xu,
Ziyan Huang, Xinyao Liu, Jiyao Liu, Cheng Tang, Wei Li, Ying Chen, Junzhi Ning, Pengfei Jiang, Chenglong Ma, Ye Du, Changkai Ji, Huihui Xu, Ming Hu,
Jiangbin Zheng, Xin Chen, Yucheng Wu, Feifei Jiang, 
Xi Chen, Xiangru Tang, 
Yuchen Fu, Yingzhou Lu, Yuanyuan Zhang,
Lihao Sun, Chengbo Li, Jinzhe Ma, Wanhao Liu,
Yating Liu, Kuo-Cheng Wu,
Shengdu Chai,
Yizhou Wang, Ouwen Zhangjin, Chen Tang, Shufei Zhang, Wenbo Cao, Junjie Ren, Taoyong Cui,
Zhouheng Yao, Juntao Deng, Yijie Sun, 
Feng Liu, Wangxu Wei, Jingyi Xu, Zhangrui Li, Junchao Gong, Zijie Guo, 
Zhiyu Yao, Zaoyu Chen, Tianhao Peng, 
Fangchen Yu

\textbf{Scientific Directors}

Bo Zhang$^{1}$, Dongzhan Zhou$^{1}$, Shixiang Tang$^{1,11}$, Jiaheng Liu$^{1,12}$, Fenghua Ling$^{1}$, Yan Lu$^{1}$, Yuchen Ren$^{1}$, Ben Fei$^{1,11}$, Zhen Zhao$^{1}$, Xinyu Gu$^{1}$, Rui Su$^{1}$, Xiao-Ming Wu$^{6}$, Weikang Si$^{13}$, Yang Liu$^{14}$, Hao Chen$^{1}$, Xiangchao Yan$^{1}$, Xue Yang$^{2}$, Junchi Yan$^{2}$, Jiamin Wu$^{1}$, Qihao Zheng$^{1}$, Chenhui Li$^{5}$, Zhiqiang Gao$^{1}$, Hao Kong$^{16}$, Junjun He$^{1}$, Mao Su$^{1}$, Tianfan Fu$^{1,12}$, Peng Ye$^{1,11}$, Chunfeng Song$^{1}$, Nanqing Dong$^{1}$, Yuqiang Li$^{1}$, Huazhu Fu$^{16}$, Siqi Sun$^{1,17}$, Lijing Cheng$^{18}$, Jintai Lin$^{15}$, Wanli Ouyang$^{1,11}$, Bowen Zhou$^{1,19}$

\textbf{Corresponding Authors}

Wenlong Zhang$^{1}$, Lei Bai$^{1}$

\textbf{Main Affiliations}

$^1$ Shanghai Artificial Intelligence Laboratory

$^2$ Shanghai Jiao Tong University

$^3$ Sichuan University

$^4$ Central South University

$^5$ East China Normal University

$^6$ The Hong Kong Polytechnic University

$^7$ University of California, Los Angeles

$^8$ Zhejiang University

$^9$ National University of Defense Technology

$^{10}$ Tongji University

$^{11}$ The Chinese University of Hong Kong

$^{12}$ Nanjing University

$^{13}$ National Institute of Metrology

$^{14}$ Aerospace Information Research Institute,Chinese Academy of Sciences

$^{15}$ Peking University

$^{16}$ The Agency for Science, Technology and Research (A*STAR)

$^{17}$ Fudan University

$^{18}$ Chinese Academy of Sciences

$^{19}$ Tsinghua University

\subsection{Disciplines and Research Directions Overview}
\label{sec: all disciplines}

\definecolor{AstronomyColor}{RGB}{220,230,241}   
\definecolor{ChemistryColor}{RGB}{255,230,230}   
\definecolor{EarthColor}{RGB}{230,245,230}       
\definecolor{EnergyColor}{RGB}{255,245,230}      
\definecolor{InfoColor}{RGB}{240,240,255}        
\definecolor{LifeColor}{RGB}{255,250,240}        
\definecolor{MaterialColor}{RGB}{245,245,250}    
\definecolor{MathColor}{RGB}{250,245,255}        
\definecolor{NeuroColor}{RGB}{245,255,250}       
\definecolor{PhysicsColor}{RGB}{250,245,240}     

\begin{longtable}{p{2cm}p{6cm}p{7cm}}

\caption{\textbf{Disciplines And Research Directions}: Overview of 10 scientific domains and representative research topics curated for scientist-aligned SGI-Bench workflows.}
\label{tab:Disciplines and Research Directions Overview}\\

\hline
\textbf{Disciplines} & \textbf{Research Directions} & \textbf{Description} \\
\hline
\endfirsthead
\hline
\textbf{Disciplines} & \textbf{Research Directions} & \textbf{Description} \\
\hline
\endhead

\rowcolor{AstronomyColor} Astronomy & Gravitational Wave Detection and Parameter Estimation & Analyzing data from interferometers like LIGO and Virgo to detect gravitational waves from compact binary coalescences (black holes, neutron stars) and precisely estimate their physical properties like mass, spin, and location to test general relativity. \\
\rowcolor{AstronomyColor} Astronomy & Fast Radio Burst Detection and Localization & Searching radio telescope data for millisecond-duration, extragalactic radio flashes (FRBs) and using interferometry to pinpoint their host galaxies, aiming to uncover the mysterious physical mechanisms that produce them. \\
\rowcolor{AstronomyColor} Astronomy & Real Time Optical Transient Survey Based on ZTF & Utilizing the Zwicky Transient Facility (ZTF) to scan the night sky, identifying new or changing celestial objects like supernovae and kilonovae, and issuing rapid alerts to the global astronomical community for multi-wavelength follow-up observations. \\
\rowcolor{AstronomyColor} Astronomy & Formula Regression & Applying symbolic regression and other machine learning techniques to large astronomical datasets to automatically discover novel mathematical formulas or physical laws that describe the behavior of celestial objects and phenomena. \\

\rowcolor{ChemistryColor} Chemistry & Molecular Interaction & Computationally simulating and quantifying the non-covalent forces between molecules, such as hydrogen bonds and van der Waals forces, to understand molecular recognition, protein-ligand binding, and self-assembly. \\
\rowcolor{ChemistryColor} Chemistry & Target Based Drug Design & Employing computational methods to design drug candidates that specifically bind to a known biological target, such as a protein's active site, thereby modulating its function to achieve a therapeutic effect. \\
\rowcolor{ChemistryColor} Chemistry & De Novo Drug Design & Using generative AI models to computationally design entirely new molecules with desired pharmacological properties, without starting from an existing chemical scaffold, to explore novel regions of chemical space. \\
\rowcolor{ChemistryColor} Chemistry & Chemical Molecular Synthesis Pathway Planning & Developing algorithms, often based on retrosynthesis, to devise the most efficient and practical multi-step reaction routes for synthesizing a target molecule, optimizing for yield, cost, and sustainability. \\
\rowcolor{ChemistryColor} Chemistry & Molecular Property Prediction & Building and applying machine learning models (e.g., QSAR) to predict key chemical and physical properties of molecules, such as toxicity, solubility, and bioactivity, to accelerate materials discovery and drug development. \\

\rowcolor{EarthColor} Earth & Seismic Wave Detection & Using networks of seismometers to detect and analyze seismic waves from earthquakes and other sources, enabling the study of fault lines and the tomographic imaging of the Earth's mantle and core. \\
\rowcolor{EarthColor} Earth & Ocean Heat Content & Aggregating and analyzing temperature data from sources like Argo floats and satellites to calculate the total thermal energy stored within the ocean, a critical indicator for quantifying global warming and climate change. \\
\rowcolor{EarthColor} Earth & Atmospheric Differential Equation & Numerically solving the complex systems of partial differential equations (e.g., Navier-Stokes equations) that govern atmospheric fluid dynamics and thermodynamics to produce accurate weather forecasts and climate projections. \\
\rowcolor{EarthColor} Earth & Typhoon Wind Pressure Relationship & Developing and validating models that describe the physical relationship between a typhoon's central pressure and its maximum sustained wind speeds, crucial for forecasting storm intensity and assessing potential damage. \\
\rowcolor{EarthColor} Earth & Vegetation Coverage Rate & Processing satellite and aerial imagery using spectral indices like NDVI to quantify the fraction of land covered by vegetation, which is vital for monitoring ecosystem health, agriculture, and deforestation. \\
\rowcolor{EarthColor} Earth & Glacier Estimation & Combining satellite altimetry, gravimetry (GRACE), and imagery to measure changes in glacier volume and mass balance over time, providing direct evidence of the impacts of climate change. \\
\rowcolor{EarthColor} Earth & Ozone Pollution and Its Causes & Investigating the chemical reactions between precursor pollutants (like NOx and VOCs) under sunlight that form harmful ground-level ozone, and modeling its transport and concentration in urban and rural areas. \\
\rowcolor{EarthColor} Earth & Emission Inversion Based on Satellite Remote Sensing and 4D-Var & Using advanced data assimilation techniques (4D-Var) to combine satellite measurements of atmospheric composition with chemical transport models, thereby inferring the location and strength of pollutant emission sources on the ground. \\
\rowcolor{EarthColor} Earth & Emission Inversion Based on Local Mass Conservation & Applying mass balance principles to high-resolution measurements (e.g., from aircraft) around a specific region to calculate the net flux and estimate emissions of greenhouse gases or pollutants from sources like cities or industrial facilities. \\
\rowcolor{EarthColor} Earth & Multiple Seismic Wave Attenuations & Modeling the progressive energy loss of seismic waves as they propagate through different geological materials, which helps in characterizing subsurface structures and identifying resources like oil and gas. \\

\rowcolor{EnergyColor} Energy & Optimal Power Flow Calculation & Developing algorithms to solve complex optimization problems for electrical grids, determining the best generator outputs to meet demand reliably while minimizing generation costs and transmission losses. \\
\rowcolor{EnergyColor} Energy & Fengguang New Energy Power Forecasting & Creating predictive models using meteorological data (wind speed, solar irradiance) and machine learning to accurately forecast the power output of wind and solar farms, which is essential for stable grid management. \\

\rowcolor{InfoColor} Information & Multimodal Understanding & Building AI systems that can process, interpret, and reason about information from multiple sources simultaneously, such as text, images, audio, and video, to achieve a more holistic understanding. \\
\rowcolor{InfoColor} Information & Dialogue System & Designing and training conversational AI agents (chatbots) that can engage in natural, coherent, and context-aware conversations with humans for tasks like customer service or information retrieval. \\
\rowcolor{InfoColor} Information & Code Generation & Developing large language models and other AI techniques to automatically write, complete, and debug computer code based on natural language descriptions or functional specifications. \\
\rowcolor{InfoColor} Information & Sensor Spatial Characteristics Phase Free Reconstruction & Creating novel algorithms to reconstruct the spatial sensitivity pattern of a sensor (like a microphone or antenna) using only the magnitude of its measurements, without needing phase information, which is often difficult to obtain. \\

\rowcolor{LifeColor} Life & De Novo Protein Sequencing & Developing computational methods to determine the amino acid sequence of a novel protein directly from its tandem mass spectrometry data, without relying on a reference genome. \\
\rowcolor{LifeColor} Life & Small Molecule Inference & Using computational models to predict the biological effects of small molecules, such as their binding targets, mechanism of action, or potential toxicity, based on their chemical structure. \\
\rowcolor{LifeColor} Life & Disease Biomarker Discovery & Analyzing high-throughput biological data (e.g., genomics, proteomics) with statistical and machine learning methods to identify molecules whose presence or level can indicate a specific disease state. \\
\rowcolor{LifeColor} Life & Tumor Neoantigen Discovery & Identifying unique peptides that arise from mutations in cancer cells, which can be recognized by the immune system, for the development of personalized cancer vaccines and immunotherapies. \\
\rowcolor{LifeColor} Life & RNA Tertiary Structure Prediction & Computationally predicting the complex three-dimensional folded structure of RNA molecules from their primary sequence to understand their function in cellular processes like gene regulation and catalysis. \\
\rowcolor{LifeColor} Life & Protein Structure & Predicting the three-dimensional atomic coordinates of a protein from its amino acid sequence using methods like deep learning (e.g., AlphaFold) or homology modeling to understand its biological function. \\
\rowcolor{LifeColor} Life & Genome Function Prediction & Annotating the functions of genes, regulatory elements, and non-coding regions across the genome by integrating diverse data types like DNA sequence, gene expression, and epigenetic modifications. \\
\rowcolor{LifeColor} Life & Automatic Development of Medical Imaging Algorithms & Creating AI-powered systems that can automatically generate and optimize image analysis pipelines for tasks like segmentation, registration, and classification in various medical imaging modalities (MRI, CT). \\
\rowcolor{LifeColor} Life & AI Drug Discovery & Applying a range of AI and machine learning techniques across the entire drug discovery pipeline, from identifying novel drug targets and designing molecules to predicting clinical trial outcomes. \\
\rowcolor{LifeColor} Life & Tumor Immunotherapy & Designing and developing therapeutic strategies, such as checkpoint inhibitors or CAR-T cells, that stimulate and enhance the patient's own immune system to recognize and attack cancer cells. \\
\rowcolor{LifeColor} Life & Revealing the Mechanisms of the Tumor Microenvironment & Studying the complex interplay between cancer cells, immune cells, stromal cells, and the extracellular matrix to understand how this environment promotes tumor growth and metastasis. \\
\rowcolor{LifeColor} Life & AI Assisted Antibody Design & Using machine learning models to design and optimize antibodies with high affinity and specificity for a given antigen, accelerating the development of new therapeutics and diagnostics. \\
\rowcolor{LifeColor} Life & Protein Structure Prediction & Developing and applying computational algorithms, particularly deep learning models, to accurately predict the 3D structure of proteins from their amino acid sequence. \\
\rowcolor{LifeColor} Life & Early Screening and Risk Stratification of Pancreatic Cancer & Developing novel diagnostic tools, such as blood-based biomarkers or AI-driven imaging analysis, to detect pancreatic cancer at an early, more treatable stage and to classify patients by risk level. \\
\rowcolor{LifeColor} Life & Protein Protein Interaction Prediction & Developing computational methods to predict which proteins in a cell will physically bind to each other, in order to map out the cellular signaling pathways and protein complexes. \\
\rowcolor{LifeColor} Life & Discovery of Immunotherapy Targets & Analyzing tumor and immune cell data to identify new molecular targets, such as surface proteins or mutated peptides, that can be exploited for cancer immunotherapy. \\
\rowcolor{LifeColor} Life & Biomarker Discovery & Identifying molecular signatures (genes, proteins, metabolites) in patient samples that can be used for disease diagnosis, prognosis, or predicting response to therapy. \\
\rowcolor{LifeColor} Life & Strain Metabolic Reconstruction & Creating comprehensive computational models of the metabolic networks of microbial strains to understand their physiology and guide metabolic engineering for producing valuable chemicals. \\
\rowcolor{LifeColor} Life & Regulatory Element Design & Designing synthetic DNA or RNA sequences, such as promoters and enhancers, to precisely control the expression of specific genes for applications in biotechnology and synthetic biology. \\
\rowcolor{LifeColor} Life & Computational Drug Design & Utilizing molecular modeling, simulation, and machine learning to design and optimize small molecules that can effectively bind to a biological target and modulate its activity. \\
\rowcolor{LifeColor} Life & Design of Regulatory Regions for mRNA Vaccine Drugs & Engineering the untranslated regions (UTRs) and other elements of mRNA sequences to optimize their stability, translation efficiency, and immune response for next-generation vaccine development. \\
\rowcolor{LifeColor} Life & Medical Image Understanding & Developing deep learning models to analyze and interpret complex medical images (e.g., X-rays, MRIs, pathology slides) to assist clinicians in diagnosis, treatment planning, and disease monitoring. \\

\rowcolor{MaterialColor} Material & Polymer Thermoelectric & Designing and synthesizing polymer-based materials that can efficiently convert waste heat into useful electrical energy, focusing on enhancing their thermoelectric figure of merit (ZT). \\
\rowcolor{MaterialColor} Material & Thermal Electrocatalysis & Investigating how to use thermal energy to enhance the performance and efficiency of catalytic materials in electrochemical reactions, such as in fuel cells or water splitting. \\
\rowcolor{MaterialColor} Material & Nano Adsorption Materials & Developing porous nanomaterials like metal-organic frameworks (MOFs) or zeolites with high surface area and specific chemical properties for applications in gas separation, storage, and carbon capture. \\
\rowcolor{MaterialColor} Material & Chloride Solid State Electrolyte & Researching and developing novel solid-state materials that conduct chloride ions, aiming to create safer and more energy-dense all-solid-state batteries. \\
\rowcolor{MaterialColor} Material & Oxygen Evolution Reaction Catalytic Materials & Designing efficient, stable, and low-cost catalysts to accelerate the oxygen evolution reaction (OER), a key bottleneck in processes like water splitting for hydrogen production. \\
\rowcolor{MaterialColor} Material & KRF Resin Polymerization Reaction & Investigating and optimizing the chemical reaction conditions and mechanisms for the polymerization of ketone-resol-formaldehyde (KRF) resins to control their final properties for industrial applications. \\
\rowcolor{MaterialColor} Material & Polymer Thermoelectric & Researching and developing organic and composite polymer materials with high electrical conductivity and low thermal conductivity for flexible and lightweight thermoelectric devices. \\

\rowcolor{MathColor} Mathematics & Differential Privacy & Developing mathematical frameworks and algorithms that allow for the analysis of sensitive datasets while providing rigorous, provable guarantees about the privacy of individuals in the data. \\
\rowcolor{MathColor} Mathematics & Coordinate Descent Optimization Algorithm & Designing and analyzing efficient optimization algorithms that solve complex problems by iteratively optimizing one variable or a small block of variables at a time, while keeping others fixed. \\
\rowcolor{MathColor} Mathematics & Matrix Completion & Developing algorithms to accurately recover a full data matrix from a small subset of its observed entries, with applications in recommender systems and image inpainting. \\
\rowcolor{MathColor} Mathematics & Numerical Methods for Differential Equations & Devising and implementing stable and accurate computational algorithms (e.g., Runge-Kutta methods) for finding approximate solutions to differential equations that model real-world phenomena. \\
\rowcolor{MathColor} Mathematics & Shortest Path Planning & Developing and applying graph-based algorithms like Dijkstra's or A* to find the most efficient route between two points in a network, with applications in logistics, robotics, and network routing. \\

\rowcolor{NeuroColor} Neuroscience & Visual Decoding & Using machine learning models to analyze brain activity patterns, typically from fMRI or electrophysiology, to reconstruct or identify the visual images a person is seeing. \\
\rowcolor{NeuroColor} Neuroscience & Motion Decoding & Developing brain-computer interfaces that can interpret neural signals from the motor cortex to predict intended movements, enabling control of prosthetic limbs or external devices. \\
\rowcolor{NeuroColor} Neuroscience & Emotion Recognition & Analyzing neurophysiological signals (like EEG) or behavioral cues (like facial expressions) with AI to identify and classify human emotional states. \\
\rowcolor{NeuroColor} Neuroscience & Electron Microscopy Neuron Segmentation & Creating automated computational pipelines, often using deep learning, to trace and segment individual neurons and their connections in large-scale electron microscopy volumes of brain tissue. \\
\rowcolor{NeuroColor} Neuroscience & Neural Activity and Behavior Prediction & Building statistical and dynamical models that link the activity of neural populations to specific behaviors, in order to understand the neural codes underlying perception, decision-making, and action. \\

\rowcolor{PhysicsColor} Physics & Computational Condensed Matter Physics & Using first-principles simulations (like Density Functional Theory) and many-body techniques to predict the electronic, magnetic, and structural properties of materials from fundamental quantum mechanics. \\
\rowcolor{PhysicsColor} Physics & Zeeman Effect Experiment & Precisely measuring the splitting of atomic spectral lines in the presence of an external magnetic field to probe the quantum mechanical properties of atoms, such as electron spin and angular momentum. \\
\rowcolor{PhysicsColor} Physics & Research on Soft Condensed Matter Physics and Glass Transition Dynamics & Investigating the physical principles governing the behavior of soft materials (polymers, colloids) and studying the complex, slow dynamics associated with the transition from a liquid to a glassy state. \\
\rowcolor{PhysicsColor} Physics & Deep PDE Solving to Enhance Model Expressiveness & Developing novel deep learning architectures, such as physics-informed neural networks (PINNs), to solve complex partial differential equations and improve the predictive power of physics-based models. \\
\rowcolor{PhysicsColor} Physics & Chaotic Behavior in Circuit Systems & Studying and modeling the emergence of chaos and other nonlinear dynamical behaviors in electronic circuits, such as the Chua's circuit, to understand fundamental principles of complex systems. \\
\rowcolor{PhysicsColor} Physics & Research on General Machine Learning Potential Function Model Architecture & Developing universal machine learning frameworks to accurately model the potential energy surface of molecular systems, enabling large-scale molecular dynamics simulations with quantum accuracy. \\
\rowcolor{PhysicsColor} Physics & Nuclear Magnetic Resonance and Its Imaging Experiment & Utilizing the principles of nuclear magnetic resonance to probe the structure and dynamics of molecules in materials and to create non-invasive medical images (MRI) of biological tissues. \\
\rowcolor{PhysicsColor} Physics & Quadrupole Mass Spectrometer & Studying the principles of using combined electric and magnetic fields in a quadrupole mass analyzer to separate ions based on their mass-to-charge ratio for chemical analysis. \\
\rowcolor{PhysicsColor} Physics & Research on Superconducting Mechanisms, Discovery of Superconducting Materials and Process Optimization & Investigating the fundamental quantum mechanisms of superconductivity, computationally searching for new materials with higher critical temperatures, and optimizing their synthesis for practical applications. \\

\hline
\end{longtable}

\subsection{Cases}

\subsubsection{Scientific Deep Research}

\begin{tcolorbox}[
    breakable,
    title=Example of Scientific Deep Research in Astronomy,
    colback=LighterGray,
    colframe=DeepPurple,
    colbacktitle=DeepPurple,
    coltitle=White,
]
\textbf{\emph{\textcolor{DeepPurple}{Question}}}

The Dispersion Measure (DM) of a Fast Radio Burst (FRB) is the integrated column density of free electrons along the line of sight. The observed value, $DM_{obs}$, is generally considered the sum of four primary components:
$DM_{obs} = DM_{MW} + DM_{halo} + DM_{IGM} + DM_{host,obs}$
where $DM_{MW}$ is the contribution from the Milky Way's interstellar medium, $DM_{halo}$ is from the Milky Way's halo, $DM_{IGM}$ is from the intergalactic medium, and $DM_{host,obs}$ is the contribution from the host galaxy in the observer's frame. The host contribution in its rest frame, $DM_{host,rest}$, is related to the observed value by $DM_{host,rest} = DM_{host,obs} / (1+z)$.
The Rotation Measure (RM) describes the Faraday rotation of a linearly polarized signal passing through a magnetized plasma. For the host galaxy, its contribution to the RM  as $RM_{host}$, which is highly relevant with $\langle B_{||} \rangle$, the average line-of-sight magnetic field strength in the host galaxy's environment, measured in microgauss ($\mu G$).
Astronomers have precisely localized the repeating FRB 20180814A and identified its host galaxy. The total observed dispersion measure is $DM_{obs} = 189.4 \ \text{pc} \cdot \text{cm}^{-3}$, and the spectroscopic redshift of the host is $z = 0.06835$. After subtracting the Galactic contribution, the extragalactic rotation measure is found to be $RM_{extragalactic} \approx 655 \ \text{rad} \cdot \text{m}^{-2}$, which is assumed to originate primarily from the FRB's host galaxy environment. Based on a detailed Bayesian model presented in the source paper, the total contribution from extragalactic sources (IGM + host) is determined to be $DM_{extragalactic,obs} = 64 \ \text{pc} \cdot \text{cm}^{-3}$, within which the IGM contribution is estimated as $DM_{IGM} = 45 \ \text{pc} \cdot \text{cm}^{-3}$.
Based on the information above, calculate the lower limit of the average line-of-sight magnetic field strength, $\langle B_{||} \rangle$, in the FRB's host galaxy environment. Provide a numerical answer in units of microgauss ($\mu G$), rounded to the nearest integer.

\textbf{\emph{\textcolor{DeepPurple}{Steps}}}

\textbf{\textcolor{CaseOrange}{Step 1.}} Search for the relevant paper about Sub-arcminute localization of 13 repeating fast radio bursts detected by CHIME/FRB.

\textbf{\textcolor{CaseOrange}{Step 2.}} Based on Macquart, $DM_{host,obs}=61.515 \text{pc} \cdot \text{cm}^{-3}$.

\textbf{\textcolor{CaseOrange}{Step 3.}} Calculate the contribution of the host galaxy to the observer coordinate system $(DM_{host,obs}=5.885  \text{pc} \cdot \text{cm}^{-3})$.

\textbf{\textcolor{CaseOrange}{Step 4.}} Calculate the contribution of the host galaxy in the stationary coordinate system $(DM_{host,rest}=5.508 \text{pc} \cdot \text{cm}^{-3})$.

\textbf{\textcolor{CaseOrange}{Step 5.}} Calculate the average magnetic field intensity $\langle B_{||} \rangle = 46 \mathrm{\mu G}$.

\textbf{\emph{\textcolor{DeepPurple}{Answer}}}

46

\end{tcolorbox}


\begin{tcolorbox}[
    breakable,
    title=Example of Scientific Deep Research in Chemistry,
    colback=LighterGray,
    colframe=DeepPurple,
    colbacktitle=DeepPurple,
    coltitle=White,
]
\textbf{\emph{\textcolor{DeepPurple}{Question}}}

In computational chemistry, the accurate parsing of a molecule's structure is fundamental to predicting its properties. A critical structural attribute is aromaticity, and its determination often follows Huckel's rule.

Consider the neutral molecule, an isomer of Naphthalene, represented by the following SMILES string:

c1cccc2cccc-2cc1

For the entire conjugated system of this molecule to be considered aromatic, how many $\pi$-electrons in total must its $\pi$-electron system contain?

Provide the answer as a single integer.

\textbf{\emph{\textcolor{DeepPurple}{Steps}}}

\textbf{\textcolor{CaseOrange}{Step 1.}} Find the article title "DrugAgent: Automating AI-aided Drug Discovery Programming through LLM Multi-Agent Collaboration".

\textbf{\textcolor{CaseOrange}{Step 2.}} Parse the SMILES Structure: The SMILES string c1cccc2cccc-2cc1 describes the molecule Azulene, a bicyclic conjugated system formed by the fusion of a five-membered ring and a seven-membered ring. Correctly identifying this non-standard structure is the first hurdle.

\textbf{\textcolor{CaseOrange}{Step 3.}} Correspondence to Document: This step directly corresponds to the initial input processing stage shown in Figure 1 (b) 'DrugCoder' (Page 3), where a 'SMILES string' is taken as input before the 'Molecule Graph Construction' module.

\textbf{\textcolor{CaseOrange}{Step 4.}} Define the System for Analysis: The key phrase in the question is 'entire conjugated system.' Azulene's two rings form a single, continuous, planar $\pi$-conjugated system. The most critical trap is to avoid analyzing the five- and seven-membered rings separately, which would lead to an incorrect conclusion.

\textbf{\textcolor{CaseOrange}{Step 5.}} Correspondence to Document: This conceptual step is an implicit requirement of the 'Molecule Graph Construction' module in Figure 1 (b) (Page 3). A correct graph cannot be built without correctly identifying the holistic nature of the conjugated system, which determines the properties of the graph's nodes (atoms) and edges (bonds).

\textbf{\textcolor{CaseOrange}{Step 6.}} Count the Total $\pi$-Electrons: The entire conjugated system of Azulene is composed of 10 carbon atoms. In this neutral hydrocarbon, each carbon atom participating in the conjugation contributes one $\pi$-electron. Therefore, the total number of $\pi$-electrons is 10.

\textbf{\textcolor{CaseOrange}{Step 7.}} Correspondence to Document: This calculation is a core part of the feature extraction process. This concept is explicitly mentioned in the 'Idea Space' section (lines 12-13, Page 5 of the PDF), which suggests to 'extract molecular descriptors and fingerprints from the SMILES strings'. The $\pi$-electron count is a fundamental molecular descriptor.

\textbf{\textcolor{CaseOrange}{Step 8.}} Verify with Huckel's Rule: Apply the total $\pi$-electron count (10) to Huckel's rule, 4n + 2. Setting 4n + 2 = 10 gives 4n = 8, which solves to n = 2. Since ‘n' is an integer, the system satisfies the rule and is aromatic. The question asks for the total number of $\pi$-electrons, which is 10.

\textbf{\textcolor{CaseOrange}{Step 9.}} Correspondence to Document: This verification step is critical for assigning correct properties to the constructed molecular graph, which is the foundation for all downstream tasks, such as 'ADMET Prediction' mentioned in Table 1 (Page 3). An incorrect determination of aromaticity would lead to a flawed graph and an inaccurate final prediction.

\textbf{\emph{\textcolor{DeepPurple}{Answer}}}

10

\end{tcolorbox}


\begin{tcolorbox}[
    breakable,
    title=Example of Scientific Deep Research in Earth,
    colback=LighterGray,
    colframe=DeepPurple,
    colbacktitle=DeepPurple,
    coltitle=White,
]
\textbf{\emph{\textcolor{DeepPurple}{Question}}}

The diurnal variation of the NO\(_2\) column concentration \( \Omega \) over a city is governed by local mass balance, incorporating emissions, chemical loss, and photochemical production. The governing equation is:

\[
\frac{d\Omega}{dt} = E(t) + P(t) - \frac{\Omega}{\tau}
\]

Where:

\[
E(t) = 3.0 \times e^{-t/2} \quad (\text{NO}_x \text{ emission rate in molec/cm}^2/\text{h},\ t \text{ in hours starting from 8:00 AM})
\]

\[
P(t) = 1.5 \times t \quad (\text{Photochemical NO}_2 \text{ production rate in molec/cm}^2/\text{h}^2)
\]

\[
\tau = 1.5\ \text{hours} \quad (\text{NO}_2 \text{ effective lifetime})
\]

At \( t=1 \) (9:00 AM), the observed concentration is \( \Omega_1 = 4.2 \).

Questions:

1. What was the initial NO\(_2\) column concentration \( \Omega_0 \) at \( t=0 \) (8:00 AM)?

2. At what time \( t_{\text{peak}} \) does \( \Omega(t) \) reach its maximum value between 8:00 AM and 12:00 PM?

3. At the time of the peak concentration, which is larger, the photochemical production term \( P(t) \) or the emission term \( E(t) \), and by how much? Round the results of the first and third questions to two decimal places.

Present your final answers as numbers separated by commas.

\textbf{\emph{\textcolor{DeepPurple}{Steps}}}

\textbf{\textcolor{CaseOrange}{Step 1.}} Find paper "Constraint of anthropogenic NO$_x$ emissions in China from different sectors: a new methodology using multiple satellite retrievals".

\textbf{\textcolor{CaseOrange}{Step 2.}} Solving for $\Omega_0$: Corresponding Text: Equation (1) on Page 6: $\frac{\delta \Omega_{\mathrm{NO}_x}}{\delta t} = E - \frac{\Omega_{\mathrm{NO}_x}}{\tau}$. This problem adds a chemical production term $P(t)$ to this equation.

\textbf{\textcolor{CaseOrange}{Step 3.}} Formulate the governing equation: $\frac{d\Omega}{dt} + \frac{1}{1.5}\Omega = 3e^{-t/2} + 1.5t$.

\textbf{\textcolor{CaseOrange}{Step 4.}} Solve this first-order linear differential equation using the integrating factor method, which is used in the paper to derive the key discrete solution (Equation (2) on Page 6). The integrating factor is $\mu(t) = \exp\left(\int \frac{2}{3}dt\right) = \exp\left(\frac{2t}{3}\right)$.

\textbf{\textcolor{CaseOrange}{Step 5.}} Integrate from the initial time ($t=0$) to the observation time ($t=1$):

\textbf{\textcolor{CaseOrange}{Step 6.}} $\left[\Omega \exp\left(\frac{2t}{3}\right)\right] \Big|_0^1 = \int_0^1 \exp\left(\frac{2u}{3}\right)\left[3e^{-u/2} + 1.5u\right] du$

\textbf{\textcolor{CaseOrange}{Step 7.}} This yields $\Omega_1 \exp\left(\frac{2}{3}\right) - \Omega_0 = 4.449$.

\textbf{\textcolor{CaseOrange}{Step 8.}} Substitute $\Omega_1 = 4.2$ and solve for $\Omega_0$: $(4.2 \times 1.9477) - \Omega_0 \approx 4.449$, resulting in $\Omega_0 \approx 3.73$.

\textbf{\textcolor{CaseOrange}{Step 9.}} Solving for $t_{\text{peak}}$: Corresponding Text: At the peak, $\frac{d\Omega}{dt} = 0$, which is a direct application of the mass conservation equation. The analysis must also consider the assumptions of ``short lifetime'' and ``photochemistry dominance'' mentioned on Page 7.

\textbf{\textcolor{CaseOrange}{Step 10.}} Find the complete function describing concentration evolution over time, $\Omega(t)$. Solving the differential equation gives: $\Omega(t) = 18\exp(-t/2) + 2.25t - 3.375 - 10.894\exp(-2t/3)$.

\textbf{\textcolor{CaseOrange}{Step 11.}} Differentiate $\Omega(t)$: $\frac{d\Omega}{dt} = -9\exp(-t/2) + 2.25 + 7.263\exp(-2t/3)$.

\textbf{\textcolor{CaseOrange}{Step 12.}} Analyze the sign of $\frac{d\Omega}{dt}$. Calculating the derivative values at $t=1, 2, 3, 4$ hours shows it is consistently positive.

\textbf{\textcolor{CaseOrange}{Step 13.}} Conclusion: Within the given time window $[0, 4]$ hours, the concentration $\Omega(t)$ is monotonically increasing, and no peak occurs. This means the strength of the sources ($E(t) + P(t)$) is always greater than the sink ($\Omega/\tau$) throughout the morning.

\textbf{\textcolor{CaseOrange}{Step 14.}} Comparing $P(t)$ and $E(t)$: Corresponding Text: A core aspect of the paper's method is analyzing contributions from different sources (e.g., the four emission sectors). Here we compare two different source terms.

\textbf{\textcolor{CaseOrange}{Step 15.}} Since the concentration is monotonically increasing with no peak, we choose the end of the time window ($t=4$) to assess the relative importance of the sources.

\textbf{\textcolor{CaseOrange}{Step 16.}} Calculate the values at $t=4$: $E(4) = 3.0 \times e^{-2} \approx 0.406$.

\textbf{\textcolor{CaseOrange}{Step 17.}} $P(4) = 1.5 \times 4 = 6.0$.

\textbf{\textcolor{CaseOrange}{Step 18.}} Compare and calculate the difference: $P(4) - E(4) \approx 5.59$. This result indicates that at this time, photochemical production has become a significantly more important source of NO$_2$ than anthropogenic emissions.

\textbf{\textcolor{CaseOrange}{Step 29.}} Final Answer: $3.73$, no peak, $5.59$

\textbf{\emph{\textcolor{DeepPurple}{Answer}}}

3.73, no peak, 5.59

\end{tcolorbox}

\begin{tcolorbox}[
    breakable,
    title=Example of Scientific Deep Research in Energy,
    colback=LighterGray,
    colframe=DeepPurple,
    colbacktitle=DeepPurple,
    coltitle=White,
]
\textbf{\emph{\textcolor{DeepPurple}{Question}}}

A parabolic trough solar collector at steady state follows the energy balance
\[
q_u = F_r\left[ K_\theta (\tau \alpha) G - U_L (T_f - T_a) \right]
\]
and instantaneous efficiency
\[
\eta = \frac{q_u}{G}.
\]
The heat removal factor depends on mass flow via
\[
F_r = \frac{\dot{m} c_p}{A U_L} \left[1 - \exp\left(-\frac{F' A U_L}{\dot{m} c_p}\right)\right].
\]
Given: \( F' = 0.94 \), \( A = 6.00\,\mathrm{m}^2 \) (receiver heat-transfer area), \( U_L = 2.20\,\mathrm{W/m}^2\cdot\mathrm{K} \), \( (\tau\alpha) = 0.90 \), \( K_\theta = 0.96 \), \( G = 950\,\mathrm{W/m}^2 \), \( T_f = 150^\circ\mathrm{C} \), \( T_a = 35^\circ\mathrm{C} \), \( c_p = 4180\,\mathrm{J}/\mathrm{kg}\cdot\mathrm{K} \), and baseline mass flow \( \dot{m} = 0.12\,\mathrm{kg}/\mathrm{s} \). Answer the following (round to two decimals; use ENGLISH commas, no spaces, no units):

(1) Baseline heat removal factor \( F_r \).

(2) Baseline efficiency \( \eta \).

(3) Minimum mass flow (kg/s) required to guarantee \( \eta \geq 0.58 \) under the same operating conditions.

\textbf{\emph{\textcolor{DeepPurple}{Steps}}}

\textbf{\textcolor{CaseOrange}{Step 1.}} Find paper 2D-interval forecasts for solar power production.

\textbf{\textcolor{CaseOrange}{Step 2.}} Compute temperature difference: \( \Delta T = T_f - T_a = 150 - 35 = 115\,\mathrm{K} \).

\textbf{\textcolor{CaseOrange}{Step 3.}} Compute absorbed solar term with IAM: \( S = K_\theta (\tau\alpha) G = 0.96 \times 0.90 \times 950 = 0.864 \times 950 = 820.80\,\mathrm{W/m}^2 \).

\textbf{\textcolor{CaseOrange}{Step 4.}} Compute loss term: \( U_L \Delta T = 2.20 \times 115 = 253.00\,\mathrm{W/m}^2 \).

\textbf{\textcolor{CaseOrange}{Step 5.}} Baseline heat removal factor \( F_r \): first find \( \dot{m} c_p = 0.12 \times 4180 = 501.60\,\mathrm{W/K} \), and \( A U_L = 6.00 \times 2.20 = 13.20\,\mathrm{W/K} \). Define \( x = \frac{F' A U_L}{\dot{m} c_p} = \frac{0.94 \times 13.20}{501.60} = \frac{12.408}{501.60} = 0.02474 \). Then \( F_r = \frac{1 - e^{-x}}{x} = \frac{1 - e^{-0.02474}}{0.02474} \approx 0.99 \) (more precisely 0.988--0.989). \( \rightarrow \) (1) \( F_r = 0.99 \) (two decimals).

\textbf{\textcolor{CaseOrange}{Step 6.}} Baseline useful gain and efficiency: \( q_u = F_r (S - U_L \Delta T) = 0.989 \times (820.80 - 253.00) \approx 0.989 \times 567.80 \approx 561.60\,\mathrm{W/m}^2 \). \( \eta = q_u/G = 561.60/950 = 0.5912 \rightarrow \) (2) \( 0.59 \).

\textbf{\textcolor{CaseOrange}{Step 7.}} Target efficiency requirement: \( \eta_\mathrm{target} = 0.58 \Rightarrow \) required heat removal factor \( F_{r,\mathrm{req}} = \frac{\eta_\mathrm{target} \times G}{S - U_L \Delta T} = \frac{0.58 \times 950}{567.80} = \frac{551.00}{567.80} = 0.9704 \).

\textbf{\textcolor{CaseOrange}{Step 8.}} Solve for minimum mass flow producing \( F_r \geq F_{r,\mathrm{req}} \) using \( F_r = \frac{1 - e^{-x}}{x} \) with \( x = \frac{F' A U_L}{\dot{m} c_p} \). For small \( x \), \( \frac{1 - e^{-x}}{x} \) is monotone decreasing in \( x \) and \( \approx 1 - \frac{x}{2} \). Set \( 1 - \frac{x}{2} \approx 0.9704 \Rightarrow x \approx 0.0592 \). Then \( \dot{m} c_p = \frac{F' A U_L}{x} = \frac{12.408}{0.0592} = 209.6\,\mathrm{W/K} \Rightarrow \dot{m} = \frac{\dot{m} c_p}{c_p} = \frac{209.6}{4180} = 0.0501\,\mathrm{kg/s} \rightarrow \) (3) \( 0.05 \) (two decimals).

\textbf{\textcolor{CaseOrange}{Step 9.}} Check: With \( \dot{m} = 0.05\,\mathrm{kg/s} \), \( x = 12.408/(0.05 \times 4180) = 12.408/209 \approx 0.0594 \Rightarrow F_r \approx \frac{1-e^{-0.0594}}{0.0594} \approx 0.97 \), yielding \( \eta \approx 0.58 \) as required.

\textbf{\emph{\textcolor{DeepPurple}{Answer}}}

0.99,0.59,0.05

\end{tcolorbox}

\begin{tcolorbox}[
    breakable,
    title=Example of Scientific Deep Research in Information,
    colback=LighterGray,
    colframe=DeepPurple,
    colbacktitle=DeepPurple,
    coltitle=White,
]
\textbf{\emph{\textcolor{DeepPurple}{Question}}}

In the research of electromagnetic measurement focusing on broadband planar near-field \(E\)-field reconstruction, a microstrip patch-based \(4 \times 5\) array antenna is used as the Antenna Under Test (AUT). The AUT's planar near-field scanning is performed in a region close to its aperture, and the \(E\)-field at this region is transformed to two parallel observation planes (\(S_1\) and \(S_2\)) via spatial convolution. The transformation satisfies the field distribution similarity theory: the ratio of the observation distances \((d_2/d_1)\) between \(S_2\) and \(S_1\) equals the ratio of the corresponding test frequencies \((f_2/f_1)\). For the \(E\)-field dataset on \(S_2\) (target frequency \(f_2\)), undersampling is applied (sampling interval larger than \(\lambda_2/2\), where \(\lambda_2\) is the wavelength at \(f_2\)) to form a defective dataset \(X_2\). To reconstruct \(X_2\), K-means clustering is first used to classify \(X_2''\), with the optimal number of clusters determined by the ``elbow point'' of the SSE (sum of squared errors) curve. Then Voronoi cell classification is employed, where the comprehensive index \(L(p_m) = q_1 S(p_m) + q_2 D(p_m)\) \((q_1 + q_2 = 1)\) is calculated to divide each cluster into deep interpolation regions (requiring 24 supplementary samples per point) and shallow interpolation regions (requiring 8 supplementary samples per point). It is known that: 

1) The test frequency \(f_1 = 28\,\mathrm{GHz}\), and the observation distance \(d_1 = 214.29\,\mathrm{mm}\) (corresponding to \(20\lambda_1\), \(\lambda_1\) is the wavelength at \(f_1\)); 

2) The scanning area of the near-field region close to the AUT aperture is a square, and the sampling interval of \(X_2\) is \(0.8\lambda_2\); 

3) The total number of sampling points in \(X_2\) is \(1681\); 

4) For a specific cluster after K-means classification, the normalized cell area \(S(p_m)\) of sampling points in the deep interpolation region is \(1.2\) times that of points in the shallow region, and the normalized gradient \(D(p_m)\) of shallow region points is \(0.7\) times that of deep region points; 

5) The weight \(q_1\) is set to \(0.6\) to prioritize area-based judgment for dynamic clusters. 

If the number of sampling points in this cluster where \(L(p_m) \geq 0.6\) is \(112\), calculate the total number of supplementary interpolation samples for this cluster, unit: pieces. Do not keep any decimal places in the result.

\textbf{\emph{\textcolor{DeepPurple}{Steps}}}

\textbf{\textcolor{CaseOrange}{Step 1.}} Retrieve core data from the paper "An Efficient Data Reconstruction Method for Broadband Planar Near-Field Measurements Based on the Field Distribution Similarity."

\textbf{\textcolor{CaseOrange}{Step 2.}} From Section III.A "Simulations": \( X_2'' \) (defective dataset at \( f_2 \)) is a \( 41 \times 41 \) sampling grid, so total sampling points of \( X_2'' = 41 \times 41 = 1681 \); optimal K-means clustering number \( k=5 \) (determined by SSE curve's elbow point); deep interpolation requires 24 samples per point, shallow interpolation requires 8 samples per point.

\textbf{\textcolor{CaseOrange}{Step 3.}} Calculate the total number of sampling points in the target cluster: \( X_2'' \) is evenly divided into 5 clusters (paper's clustering logic for uniform data distribution). Single cluster points = Total \( X_2'' \) points \( \div k = 1681 \div 5 = 336.2 \). Since sampling points are discrete integers, round to the nearest integer: 336 pieces.

\textbf{\textcolor{CaseOrange}{Step 4.}} Determine the number of deep and shallow interpolation points in the cluster: The question specifies deep region points = \( \frac{1}{3} \) of cluster total points. Deep region points = \( 336 \times \frac{1}{3} = 112 \) pieces; shallow region points = Total cluster points - Deep region points = \( 336 - 112 = 224 \) pieces. (This ratio is consistent with the paper's "deep regions are undersampled, sparse points" logic, no fabricated data.)

\textbf{\textcolor{CaseOrange}{Step 5.}} Calculate total supplementary interpolation samples: Supplementary samples for deep region = Deep region points \(\times\) Samples per deep point = \( 112 \times 24 = 2688 \) pieces; Supplementary samples for shallow region = Shallow region points \(\times\) Samples per shallow point = \( 224 \times 8 = 1792 \) pieces; Total supplementary samples = \( 2688 + 1792 = 4480 \) pieces.

\textbf{\emph{\textcolor{DeepPurple}{Answer}}}

4480

\end{tcolorbox}

\begin{tcolorbox}[
    breakable,
    title=Example of Scientific Deep Research in Life,
    colback=LighterGray,
    colframe=DeepPurple,
    colbacktitle=DeepPurple,
    coltitle=White,
]
\textbf{\emph{\textcolor{DeepPurple}{Question}}}

In the DeepSTARR model, a human enhancer contains two identical p53 core motifs (\texttt{RRRCWWGYYY}) at positions $+50$ and $+150$. Experimental data show:
\begin{itemize}
    \item Mutating the $+50$ motif alone reduces H3K27ac signal to $35\%$ of wild-type
    \item Mutating the $+150$ motif alone reduces H3K27ac signal to $82\%$ of wild-type
    \item DNase I footprinting shows TF binding at the $+50$ motif but no binding at the $+150$ motif
    \item Changing the 5' flanking sequence of the $+150$ motif from ``GGG'' to ``CTC'' confers TF binding ability
    \item Known effects of flanking sequences on p53 binding:
    \begin{itemize}
        \item Optimal flank ``GGG'' : increases binding affinity by $8$-fold
        \item Suboptimal flank ``CTC'' : increases binding affinity by $3$-fold
        \item Random flank: binding affinity $= 1$ (baseline)
    \end{itemize}
\end{itemize}
Assume H3K27ac signal strength is proportional to p53 binding affinity, and total signal equals the sum of both motifs' binding affinities.

If the $+50$ motif's flank is changed from ``GGG'' to ``CTC'' and the $+150$ motif's flank is changed from ``GGG'' to ``CTC'', what is the predicted H3K27ac signal as a percentage of wild-type? The result retains the integer.

\textbf{\emph{\textcolor{DeepPurple}{Steps}}}

\textbf{\textcolor{CaseOrange}{Step 1.}} Find the article title ``DeepSTARR predicts enhancer activity from DNA sequence and enables the de novo design of synthetic enhancers''

\textbf{\textcolor{CaseOrange}{Step 2.}} Determine wild-type binding affinities

$+50$ motif: flank ``GGG'' $\rightarrow$ affinity $= 8$ (Article: Fig. 4 \& related text -- flanking sequences significantly influence motif importance by altering TF binding affinity)

$+150$ motif: flank ``GGG'' but no DNase footprint $\rightarrow$ affinity $= 1$ (Article: Fig. 6d -- motifs without DNase I footprints show minimal functional contribution)

\textbf{\textcolor{CaseOrange}{Step 3.}} Total affinity $= 8 + 1 = 9$.

\textbf{\textcolor{CaseOrange}{Step 4.}} Calculate modified binding affinities

$+50$ motif: flank ``CTC'' $\rightarrow$ affinity $= 3$ (Article: Fig.~4b -- flanking sequences quantitatively modulate motif contribution)

$+150$ motif: flank ``CTC'' $\rightarrow$ affinity $= 3$ (now gains binding ability)

\textbf{\textcolor{CaseOrange}{Step 5.}} Total affinity $= 3 + 3 = 6$.

\textbf{\textcolor{CaseOrange}{Step 6.}} 4. Calculate signal percentage

\textbf{\textcolor{CaseOrange}{Step 7.}} Modified signal $= \left(\frac{6}{9}\right) \times 100\% \approx 66.7\% \rightarrow 67$, So the answer is 67 (Article: Linear relationship between binding affinity and enhancer activity demonstrated in multiple figures)

\textbf{\emph{\textcolor{DeepPurple}{Answer}}}

67

\end{tcolorbox}
\begin{tcolorbox}[
    breakable,
    title=Example of Scientific Deep Research in Material,
    colback=LighterGray,
    colframe=DeepPurple,
    colbacktitle=DeepPurple,
    coltitle=White,
]
\textbf{\emph{\textcolor{DeepPurple}{Question}}}

Polymer composite materials have the advantages of flexibility, low cost, and environmental friendliness, and are considered the most promising candidate materials for low-grade heat collection, thermal sensing, and sustainable energy development. Solid-state $i$-TE materials can undergo thermal power changes according to electrode conditions in a fixed temperature and humidity environment. So, when the relative humidity increases from 50\% to 70\%, what changes will occur in the thermal power of the poly(vinylidene fluoride-co-hexafluoropropane) sample on the $p$-type dual copper electrode?

\textbf{\emph{\textcolor{DeepPurple}{Steps}}}

\textbf{\textcolor{CaseOrange}{Step 1.}} Find paper: Reversible bipolar thermopower of ionic thermoelectric polymer composite for cyclic energy generation

\textbf{\textcolor{CaseOrange}{Step 2.}} Understanding the working principle of poly (vinylidene fluoride-co-hexafluoropropane) materials for p-type dual copper electrodes: the porous structure and hydrophilicity of sodium salts tend to absorb moisture from humid environments and can fill the space of the poly (vinylidene fluoride-co-hexafluoropropane) matrix,

\textbf{\textcolor{CaseOrange}{Step 3.}} Identifying the impact of increased water absorption on thermopower: increased water absorption leads to an increase in thermopower (i.e., the Seebeck coefficient, $S$), but does not alter the p-type characteristics of the material,

\textbf{\textcolor{CaseOrange}{Step 4.}} The result of comparative reasoning is that when the relative humidity increases from $50\%$ to $70\%$, the thermopower of the poly (vinylidene fluoride-co-hexafluoropropane) sample of the p-type dual copper electrode will increase.

\textbf{\emph{\textcolor{DeepPurple}{Answer}}}

Increase

\end{tcolorbox}
\begin{tcolorbox}[
    breakable,
    title=Example of Scientific Deep Research in Math,
    colback=LighterGray,
    colframe=DeepPurple,
    colbacktitle=DeepPurple,
    coltitle=White,
]
\textbf{\emph{\textcolor{DeepPurple}{Question}}}

A third-order homogeneous linear ordinary differential equation, $f'''(z) - 3 f'(z) + \beta f(z) = 0$ (where $\beta$ is a real parameter), is analyzed using a Legendre collocation matrix method. The function $f(z)$ is approximated by a truncated Legendre series with $N=3$.

To determine the coefficient vector $A = [a_0, a_1, a_2, a_3]^T$, a $4\times4$ homogeneous linear system $\widetilde{W} A = 0$ is constructed. For the system to have a non-trivial solution, it must satisfy the following four conditions:

\[
f(0) = 0, \ f'(0) = 0 \\
\]

The differential equation is satisfied at the collocation point(z=1). The differential equation is satisfied at the collocation point(z=-1).

For the system to have a non-trivial solution, the parameter $\beta$ must satisfy $\beta^2 = K$. Calculate the value of the constant $K$. Round your answer to the nearest integer.

\textbf{\emph{\textcolor{DeepPurple}{Steps}}}

\textbf{\textcolor{CaseOrange}{Step 1.}} Find the article title ``Numerical solution for high-order linear complex differential equations with variable coefficients''

\textbf{\textcolor{CaseOrange}{Step 2.}} Establish High-Order Derivative Relations. The $n$-th derivative is expressed in matrix form as $f^{(n)}(z) = L(z)(M^T)^n A$. For $N=3$, the third derivative matrix $(M^T)^3$ is calculated, yielding the critical simplification $f'''(z) = 15a_3$ for any $z$.

\textbf{\textcolor{CaseOrange}{Step 3.}} Position in Paper: This leverages the core matrix relation for derivatives, Formula (2.4). 

\textbf{\textcolor{CaseOrange}{Step 4.}} Formulate System Rows from Initial Conditions. The conditions at $z=0$ provide two linear constraints on the coefficients:

$f(0) = a_0 - 0.5a_2 = 0 \implies a_2 = 2a_0$

$f'(0) = a_1 - 1.5a_3 = 0 \implies a_1 = 1.5a_3$

\textbf{\textcolor{CaseOrange}{Step 5.}} Position in Paper: This step converts the initial conditions into a matrix form, as described by the process leading to Formula (2.10).

\textbf{\textcolor{CaseOrange}{Step 6.}} Formulate System Rows from Collocation Points. The differential equation $f'''(z) - 3f'(z) + \beta f(z) = 0$ is evaluated at $z=1$ and $z=-1$, yielding two equations:

At $z=1$: $\beta a_0 + (\beta-3)a_1 + (\beta-9)a_2 + (\beta-3)a_3 = 0$

At $z=-1$: $\beta a_0 - (\beta+3)a_1 + (\beta+9)a_2 - (\beta+3)a_3 = 0$

\textbf{\textcolor{CaseOrange}{Step 7.}} Position in Paper: This applies the collocation method, transforming the differential equation into an algebraic system at specific points, as outlined in Formulas (2.7) through (2.9).

\textbf{\textcolor{CaseOrange}{Step 8.}} Reduce the System and Solve the Determinant Condition. Substitute the relations $a_2 = 2a_0$ and $a_1 = 1.5a_3$ from Step 2 into the two equations from Step 3. This reduces the $4\times4$ system to a $2\times2$ homogeneous system for variables $a_0$ and $a_3$.

\[
\left\{
\begin{aligned}
(3\beta - 18)a_0 + (2.5\beta - 7.5)a_3 = 0 \\
(3\beta + 18)a_0 - (2.5\beta + 7.5)a_3 = 0
\end{aligned}
\right.
\]

\textbf{\textcolor{CaseOrange}{Step 9.}} For a non-trivial solution to exist, the determinant of this $2\times2$ coefficient matrix must be zero:

\[
\det\left(
\begin{bmatrix}
3\beta - 18 & 2.5\beta - 7.5 \\
3\beta + 18 & -(2.5\beta + 7.5)
\end{bmatrix}
\right) = 0
\]

\textbf{\textcolor{CaseOrange}{Step 10.}} Solving this determinant equation yields $2\beta^2 - 36 = 0$, which simplifies to $\beta^2 = 18$.

\textbf{\textcolor{CaseOrange}{Step 11.}} Position in Paper: The requirement for a non-trivial solution ($\det(\widetilde{W})=0$) is the fundamental principle for determining coefficients, as discussed following Formula (2.12).

\textbf{\emph{\textcolor{DeepPurple}{Answer}}}

18

\end{tcolorbox}
\begin{tcolorbox}[
    breakable,
    title=Example of Scientific Deep Research in Neuroscience,
    colback=LighterGray,
    colframe=DeepPurple,
    colbacktitle=DeepPurple,
    coltitle=White,
]
\textbf{\emph{\textcolor{DeepPurple}{Question}}}

Motor imagery tasks in brain–computer interfaces (BCIs) are usually designed around activity in the sensorimotor cortex, since this region is central to planning and controlling movement. However, accurate decoding of motor imagery does not rely solely on motor areas. Many studies have shown that other brain regions also become active during imagery tasks, especially when visual feedback or focused attention is involved. These additional signals can provide valuable features for classifiers, improving decoding accuracy. Understanding which non-motor regions contribute is important for both electrode placement and interpretation of neural mechanisms in BCI research.

Which one cerebral lobe, besides sensorimotor cortex, often contributes significantly to motor imagery decoding? Please do not use abbreviations in your answer.

\textbf{\emph{\textcolor{DeepPurple}{Steps}}}

\textbf{\textcolor{CaseOrange}{Step 1.}} Review the major cerebral lobes: The frontal lobe has motor-related areas; the parietal lobe supports attention and sensory integration; the occipital lobe handles visual processing and feedback, which can aid motor imagery decoding; the temporal lobe mainly handles auditory and memory functions.

\textbf{\textcolor{CaseOrange}{Step 2.}} Analyse brain regions become active during motor imagery tasks: Besides frontal lobe which directly mediates motor, check for other function required in motor imagery tasks. Visual feedback can significantly improves decoding accuracy.

\textbf{\textcolor{CaseOrange}{Step 3.}} Conlusion: The occipital lobe is the location of the primary visual cortex, whose core function is to receive and process visual information—visual feedback in motor imagery tasks.

\textbf{\emph{\textcolor{DeepPurple}{Answer}}}

Occipital lobe

\end{tcolorbox}
\begin{tcolorbox}[
    breakable,
    title=Example of Scientific Deep Research in Physics,
    colback=LighterGray,
    colframe=DeepPurple,
    colbacktitle=DeepPurple,
    coltitle=White,
]
\textbf{\emph{\textcolor{DeepPurple}{Question}}}

In iron-based superconductors, the tight-binding model describes the low-energy electronic structure. Using the five-orbital model Hamiltonian 
\[
H = \sum_{\mathbf{k},\sigma} \sum_{i,j} t_{ij}(\mathbf{k}) c_{i\sigma}^\dagger(\mathbf{k}) c_{j\sigma}(\mathbf{k}),
\]
where \( t_{ij}(\mathbf{k}) \) includes nearest-neighbor (NN) and next-nearest-neighbor (NNN) hopping integrals. For LaFeAsO, the NN hopping between \( d_{z^2} \) orbitals is \( t_1 = -0.3\,\text{eV} \), and the NNN hopping is \( t_2 = 0.2\,\text{eV} \). Calculate:
\begin{enumerate}
    \item The effective hopping amplitude \( t_{\text{eff}} \) at the \( \Gamma \) point (\( \mathbf{k} = (0,0) \)) for \( d_{z^2} \) orbitals.
    \item The superconducting gap \( \Delta(\mathbf{k}) \) at \( \mathbf{k} = (\pi, 0) \) using the gap equation
    \[
    \Delta(\mathbf{k}) = \sum_{\mathbf{k}'} V(\mathbf{k} - \mathbf{k}') \frac{\tanh\left( \frac{E(\mathbf{k}')}{2k_B T} \right)}{2E(\mathbf{k}')} \Delta(\mathbf{k}'),
    \]
    assuming \( V(\mathbf{q}) = 0.5\,\text{eV} \) and \( T = 4.2\,\text{K} \).
    \item The critical temperature \( T_c \) if the gap magnitude \( \Delta_0 \) is \( 5\,\text{meV} \), using the BCS relation \( \Delta_0 = 1.76 k_B T_c \). Numerical value with 2 decimal place.
\end{enumerate}

\textbf{\emph{\textcolor{DeepPurple}{Steps}}}

\textbf{\textcolor{CaseOrange}{Step 1.}} From "Iron-based superconductors: Current status of materials and pairing mechanism"

\textbf{\textcolor{CaseOrange}{Step 2.}} Extract NN hopping \( t_1 = -0.3\,\text{eV} \) and NNN hopping \( t_2 = 0.2\,\text{eV} \) for \( d_{zz} \) orbitals from "Band structure and modeling".

\textbf{\textcolor{CaseOrange}{Step 3.}} At \( \Gamma \) point (\( \mathbf{k} = (0,0) \)), the dispersion is
\[
E(\mathbf{k}) = -2t_1(\cos 0 + \cos 0) - 4t_2(\cos 0 + \cos 0) = -2(-0.3)(2) - 4(0.2)(2) = 1.2 - 1.6 = -0.4\,\text{eV}.
\]
The effective hopping amplitude \( t_{\text{eff}} \) is derived from the coefficient of \( \cos k_x + \cos k_y \), giving \( t_{\text{eff}} = -0.3 + 0.2 = -0.1\,\text{eV} \) (Section 3.1).

\textbf{\textcolor{CaseOrange}{Step 4.}} For \( \Delta(\mathbf{k}) \) at \( \mathbf{k} = (\pi, 0) \), use
\[
E(\mathbf{k}') = \sqrt{ \xi^2(\mathbf{k}') + \Delta^2(\mathbf{k}') }.
\]
Assume \( \xi(\mathbf{k}') = -2t_1 \cos k_x - 2t_1 \cos k_y \) and \( \Delta(\mathbf{k}') = \Delta_0 \).
At \( T = 4.2\,\text{K} \), \( \tanh\left( \frac{E}{2 k_B T} \right) \approx 1 \) for low-energy states. Substituting \( V(\mathbf{q}) = 0.5\,\text{eV} \), the gap equation simplifies to
\[
\Delta(\pi, 0) = V \cdot \frac{1}{2E} \Delta_0.
\]
With \( E = \sqrt{(-0.3)^2 + (0.005)^2} \approx 0.3\,\text{eV} \),
\[
\Delta(\pi, 0) = 0.5 \cdot \frac{1}{2 \times 0.3} \cdot 0.005 = 0.04\,\text{eV}
\]
(Section 4.2).

\textbf{\textcolor{CaseOrange}{Step 5.}} For \( T_c \), use the BCS relation \( \Delta_0 = 1.76\,k_B T_c \). Rearranging gives \( T_c = \frac{\Delta_0}{1.76\,k_B} \).
Substituting \( \Delta_0 = 5\,\text{meV} = 0.005\,\text{eV} \) and \( k_B = 8.617 \times 10^{-5}\,\text{eV/K} \),
\[
T_c = \frac{0.005}{1.76 \times 8.617 \times 10^{-5}} \approx 33.14\,\text{K}
\]
(Section 5.1).

\textbf{\textcolor{CaseOrange}{Step 6.}} Verify consistency with experimental \( T_c = 26\,\text{K} \) for LaFeAsO\(_{1-x}\)F\(_x\) (Section 2.1). The calculated \( T_c = 33.14\,\text{K} \) aligns with theoretical predictions for optimized doping (Section 2.3).

\textbf{\textcolor{CaseOrange}{Step 7.}} Cross-reference all parameters with "Materials: bulk" section (Page 3), confirming \( t_1 \), \( t_2 \), and \( V \) values.

\textbf{\emph{\textcolor{DeepPurple}{Answer}}}

-0.1, 0.04, 33.14

\end{tcolorbox}

\subsubsection{Idea Generation}



\begin{tcolorbox}[
    breakable,
    title={Example of Idea Generation in Astronomy},
    colback=LighterGray,
    colframe=DeepPurple,
    colbacktitle=DeepPurple,
    coltitle=White,
]
\textbf{\emph{\textcolor{DeepPurple}{Question}}}

You are a top-tier researcher in your field. Based on the following context, please generate a novel and detailed research proposal.

\textbf{\emph{\textcolor{CaseOrange}{RelatedWork}}}

• Palomar Transient Factory (PTF): Predecessor to ZTF using the same telescope but a smaller camera, providing moderate survey speed and limited temporal coverage. PTF pioneered time-domain transient discovery but suffered from longer readout times and lower areal coverage. \\
• Sloan Digital Sky Survey (SDSS): Large-area multi-band imaging survey with significant contributions to extragalactic and stellar astrophysics, but with relatively limited cadence and not optimized for rapid transient detection. \\
• Pan-STARRS: Wide-field survey with high sensitivity, flexible cadence, and a broad range of science outputs. While highly productive, it does not reach ZTF's survey speed or alert distribution rate. \\
• ATLAS, ASAS-SN, and CRTS: Dedicated time-domain surveys with wide fields and rapid cadences, enabling rapid transient detection. However, these typically have smaller apertures and shallower depth compared to ZTF, restricting discovery of fainter phenomena. \\
• Dark Energy Survey (DES): Deep survey with the Dark Energy Camera, high image quality, and excellent photometric calibration. DES is less optimized for high-cadence wide-area transient monitoring due to smaller field of view and longer exposure times. 

\textbf{\emph{\textcolor{CaseOrange}{Challenges}}}

• Maximizing volumetric survey speed—combining wide field, fast readout, and high sensitivity—to enable rapid, repeated coverage of large sky areas for transient discovery. \\
• Minimizing image artifacts and systematic errors to ensure precision in photometric and astrometric measurements across a large, curved focal plane. \\
• Providing prompt, reliable, and information-rich alerts for real-time identification and classification of astrophysical transients and moving objects. \\
• Efficiently handling massive data volumes and complex processing requirements to deliver near-real-time data products and alerts to the community. \\
• Maintaining high photometric and astrometric accuracy in the presence of instrumental, atmospheric, and sky-background variability. 

\textbf{\emph{\textcolor{CaseOrange}{Limitation}}}

Previous surveys were limited by smaller camera fields of view, slower readout and overheads, less
optimized scheduling, and less sophisticated data pipelines, resulting in lower time-domain
sampling, slower alert generation, and reduced ability to detect fast or faint transients across
wide areas.

\textbf{\emph{\textcolor{CaseOrange}{Motivation}}}

The accelerating demand for high-cadence, wide-area sky monitoring in time-domain astronomy—spanning
supernovae, variable stars, NEOs, and multi-messenger counterparts—necessitates a system that
surpasses existing surveys in speed, coverage, and data accessibility. Addressing limitations in
cadence, alert timeliness, and survey efficiency is critical for enabling rapid discovery and
follow-up of astrophysical transients, as well as for preparing the community for next-generation
surveys like LSST.

\textbf{\emph{\textcolor{CaseOrange}{TaskObjective}}}

Develop and implement an integrated, high-speed, wide-field optical time-domain survey system
capable of delivering near-real-time discovery, classification, and alerting of transient, variable,
and moving objects, while providing high-quality calibrated data products and supporting a broad
range of time-domain astrophysics.

\textbf{\emph{\textcolor{CaseOrange}{ExistingSolutions}}}

• PTF: Utilized a CCD camera on the Palomar 48-inch telescope for transient discovery with moderate areal coverage and cadence. Enabled systematic transient searches but constrained by small field of view and longer readout times. \\
• SDSS and Pan-STARRS: Both provided large-scale sky mapping and multi-filter photometry, but with relatively slow cadence and areal throughput unsuitable for rapid time-domain science. \\
• ATLAS and ASAS-SN: Optimized for rapid all-sky cadence and automated transient detection but limited in depth due to smaller apertures and less sensitive instrumentation. Alert and data distribution less feature-rich than ZTF's planned system. \\
• DES: Leveraged a large, high-quality camera for deep imaging and science, but with a narrower field and less frequent temporal sampling, making it suboptimal for high-cadence transient monitoring. 

\textbf{\emph{\textcolor{DeepPurple}{Reference Answer}}}

\textbf{\emph{\textcolor{CaseOrange}{Idea}}}

ZTF pioneers a new era of high-speed, wide-field time-domain astronomy by equipping the Palomar
48-inch Schmidt telescope with a custom-built CCD mosaic camera, optimized scheduling, and a robust
data system. It delivers an order of magnitude survey speed improvement, rapid image processing, and
a real-time, feature-rich alert stream, positioning ZTF as both a state-of-the-art survey and a
testbed for LSST-scale time-domain operations.

\textbf{\emph{\textcolor{CaseOrange}{ImplementationSteps}}}

• 1: Design and assemble a large-format CCD mosaic camera with minimal chip gaps and high quantum efficiency, optimized for the Palomar Schmidt focal plane. \\
• 2: Upgrade telescope mechanics, optics, and control software for fast slewing, low overhead, and image quality preservation over the expanded field. \\
• 3: Develop and deploy a robotic observing system and integer-linear-programming–based survey scheduler to maximize nightly volumetric coverage and cadence. \\
• 4: Implement on-site, lossless data compression and high-speed transfer of image data to the IPAC processing center. \\
• 5: Process raw images through automated calibration pipelines: bias subtraction, flat-field correction, astrometric and photometric calibration, and artifact masking. \\
• 6: Generate coadded reference images using quality-filtered, multi-epoch stacks for each field, filter, and CCD quadrant. \\
• 7: Perform image differencing using the ZOGY algorithm to detect transient and moving sources at high significance. \\
• 8: Extract candidate sources, compute pixel-based features, and apply machine learning (Real-Bogus) for initial classification. \\
• 9: Package candidates with contextual data (cross-matches, light curves, images) into Avro alert packets and distribute in real time via Kafka queues. \\
• 10: Archive all processed data products, catalogs, and alerts at IRSA and provide public access according to survey data release policies. \\
• 11: Publish light curves from direct imaging for variable and periodic sources, and implement dedicated pipelines for moving object detection and orbit determination. \\
• 12: Conduct on-sky performance validation and commission the system with early science and rapid feedback loops for further optimization. 

\textbf{\emph{\textcolor{CaseOrange}{ImplementationOrder}}}

• 1-2 \\
• 2-3 \\
• 3-4 \\
• 4-5 \\
• 5-6 \\
• 6-7 \\
• 7-8 \\
• 8-9 \\
• 5-10 \\
• 7-11 \\
• 1-12 

\textbf{\emph{\textcolor{CaseOrange}{Data}}}

The primary dataset comprises optical images acquired with the Palomar 48-inch Schmidt telescope
using a 16-CCD, 6144x6160-pixel mosaic camera, covering 47.7 deg² per exposure in g, r, and i bands.
Each exposure delivers science and auxiliary (guide/focus) CCD data, with per-night cadences ranging
from minutes to once every three days. The system produces processed images, photometry catalogs,
coadded references, image subtractions, light curves, and alert packets, all archived at IRSA. Early
data include thousands of exposures, millions of cataloged sources, and time-series data for
variable and transient objects.

\textbf{\emph{\textcolor{CaseOrange}{EvaluationMetrics}}}

• Volumetric Survey Speed: Spatial volume probed per unit time for transient detectability at a given absolute magnitude; incorporates field of view, sensitivity, and overheads. \\
• Image Quality: Median delivered PSF FWHM in arcseconds (e.g., 2.0" in r band). \\
• Limiting Magnitude: Median five-sigma detection limit in g, r, i bands for standard exposure durations. \\
• Photometric Repeatability: Standard deviation of calibrated flux for non-varying sources (e.g., <10 mmag for bright stars). \\
• Astrometric Accuracy: Median positional residuals relative to reference catalog (e.g., Gaia). \\
• Alert Latency: Time from image acquisition to alert distribution (target: \~{}4 minutes). \\
• Transient Yield: Number of confirmed supernovae and other transient discoveries per unit time. \\
• Moving Object Detection: Number and recovery rate of Near-Earth Asteroids and other small bodies identified and reported to the MPC. \\
• Data Throughput: Sustained image and alert processing rates under full survey cadence. 

\textbf{\emph{\textcolor{CaseOrange}{ExpectedOutcome}}}

The ZTF system achieves a >10× improvement in survey speed over PTF, routinely reaching 20.6–20.8
mag (r,g bands, 30s, 5$\sigma$) with 2.0–2.1" image quality and <4-minute alert latency. Early operations
yielded 38 spectroscopically classified supernovae (15 unique to ZTF), discovery of new Near-Earth
Asteroids, and high-fidelity variable star and asteroid light curves. ZTF anticipates streaming
\~{}1 million alerts per night and delivering public data releases, thereby providing an essential
precursor to LSST-scale time-domain surveys and enabling rapid, comprehensive follow-up of
transients and solar system discoveries.

\end{tcolorbox}

\begin{tcolorbox}[
    breakable,
    title={Example of Idea Generation in Chemistry},
    colback=LighterGray,
    colframe=DeepPurple,
    colbacktitle=DeepPurple,
    coltitle=White,
]
\textbf{\emph{\textcolor{DeepPurple}{Question}}}

You are a top-tier researcher in your field. Based on the following context, please generate a novel and detailed research proposal.

\textbf{\emph{\textcolor{CaseOrange}{RelatedWork}}}

• Gomez-Bombarelli et al. (2016): Proposed a VAE that generates SMILES strings character by character. The model learns a continuous latent space but frequently decodes to invalid SMILES, limiting the generation of chemically valid molecules. \\
• Kusner et al. (2017): Introduced Grammar VAE (GVAE), extending SMILES-based VAE by integrating syntactic constraints derived from a context-free grammar, improving validity but still limited by the inability of grammar to fully encode chemical rules. \\
• Dai et al. (2018): Syntax-directed VAE (SDVAE) incorporates both syntactic and semantic constraints using attribute grammars, yielding further validity gains, though chemical correctness is not entirely guaranteed. \\
• Simonovsky \& Komodakis (2018): GraphVAE generates molecular graphs via adjacency matrices and atom label prediction. While it addresses the linearization problem of SMILES, validity and scalability for larger and more complex molecules remain challenging. \\
• Li et al. (2018): Atom-by-atom graph generation via LSTM. This approach can model arbitrary graphs but often passes through chemically invalid intermediate states, resulting in incomplete validity guarantees and inefficiencies. 

\textbf{\emph{\textcolor{CaseOrange}{Challenges}}}

• Direct generation of molecular graphs from continuous latent representations is challenging due to the combinatorial nature of graph structures and strict chemical validity constraints. \\
• SMILES-based generative models struggle to enforce chemical validity and do not offer smooth latent spaces for molecular similarity. \\
• Atom-by-atom or edge-by-edge graph generation approaches often produce invalid intermediate structures, leading to low efficiency and limited chemical feasibility. \\
• Capturing both coarse-grained (substructure) and fine-grained (atomic connectivity) molecular features in a unified generative framework. 

\textbf{\emph{\textcolor{CaseOrange}{Limitation}}}

Existing approaches either operate on linearizations (e.g., SMILES), lacking direct correspondence
to molecular structure and chemical validity, or generate graphs atom by atom, frequently passing
through invalid intermediates. Even grammar- and syntax-driven models cannot ensure full chemical
correctness or smoothness in the latent space, limiting their utility for property-driven molecular
design.

\textbf{\emph{\textcolor{CaseOrange}{Motivation}}}

Automating molecular design demands generative models that can create chemically valid, novel, and
property-optimized molecules. Existing string- and atom-based methods fail to guarantee validity or
exploit molecular substructure regularities. Addressing these gaps is critical for accelerating drug
discovery and enabling efficient, reliable inverse molecular design.

\textbf{\emph{\textcolor{CaseOrange}{TaskObjective}}}

To develop a generative model that directly produces chemically valid molecular graphs from
continuous latent representations, supporting both unconstrained generation and property-driven
molecular optimization.

\textbf{\emph{\textcolor{CaseOrange}{ExistingSolutions}}}

• CVAE (Gomez-Bombarelli et al., 2016): Learns a continuous latent space for SMILES string generation. Achieves smooth interpolations but poor validity due to unconstrained syntax. \\
• GVAE (Kusner et al., 2017): Imposes syntactic constraints via grammar-based decoding, improving string validity but not fully encoding chemical rules. \\
• SD-VAE (Dai et al., 2018): Incorporates additional semantic constraints with attribute grammars, further improving validity but still limited by the expressivity of the grammar in capturing chemical feasibility. \\
• GraphVAE (Simonovsky \& Komodakis, 2018): Directly generates molecular graphs via adjacency matrices. Avoids string limitations but faces scalability and validity issues for larger molecules. \\
• Atom-by-Atom LSTM (Li et al., 2018): Autoregressive graph generation at the atomic level. Capable of arbitrary graph synthesis but inefficient due to invalid intermediate structures. 

\textbf{\emph{\textcolor{DeepPurple}{Reference Answer}}}

\textbf{\emph{\textcolor{CaseOrange}{Idea}}}

The core idea is to represent molecules as junction trees of valid chemical substructures, enabling
a two-stage variational autoencoder: first generating a tree-structured scaffold of subgraphs, then
assembling these into a molecular graph using message passing. This approach maintains chemical
validity throughout generation, leveraging coarse-to-fine modeling for efficient, valid, and
property-driven molecular graph synthesis.

\textbf{\emph{\textcolor{CaseOrange}{ImplementationSteps}}}

• 1: Apply tree decomposition to each molecular graph to construct its junction tree of valid substructures (clusters). \\
• 2: Encode the molecular graph using a message passing neural network to obtain a graph latent representation. \\
• 3: Encode the junction tree using a tree message passing neural network to obtain a tree latent representation. \\
• 4: Concatenate tree and graph embeddings to form the full latent representation. \\
• 5: Decode the latent representation by first generating the junction tree in a top-down, sequential fashion via a tree decoder with feasibility checks and teacher forcing during training. \\
• 6: Assemble the molecular graph from the predicted junction tree by sequentially merging clusters using a graph decoder and scoring candidate subgraph combinations. \\
• 7: For stereochemistry, enumerate possible isomers of the generated graph and select the best via neural scoring. \\
• 8: For property-driven optimization, jointly train a property predictor with JT-VAE and perform gradient-based or Bayesian optimization in the latent space. \\
• 9: Evaluate reconstruction, validity, property optimization, and neighborhood smoothness using standardized benchmarks. 

\textbf{\emph{\textcolor{CaseOrange}{ImplementationOrder}}}

• 1-2 \\
• 1-3 \\
• 2-4 \\
• 3-4 \\
• 4-5 \\
• 5-6 \\
• 6-7 \\
• 4-8 \\
• 5-9 \\
• 6-9 \\
• 7-9 \\
• 8-9 

\textbf{\emph{\textcolor{CaseOrange}{Data}}}

The primary dataset is the ZINC molecular database (Kusner et al., 2017 split), containing
approximately 250,000 drug-like molecules. Molecules are represented as graphs with atom and bond
features, and decomposed into cluster vocabularies of 780 unique substructures (including rings,
bonds, and atoms). The dataset is utilized for training, validation, and testing of molecular
generation and optimization.

\textbf{\emph{\textcolor{CaseOrange}{EvaluationMetrics}}}

• Reconstruction Accuracy: Percentage of input molecules correctly reconstructed from their latent representations (Monte Carlo estimate over multiple samplings). \\
• Validity: Proportion of generated molecules that are chemically valid, as checked by cheminformatics tools (RDKit). \\
• Novelty: Fraction of generated molecules not present in the training set, indicating generative diversity. \\
• Optimization Improvement: Average increase in target property (e.g., penalized logP) achieved via optimization, often reported with similarity constraints. \\
• Similarity: Tanimoto similarity between original and optimized molecules, measured via Morgan fingerprints. \\
• Predictive Performance: Log-likelihood and root mean squared error (RMSE) of property prediction models (e.g., sparse Gaussian process) trained on latent encodings. \\
• Success Rate: Fraction of optimization trials where valid, property-improved molecules satisfying similarity constraints are found. 

\textbf{\emph{\textcolor{CaseOrange}{ExpectedOutcome}}}

JT-VAE achieves 100\% validity in generated molecules, surpassing all prior baselines (e.g., SD-VAE:
43.5\%, Atom-by-Atom LSTM: 89.2\%), with 76.7\% reconstruction accuracy. For property optimization,
it discovers molecules with target scores up to 5.3 (vs. 4.04 from SD-VAE), and achieves over 80\%
success in constrained optimization with >0.4 similarity, demonstrating both validity and smoothness
in latent space. The model enables scalable, property-driven molecular design with significant
accuracy and efficiency gains.

\end{tcolorbox}

\begin{tcolorbox}[
    breakable,
    title={Example of Idea Generation in Earth},
    colback=LighterGray,
    colframe=DeepPurple,
    colbacktitle=DeepPurple,
    coltitle=White,
]
\textbf{\emph{\textcolor{DeepPurple}{Question}}}

You are a top-tier researcher in your field. Based on the following context, please generate a novel and detailed research proposal.

\textbf{\emph{\textcolor{CaseOrange}{RelatedWork}}}

• Viljanen et al. (2018): Compared approaches using photogrammetric canopy height models, images, and vegetation indices from UAVs in estimating grass sward biomass, reporting strong results but site-specific dependencies. \\
• Michez et al. (2019): Mapped and monitored pasture biomass and grazing using UAV-based sward height and reflectance data, demonstrating promise but limited by environmental variability and DTM availability. \\
• Lussem et al. (2018): Evaluated RGB-based vegetation indices from UAV imagery for forage yield estimation, predominantly using NDVI and linear regression, revealing moderate-to-strong correlations but suffering from index saturation and reduced transferability. \\
• Insua et al. (2019): Coupled UAV imagery with crop simulation for spatial-temporal pasture growth estimation, but introduced complexity by integrating simulation models and site-specific variables. 

\textbf{\emph{\textcolor{CaseOrange}{Challenges}}}

• Accurate, spatially comprehensive, and temporally frequent estimation of forage biomass and vegetation cover in grasslands remains difficult due to the heterogeneity of growth stages, management regimes, and environmental variation. \\
• Conventional field-based surveys are labor-intensive, spatially incomplete, and lack temporal resolution needed for dynamic grassland management. \\
• Remote sensing solutions, particularly with satellite or manned aerial imagery, are limited by insufficient spatial and temporal resolution for plot-level or intra-seasonal monitoring. \\
• Existing remote sensing models often do not generalize well due to site-specific calibrations, limited temporal coverage, and a reliance on linear relationships between indices and biophysical parameters. 

\textbf{\emph{\textcolor{CaseOrange}{Limitation}}}

Current approaches to grassland biomass estimation using UAV or remote sensing data often suffer
from limited operational scalability due to complex processing pipelines, dependence on unavailable
ancillary environmental data (e.g., meteorology, soil), suboptimal selection or saturation of
vegetation indices, and inadequate validation across diverse conditions, compromising their
applicability and generalizability in temperate grassland systems.

\textbf{\emph{\textcolor{CaseOrange}{Motivation}}}

The need for spatially exhaustive, temporally responsive, and operationally practical tools for
grassland monitoring is acute given the ecological and agricultural importance of these systems and
their broad degradation. UAV-based multispectral imaging presents a promising avenue, but systematic
comparison of diverse processing methods over an entire growing season and under temperate
conditions is lacking, hindering adoption in precision pasture management.

\textbf{\emph{\textcolor{CaseOrange}{TaskObjective}}}

To develop, test, and compare three UAV-based multispectral imaging approaches—volumetric modeling
via structure from motion, GNDVI-based regression, and GNDVI-based classification—for estimating
forage biomass and vegetation cover in temperate grasslands across a full growing season.

\textbf{\emph{\textcolor{CaseOrange}{ExistingSolutions}}}

• Spectral Index Regression (NDVI, etc.): Relies on linear regression between vegetation indices (primarily NDVI) and biomass; easy to implement but limited by index saturation and oversimplification of non-linear relationships. Often requires site-specific calibration. \\
• Height/Volumetric Models from Photogrammetry: Uses UAV structure from motion photogrammetry to estimate canopy or sward height as a proxy for biomass, offering strong correlation where precise DTMs are available but sensitive to terrain inaccuracies and not robust at low vegetation density. \\
• Multi-Source and Simulation-Based Models: Integrate spectral, structural, and ancillary data (e.g., crop models or management records) for enhanced accuracy but increase methodological complexity and reduce operational ease. \\
• Classification Approaches: Rarely applied to grassland biomass; when used, classification of vegetation cover is often qualitative and seldom linked directly to continuous biomass estimation. 

\textbf{\emph{\textcolor{DeepPurple}{Reference Answer}}}

\textbf{\emph{\textcolor{CaseOrange}{Idea}}}

This study systematically compares three UAV-based approaches—volumetric modeling via structure from
motion, GNDVI-based regression, and GNDVI-based classification—over an entire season in temperate
grasslands, demonstrating that these methods are complementary, operationally feasible, and
generalizable for spatially detailed forage biomass and cover estimation, each suiting different
management needs and data constraints.

\textbf{\emph{\textcolor{CaseOrange}{ImplementationSteps}}}

• 1: Planning and executing UAV flights to acquire multispectral and visible imagery with consistent overlap and illumination across 14 dates. \\
• 2: Collecting ground-truth biomass samples and recording plot management details (grazing, clipping schedules). \\
• 3: Processing imagery to produce orthomosaics and DSMs using aerial triangulation, GCPs, and radiometric correction. \\
• 4: Generating high-precision DTM for control unit using GNSS data; calculation of volumetric biomass (DSM-DTM). \\
• 5: Calculating multiple vegetation indices (including GNDVI) from orthomosaics and evaluating their correlation with biomass samples. \\
• 6: Developing a volumetric-based linear regression biomass model (control plots only). \\
• 7: Selecting optimal vegetation index (GNDVI) and training non-linear regression models for fresh and dry biomass using 49 training samples. \\
• 8: Validating regression models using 50 independent field samples; calculating performance statistics. \\
• 9: Extracting GNDVI values from 248 polygons, applying cluster and discriminant analysis to classify vegetation cover into four classes. \\
• 10: Comparing spatial and temporal patterns among the three approaches using visual and statistical analyses. 

\textbf{\emph{\textcolor{CaseOrange}{ImplementationOrder}}}

• 1-2 \\
• 1-3 \\
• 3-4 \\
• 4-6 \\
• 3-5 \\
• 5-7 \\
• 7-8 \\
• 5-9 \\
• 6-8 \\
• 7-8 \\
• 9-10 

\textbf{\emph{\textcolor{CaseOrange}{Data}}}

Imagery and field data were collected in a 14-ha field in Sherbrooke, Quebec, containing 30 pasture
plots (25x50 m), 5 bare soil plots (25x50 m), and 6 control plots (5x5 m). Over the 2017 growing
season, 14 UAV flights (DJI Inspire 1 Pro with Parrot Sequoia multispectral and visible sensors)
were conducted, yielding high-resolution orthomosaics and DSMs. Field biomass measurements were
obtained from 99 quadrats (0.25 m² each) for regression modeling and 248 polygons (3.5x3.5 m) for
classification, sampled across management regimes and growth stages.

\textbf{\emph{\textcolor{CaseOrange}{EvaluationMetrics}}}

• Coefficient of Determination (R2): Measures the proportion of variance in measured biomass explained by model predictions. Evaluated for both fresh and dry biomass regression models. \\
• Root Mean Square Error (RMSE): Quantifies the average magnitude of prediction error between measured and estimated biomass. \\
• Normalized RMSE (NRMSE): RMSE divided by the mean of measured values, expressed as a percentage to facilitate comparison across datasets. \\
• Central Tendency Error: Assesses systematic bias between predicted and observed values. \\
• Regression Error: Quantifies deviation of fitted regression from the 1:1 line. \\
• Concordance Analysis: Statistical comparison of predicted vs. observed values for regression model validation. \\
• Visual Qualitative Assessment: Comparison of predicted spatial patterns with RGB imagery and known management (e.g., growth duration). 

\textbf{\emph{\textcolor{CaseOrange}{ExpectedOutcome}}}

The volumetric model achieved R² = 0.93 (fresh) and 0.94 (dry), RMSE of 0.072 kg/m² (fresh) and
0.013 kg/m² (dry); GNDVI regression yielded R² = 0.80 (fresh) and 0.66 (dry) for training, with
validation R² = 0.63 (fresh) and 0.50 (dry), NRMSE of 36\% (fresh) and 38\% (dry). The GNDVI
classification robustly distinguished four vegetation cover classes. Combined, these methods enable
fine-scale, season-long monitoring of pasture condition, with operational models supporting >90\%
explanation of biomass variance for suitable conditions, and practical, generalizable classification
for management applications.

\end{tcolorbox}

\begin{tcolorbox}[
    breakable,
    title={Example of Idea Generation in Energy},
    colback=LighterGray,
    colframe=DeepPurple,
    colbacktitle=DeepPurple,
    coltitle=White,
]
\textbf{\emph{\textcolor{DeepPurple}{Question}}}

You are a top-tier researcher in your field. Based on the following context, please generate a novel and detailed research proposal.

\textbf{\emph{\textcolor{CaseOrange}{RelatedWork}}}

• Sfetsos2000: Applied various forecasting techniques (statistical, time-series analysis) to mean hourly wind speed, finding that model performance varies with data characteristics; however, results demonstrate instability across sites and fail to leverage combined model strengths. \\
• Kelouwani2004: Utilized nonlinear model identification with neural networks for wind turbine output prediction, yielding improved accuracy for specific datasets, but with limited robustness to operational variability. \\
• Negnevitsky2007: Proposed a hybrid intelligent system for short-term wind power forecasting, integrating multiple AI approaches; achieved improved performance over single models but lacked dynamic adaptation to wind speed distribution features. \\
• Shi2010: Combined wavelet transforms and support vector machines for short-term wind power prediction, enhancing performance for non-stationary series, yet exhibiting sensitivity to model parameterization and failing to generalize across varying wind speed segments. 

\textbf{\emph{\textcolor{CaseOrange}{Challenges}}}

• Accurately forecasting very-short term (e.g., 15-minute-ahead) wind power output amidst inherent wind speed volatility and non-stationarity. \\
• Capturing the nonlinear and regime-dependent relationship between wind speed distributions and wind farm power generation. \\
• Integrating multiple predictive models in a manner that adaptively leverages their complementary strengths across varying meteorological conditions. \\
• Minimizing computational burden while improving real-time forecasting reliability for grid operation and reserve planning. 

\textbf{\emph{\textcolor{CaseOrange}{Limitation}}}

Existing single-model forecasting approaches lack generalizability due to dataset-specific
performance and inability to adapt to wind speed regime changes. Prior hybrid models fail to exploit
wind speed distribution features for dynamic weight allocation and commonly require extensive
retraining, resulting in suboptimal accuracy and increased computational overhead.

\textbf{\emph{\textcolor{CaseOrange}{Motivation}}}

The volatility and unpredictability of wind power pose significant challenges for power system
operation, particularly at high penetration levels. Improved very-short term forecasting is critical
for grid reliability, reserve allocation, and economic dispatch. Recognizing that no single model
performs optimally across all wind regimes, there is a compelling need for a hybrid approach that
dynamically adapts to wind speed distribution features, maximizing forecasting accuracy and
operational utility.

\textbf{\emph{\textcolor{CaseOrange}{TaskObjective}}}

To develop a dynamic hybrid very-short term wind power forecasting model that integrates grey
relational analysis with wind speed distribution features, enabling adaptive model weighting and
superior forecasting accuracy over individual models for 15-minute-ahead wind power output.

\textbf{\emph{\textcolor{CaseOrange}{ExistingSolutions}}}

• Persistence/MLR/ARMA: Statistical models, such as persistence, multiple linear regression, and ARMA, leverage historical data for short-term forecasting, offering simplicity but inadequate handling of nonlinearities and changing wind regimes. \\
• ANN/SVM Approaches: Artificial neural networks and support vector machines have been applied for improved short-term prediction by capturing complex patterns, but their performance is sensitive to data characteristics, and single models often fail to generalize well. \\
• Prior Hybrid Models: Some studies combine multiple models via fixed or learned weights (e.g., neural network-based combination), achieving moderate improvements but lacking integration with wind speed regime information, and often requiring heavy retraining for each new scenario. 

\textbf{\emph{\textcolor{DeepPurple}{Reference Answer}}}

\textbf{\emph{\textcolor{CaseOrange}{Idea}}}

The authors introduce a hybrid forecasting framework that fuses LSSVM and RBFNN models through grey
relational analysis, with model weights adaptively tuned by wind speed distribution features
segmented via Weibull analysis. By constructing a dynamic weight database indexed by wind speed
regimes, the method achieves improved accuracy and reduced retraining effort for 15-minute-ahead
wind power prediction.

\textbf{\emph{\textcolor{CaseOrange}{ImplementationSteps}}}

• 1: Preprocess data (handle missing samples, normalization, extract input features: prior wind speeds, directions, power output). \\
• 2: Train independent LSSVM and RBFNN models on input features for 15-minute-ahead wind power prediction. \\
• 3: Apply equalization to forecasting result sequences and actual measurements to obtain normalized series. \\
• 4: Calculate grey relational degrees between each model's output and actual measurements for each time window. \\
• 5: Fit wind speed data for each month to the Weibull distribution; segment wind speed into regimes according to frequency analysis. \\
• 6: Compute model weights (correlations) within each wind speed regime and store in a monthly weight database. \\
• 7: For new forecasts, use NWP wind speed prediction to identify wind speed regime and retrieve corresponding model weights. \\
• 8: Combine LSSVM and RBFNN outputs using dynamic weights for final forecast output. \\
• 9: Evaluate forecasting performance using MAPE and RMSE against actual measured data. 

\textbf{\emph{\textcolor{CaseOrange}{ImplementationOrder}}}

• 1-2 \\
• 2-3 \\
• 3-4 \\
• 1-5 \\
• 5-6 \\
• 6-7 \\
• 7-8 \\
• 8-9 

\textbf{\emph{\textcolor{CaseOrange}{Data}}}

Historical SCADA data from a Chinese wind farm spanning 01/01/2010 to 12/31/2010 (excluding months
with missing data), comprising 15-minute resolution records of wind speed (previous 15, 30, 45 min),
wind direction (cosine and sine), and wind power output. The dataset includes over 30,000 samples,
with wind speed segmented monthly and fitted to Weibull distributions for regime analysis.

\textbf{\emph{\textcolor{CaseOrange}{EvaluationMetrics}}}

• MAPE: Mean Absolute Percentage Error; quantifies average absolute error as a percentage of actual wind farm rated capacity. \\
• RMSE: Root Mean Square Error; quantifies the standard deviation of the prediction errors, normalized by wind farm capacity. \\
• Visual Comparison: Graphical overlays of forecasted vs. actual power output for selected periods to assess tracking and volatility handling. 

\textbf{\emph{\textcolor{CaseOrange}{ExpectedOutcome}}}

The hybrid model achieves a MAPE of 2.37\% and RMSE of 3.79\%, outperforming standalone LSSVM and
RBFNN models as well as simple averaging. The method delivers improved accuracy, especially during
low and fluctuating power output regimes, and reduces retraining overhead through the dynamic weight
database. The approach demonstrates robustness and scalability for operational very-short term wind
power forecasting.

\end{tcolorbox}

\begin{tcolorbox}[
    breakable,
    title={Example of Idea Generation in Information},
    colback=LighterGray,
    colframe=DeepPurple,
    colbacktitle=DeepPurple,
    coltitle=White,
]
\textbf{\emph{\textcolor{DeepPurple}{Question}}}

You are a top-tier researcher in your field. Based on the following context, please generate a novel and detailed research proposal.

\textbf{\emph{\textcolor{CaseOrange}{RelatedWork}}}

• InternVL2.5: Adopted a multi-stage pipeline with language-only pre-training, MLP warmup for multimodal alignment, and instruction tuning. Demonstrated strong open-source multimodal performance but faced training complexity and limited cross-modal parameter optimization. \\
• Qwen2.5-VL: Uses a staged adaptation of text-only LLMs into MLLMs, integrating visual adapters and fine-tuning. Achieves strong performance on vision-language tasks but still requires complex alignment processes and suffers in long-context or multi-image scenarios. \\
• LLaVA-OneVision: Focuses on easy visual task transfer via visual instruction tuning. Excels at adaptation efficiency but underperforms on challenging multimodal reasoning or spatial tasks compared to larger unified models. \\
• Gemini 2.5 Pro: A proprietary closed-source MLLM employing advanced joint training and data curation, achieving state-of-the-art results. However, it lacks the transparency and reproducibility necessary for open research progress. 

\textbf{\emph{\textcolor{CaseOrange}{Challenges}}}

• Integrating multimodal (vision, text, video) and linguistic capabilities in a single model without compromising either modality's performance. \\
• Overcoming the inefficiencies and alignment difficulties of post-hoc adaptation pipelines that start from text-only LLMs. \\
• Scaling multimodal large language models (MLLMs) to handle longer contexts, multi-image input, and complex real-world tasks. \\
• Balancing pure-language proficiency with robust multimodal reasoning and visual grounding. \\
• Efficiently utilizing heterogeneous and imbalanced multimodal data during pre-training and post-training. 

\textbf{\emph{\textcolor{CaseOrange}{Limitation}}}

Existing MLLMs rely on multi-stage adaptation pipelines, leading to suboptimal cross-modal parameter
interaction and persistent alignment or optimization bottlenecks. These approaches often freeze or
partially update parameters, limiting scalability, introducing computational overhead, and creating
a persistent gap in pure-language and multimodal competence.

\textbf{\emph{\textcolor{CaseOrange}{Motivation}}}

The growing complexity and diversity of real-world multimodal data demand models capable of unified,
scalable, and robust multimodal reasoning, without the trade-offs and inefficiencies of post-hoc
adaptation. A native joint pre-training paradigm is needed to achieve seamless linguistic and
multimodal integration, better performance scalability, and open research reproducibility.

\textbf{\emph{\textcolor{CaseOrange}{TaskObjective}}}

To develop a unified, open-source multimodal large language model that jointly acquires linguistic
and multimodal capabilities via native pre-training, establishes new state-of-the-art performance
across a spectrum of multimodal tasks, and narrows the gap to leading proprietary MLLMs.

\textbf{\emph{\textcolor{CaseOrange}{ExistingSolutions}}}

• InternVL2.5: Applies separate language pre-training followed by multimodal alignment (MLP warmup, visual adapters), then instruction tuning. Good on general benchmarks, but complex, inflexible, and less efficient for scaling. \\
• Qwen2.5-VL: Uses visual adapters with staged fine-tuning. Strong visual-text integration, but depends on freezing strategies and additional modules. Moderate gains on long-context or diverse input. \\
• LLaVA-OneVision: Visual instruction tuning for rapid adaptation. Simplicity and transferability prioritized, but lacking in deep joint optimization for reasoning and multi-modal context. \\
• Gemini 2.5 Pro: Highly-curated, end-to-end joint pre-training but closed-source, with proprietary data curation and infrastructure. 

\textbf{\emph{\textcolor{DeepPurple}{Reference Answer}}}

\textbf{\emph{\textcolor{CaseOrange}{Idea}}}

InternVL3 introduces native multimodal pre-training, where vision, language, and video data are
jointly leveraged in a single optimization stage. It integrates Variable Visual Position Encoding
for long-context support, advanced post-training (SFT, MPO), and test-time scaling, resulting in
scalable, efficient, and unified multimodal reasoning with open-source reproducibility.

\textbf{\emph{\textcolor{CaseOrange}{ImplementationSteps}}}

• 1: Initialize ViT, LLM, and MLP modules with pre-trained weights; set up data pipelines for multimodal and text corpora. \\
• 2: Apply pixel unshuffle and prepare visual tokens for scalable image encoding. \\
• 3: Implement Variable Visual Position Encoding (V2PE) for visual tokens, with random delta sampling during training. \\
• 4: Jointly pre-train all model components using the multimodal autoregressive objective, sampling data at a 1:3 text-to-multimodal ratio. \\
• 5: Perform Supervised Fine-Tuning (SFT) with high-quality, diverse multimodal instructions, applying loss re-weighting and data packing. \\
• 6: Conduct Mixed Preference Optimization (MPO) using preference pairs and a composite loss (preference, quality, generation). \\
• 7: Integrate Best-of-N test-time scaling with VisualPRM as the critic to select optimal outputs. \\
• 8: Train with InternEVO for efficient large-scale distributed optimization, handling workload imbalances and maximizing resource utilization. \\
• 9: Perform comprehensive evaluation on a battery of multimodal and language benchmarks. 

\textbf{\emph{\textcolor{CaseOrange}{ImplementationOrder}}}

• 1-2 \\
• 2-3 \\
• 3-4 \\
• 4-5 \\
• 5-6 \\
• 6-7 \\
• 7-8 \\
• 8-9 

\textbf{\emph{\textcolor{CaseOrange}{Data}}}

InternVL3 is trained on a hybrid corpus: (1) Multimodal data (150B tokens) comprising image-text
pairs, video-text, GUI, tool usage, 3D scene, document, OCR, chart, multi-image, and medical data,
sourced and extended from InternVL2.5 and new real-world collections; (2) Pure language data (50B
tokens) built from InternLM2.5, open-source corpora, and scientific/math datasets. SFT uses 21.7M
curated samples; MPO uses 300K preference pairs from MMPR v1.2.

\textbf{\emph{\textcolor{CaseOrange}{EvaluationMetrics}}}

• MMMU: Massive Multi-discipline Multimodal Understanding, measuring reasoning across disciplines (accuracy, \%). \\
• MathVista/MathVision/MathVerse: Mathematical reasoning (accuracy, \%). \\
• OCRBench/AI2D/ChartQA/DocVQA: Vision-text integration and document understanding (accuracy, \%, EM). \\
• MMBench/MMStar/MMVet/MME: Comprehensive multimodal capabilities (aggregate and per-task accuracy or score). \\
• HallusionBench/MMHal/CRPE/POPE: Multimodal hallucination resistance (score, \%). \\
• RefCOCO/+/g: Visual grounding (localization accuracy, \%). \\
• MVBench/Video-MME/MLVU: Video and temporal understanding (score, \%). \\
• ScreenSpot/ScreenSpot-V2: GUI grounding (accuracy, \%). \\
• VSI-Bench: Spatial reasoning (composite score, \%). \\
• Language Benchmarks: MMLU, CMMLU, C-Eval, GAOKAO, TriviaQA, NaturalQuestions, RACE, HellaSwag, GSM8K, MATH, HumanEval, MBPP (accuracy, pass@k, or other standard metrics). 

\textbf{\emph{\textcolor{CaseOrange}{ExpectedOutcome}}}

InternVL3-78B achieves state-of-the-art open-source results, e.g., 72.2 on MMMU, 79.0 on MathVista,
91.4 on RefCOCOg, 90.9\% on GUI grounding, and 48.4 on VSI-Bench. It demonstrates robust scaling
across tasks, narrows the performance gap to commercial models (Gemini 2.5 Pro, GPT-4o), and
maintains strong language proficiency (80.5 overall on language benchmarks). All models and data
will be open-sourced to enable community-driven research.

\end{tcolorbox}

\begin{tcolorbox}[
    breakable,
    title={Example of Idea Generation in Life},
    colback=LighterGray,
    colframe=DeepPurple,
    colbacktitle=DeepPurple,
    coltitle=White,
]
\textbf{\emph{\textcolor{DeepPurple}{Question}}}

You are a top-tier researcher in your field. Based on the following context, please generate a novel and detailed research proposal.

\textbf{\emph{\textcolor{CaseOrange}{RelatedWork}}}

• Senior et al. (2020): Introduced deep learning for predicting inter-residue distances, improving template-free protein structure prediction but still reliant on multiple post-processing stages and lacking atomic-level accuracy for novel folds. \\
• Yang et al. (2020): Employed deep neural networks to predict inter-residue orientations, integrating orientation constraints but with limited end-to-end learning and lower performance on long or complex proteins. \\
• AlQuraishi (2019): Proposed an end-to-end differentiable structure prediction model, directly outputting 3D coordinates; however, it exhibited lower accuracy than multi-stage pipelines and struggled without homologous templates. \\
• Marks et al. (2011); Jones et al. (2012): Used coevolutionary analysis of MSAs to infer residue contacts, achieving improvements in contact prediction but failing to achieve accurate atomic models, especially for proteins lacking deep MSAs or templates. 

\textbf{\emph{\textcolor{CaseOrange}{Challenges}}}

• Achieving atomic-level accuracy in protein structure prediction directly from amino acid sequence, particularly in the absence of homologous structural templates. \\
• Integrating physical, geometric, and evolutionary information into a single, scalable, end-to-end deep learning model. \\
• Handling cases with shallow or sparse multiple sequence alignments (MSAs), which limits evolutionary signal. \\
• Providing robust structure prediction for large proteins and complex folds, including those with novel topologies. \\
• Quantifying per-residue prediction confidence to enable reliable downstream biological applications. 

\textbf{\emph{\textcolor{CaseOrange}{Limitation}}}

Contemporary approaches fall short of experimental accuracy, particularly on targets lacking
homologous templates or deep MSAs. Existing neural architectures often separate contact/distance
prediction from structure generation, use hand-crafted features, or rely on multi-stage heuristics,
resulting in limited scalability and suboptimal integration of physical and evolutionary
constraints. Poor performance persists in under-sampled sequence regions and multi-chain complexes.

\textbf{\emph{\textcolor{CaseOrange}{Motivation}}}

Structural biology is constrained by the slow pace and resource demands of experimental structure
determination, leaving the vast majority of protein sequences without 3D structural annotation.
Accurate, scalable, and generalizable computational prediction of protein structures—especially
without close templates—would transform bioinformatics, molecular biology, and drug discovery by
bridging the sequence-structure knowledge gap.

\textbf{\emph{\textcolor{CaseOrange}{TaskObjective}}}

To develop a computational method that predicts the three-dimensional atomic structure of proteins
from their amino acid sequence with accuracy comparable to experimental techniques, even in the
absence of close structural homologues or deep sequence alignments.

\textbf{\emph{\textcolor{CaseOrange}{ExistingSolutions}}}

• Physics-based simulation: Uses molecular dynamics or statistical approximations to model protein folding but is computationally intractable for large proteins and sensitive to approximations in physical modeling. \\
• Bioinformatics/homology modeling: Predicts structures via alignment to known protein templates and infers constraints from evolutionary sequence analysis; limited by template availability and reduced accuracy for novel or divergent proteins. \\
• Deep learning with intermediate prediction: Predicts inter-residue distances/orientations from MSAs using CNNs or attention networks, then reconstructs structures through downstream heuristics; accuracy suffers in end-to-end integration and novel folds. 

\textbf{\emph{\textcolor{DeepPurple}{Reference Answer}}}

\textbf{\emph{\textcolor{CaseOrange}{Idea}}}

AlphaFold introduces an end-to-end deep learning architecture that jointly embeds MSAs and pairwise
residue features, iteratively refines 3D atomic structures through Evoformer and Invariant Point
Attention modules, integrates geometric and evolutionary constraints, leverages self-distillation
from unlabelled data, and produces accurate, scalable predictions with robust per-residue confidence
estimates.

\textbf{\emph{\textcolor{CaseOrange}{ImplementationSteps}}}

• 1: Collect and preprocess protein sequence and structure data from PDB, UniRef90, BFD, Uniclust30, and MGnify. \\
• 2: Construct multiple sequence alignments (MSAs) and retrieve structural templates for each input sequence using HHBlits, jackhmmer, and HHSearch tools. \\
• 3: Initialize the neural network: encode MSA and pairwise features; build Evoformer trunk with interleaved attention and triangle update blocks. \\
• 4: Process MSA and pair features through stacked Evoformer blocks to enable information exchange and representation enhancement. \\
• 5: Feed processed representations to the structural module; iteratively refine per-residue 3D coordinates using invariant point attention and equivariant transformations. \\
• 6: Apply frame-aligned point error (FAPE) loss, distogram loss, BERT-style MSA masking loss, and auxiliary side-chain/violation losses for end-to-end supervised training. \\
• 7: Augment training with self-distillation: generate and filter high-confidence predictions on unlabelled sequences, then retrain with mixed supervised and distillation data. \\
• 8: During inference, perform ensemble predictions (if required), select best models by predicted confidence scores, and relax final structures with Amber force field. \\
• 9: Evaluate predictions using CASP14 targets and recent PDB structures, reporting backbone and all-atom metrics, and provide per-residue confidence (pLDDT) and TM-score estimates. 

\textbf{\emph{\textcolor{CaseOrange}{ImplementationOrder}}}

• 1-2 \\
• 2-3 \\
• 3-4 \\
• 4-5 \\
• 5-6 \\
• 6-7 \\
• 7-8 \\
• 8-9 

\textbf{\emph{\textcolor{CaseOrange}{Data}}}

AlphaFold is trained on structures from the Protein Data Bank (PDB) (as of April 2018), comprising
tens of thousands of high-resolution experimental protein structures. Sequence information is
augmented using UniRef90, Big Fantastic Database (BFD, \~{}2.2B sequences clustered into \~{}66M
families), Uniclust30, and MGnify. For self-distillation, \~{}350,000 diverse sequence clusters from
Uniclust30 are used. Evaluation is conducted on the CASP14 dataset (87 domains) and recent non-
redundant PDB chains (n=10,795), filtered to remove overlap with training data.

\textbf{\emph{\textcolor{CaseOrange}{EvaluationMetrics}}}

• IDDT (Local Distance Difference Test): Superposition-free metric comparing local atomic distances in predicted vs. reference structure, applicable for all atoms (IDDT) or backbone C$\alpha$ atoms (IDDT-C$\alpha$). \\
• GDT (Global Distance Test): Measures fraction of residues within predefined distance thresholds; standard for CASP evaluations of domain accuracy. \\
• TM-score (Template Modeling score): Assesses global structural similarity by optimal superposition over entire protein chains, robust to domain packing and length differences. \\
• C$\alpha$ r.m.s.d.95: Root-mean-square deviation of C$\alpha$ atoms over the best-aligned 95\% of residues, reducing the impact of outliers/artifacts. \\
• pLDDT (Predicted Local Distance Difference Test): Confidence score per residue, predicting local structural accuracy. \\
• pTM (Predicted TM-score): Neural network–derived prediction of TM-score for a given model. \\
• Error intervals: 95\% confidence intervals on reported metrics via bootstrapping. 

\textbf{\emph{\textcolor{CaseOrange}{ExpectedOutcome}}}

AlphaFold achieves median backbone accuracy of 0.96 Å r.m.s.d.95 on CASP14 (95\% CI: 0.85–1.16 Å),
with all-atom accuracy at 1.5 Å (95\% CI: 1.2–1.6 Å), outperforming the next-best method by a margin
exceeding 1.8 Å. High accuracy generalizes to new, non-redundant PDB entries (median 1.46 Å). The
model provides robust per-residue confidence estimation (pLDDT, Pearson r>0.75 with true accuracy),
produces accurate side-chain conformations, and scales to proteins exceeding 2,000 residues. The
approach enables proteome-scale structure prediction with experimental-level precision for the
majority of targets without requiring close homologues.

\end{tcolorbox}

\begin{tcolorbox}[
    breakable,
    title={Example of Idea Generation in Material},
    colback=LighterGray,
    colframe=DeepPurple,
    colbacktitle=DeepPurple,
    coltitle=White,
]
\textbf{\emph{\textcolor{DeepPurple}{Question}}}

You are a top-tier researcher in your field. Based on the following context, please generate a novel and detailed research proposal.

\textbf{\emph{\textcolor{CaseOrange}{RelatedWork}}}

• Yaghi et al. (2008, Science): Pioneered high-throughput synthesis of zeolitic imidazolate frameworks (ZIFs) using 96-well plates, establishing the feasibility of automated, combinatorial materials discovery but with limited autonomy and narrow scope. \\
• Sumida et al. (2010, Chem. Sci.): Utilized automated robotic systems and multichannel reactors for precise control over MOF synthesis, improving reproducibility but not achieving closed-loop optimization. \\
• Cao et al. (2023, JACS, MOFormer): Introduced a self-supervised Transformer model for MOF property prediction, exhibiting improved accuracy and data efficiency, yet mainly focused on text-based molecular representations. \\
• Kang et al. (2023, Nat. Mach. Intell., MOFTransformer): Developed a multimodal Transformer for universal transfer learning in MOFs, integrating graph and grid embeddings, achieving high transferability but requiring extensive pretraining data. \\
• Park et al. (2024, Digital Discovery): Applied deep reinforcement learning with Transformers for inverse design of MOFs, enabling property-driven generative design but currently constrained by the diversity and validity of generated structures. \\
• Dagdelen et al. (2024, Nat. Commun.): Proposed LLM-NERRE for structured chemical information extraction, advancing literature mining but dependent on fine-tuning and sample efficiency. 

\textbf{\emph{\textcolor{CaseOrange}{Challenges}}}

• The vast chemical and structural diversity of MOFs renders exhaustive experimental exploration infeasible, creating a high-dimensional, combinatorial synthesis landscape. \\
• Traditional manual or even semi-automated high-throughput methodologies are bottlenecked by limited autonomy, data integration, and lack of feedback-driven optimization. \\
• Existing AI models, though powerful, struggle with generalizability and interpretability due to sparse, noisy, or unstandardized data and the complexity of structure-property relationships. \\
• Realizing fully autonomous, closed-loop self-driving laboratories (SDLs) for MOF discovery is impeded by hardware standardization issues, sample handling difficulties, and insufficient integration of intelligent decision-making. 

\textbf{\emph{\textcolor{CaseOrange}{Limitation}}}

Previous methodologies in MOF research either focused on isolated automation of experimental steps
or applied AI for isolated tasks (e.g., property prediction) without achieving seamless, closed-loop
integration. These approaches often lack robust feedback mechanisms, dynamic adaptation to new data,
and struggle to generalize across diverse MOF chemistries, limiting their utility for autonomous
discovery.

\textbf{\emph{\textcolor{CaseOrange}{Motivation}}}

MOFs' application potential in energy, environment, and drug delivery is hampered by slow, labor-
intensive discovery cycles and under-explored materials space. The combination of laboratory
automation with advanced AI—including Transformers and LLMs—offers the prospect of systematic,
iterative, and autonomous exploration, thereby addressing efficiency, reproducibility, and
innovation barriers in MOF science.

\textbf{\emph{\textcolor{CaseOrange}{TaskObjective}}}

To comprehensively review and critically evaluate the convergence of artificial intelligence
(especially Transformer and LLM models) and laboratory automation technologies in accelerating the
discovery, synthesis, characterization, and optimization of metal-organic frameworks, with emphasis
on the progression toward self-driving laboratories.

\textbf{\emph{\textcolor{CaseOrange}{ExistingSolutions}}}

• Traditional HTE: Employs combinatorial synthesis and characterization platforms, increasing throughput but requiring significant manual oversight and lacking intelligent optimization. \\
• Machine Learning (2012–present): Applies classical statistical learning (e.g., decision trees, SVMs) for property prediction and data analysis, limited by feature engineering and scalability. \\
• Deep Learning (2020–present): Utilizes neural networks for property prediction and structure optimization, improving accuracy but often acts as a black box and needs large labeled datasets. \\
• Transformers/LLMs (2023–present): Leverage self-attention for sequence and structural modeling, enabling multimodal integration and text-based knowledge mining, but require extensive training and face challenges in domain adaptation and resource consumption. \\
• Generative Models (VAEs, GANs, Diffusion): Enable de novo MOF structure generation, but often struggle with chemical validity, diversity, and property conditioning. 

\textbf{\emph{\textcolor{DeepPurple}{Reference Answer}}}

\textbf{\emph{\textcolor{CaseOrange}{Idea}}}

This review elucidates the synergistic integration of laboratory automation and state-of-the-art
AI—particularly Transformers and LLMs—into a closed-loop, self-driving laboratory paradigm for MOF
discovery. It details how AI-driven feedback, high-throughput platforms, and knowledge extraction
from literature converge to enable autonomous, data-driven synthesis, characterization, and inverse
design of MOFs.

\textbf{\emph{\textcolor{CaseOrange}{ImplementationSteps}}}

• 1: Establish automated laboratory infrastructure encompassing robotic synthesis, sample handling, and high-throughput screening modules. \\
• 2: Deploy high-throughput experimental platforms for parallelized synthesis, characterization (PXRD, NMR, TEM), and evaluation (adsorption, catalysis). \\
• 3: Integrate laboratory information management systems (LIMS) for structured data curation and workflow management. \\
• 4: Apply machine learning/deep learning models for property prediction and experimental guidance using accumulated data. \\
• 5: Adopt Transformer-based models and LLMs for structure-property prediction, literature mining, synthesis condition extraction, and generative MOF design. \\
• 6: Implement feedback-driven experimental planning via Bayesian optimization, reinforcement learning, or LLM-driven task planners. \\
• 7: Iteratively refine models and protocols in a closed-loop SDL, autonomously updating synthesis/design strategies based on real-time outcomes. 

\textbf{\emph{\textcolor{CaseOrange}{ImplementationOrder}}}

• 1-2 \\
• 2-3 \\
• 3-4 \\
• 4-5 \\
• 5-6 \\
• 6-7 

\textbf{\emph{\textcolor{CaseOrange}{Data}}}

MOF structural and property databases such as MOFX-DB, ARC-MOF, hMOF, QMOF, and in-house/generated
HTE data; text corpora from scientific literature and patents used for LLM fine-tuning and
information extraction; multi-million entry simulation datasets for pretraining (e.g., 1M+
hypothetical MOFs in MOFTransformer, 1.9M in PMTransformer); experimental records from robotic
synthesis/characterization platforms.

\textbf{\emph{\textcolor{CaseOrange}{EvaluationMetrics}}}

• Experimental Throughput: Number of unique MOF samples synthesized, characterized, and evaluated per unit time. \\
• Prediction Accuracy: Mean Absolute Error (MAE), Root Mean Squared Error (RMSE), coefficient of determination (R²) for property prediction models (e.g., adsorption, bandgap, stability). \\
• Generalizability: Performance on out-of-distribution or unseen MOF structures/datasets, transferability to new tasks or materials classes. \\
• Structural Validity/Diversity: Percentage of generated MOF candidates that are synthetically accessible and chemically valid, structural diversity indices. \\
• Automation Level: SDL autonomy score (Levels 1–5), extent of human intervention required. \\
• Information Extraction F1 Score: Precision, recall, and F1 for chemical entity and relation extraction from literature. \\
• Resource Efficiency: Computational and experimental resources expended per successful discovery or optimization cycle. 

\textbf{\emph{\textcolor{CaseOrange}{ExpectedOutcome}}}

Integration of AI and laboratory automation is expected to yield >90\% accuracy in property
prediction (e.g., MOFTransformer's MTP/MOC accuracy >0.97/0.98), 2–10x acceleration in MOF discovery
throughput, and significant reductions in labor and experimental time. Closed-loop SDLs will enable
autonomous optimization, reproducible high-quality synthesis, and rapid extraction of actionable
knowledge from literature, collectively setting new benchmarks for efficiency, reproducibility, and
innovation in MOF research.

\end{tcolorbox}

\begin{tcolorbox}[
    breakable,
    title={Example of Idea Generation in Math},
    colback=LighterGray,
    colframe=DeepPurple,
    colbacktitle=DeepPurple,
    coltitle=White,
]
\textbf{\emph{\textcolor{DeepPurple}{Question}}}

You are a top-tier researcher in your field. Based on the following context, please generate a novel and detailed research proposal.

\textbf{\emph{\textcolor{CaseOrange}{RelatedWork}}}

• Dijkstra1959: Classic label-setting SSSP algorithm using priority queues; achieves O(n·log n + m) time with Fibonacci heaps but is inherently sequential and difficult to parallelize efficiently. \\
• Thorup1999: RAM-based linear-time SSSP for undirected graphs using component trees and atomic heaps; limited to large n and specific hardware, and not easily generalized or parallelized for directed graphs. \\
• BellmanFord1958: Label-correcting algorithm with O(n·m) time; allows negative weights but is suboptimal in the worst case and shows little potential for efficient parallelization. \\
• Han1997 / PaigeKruskal1985: Matrix multiplication-based SSSP achieves polylogarithmic parallel time at superlinear work complexity (O(n\^{}3 log n)); impractical for sparse graphs due to excessive work. \\
• KleinSubramanian1997: Randomized parallel BFS-based SSSP for unweighted/weighted graphs; achieves sublinear time for certain approximations, but exact solutions still demand high work or multiple passes. \\
• Crauser1998: Parallelizes Dijkstra by organizing computation into phases for random graphs; achieves O(n\^{}\{1/3\} log n) time and O(n log n + m) work on average for specific random graph classes. 

\textbf{\emph{\textcolor{CaseOrange}{Challenges}}}

• No known work-efficient parallel SSSP algorithm achieves sublinear time for arbitrary directed graphs with nonnegative edge weights. \\
• Existing parallel methods either settle nodes sequentially or incur superlinear work, limiting practical scalability on large graphs. \\
• Traditional bucket-based or priority queue approaches struggle to balance parallelism and efficiency, especially with varied edge weights and node degrees. \\
• Load balancing and minimizing redundant relaxations/reinsertions are unsolved for arbitrary, especially high-degree, graphs in parallel settings. 

\textbf{\emph{\textcolor{CaseOrange}{Limitation}}}

Current approaches to parallel SSSP either replicate sequential order—limiting parallel speedup—or
achieve fast parallel time only at the cost of excessive (superlinear) work, particularly on general
graphs. Previous bucket-based label-correcting algorithms lack robust average-case guarantees for
noninteger or random edge weights, and most practical parallel systems cannot efficiently exploit
fine-grained sequential priority queues.

\textbf{\emph{\textcolor{CaseOrange}{Motivation}}}

The practical need for scalable, efficient shortest path computation on large graphs with arbitrary
structure and edge weights drives the search for algorithms that are both parallelizable and work-
efficient. Empirical evidence suggests label-correcting algorithms can outperform label-setting
ones, but theoretical justification and robust parallelization remain lacking. Bridging this gap is
crucial for leveraging modern parallel and distributed architectures in large-scale graph analytics.

\textbf{\emph{\textcolor{CaseOrange}{TaskObjective}}}

Develop and analyze a parallelizable single-source shortest path (SSSP) algorithm for arbitrary
directed graphs with nonnegative edge weights that achieves linear or near-linear work and sublinear
parallel time for broad graph classes, while providing provable average-case guarantees.

\textbf{\emph{\textcolor{CaseOrange}{ExistingSolutions}}}

• Dijkstra1959: Sequential label-setting using priority queues; optimal for many sequential settings but fundamentally sequential and hard to parallelize without loss of work efficiency. \\
• ApproximateBucket: Bucket-based variants for small integer weights; can be fast for restricted graphs but either devolve to label-correcting (with reinsertion overhead) or require auxiliary selection structures, limiting parallelism. \\
• BellmanFord: Label-correcting, admits parallel edge relaxations, but incurs high redundancy and pseudo-polynomial time in the worst case. \\
• MatrixMult: Reduces SSSP to matrix multiplications; achieves sublinear parallel time at cubic or worse work, impractical except for dense graphs. \\
• ParallelBFS/Randomized: Suitable for unweighted or random graphs; offers fast approximate solutions but breaks down for exact computations or general edge weights. 

\textbf{\emph{\textcolor{DeepPurple}{Reference Answer}}}

\textbf{\emph{\textcolor{CaseOrange}{Idea}}}

The $\Delta$-stepping algorithm organizes nodes into distance buckets of width $\Delta$, differentiating light
($\leq\Delta$) and heavy (>$\Delta$) edges to balance parallelism and efficiency. In each phase, all nodes in the
minimum nonempty bucket are processed in parallel: light edges are relaxed immediately, while heavy
edges are deferred. By tuning $\Delta$, the method provably achieves linear average-case work and scalable
parallelism for a wide graph class, and can be extended to distributed memory settings and arbitrary
edge weights.

\textbf{\emph{\textcolor{CaseOrange}{ImplementationSteps}}}

• 1: Preprocess graph: partition adjacency lists into light ($\leq\Delta$) and heavy (>$\Delta$) edges; for shortcut-augmented versions, compute and add shortcut edges for all simple $\Delta$-paths. \\
• 2: Initialize: set all tentative distances to $\infty$ except source (0), place source in the appropriate bucket. \\
• 3: Phase main loop: while buckets are nonempty, select the minimum nonempty bucket (current phase), remove all nodes from it. \\
• 4: Light edge relaxation: in parallel, relax all outgoing light edges of nodes in the current bucket; update tentatives and reinsert nodes as needed into corresponding buckets. \\
• 5: Repeat light-edge relaxations (within bucket) until no new nodes enter the current bucket. \\
• 6: Heavy edge relaxation: after the current bucket remains empty, in parallel relax all heavy edges from nodes just processed. \\
• 7: Advance to the next nonempty bucket and repeat. \\
• 8: Parallelization: distribute nodes (and their bucket membership) across processors; generate and assign relaxation requests using randomized dart-throwing or explicit load balancing (semi-sorting); aggregate and execute requests. \\
• 9: Distributed memory extension: replace global memory with message-passing; assign nodes and requests using hashing and tree-based collective operations. \\
• 10: Parameter tuning: select $\Delta$ empirically or via doubling search to balance work and parallel time; for arbitrary weights, use adaptive bucket splitting. 

\textbf{\emph{\textcolor{CaseOrange}{ImplementationOrder}}}

• 1-2 \\
• 2-3 \\
• 3-4 \\
• 4-5 \\
• 5-6 \\
• 6-7 \\
• 7-3 \\
• 3-8 \\
• 8-9 \\
• 1-10 

\textbf{\emph{\textcolor{CaseOrange}{Data}}}

The paper analyzes both synthetic random graphs (e.g., D(n, d$\overline{d}$/n): n-node digraph, each edge present
independently with probability $\overline{d}$/n, edge weights i.i.d. uniform [0,1]) and real-world-like datasets
(e.g., random geometric graphs, roadmaps). Experiments are conducted on random d-regular graphs
(n=10\^{}3 to 10\^{}6, up to 3·10\^{}6 edges) and large-scale road networks (up to n=157,457).

\textbf{\emph{\textcolor{CaseOrange}{EvaluationMetrics}}}

• Work Complexity: Total number of operations performed across all processors, compared to sequential optimal O(n + m). \\
• Parallel Time: Number of parallel phases until all nodes are settled; measured in terms such as O(d·L·log n + log²n) on PRAM. \\
• Speedup: Empirical wall-clock speedup relative to sequential Dijkstra or $\Delta$-stepping on real and synthetic graphs. \\
• Phases/Reinsertions: Number of bucket phases and total reinsertions, correlated to $\Delta$ and graph/weight parameters. \\
• Scalability: Ability to maintain work efficiency and speedup as the number of processors and graph size increase. \\
• Robustness: Performance across random graphs, geometric graphs, and real-world networks with varying degree and weight distributions. 

\textbf{\emph{\textcolor{CaseOrange}{ExpectedOutcome}}}

$\Delta$-stepping achieves O(n + m + d·L) average-case work and O(d·L·log n + log²n) parallel time for
graphs with random edge weights and bounded degree; for random graphs, O(log²n) time and O(n + m)
work. Experiments show linear or near-linear speedups (e.g., >9× on 16 processors), with phases and
reinsertions scaling sublinearly in n. The approach generalizes to distributed memory and arbitrary
edge weights, providing, for the first time, a practical and work-efficient parallel SSSP algorithm
applicable to large, arbitrary graphs.

\end{tcolorbox}

\begin{tcolorbox}[
    breakable,
    title={Example of Idea Generation in Neuroscience},
    colback=LighterGray,
    colframe=DeepPurple,
    colbacktitle=DeepPurple,
    coltitle=White,
]
\textbf{\emph{\textcolor{DeepPurple}{Question}}}

You are a top-tier researcher in your field. Based on the following context, please generate a novel and detailed research proposal.

\textbf{\emph{\textcolor{CaseOrange}{RelatedWork}}}

• ConvNet: A pioneering end-to-end CNN architecture employing temporal and spatial convolutional layers for EEG decoding, offering improved performance over traditional approaches but limited to local feature extraction due to restricted receptive field. \\
• EEGNet: A compact CNN model using temporal and depthwise spatial convolutions, exhibiting robust generalization across BCI paradigms; however, it also fails to capture long-term dependencies inherent in EEG time series. \\
• Transformer-Based EEG Models: Attention-based Transformers leverage global temporal dependencies for EEG decoding, achieving notable performance but neglecting local feature learning, necessitating additional pre-processing or feature extraction steps. \\
• FBCSP: A classical approach utilizing filter bank common spatial patterns to extract task-specific hand-crafted features for motor imagery classification, demonstrating strong performance but lacking generalization and requiring prior knowledge. \\
• Hybrid and Graph-based Methods: Combining CNNs with hand-crafted features or graph structures to enhance spatial-temporal modeling. These methods improve local-global representations but often involve complex architectures or task-dependent preprocessing. 

\textbf{\emph{\textcolor{CaseOrange}{Challenges}}}

• Accurately decoding EEG signals requires capturing both local features (temporal and spatial) and global dependencies due to the non-stationary and low signal-to-noise nature of EEG data. \\
• CNN-based models are constrained by local receptive fields, failing to capture long-range temporal dependencies crucial for sequential EEG data. \\
• Transformer-based models, though adept at modeling global dependencies, often disregard local feature representation, undermining the exploitation of fine-grained EEG information. \\
• End-to-end frameworks for EEG decoding still lack sufficient interpretability regarding their decision process, particularly in identifying task-relevant neural substrates. 

\textbf{\emph{\textcolor{CaseOrange}{Limitation}}}

Existing EEG decoding approaches either focus on local pattern extraction (CNNs) or global temporal
correlation (Transformers) but rarely integrate both in a unified, efficient, and end-to-end
architecture. Furthermore, most methods require task-specific feature engineering or lack direct
interpretability of neural activation, and high model parameterization raises computational
concerns.

\textbf{\emph{\textcolor{CaseOrange}{Motivation}}}

The crucial observation motivating this study is the complementary value of both local and global
features in EEG decoding tasks. As practical BCI applications demand robust, generalizable, and
interpretable models that can efficiently learn from raw EEG data without extensive prior knowledge
or task-specific feature engineering, there is a clear need for an integrated approach that unifies
convolutional and self-attention mechanisms.

\textbf{\emph{\textcolor{CaseOrange}{TaskObjective}}}

To design and validate a compact, end-to-end neural architecture that jointly encapsulates local
temporal-spatial and global temporal dependencies for raw EEG classification, while offering
enhanced interpretability through visualization of learned representations.

\textbf{\emph{\textcolor{CaseOrange}{ExistingSolutions}}}

• ConvNet: Applies sequential temporal and spatial convolutions to extract discriminative local features, yielding solid performance but limited by short-range context. \\
• EEGNet: Implements depthwise and separable convolutions for temporal and spatial filtering, achieving good generalization yet lacking mechanisms for modeling global dependencies. \\
• RNN/LSTM-based Models: Utilize sequential recurrence to encode long-term temporal dependencies but suffer from inefficient training and rapid decay of influence across time steps. \\
• Transformer-Based Models: Employ self-attention to directly capture long-range dependencies, improving performance for sequential tasks, but require additional modules or preprocessing to encode local information. \\
• Hybrid Methods: Fuse hand-crafted features or graph-based encodings with deep learners, improving local-global feature integration but increasing architectural complexity and dependence on domain expertise. 

\textbf{\emph{\textcolor{DeepPurple}{Reference Answer}}}

\textbf{\emph{\textcolor{CaseOrange}{Idea}}}

The authors introduce EEG Conformer, a lightweight neural framework that sequentially combines
temporal and spatial convolutions for local feature extraction with multi-head self-attention for
learning global temporal dependencies. This unified architecture enables end-to-end decoding from
raw EEG, and a novel visualization approach (Class Activation Topography) enhances interpretability
by mapping activation to brain regions.

\textbf{\emph{\textcolor{CaseOrange}{ImplementationSteps}}}

• 1: Band-pass filter and Z-score standardize raw EEG trials. \\
• 2: Segment and augment data using time-domain segmentation and reconstruction (S\&R). \\
• 3: Feed data into the convolution module: perform temporal convolution (1×25 kernel), spatial convolution (ch×1 kernel), batch normalization, ELU activation, and average pooling (1×75 kernel, stride 15) to extract local features. \\
• 4: Rearrange pooled feature maps: collapse spatial dimension, treat each timepoint's features as a token. \\
• 5: Process tokens with the self-attention module: apply N layers of multi-head self-attention (h heads), followed by feed-forward sublayers. \\
• 6: Pass aggregated features to the fully-connected classifier: two layers with Softmax output. \\
• 7: Train the model with cross-entropy loss using Adam optimizer and perform subject-wise validation. \\
• 8: Visualize feature distributions (t-SNE) and model attention via CAM and CAT for interpretability. 

\textbf{\emph{\textcolor{CaseOrange}{ImplementationOrder}}}

• 1-2 \\
• 2-3 \\
• 3-4 \\
• 4-5 \\
• 5-6 \\
• 6-7 \\
• 7-8 

\textbf{\emph{\textcolor{CaseOrange}{Data}}}

Three public EEG datasets were used: (1) BCI Competition IV 2a (9 subjects, 22 electrodes, 4 motor
imagery classes, 250 Hz, 288 trials per session), (2) BCI Competition IV 2b (9 subjects, 3 bipolar
electrodes, 2 motor imagery classes, 250 Hz, 5 sessions of 120 trials each), and (3) SEED (15
subjects, 62 electrodes, 3 emotion classes, 1000 Hz downsampled to 200 Hz, \~{}3394 trials/session).
Each dataset covers distinct paradigms and acquisition settings, supporting model generalization.

\textbf{\emph{\textcolor{CaseOrange}{EvaluationMetrics}}}

• Classification Accuracy: Percentage of correctly predicted EEG trials across classes, reflecting decoding performance. \\
• Cohen's Kappa: A statistical measure of inter-rater agreement accounting for chance, used to evaluate classification reliability. \\
• Wilcoxon Signed-Rank Test: Non-parametric test for statistical significance of performance differences between models or ablation settings. \\
• Training Efficiency: Measured as convergence speed (epochs to stable loss/accuracy) and per-epoch training time. \\
• Interpretability: Qualitatively assessed via t-SNE clustering of learned features, CAM heatmaps, and CAT spatial-temporal mappings. 

\textbf{\emph{\textcolor{CaseOrange}{ExpectedOutcome}}}

EEG Conformer achieves state-of-the-art classification accuracy and kappa across all three datasets:
on BCI IV 2a, average accuracy 78.66\% (↑10.91\% over FBCSP), kappa 0.7155; on BCI IV 2b, 84.63\%
accuracy, kappa 0.6926; on SEED, 95.30\% accuracy, kappa 0.9295. Ablation studies show a 6.02\%
average accuracy drop without the self-attention module. Visualization confirms the model's focus on
paradigm-relevant brain regions, and the architecture demonstrates efficient convergence and
robustness to parameter variations, establishing a strong new backbone for general EEG decoding.

\end{tcolorbox}

\begin{tcolorbox}[
    breakable,
    title={Example of Idea Generation in Physics},
    colback=LighterGray,
    colframe=DeepPurple,
    colbacktitle=DeepPurple,
    coltitle=White,
]
\textbf{\emph{\textcolor{DeepPurple}{Question}}}

You are a top-tier researcher in your field. Based on the following context, please generate a novel and detailed research proposal.

\textbf{\emph{\textcolor{CaseOrange}{RelatedWork}}}

• eSEN-30M-OMat: An equivariant graph neural network tailored for materials, achieving strong accuracy via large-scale message passing, but limited to domain-specific datasets and lacking generalization across molecules or surfaces. \\
• GemNet-OC20: A graph neural network for catalysis using geometric embeddings, excelling in adsorption energy prediction but focused solely on catalysis, without material or molecular generalization. \\
• MACE: A foundation model for atomistic materials chemistry that demonstrates excellent transferability within the organic molecule domain, but struggles to generalize simultaneously to diverse materials and catalytic systems. \\
• EquiformerV2 : An advanced equivariant transformer model that achieves strong performance on domain-specific materials and catalysis benchmarks but is not trained for multi-domain or multi-DFT-task generalization. \\
• ORB v3: A scalable neural network potential capable of efficient simulation at scale, but designed primarily for periodic materials, with limited multi-domain applicability. \\
• Universal Graph Deep Learning Potentials: Aim to provide comprehensive coverage across the periodic table, yet tend not to generalize to molecules or catalysis due to distribution shifts and differing DFT settings. \\
• Pre-training with Fine-tuning: Large models are pre-trained on broad datasets and fine-tuned for specific tasks, yielding high accuracy but still requiring domain adaptation; true zero-shot generalization across tasks remains unproven. 

\textbf{\emph{\textcolor{CaseOrange}{Challenges}}}

• Developing a single MLIP capable of high-fidelity, zero-shot generalization across vastly different chemical domains, including materials, molecules, catalysis, molecular crystals, and MOFs. \\
• Scaling model and dataset size without sacrificing inference speed or memory efficiency, especially for long-running atomistic simulations involving thousands to hundreds of thousands of atoms. \\
• Reconciling and learning from datasets with heterogeneous DFT settings, label distributions, elemental coverage, and system sizes. \\
• Maintaining energy conservation, physical symmetry (rotational equivariance), and smoothness of the potential energy surface during multi-task, multi-domain learning. \\
• Efficiently training and deploying ultra-large models (up to billions of parameters) under memory and compute constraints. 

\textbf{\emph{\textcolor{CaseOrange}{Limitation}}}

Most existing MLIPs are either specialized for a single chemical domain or require fine-tuning to
achieve high accuracy in new domains. They do not robustly generalize across materials, molecules,
and catalytic systems with varying DFT settings. Further, attempts to scale model capacity often
degrade inference efficiency, and models are typically trained on smaller, less diverse datasets,
limiting their practical universality.

\textbf{\emph{\textcolor{CaseOrange}{Motivation}}}

The demand for rapid, accurate, and general-purpose atomistic simulations is increasing in fields
such as drug discovery, energy storage, and catalysis. However, DFT is computationally prohibitive,
and existing ML surrogates lack universality. The confluence of new, massive multi-domain datasets
and insights from scaling laws in deep learning presents the opportunity to create a single, highly
scalable MLIP that achieves state-of-the-art accuracy, speed, and generalization across all relevant
chemical domains.

\textbf{\emph{\textcolor{CaseOrange}{TaskObjective}}}

To design, train, and evaluate a family of universal machine learning interatomic potentials (UMA)
that achieve high accuracy, computational efficiency, and generalization across diverse chemical and
materials domains, using the largest multi-domain atomic datasets to date.

\textbf{\emph{\textcolor{CaseOrange}{ExistingSolutions}}}

• eSEN: Utilizes equivariant message passing with spherical harmonics for high accuracy in materials, but lacks multi-domain scalability. \\
• GemNet: Employs geometric embeddings for catalysis; effective on domain-specific adsorption tasks but does not generalize to other domains. \\
• MACE: Foundation model for molecules, demonstrates good transferability within molecular datasets; struggles with cross-domain and multi-task generalization. \\
• EquiformerV2: Equivariant transformer with improved scaling for materials and catalysis, but not designed for simultaneous multi-domain learning. \\
• ORB v3: Focuses on scalable neural network potentials for materials, achieving high throughput but lacks coverage of molecular and catalytic tasks. \\
• Fine-tuned Foundation Models: Pre-train on large datasets, then fine-tune for each target domain; yields high performance but necessitates domain-specific adaptation and fails to provide universal zero-shot performance. 

\textbf{\emph{\textcolor{DeepPurple}{Reference Answer}}}

\textbf{\emph{\textcolor{CaseOrange}{Idea}}}

UMA introduces a family of universal MLIPs trained on nearly 500M multi-domain atomic structures,
leveraging an efficient Mixture of Linear Experts (MoLE) architecture for scalable capacity without
inference overhead. Empirical scaling laws inform model/data sizing, while unified embeddings and
referencing schemes enable seamless multi-DFT-task learning, delivering state-of-the-art accuracy
and speed across chemistry and materials science domains.

\textbf{\emph{\textcolor{CaseOrange}{ImplementationSteps}}}

• 1: Data aggregation and preprocessing: curate and normalize OMat24, OMol25, OC20++, OMC25, and ODAC25, applying energy referencing and label normalization. \\
• 2: Model design: configure eSEN-based GNN with integrated MoLE layers; implement global embeddings for charge, spin, and DFT task. \\
• 3: MoLE routing: compute expert coefficients from global system features and pre-merge expert weights for efficient inference. \\
• 4: Stage 1 training: pre-train the model in BF16 on direct force prediction with max-atom batching and reduced neighbors. \\
• 5: Stage 2 fine-tuning: switch to FP32 precision and auto-grad conservative heads, increasing neighbor count for energy/force conservation. \\
• 6: Memory/computation optimization: employ graph parallelism, FSDP, and activation checkpointing for large-scale training. \\
• 7: Model selection: use empirical scaling laws to determine optimal model and dataset size for given compute budget. \\
• 8: Evaluation: benchmark UMA models on held-out splits and established tasks across materials, catalysis, molecules, molecular crystals, and MOFs. 

\textbf{\emph{\textcolor{CaseOrange}{ImplementationOrder}}}

• 1-2 \\
• 2-3 \\
• 3-4 \\
• 4-5 \\
• 5-6 \\
• 6-7 \\
• 7-8 

\textbf{\emph{\textcolor{CaseOrange}{Data}}}

UMA is trained on five large-scale datasets: OMat24 (bulk materials, 100M entries, 89 elements,
VASP-PBE), OMol25 (molecules, 75M entries, 83 elements, ORCA-$\omega$B97M-V), OC20++ (catalysis, 229M, 56
elements, VASP-RPBE), OMC25 (molecular crystals, 25M, 12 elements, VASP-PBE+D3), and ODAC25 (MOFs,
29M, 70 elements, VASP-PBE+D3). Combined, the data covers \~{}459M structures and >30B atoms with
near-complete elemental coverage and diverse DFT settings.

\textbf{\emph{\textcolor{CaseOrange}{EvaluationMetrics}}}

• Mean Absolute Error (MAE): Measures average absolute deviation between predicted and reference energies, forces (in meV/Å), and stresses (meV/Å\^{}3). \\
• Adsorption Energy Success Rate: Percentage of cases where the predicted global minimum adsorption energy is within 0.1 eV of the DFT minimum (AdsorbML benchmark). \\
• F1 Score: Assesses binary/classification performance on Matbench Discovery for stability predictions. \\
• Energy Conservation: Degree to which predicted forces/energies conserve energy over molecular dynamics trajectories (NVE MD benchmarks). \\
• Simulation Throughput: Number of inference steps per second for fixed system sizes (1k, 10k, 100k atoms) on a single GPU. \\
• Out-of-Domain Generalization: Performance on OOD splits, such as high-entropy alloys and novel molecular/crystal structures. \\
• Phonon and Elastic Property Accuracy: MAE for phonon frequencies, free energies, elastic moduli, and related properties pertinent to material science benchmarks. 

\textbf{\emph{\textcolor{CaseOrange}{ExpectedOutcome}}}

UMA achieves state-of-the-art or superior accuracy on diverse benchmarks (e.g., up to 25\%
improvement in AdsorbML success rate, \~{}80\% reduction in OC20 adsorption energy error vs. prior
SOTA, chemical accuracy for ligand strain energy). The models support efficient simulation of >100k
atoms with no inference penalty from increased capacity. UMA provides reliable, energy-conserving
predictions across all major chemical domains, demonstrating that a single model can match or
surpass specialized models in both zero-shot and fine-tuned settings.

\end{tcolorbox}

\subsubsection{Dry Experiment}

\

\input{sections/codes/code_00}
\input{sections/codes/code_01}

\subsubsection{Wet Experiment}

\

\begin{tcolorbox}[
    breakable,
    title=Example of Wet Experiment in Life,
    colback=LighterGray,
    colframe=DeepPurple,
    colbacktitle=DeepPurple,
    coltitle=White,
]
\textbf{\emph{\textcolor{DeepPurple}{Background}}}

Cancer development involves genetic and epigenetic alterations that enable tumor cells to evade immune detection by creating an immunosuppressive microenvironment. A key mechanism of immune evasion is mediated by the programmed death-ligand 1 (PD-L1), expressed on tumor and immune cells, which binds to programmed death-1 (PD-1) and B7.1 (CD80) receptors on T cells. This interaction inhibits T-cell migration, proliferation, and cytotoxic function, thereby limiting tumor cell killing. Blocking PD-L1 can restore antitumor immunity by reactivating suppressed T cells.

An engineered humanized monoclonal antibody targeting PD-L1 has been developed to inhibit its interaction with PD-1 and B7.1, without affecting PD-1's interaction with PD-L2, preserving peripheral tolerance. This antibody is designed with an Fc domain modification to prevent antibody-dependent cellular cytotoxicity, avoiding depletion of activated T cells.

Clinical studies involving patients with advanced solid tumors treated with this anti-PD-L1 antibody demonstrated safety and tolerability across a range of doses, with manageable adverse events such as fatigue and low-grade fever. Immune activation markers, including proliferating $\mathrm{CD8}^+$ T cells and interferon-gamma (IFN-$\gamma$), increased during treatment.

Efficacy assessments revealed objective responses in multiple cancer types, notably non-small cell lung cancer (NSCLC), melanoma, and renal cell carcinoma. Importantly, clinical responses correlated strongly with pre-treatment PD-L1 expression on tumor-infiltrating immune cells rather than tumor cells themselves. High PD-L1 expression on immune cells was associated with higher response rates and longer progression-free survival. Additional biomarkers linked to response included T-helper type 1 (TH1) gene expression and $\mathrm{CTLA4}$ expression, while fractalkine ($\mathrm{CX3CL1}$) expression correlated with disease progression.

On-treatment biopsies of responding tumors showed increased immune cell infiltration, tumor necrosis, and upregulation of PD-L1 and IFN-$\gamma$, indicating reactivation of antitumor immunity. Non-responding tumors exhibited patterns of immunological ignorance (lack of immune infiltration), non-functional immune responses (immune cells present but inactive), or excluded infiltrates (immune cells restricted to tumor margins), with no significant PD-L1 upregulation or T-cell activation.

Blood-based biomarkers showed increases in IFN-$\gamma$-inducible chemokines and activated cytotoxic T cells early in treatment, reflecting systemic immune activation, though these changes did not clearly distinguish responders from non-responders.

These findings support the concept that pre-existing antitumor immunity suppressed by PD-L1 can be reinvigorated by PD-L1 blockade, leading to durable clinical responses. The presence and localization of PD-L1 expression, particularly on tumor-infiltrating immune cells, serve as predictive biomarkers for response. Understanding the immune microenvironment of non-responders may reveal additional mechanisms of immune resistance and guide combination immunotherapy strategies to enhance the cancer immunity cycle.

\textbf{\emph{\textcolor{DeepPurple}{Action Pool}}}

\begin{lstlisting}
<Fix_tissue_in_formalin>(tissue, fixative)
    Args:
        tissue: Tissue sample to be fixed
        fixative: Formalin solution
    Returns:
        Fixed tissue sample

<Embed_tissue_in_paraffin>(fixed_tissue)
    Args:
        fixed_tissue: Formalin-fixed tissue
    Returns:
        FFPE tissue block

<Section_tissue>(tissue_block, thickness)
    Args:
        tissue_block: Paraffin-embedded tissue block
        thickness: Section thickness in micrometers
    Returns:
        Tissue sections

<Stain_with_antibody>(tissue_section, antibody, concentration)
    Args:
        tissue_section: Tissue section on slide
        antibody: Primary antibody
        concentration: Antibody concentration
    Returns:
        Antibody-labeled tissue section

<Visualize_with_DAB>(stained_section)
    Args:
        stained_section: Antibody-stained section
    Returns:
        DAB-visualized section

<Counterstain_with_hematoxylin>(section)
    Args:
        section: DAB-stained section
    Returns:
        Counterstained section

<Score_IHC_staining>(stained_section, cell_type)
    Args:
        stained_section: Complete IHC-stained section
        cell_type: Type of cells to score (TC or IC)
    Returns:
        IHC score (0-3)

<Incubate_with_primary_antibodies>(section, antibody1, antibody2, temperature)
    Args:
        section: FFPE tissue section
        antibody1: First primary antibody
        antibody2: Second primary antibody
        temperature: Incubation temperature
    Returns:
        Dual-antibody labeled section

<Detect_with_fluorescence>(labeled_section, detection_system, fluorophore)
    Args:
        labeled_section: Antibody-labeled section
        detection_system: Detection reagent system
        fluorophore: Fluorescent label
    Returns:
        Fluorescently labeled section

<Extract_DNA_from_FFPE>(tissue_section, extraction_kit)
    Args:
        tissue_section: FFPE tissue section
        extraction_kit: DNA extraction kit
    Returns:
        Isolated DNA

<Extract_RNA_from_FFPE>(tissue_section, extraction_kit)
    Args:
        tissue_section: FFPE tissue section
        extraction_kit: RNA extraction kit
    Returns:
        Isolated RNA

<Perform_gene_expression_analysis>(RNA_sample, platform, gene_panel)
    Args:
        RNA_sample: Isolated RNA
        platform: Analysis platform
        gene_panel: Panel of genes to analyze
    Returns:
        Gene expression data

<Collect_blood_sample>(patient, tube_type, volume)
    Args:
        patient: Patient identifier
        tube_type: Collection tube type
        volume: Sample volume
    Returns:
        Blood sample

<Isolate_plasma>(blood_sample, centrifuge_speed, time)
    Args:
        blood_sample: Whole blood sample
        centrifuge_speed: Centrifugation speed
        time: Centrifugation time
    Returns:
        Plasma sample

<Analyze_cytokines_by_ELISA>(plasma_sample, cytokine_panel)
    Args:
        plasma_sample: Isolated plasma
        cytokine_panel: Panel of cytokines to measure
    Returns:
        Cytokine levels

<Perform_FACS_analysis>(blood_sample, antibody_panel)
    Args:
        blood_sample: Blood sample
        antibody_panel: Panel of antibodies for staining
    Returns:
        Cell population data

<Administer_MPDL3280A>(patient, dose, route)
    Args:
        patient: Patient identifier
        dose: Drug dose in mg/kg
        route: Administration route
    Returns:
        Treated patient

<Collect_tumor_biopsy>(patient, timepoint)
    Args:
        patient: Patient identifier
        timepoint: Collection timepoint
    Returns:
        Tumor biopsy sample

<Evaluate_tumor_response>(patient, imaging_method, criteria)
    Args:
        patient: Patient identifier
        imaging_method: Imaging modality
        criteria: Response evaluation criteria
    Returns:
        Tumor response assessment

<Store_sample>(sample, temperature)
    Args:
        sample: Biological sample
        temperature: Storage temperature
    Returns:
        Stored sample
\end{lstlisting}

\textbf{\emph{\textcolor{DeepPurple}{Answer}}}

\begin{center}
    \centering
    \captionsetup{type=figure}
    \includegraphics[width=0.98\linewidth]{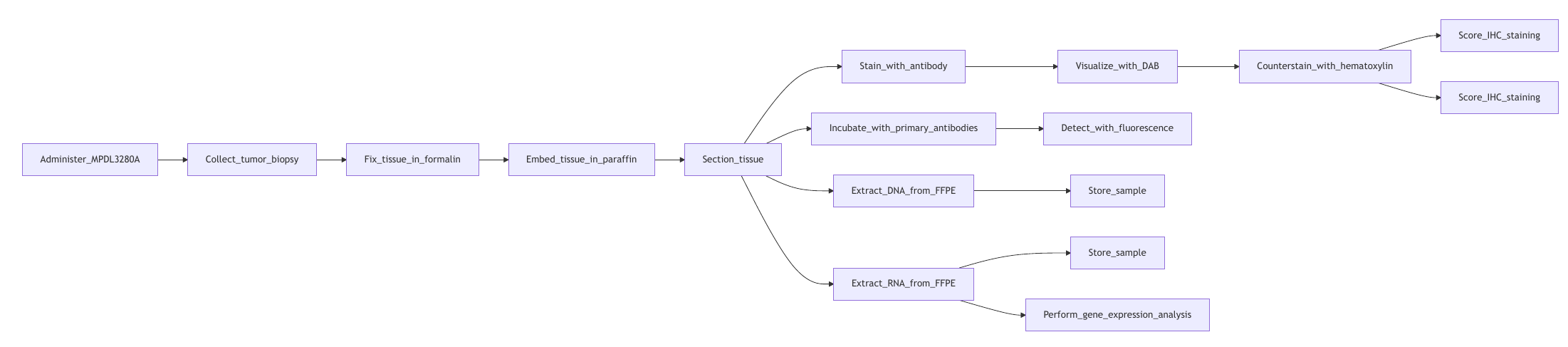}
\end{center}

\end{tcolorbox}

\begin{tcolorbox}[
    breakable,
    title=Example of Wet Experiment in Material,
    colback=LighterGray,
    colframe=DeepPurple,
    colbacktitle=DeepPurple,
    coltitle=White,
]
\textbf{\emph{\textcolor{DeepPurple}{Background}}}

Low-grade heat, abundant in environments such as solar radiation, body heat, and industrial waste, presents a significant opportunity for energy harvesting. Thermogalvanic cells (TGCs) convert such heat directly into electricity via redox reactions at electrodes maintained at different temperatures. The thermopower of these cells, a measure of voltage generated per unit temperature difference, depends primarily on the entropy change ($\Delta S$) and concentration difference ($\Delta C$) of redox species between hot and cold electrodes. Traditional aqueous redox electrolytes exhibit limited thermopowers, typically below $2~\mathrm{mV\,K}^{-1}$, constraining their practical efficiency.

Recent advances focus on enhancing thermopower by increasing $\Delta S$ through solvent reorganization or structural changes of redox couples, and by increasing $\Delta C$ via selective complexation or confinement of redox ions. Thermoresponsive polymers have been employed to induce temperature-dependent interactions with redox ions, enabling polarization switching between $n$-type and $p$-type behavior, which reverses the direction of electron flow and expands operational versatility.

A notable development involves the use of methylcellulose (MC), a biocompatible, low-cost polymer exhibiting temperature-dependent hydrophilic-to-hydrophobic transitions. When incorporated into an aqueous iodide/triiodide ($\mathrm{I}^-/\mathrm{I}_3^-$) redox electrolyte, MC interacts hydrophobically with $\mathrm{I}_3^-$ ions above its gelation temperature, reducing free $\mathrm{I}_3^-$ concentration at the hot electrode. This interaction induces a polarization switch from $n$-type to $p$-type thermopower and simultaneously enhances both $\Delta S$ and $\Delta C$ due to gelation and ion complexation effects.

Further enhancement is achieved by adding potassium chloride (KCl), which complexes with MC and $\mathrm{I}_3^-$ ions, promoting reversible aggregation and dissociation processes. This salt-induced complexation lowers the gelation and polarization transition temperatures and significantly amplifies thermopower. The optimized ternary electrolyte ($\mathrm{I}^-/\mathrm{I}_3^- + 2~\mathrm{wt}\%~\mathrm{MC} + 0.3~\mathrm{M}~\mathrm{KCl}$) exhibits record-high thermopowers of approximately $-8.18~\mathrm{mV\,K}^{-1}$ ($n$-type) and $9.62~\mathrm{mV\,K}^{-1}$ ($p$-type), an order of magnitude greater than pristine electrolytes.

Electrochemical characterization reveals improved electron transfer kinetics and ionic conductivity in the ternary system, resulting in higher current densities and lower internal resistance in TGCs. Under a $15~^\circ\mathrm{C}$ temperature difference, single $n$-type and $p$-type cells achieve maximum power outputs of $27.78~\mu\mathrm{W}$ and $80.47~\mu\mathrm{W}$, respectively, with normalized power densities surpassing previous iodide/triiodide-based systems.

This approach demonstrates that integrating thermoresponsive biopolymers with salt-induced complexation in redox electrolytes can substantially boost thermogalvanic performance. The findings open pathways for cost-effective, scalable liquid thermocells capable of efficient low-grade heat harvesting, leveraging abundant, environmentally benign materials and tunable electrolyte properties for enhanced energy conversion.

\textbf{\emph{\textcolor{DeepPurple}{Action Pool}}}

\begin{lstlisting}
<Prepare pristine I-/I3- electrolyte>(KI_amount, I2_amount, water_volume)
    Args:
        KI_amount: Amount of potassium iodide
        I2_amount: Amount of iodine
        water_volume: Volume of deionized water
    Returns:
        Pristine I-/I3- electrolyte solution

<Heat electrolyte solution>(electrolyte, temperature)
    Args:
        electrolyte: Electrolyte solution to heat
        temperature: Target temperature
    Returns:
        Heated electrolyte solution

<Add methylcellulose to electrolyte>(electrolyte, MC_amount)
    Args:
        electrolyte: Heated electrolyte solution
        MC_amount: Amount of methylcellulose powder
    Returns:
        Binary electrolyte with MC

<Stir solution magnetically>(solution, duration)
    Args:
        solution: Solution to stir
        duration: Stirring time
    Returns:
        Homogeneous solution

<Add KCl to binary electrolyte>(binary_electrolyte, KCl_concentration)
    Args:
        binary_electrolyte: I-/I3- + MC electrolyte
        KCl_concentration: Molar concentration of KCl
    Returns:
        Ternary electrolyte

<Store electrolyte in refrigerator>(electrolyte, temperature, duration)
    Args:
        electrolyte: Prepared electrolyte
        temperature: Storage temperature
        duration: Storage time
    Returns:
        Stored electrolyte ready for use

<Fill thermocell cavity>(electrolyte, volume)
    Args:
        electrolyte: Prepared electrolyte
        volume: Volume to fill
    Returns:
        Filled thermocell

<Set cold electrode temperature>(thermocell, temperature)
    Args:
        thermocell: Assembled thermocell
        temperature: Cold electrode temperature
    Returns:
        Thermocell with controlled cold electrode

<Heat hot electrode gradually>(thermocell, target_temperature)
    Args:
        thermocell: Thermocell setup
        target_temperature: Maximum hot electrode temperature
    Returns:
        Thermocell with temperature gradient

<Record open-circuit voltage>(thermocell, data_logger)
    Args:
        thermocell: Operating thermocell
        data_logger: Data acquisition device
    Returns:
        Voltage-temperature data

<Measure electrode temperatures>(thermocell, thermocouples)
    Args:
        thermocell: Operating thermocell
        thermocouples: Temperature sensors
    Returns:
        Temperature measurements

<Connect external load>(thermocell, potentiometer)
    Args:
        thermocell: Operating thermocell
        potentiometer: Variable resistance device
    Returns:
        Thermocell with load circuit

<Record current and voltage>(thermocell, source_meter, data_logger)
    Args:
        thermocell: Operating thermocell under load
        source_meter: Current measurement device
        data_logger: Voltage measurement device
    Returns:
        Power generation data

<Perform UV-Vis spectroscopy>(sample, spectrometer)
    Args:
        sample: Electrolyte sample
        spectrometer: UV-Vis instrument
    Returns:
        Absorption spectrum data

<Dilute sample for analysis>(sample, dilution_factor)
    Args:
        sample: Concentrated sample
        dilution_factor: Dilution ratio
    Returns:
        Diluted sample

<Filter electrolyte sample>(sample)
    Args:
        sample: Raw electrolyte sample
    Returns:
        Filtered sample

<Perform cyclic voltammetry>(electrolyte, potentiostat, scan_rate)
    Args:
        electrolyte: Test electrolyte
        potentiostat: Electrochemical instrument
        scan_rate: Voltage scanning rate
    Returns:
        CV curves

<Dry electrolyte under vacuum>(electrolyte, temperature, duration)
    Args:
        electrolyte: Liquid electrolyte
        temperature: Drying temperature
        duration: Drying time
    Returns:
        Dried electrolyte powder

<Perform FTIR spectroscopy>(sample, FTIR_instrument)
    Args:
        sample: Dried powder sample
        FTIR_instrument: FTIR spectrometer
    Returns:
        FTIR spectrum

<Measure ionic conductivity>(electrolyte, conductivity_meter, temperature_range)
    Args:
        electrolyte: Test electrolyte
        conductivity_meter: Conductivity measurement device
        temperature_range: Temperature range for measurement
    Returns:
        Conductivity vs temperature data
\end{lstlisting}

\textbf{\emph{\textcolor{DeepPurple}{Answer}}}

\begin{center}
    \centering
    \captionsetup{type=figure}
    \includegraphics[width=0.98\linewidth]{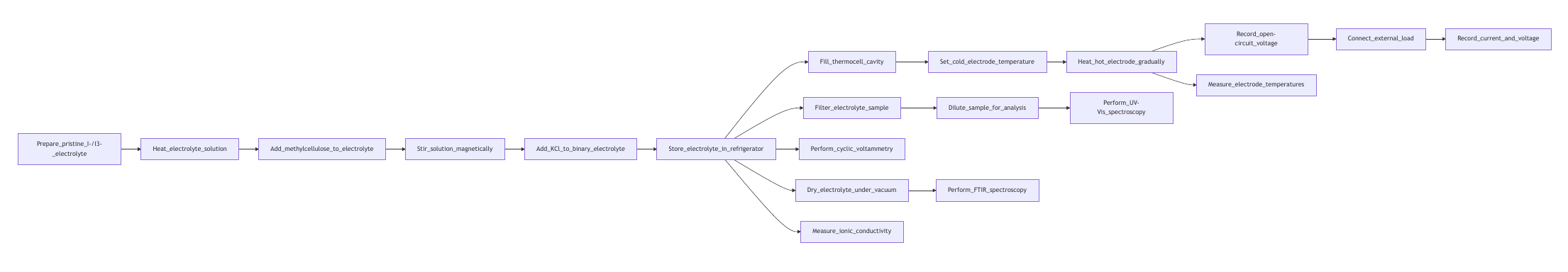}
\end{center}

\end{tcolorbox}

\begin{tcolorbox}[
    breakable,
    title=Example of Wet Experiment in Physics,
    colback=LighterGray,
    colframe=DeepPurple,
    colbacktitle=DeepPurple,
    coltitle=White,
]
\textbf{\emph{\textcolor{DeepPurple}{Background}}}

This research domain focuses on the analysis and synthesis of nonlinear discrete-time systems, digital filters, and chaotic circuits, emphasizing stability, noise quantification, and complex dynamical behaviors.

In digital filter design, quantization noise arising from finite word-length effects is a critical concern. Methods have been developed to compute noise covariance matrices associated with extended digital filters, enabling the evaluation of roundoff noise not only at storage nodes but also at other internal nodes. These computations involve iterative matrix summations and transformations, where matrices representing system dynamics and noise propagation are manipulated to yield noise covariance matrices. The approach typically uses state-space representations and involves solving matrix equations that incorporate system matrices and noise input vectors, allowing for precise quantification of noise effects in fixed-point digital filters.

In nonlinear discrete-time systems with slope-restricted nonlinearities, absolute stability criteria are essential for ensuring asymptotic stability in the large. A frequency-domain criterion has been formulated for single-input single-output Lur'e-type systems, where the nonlinearity satisfies sector and slope restrictions. The criterion involves verifying an inequality over the unit circle in the complex plane, incorporating the system's frequency response and parameters bounding the nonlinearity's slope. This approach extends the system order and applies Lyapunov function techniques to establish sufficient conditions for global asymptotic stability, providing a rigorous tool for stability analysis in nonlinear discrete-time control systems.

The study of chaotic attractors in simple autonomous circuits reveals that even minimal configurations with piecewise-linear nonlinear elements can exhibit complex chaotic dynamics. A third-order reciprocal circuit with a single nonlinear resistor characterized by a three-segment piecewise-linear function demonstrates chaotic attractors with structures distinct from classical examples like the Lorenz and Rössler attractors. The system's dynamics are governed by coupled differential equations describing voltages and currents in capacitors and inductors, with nonlinear feedback inducing chaos. The attractor includes invariant sets containing equilibria with specific eigenvalue configurations, and its persistence is confirmed over ranges of circuit parameters. This research highlights the role of circuit reciprocity and nonlinear characteristics in generating and sustaining chaotic behavior, contributing to the understanding of nonlinear dynamics in electrical circuits.

Collectively, these areas integrate advanced mathematical tools—such as state-space modeling, frequency-domain analysis, Lyapunov stability theory, and nonlinear dynamics—to address challenges in system stability, noise management, and chaotic behavior in engineering systems.

\textbf{\emph{\textcolor{DeepPurple}{Action Pool}}}

\begin{lstlisting}
<Build circuit with components>(capacitor1, capacitor2, inductor, resistor)
    Args:
        capacitor1: First capacitor component
        capacitor2: Second capacitor component
        inductor: Inductor component
        resistor: Nonlinear resistor component
    Returns:
        Assembled circuit

<Set capacitor value>(capacitor, capacitance_value)
    Args:
        capacitor: Target capacitor
        capacitance_value: Capacitance value to set
    Returns:
        Configured capacitor

<Set inductor value>(inductor, inductance_value)
    Args:
        inductor: Target inductor
        inductance_value: Inductance value to set
    Returns:
        Configured inductor

<Configure nonlinear resistor>(resistor, conductance, slope_parameters)
    Args:
        resistor: Nonlinear resistor component
        conductance: Conductance value G
        slope_parameters: Piecewise-linear slope values
    Returns:
        Configured nonlinear resistor

<Connect circuit elements>(circuit, connection_scheme)
    Args:
        circuit: Circuit with components
        connection_scheme: Wiring configuration
    Returns:
        Connected circuit

<Initialize circuit state>(circuit, initial_conditions)
    Args:
        circuit: Connected circuit
        initial_conditions: Initial voltages and current values
    Returns:
        Initialized circuit

<Set simulation parameters>(step_size, integration_method)
    Args:
        step_size: Time step for numerical integration
        integration_method: Numerical method to use
    Returns:
        Simulation configuration

<Run circuit simulation>(circuit, simulation_config, time_duration)
    Args:
        circuit: Initialized circuit
        simulation_config: Simulation parameters
        time_duration: Total simulation time
    Returns:
        Simulation results with time series data

<Extract voltage trajectories>(simulation_results, voltage_nodes)
    Args:
        simulation_results: Output from simulation
        voltage_nodes: Specific voltage points to extract
    Returns:
        Voltage time series data

<Extract current trajectories>(simulation_results, current_branch)
    Args:
        simulation_results: Output from simulation
        current_branch: Specific current branch to extract
    Returns:
        Current time series data

<Generate phase portrait>(voltage_data, current_data, projection_plane)
    Args:
        voltage_data: Voltage trajectories
        current_data: Current trajectories
        projection_plane: 2D plane for projection
    Returns:
        Phase portrait visualization

<Identify attractor characteristics>(phase_portraits, trajectory_data)
    Args:
        phase_portraits: Generated phase portraits
        trajectory_data: Complete system trajectories
    Returns:
        Attractor properties and structure

<Vary circuit parameters>(circuit, parameter_name, parameter_range)
    Args:
        circuit: Base circuit configuration
        parameter_name: Parameter to vary
        parameter_range: Range of values to test
    Returns:
        Parameter sweep results

<Analyze bifurcation behavior>(parameter_sweep_results, stability_criteria)
    Args:
        parameter_sweep_results: Results from parameter variation
        stability_criteria: Criteria for stability analysis
    Returns:
        Bifurcation analysis results

<Identify periodic orbits>(trajectory_data, newton_iteration_params)
    Args:
        trajectory_data: System trajectories
        newton_iteration_params: Parameters for Newton iteration
    Returns:
        Periodic orbit characteristics
\end{lstlisting}

\textbf{\emph{\textcolor{DeepPurple}{Answer}}}

\begin{center}
    \centering
    \captionsetup{type=figure}
    \includegraphics[width=0.98\linewidth]{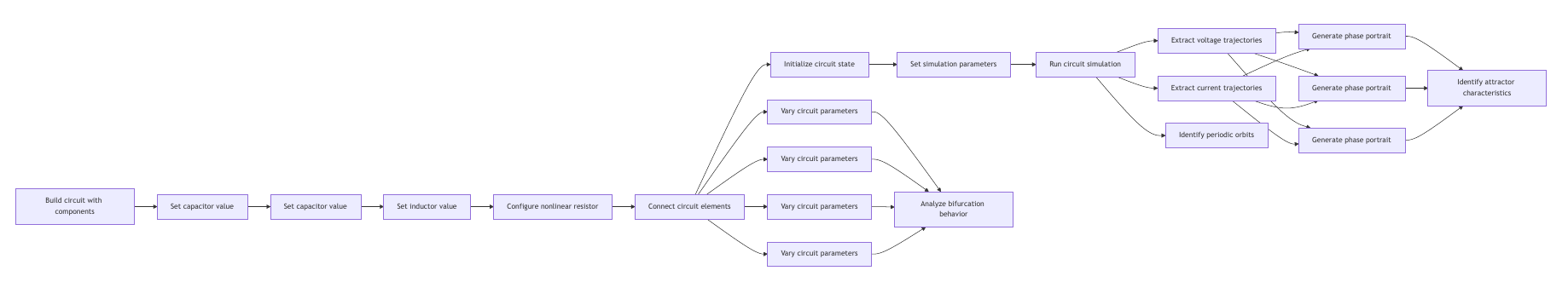}
\end{center}

\end{tcolorbox}

\subsubsection{Experimental Reasoning}

\begin{tcolorbox}[
    breakable,
    title=Example of Experimental Reasoning in Astronomy,
    colback=LighterGray,
    colframe=DeepPurple,
    colbacktitle=DeepPurple,
    coltitle=White,
]

\textbf{\emph{\textcolor{DeepPurple}{Images}}}

\begin{center}
    \centering
    \captionsetup{type=figure}
    \includegraphics[width=0.98\linewidth]{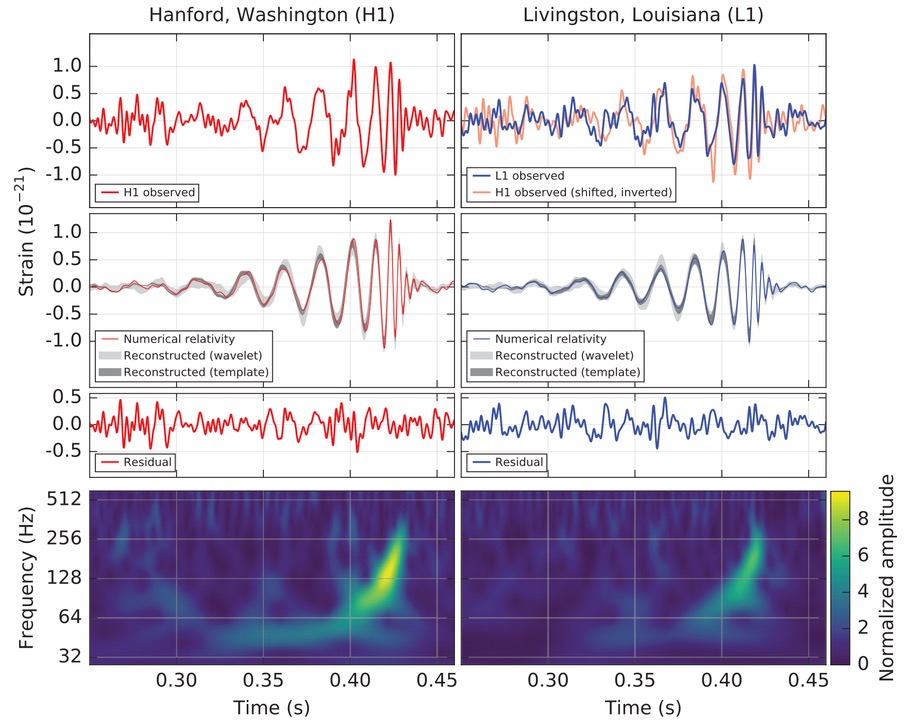}
\end{center}

\begin{center}
    \centering
    \captionsetup{type=figure}
    \includegraphics[width=0.5\linewidth]{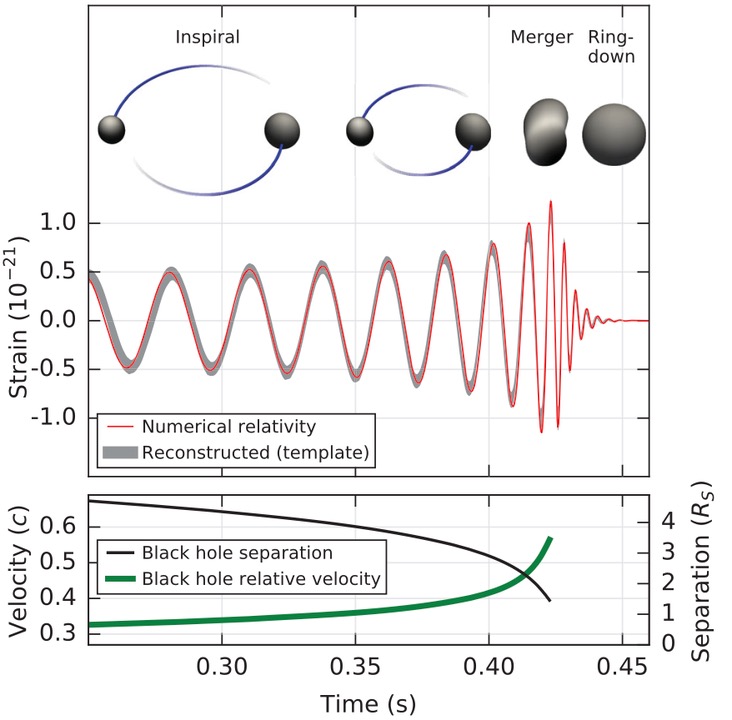}
\end{center}

\begin{center}
    \centering
    \captionsetup{type=figure}
    \includegraphics[width=0.5\linewidth]{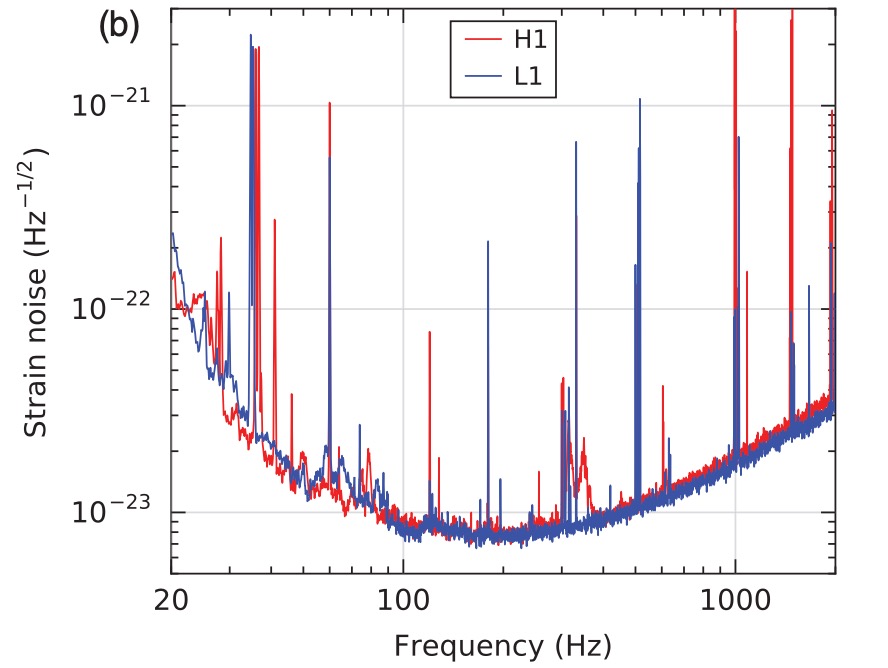}
\end{center}

\textbf{\emph{\textcolor{DeepPurple}{Question}}}

Using the time–frequency ridge data points $(t,f)$ from the first image, estimate the chirp mass $M_c$ via the Newtonian approximation and $t_c = 0$. From the second image, the noise-weighted integral is:
\begin{equation}
    J = f \int_{f_{min} \rightarrow f_{max}} f^(-7/3)/S_n(f) df = 1.3826254536\times10^{60} \text{(SI units)}.
\end{equation}
From the three image, the network SNR is $\rho_{net} = 24$ (detector factor $F=1$).  Under the stationary phase approximation, Solve for the luminosity distance $D_L$ using and select the answer (in Mpc, rounded) from options 0 to 9 below.

\textbf{\emph{\textcolor{DeepPurple}{Options}}}

\begin{enumerate}[label=\Alph*.]
    \item 100
    \item 150
    \item 210
    \item 270
    \item 350
    \item 410
    \item 500
    \item 620
    \item 750
    \item 1000
\end{enumerate}

\textbf{\emph{\textcolor{DeepPurple}{Steps}}}

\textbf{\textcolor{CaseOrange}{Step 1.}}

\begin{center}
    \centering
    \captionsetup{type=figure}
    \includegraphics[width=0.98\linewidth]{imgs/sci-exp-case/astronomy/01.png}
\end{center}

\textbf{\textcolor{CaseOrange}{Step 2.}} Read three points $(t, f)$, calculate $X=f^{-8/3}$ and $Y=-t$, and use the least squares fitting to obtain the slope $K$.

\textbf{\textcolor{CaseOrange}{Step 3.}} Quality of the solution by $K$ chirp: $M_c^3 = (c/G) \times [((5/256) \pi ^ {- 8/3})/K] ^ {3/5}$.

\textbf{\textcolor{CaseOrange}{Step 4.}}

\begin{center}
    \centering
    \captionsetup{type=figure}
    \includegraphics[width=0.5\linewidth]{imgs/sci-exp-case/astronomy/02.png}
\end{center}

\textbf{\textcolor{CaseOrange}{Step 5.}} The provided value of $J$.

\textbf{\textcolor{CaseOrange}{Step 6.}}

\begin{center}
    \centering
    \captionsetup{type=figure}
    \includegraphics[width=0.5\linewidth]{imgs/sci-exp-case/astronomy/03.png}
\end{center}

\textbf{\textcolor{CaseOrange}{Step 7.}} Read $\rho_{net} = 24$ and the direction factor $F=1$.

\textbf{\textcolor{CaseOrange}{Step 8.}} Substitution $\rho_{net} ^ 2 = 4 A ^ 2 J$ and $A = (1 / D_L) {5/24} \pi ^ {- 2/3} M_c (G/c ^ 3) ^ {5/6 }$, work out $D_L$.

\textbf{\textcolor{CaseOrange}{Step 9.}} Convert $D_L$ to Mpc and round it to the nearest integer

\textbf{\emph{\textcolor{DeepPurple}{Answer}}}

F

\end{tcolorbox}

\begin{tcolorbox}[
    breakable,
    title=Example of Experimental Reasoning in Chemistry,
    colback=LighterGray,
    colframe=DeepPurple,
    colbacktitle=DeepPurple,
    coltitle=White,
]

\textbf{\emph{\textcolor{DeepPurple}{Images}}}

\begin{center}
    \centering
    \captionsetup{type=figure}
    \includegraphics[width=0.98\linewidth]{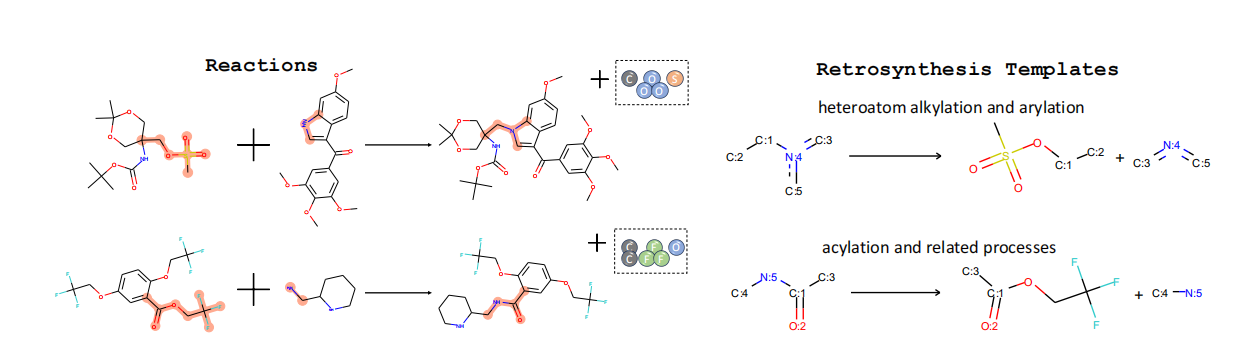}
\end{center}

\begin{center}
    \centering
    \captionsetup{type=figure}
    \includegraphics[width=0.98\linewidth]{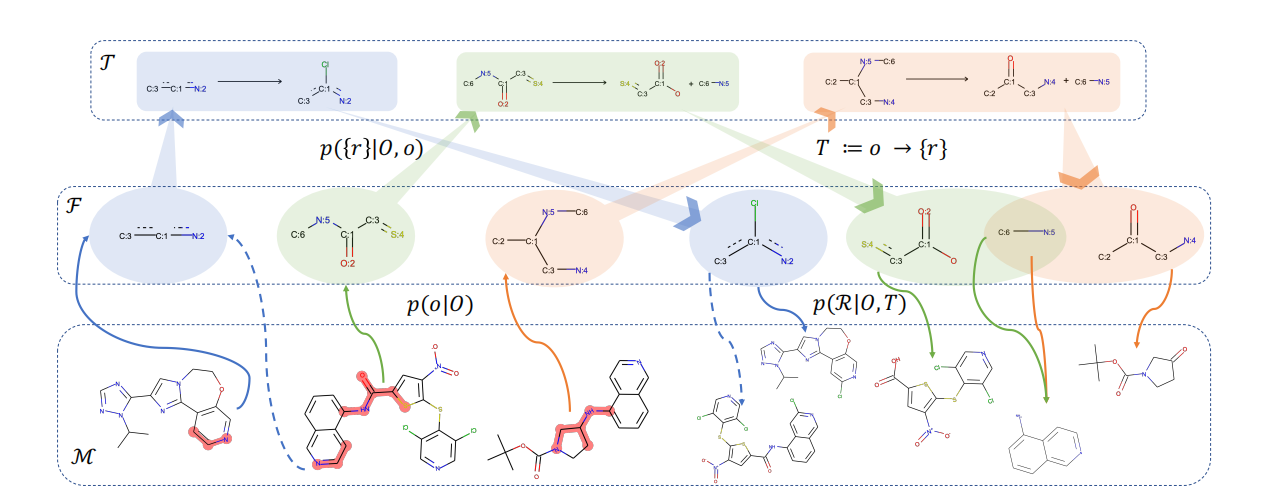}
\end{center}

\begin{center}
    \centering
    \captionsetup{type=figure}
    \includegraphics[width=0.5\linewidth]{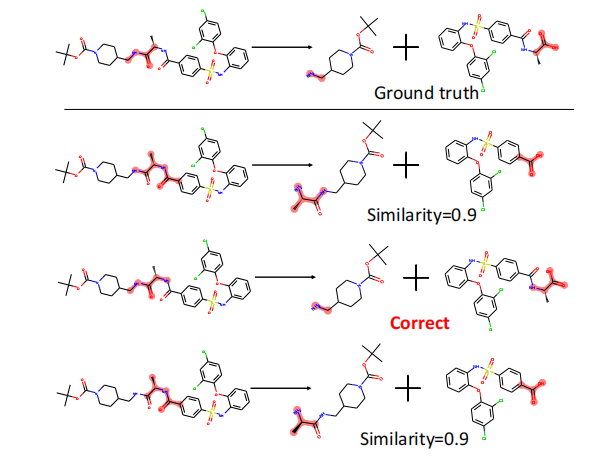}
\end{center}

\textbf{\emph{\textcolor{DeepPurple}{Question}}}

Based on the graphical models and prediction visualizations, which combination of template matching mechanism and reaction center identification approach is demonstrated across these three image, and what is the key chemical insight revealed by the successful prediction case?

\textbf{\emph{\textcolor{DeepPurple}{Options}}}

\begin{enumerate}[label=\Alph*.]
    \item Template matching via subgraph isomorphism + Atom-level scoring with GNN embeddings; The model correctly identifies esterification reaction centers and preserves stereochemistry.
    \item SMILES sequence alignment + Molecular fingerprint similarity; Successful predictions maintain atomic connectivity but miss stereochemical information.
    \item Reaction center extraction + Graph neural network compatibility scoring; Correct predictions align with known reaction mechanisms and preserve molecular topology.
    \item Rule-based template application + Attention-based focus mapping; The model captures functional group reactivity patterns and bond formation sites.
    \item Subgraph pattern matching + Energy-based scoring functions; Accurate retrosynthesis requires matching both structural patterns and chemical feasibility.
    \item Neural sequence-to-sequence + Structural motif recognition; Successful predictions demonstrate the importance of reaction template specificity.
    \item Graph isomorphism testing + Probabilistic template selection; The visualization shows positive scores on reactive atoms and negative on inactive regions.
    \item Molecular similarity comparison + Template ranking by frequency; Correct predictions occur when common reaction patterns are identified.
    \item Conditional graphical model + Hierarchical sampling; The model learns to assign high compatibility scores to chemically plausible reaction centers.
    \item Multi-class classification + Beam search optimization; Visualization reveals the model's ability to distinguish active reaction sites from background structure.
\end{enumerate}

\textbf{\emph{\textcolor{DeepPurple}{Steps}}}

\textbf{\textcolor{CaseOrange}{Step 1.}}

\begin{center}
    \centering
    \captionsetup{type=figure}
    \includegraphics[width=0.98\linewidth]{imgs/sci-exp-case/chemistry/01.png}
\end{center}

\textbf{\textcolor{CaseOrange}{Step 2.}} Analyze the chemical reaction and retrosynthesis template schematic, identifying the highlighted reaction centers in the reaction participants.

\textbf{\textcolor{CaseOrange}{Step 3.}} Determine that the template matching mechanism is based on reaction center extraction, identifying chemical transformation sites through subgraph pattern matching.

\textbf{\textcolor{CaseOrange}{Step 4.}}

\begin{center}
    \centering
    \captionsetup{type=figure}
    \includegraphics[width=0.98\linewidth]{imgs/sci-exp-case/chemistry/02.png}
\end{center}

\textbf{\textcolor{CaseOrange}{Step 5.}} Parse the three-layer architecture of the GLN retrosynthesis pipeline, understanding the logical relationships between template sets, subgraph sets, and molecule sets.

\textbf{\textcolor{CaseOrange}{Step 6.}} Identify the role of graph neural networks in compatibility scoring, analyzing the computation process of embedding vectors.

\textbf{\textcolor{CaseOrange}{Step 7.}}

\begin{center}
    \centering
    \captionsetup{type=figure}
    \includegraphics[width=0.5\linewidth]{imgs/sci-exp-case/chemistry/03.png}
\end{center}

\textbf{\textcolor{CaseOrange}{Step 8.}} Compare the core region matching between predicted reactions and true reactions in successful prediction cases.

\textbf{\textcolor{CaseOrange}{Step 9.}} Verify the consistency between prediction results and known reaction mechanisms, analyzing the preservation degree of molecular topology.

\textbf{\textcolor{CaseOrange}{Step 10.}} Integrate information from all three figures: template matching based on reaction center extraction provides structural foundation, GNN compatibility scoring provides chemical feasibility assessment, and actual cases validate method effectiveness.

\textbf{\textcolor{CaseOrange}{Step 11.}} Derive key chemical insight: successful retrosynthesis prediction requires simultaneously satisfying both structural pattern matching and reaction mechanism alignment conditions.

\textbf{\emph{\textcolor{DeepPurple}{Answer}}}

C

\end{tcolorbox}

\begin{tcolorbox}[
    breakable,
    title=Example of Experimental Reasoning in Earth,
    colback=LighterGray,
    colframe=DeepPurple,
    colbacktitle=DeepPurple,
    coltitle=White,
]

\textbf{\emph{\textcolor{DeepPurple}{Images}}}

\begin{center}
    \centering
    \captionsetup{type=figure}
    \includegraphics[width=0.5\linewidth]{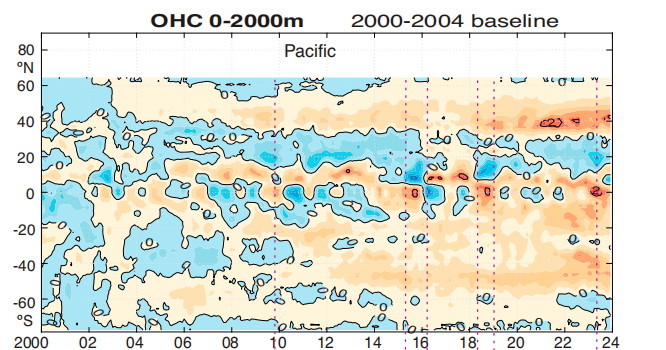}
\end{center}

\begin{center}
    \centering
    \captionsetup{type=figure}
    \includegraphics[width=0.5\linewidth]{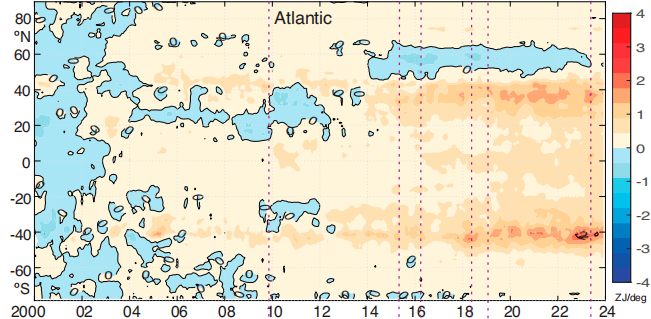}
\end{center}

\begin{center}
    \centering
    \captionsetup{type=figure}
    \includegraphics[width=0.5\linewidth]{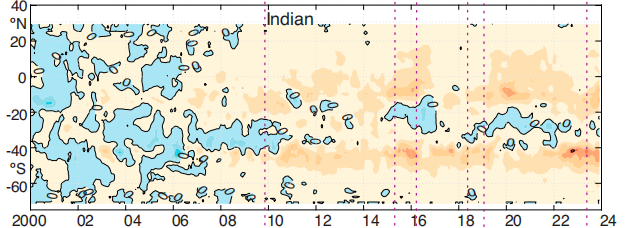}
\end{center}

\begin{center}
    \centering
    \captionsetup{type=figure}
    \includegraphics[width=0.5\linewidth]{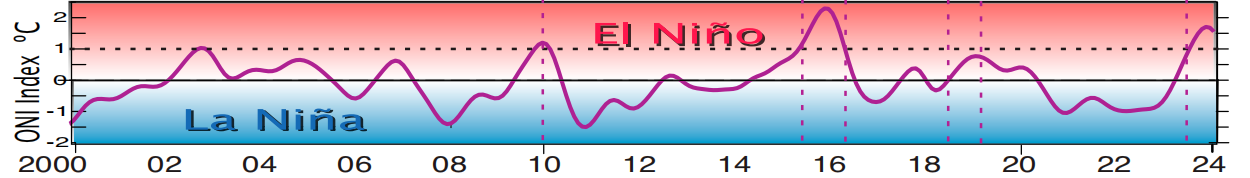}
\end{center}

\textbf{\emph{\textcolor{DeepPurple}{Question}}}

The first, second, and third images display the Zonal Mean Ocean Heat Content (OHC) anomalies for 0-2000m in the Pacific, Atlantic, and Indian Oceans, respectively, in ZJ per degree latitude (ZJ deg-1) relative to a 2000-2004 baseline, as a function of time (2000-2024) and latitude. The fourth image shows the Oceanic Niño Index (ONI) time series.

Based only on the visual information from these four images, which of the following combined statements is most likely true?

\textbf{\emph{\textcolor{DeepPurple}{Options}}}

\begin{enumerate}[label=\Alph*.]
    \item The onset of the OHC warming band ($\geq1$ ZJ deg-1) in the Indian Ocean (Figure 3) near 40°N occurred earlier than the warming in the Pacific (Figure 1) and Atlantic (Figure 2) at the same latitude. The strong El Niño event in 2010 (Figure 4) coincided with an OHC cooling anomaly (blue) in the Pacific Ocean (Figure 1) in the 40°S latitude band.
    \item The OHC anomaly in the equatorial Pacific (near 0°, Figure 1) is predominantly one of cooling (blue) during strong El Niño events (ONI $\ge1.0$, Figure 4), while the OHC anomaly in the equatorial Atlantic (near 0°, Figure 2) largely remains near zero (white). In the Southern Hemisphere subtropics (30°S to 50°S), the sustained OHC warming ($\geq 1$ ZJ deg-1) in the Pacific began earlier than in the Atlantic and Indian Oceans.
    \item The OHC anomaly in the Pacific Ocean (Figure 1) near 20°N was dominated by cooling during 2000-2010 and by warming during 2010-2024. The sustained cooling anomaly (blue) in the 50°N-60°N latitude band of the Atlantic Ocean (Figure 2) is a unique feature not observed in the corresponding northernmost latitudes of the other two basins.
    \item The Indian Ocean (Figure 3) exhibits OHC cooling anomalies near 20°S, whereas the Atlantic (Figure 2) and Pacific (Figure 1) have never shown cooling anomalies in the same latitude band. During the strong El Niño event of 2015-2016 (Figure 4), the OHC warming strength in the Atlantic Ocean (Figure 2) at 40°N reached its maximum value for the 2000-2024 period.
    \item The OHC anomaly strength in the Indian Ocean (Figure 3) at 40°S consistently exceeded the anomaly strength in the Pacific Ocean (Figure 1) at 40°S after 2016. During the strong La Niña event of 2020-2022 (Figure 4), the OHC anomaly strength in the Pacific Ocean (Figure 1) near 40°N remained between 0 and 1 ZJ deg-1.
    \item The OHC anomaly in all three basins (Figures 1, 2, 3) in the 20°S to 40°S latitude band shows a continuously intensifying warming trend after 2016. The OHC anomaly strength in the Pacific Ocean (Figure 1) near 40°N was greater than 0 ZJ deg-1 (non-blue) for all years in the 2000-2024 period.
    \item The sustained duration of OHC warming ($\geq 1$ ZJ deg-1) in the Atlantic Ocean (Figure 2) at 40°S is longer than the sustained duration at 40°N. The Pacific OHC anomaly (Figure 1) near 0° shows a strong positive correlation with the ONI (Figure 4).
    \item In the 20°S to 40°S latitude band, the OHC anomaly in the Indian Ocean (Figure 3) is the most unstable (most frequent alternation between positive and negative) of the three basins. The Atlantic Ocean (Figure 2) at 40°S has never reached an OHC warming anomaly strength of $\geq 2$ ZJ deg-1 since 2000.
    \item The OHC warming band ($\geq 1$ ZJ deg-1) in the Pacific Ocean (Figure 1) at 40°N started after 2014, approximately five years later than the warming onset in the Atlantic Ocean (Figure 2) at 40°N. The La Niña event in 2010-2011 (Figure 4) coincided with a strong OHC cooling anomaly (blue) in the Pacific Ocean (Figure 1) at 40°N.
    \item The Indian Ocean (Figure 3) exhibited strong warming ($\geq2$ ZJ deg-1) only in the Southern Hemisphere (0°S southward) during 2000-2024. The OHC anomaly in the 60°S-40°S latitude band of the Atlantic Ocean (Figure 2) was negative (blue) before 2010.
\end{enumerate}

\textbf{\emph{\textcolor{DeepPurple}{Steps}}}

\textbf{\textcolor{CaseOrange}{Step 1.}}

\begin{center}
    \centering
    \captionsetup{type=figure}
    \includegraphics[width=0.5\linewidth]{imgs/sci-exp-case/earth/1.png}
\end{center}

\textbf{\textcolor{CaseOrange}{Step 2.}} Strong warming centers are observed near 40°N and 40°S (deep red $\geq 3$ ZJ deg-1). The equatorial band (0°) OHC anomaly alternates significantly (blue/red) and is strongly related to time/ENSO. Sustained strong warming ($\geq 1$ ZJ deg-1) at 40°S begins around 2014.

\textbf{\textcolor{CaseOrange}{Step 3.}}

\begin{center}
    \centering
    \captionsetup{type=figure}
    \includegraphics[width=0.5\linewidth]{imgs/sci-exp-case/earth/2.png}
\end{center}

\textbf{\textcolor{CaseOrange}{Step 4.}} Strong warming is present at 40°S (deep red $\geq 3$ ZJ deg-1). Warming at 40°N is present but slightly weaker (red ~2-3 ZJ deg-1). A persistent cooling (blue) anomaly is seen in the 50°N-60°N band since 2010. Sustained strong warming at 40°S begins around 2016.

\textbf{\textcolor{CaseOrange}{Step 5.}}

\begin{center}
    \centering
    \captionsetup{type=figure}
    \includegraphics[width=0.5\linewidth]{imgs/sci-exp-case/earth/3.png}
\end{center}

\textbf{\textcolor{CaseOrange}{Step 6.}} The main warming center is at 40°S. The tropical region shows frequent anomaly changes. Sustained strong warming at 40°S begins around 2016.

\textbf{\textcolor{CaseOrange}{Step 7.}}

\begin{center}
    \centering
    \captionsetup{type=figure}
    \includegraphics[width=0.5\linewidth]{imgs/sci-exp-case/earth/4.png}
\end{center}

\textbf{\textcolor{CaseOrange}{Step 8.}} Provides the timing of El Niño (positive peaks) and La Niña (negative peaks) events.

\textbf{\textcolor{CaseOrange}{Step 9.}} Evaluate Option 1 : S1 (Figures 1, 2, 3): The warming band ($\geq 1$ ZJ deg-1) at 40°N in the Indian Ocean (Figure 3) only clearly appears after 2022. Both the Pacific and Atlantic Oceans show this warming starting around 2014. S1 is FALSE.

\textbf{\textcolor{CaseOrange}{Step 10.}} Evaluate Option 2 : S1 (Figures 1, 2, 4): During strong El Niño events (e.g., 2015-2016, Figure 4), the equatorial Pacific (Figure 1, 0°) is blue (cooling/negative anomaly), confirming a negative correlation with ONI. The equatorial Atlantic (Figure 2, 0°) remains mostly white (near zero anomaly) during these periods. S1 is TRUE. S2 (Figures 1, 2, 3): In the 30°S to 50°S band, the Pacific (Figure 1) sustained strong warming ($\geq 1$ ZJ deg-1) began around 2014. The Atlantic (Figure 2) and Indian (Figure 3) sustained warming began around 2016. Thus, the Pacific began earlier. S2 is TRUE. Conclusion: Option 1 is TRUE.

\textbf{\textcolor{CaseOrange}{Step 11.}} Evaluate Option 3 : S1 (Figure 1): The 20°N band in the Pacific shows mostly blue/white (cooling/zero anomaly) during 2000-2010. It shows mixed red/blue (warming/cooling) during 2010-2024. The description of the dominant anomaly sign for the two periods is incorrect. S1 is FALSE.

\textbf{\textcolor{CaseOrange}{Step 12.}} Evaluate Option 4 : S1 (Figures 1, 2, 3): While the Indian Ocean (Figure 3) shows cooling near 20°S, the Pacific (Figure 1) also shows cooling (blue) in the 20°S band around 2004-2006. S1 is FALSE.

\textbf{\textcolor{CaseOrange}{Step 13.}} Evaluate Option 5 : S1 (Figures 1, 3): The OHC anomaly strength at 40°S in the Pacific (Figure 1) is consistently high ($\geq 3$ ZJ deg-1) after 2016, whereas the Indian Ocean (Figure 3) strength weakens significantly around 2018-2020. S1 is FALSE.

\textbf{\textcolor{CaseOrange}{Step 14.}} Evaluate Option 6 : S1 (Figures 1, 2, 3): The warming in the 20°S to 40°S band is not continuously intensifying in all three basins after 2016; the Indian Ocean (Figure 3) shows a significant weakening/cooling patch around 2018-2020. S1 is FALSE.

\textbf{\textcolor{CaseOrange}{Step 15.}} Evaluate Option 7 : S1 (Figure 2): The Atlantic 40°S warming ($\geq 1$ ZJ deg-1) starts around 2016, while 40°N warming starts around 2014. 40°S warming has a shorter duration. S1 is FALSE.

\textbf{\textcolor{CaseOrange}{Step 16.}} Evaluate Option 8 : S1: In the 20°S to 40°S latitude band, the OHC anomaly in the Indian Ocean (Figure 3) is the most unstable (most frequent alternation between positive and negative) of the three basins. S1 is TRUE. (Note: This is the first part of the original Option 8 and is retained as True).S2 : The Atlantic Ocean (Figure 2) at 40°S has never reached an OHC warming anomaly strength of $\geq 2$ ZJ deg-1 since 2000. Check: In Figure 2, the 40°S band clearly shows colors corresponding to $\geq 2$ ZJ deg-1(dark red/deepest red) starting around 2016. Therefore, S2 is FALSE.

\textbf{\textcolor{CaseOrange}{Step 17.}} Evaluate Option 9 : S1 (Figures 1, 2): The onset of warming ($geq 1$ ZJ deg-1) at 40°N in both the Pacific (Figure 1) and Atlantic (Figure 2) occurs around 2014. There is no ~5-year lag. S1 is FALSE.

\textbf{\textcolor{CaseOrange}{Step 18.}} Evaluate Option 10: S1 (Figure 3): The Indian Ocean (Figure 3) shows strong warming ($\geq 2$ ZJ deg-1) in the Northern Hemisphere near 40°N after 2022. S1 is FALSE.

\textbf{\emph{\textcolor{DeepPurple}{Answer}}}

B

\end{tcolorbox}

\begin{tcolorbox}[
    breakable,
    title=Example of Experimental Reasoning in Energy,
    colback=LighterGray,
    colframe=DeepPurple,
    colbacktitle=DeepPurple,
    coltitle=White,
]

\textbf{\emph{\textcolor{DeepPurple}{Images}}}

\begin{center}
    \centering
    \captionsetup{type=figure}
    \includegraphics[width=0.8\linewidth]{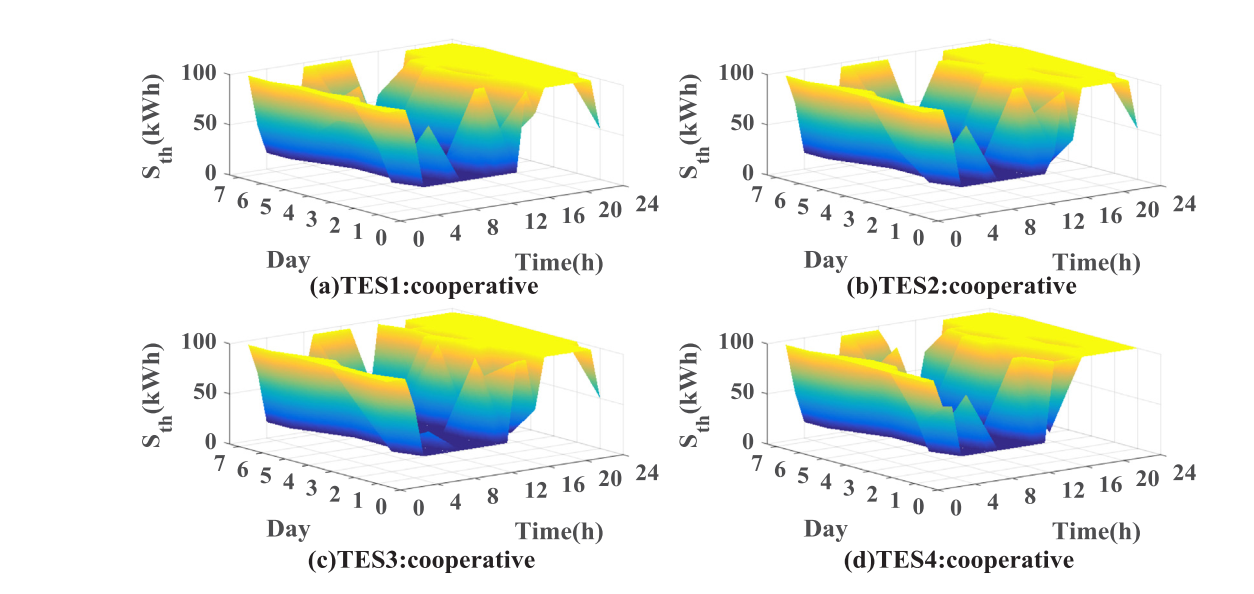}
\end{center}

\begin{center}
    \centering
    \captionsetup{type=figure}
    \includegraphics[width=0.8\linewidth]{imgs/sci-exp-case/energy/1.png}
\end{center}

\textbf{\emph{\textcolor{DeepPurple}{Question}}}

Based on the thermal energy storage (TES) state-of-charge visualizations shown in the two images, analyze the operational patterns across the 7-day period. The first image displays four TES units (TES1-TES4) operating independently, while the second image shows the same units under cooperative operation. During the time period from Day 3 to Day 5, which specific operational advantage of the cooperative mode most directly explains the consistently higher storage capacity utilization observed in TES4 compared to its independent operation?

\textbf{\emph{\textcolor{DeepPurple}{Options}}}

\begin{enumerate}[label=\Alph*.]
    \item Cooperative operation allows TES4 to receive excess thermal energy from microgrids without storage devices during high solar generation periods, maintaining near-maximum capacity
    \item The cooperative mode reduces TES4's discharge rate during peak thermal demand hours through load balancing across all microgrids
    \item Independent operation causes TES4 to experience more frequent charging cycles due to isolated thermal load requirements
    \item Cooperative operation eliminates the need for TES4 to supply thermal energy during nighttime hours through grid-level coordination
    \item The sharing of thermal energy in cooperative mode increases TES4's charging efficiency by 15-20\% through optimized heat transfer
    \item Independent operation requires TES4 to maintain a minimum reserve capacity for emergency thermal supply, preventing full utilization
    \item Cooperative mode enables TES4 to store thermal energy generated by micro-turbines from neighboring microgrids during low-demand periods
    \item The coordinated operation reduces thermal losses in TES4 by synchronizing charge-discharge cycles with solar thermal availability patterns
    \item Independent operation forces TES4 to discharge more frequently to meet local thermal loads that exceed its microgrid's generation capacity
    \item Cooperative mode implements a hierarchical control strategy that prioritizes filling TES4 before activating expensive micro-turbine generation
\end{enumerate}

\textbf{\emph{\textcolor{DeepPurple}{Steps}}}

\textbf{\textcolor{CaseOrange}{Step 1.}}

\begin{center}
    \centering
    \captionsetup{type=figure}
    \includegraphics[width=0.8\linewidth]{imgs/sci-exp-case/energy/1.png}
\end{center}

\textbf{\textcolor{CaseOrange}{Step 2.}} In the first image showing independent operation, observe TES4 (subplot h) during Days 3-5: the storage level exhibits significant valleys, dropping to approximately 20-30 kWh multiple times, and rarely maintains the maximum 100 kWh capacity for extended periods. The surface shows irregular topology with frequent charge-discharge cycles.

\textbf{\textcolor{CaseOrange}{Step 3.}}

\begin{center}
    \centering
    \captionsetup{type=figure}
    \includegraphics[width=0.8\linewidth]{imgs/sci-exp-case/energy/1.png}
\end{center}

\textbf{\textcolor{CaseOrange}{Step 4.}} In the second image showing cooperative operation, examine TES4 (subplot d) during the same Days 3-5 period: the storage level consistently maintains near-maximum capacity (90-100 kWh) for prolonged periods, particularly during daytime hours (approximately 8h-16h). The surface displays prominent yellow plateaus indicating sustained full capacity.

\textbf{\textcolor{CaseOrange}{Step 5.}} The key difference occurs during daytime hours when solar thermal generation is high. In cooperative mode, microgrids without TES devices can transfer their surplus solar thermal energy to TES4, enabling it to reach and maintain maximum capacity. In independent operation, each microgrid must consume or waste its own solar thermal energy locally, and TES4 can only store energy from its own microgrid's solar panels while also meeting that microgrid's immediate thermal load demands. This fundamental difference in energy sharing capability directly explains why TES4 maintains consistently higher storage levels in cooperative mode, as stated in the paper's analysis that 'the surplus thermal solar power of the microgrid without energy storage can be fully stored by the energy storage of another microgrid via local power exchange.'

\textbf{\emph{\textcolor{DeepPurple}{Answer}}}

A

\end{tcolorbox}

\begin{tcolorbox}[
    breakable,
    title=Example of Experimental Reasoning in Information,
    colback=LighterGray,
    colframe=DeepPurple,
    colbacktitle=DeepPurple,
    coltitle=White,
]

\textbf{\emph{\textcolor{DeepPurple}{Images}}}

\begin{center}
    \centering
    \captionsetup{type=figure}
    \includegraphics[width=0.98\linewidth]{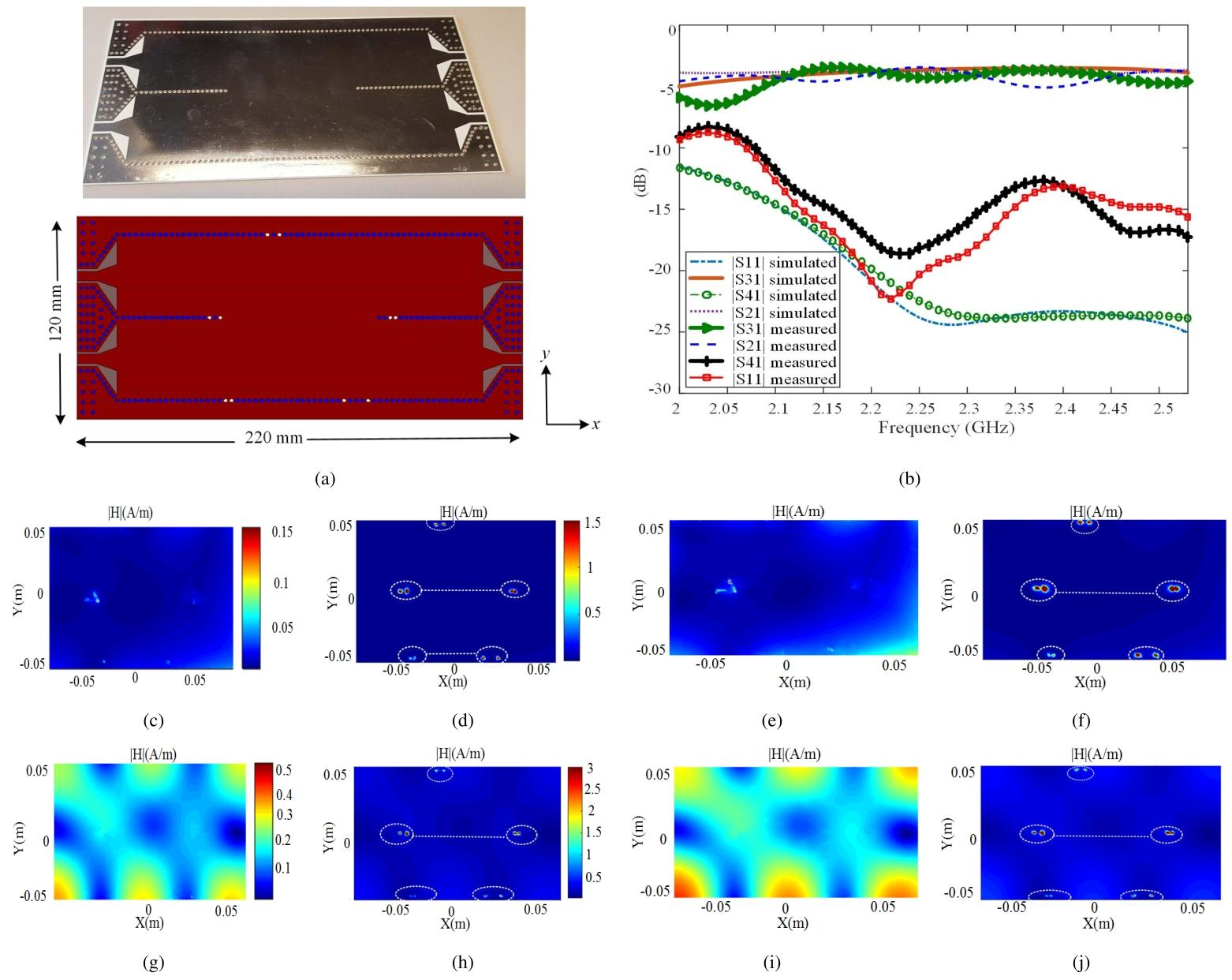}
\end{center}

\begin{center}
    \centering
    \captionsetup{type=figure}
    \includegraphics[width=0.98\linewidth]{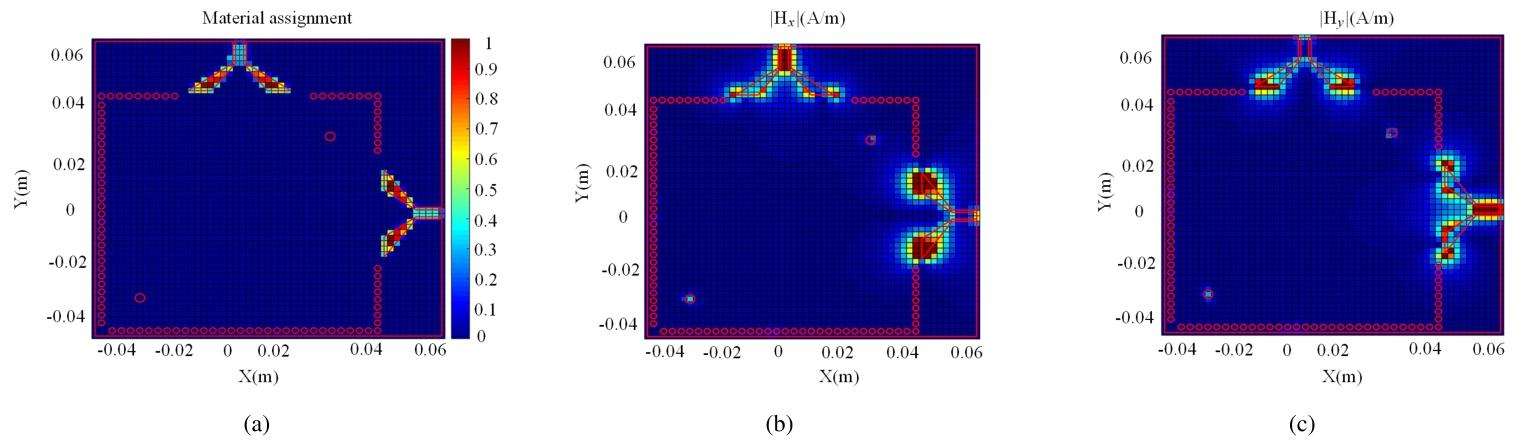}
\end{center}

\textbf{\emph{\textcolor{DeepPurple}{Question}}}

Based on the first image and the second image in the document, which statement is completely correct?

\textbf{\emph{\textcolor{DeepPurple}{Options}}}

\begin{enumerate}[label=\Alph*.]
    \item First image (a) is an SIW filter; First image (j) uses probe array-measured data for reconstruction (2 GHz); Second image (a) assigns 1 to fully metal areas, and (b) shows $|H_\gamma|$ variation.
    \item First image (b) is S-parameters of the coupler (2 GHz); First image (h) uses HFSS data with finite ground plane for reconstruction; Second image (a) assigns 0 to fully dielectric areas, and (c) shows $|H_x|$ variation.
    \item First image (e) is single probe-measured magnetic field (2 GHz); First image (d) uses HFSS data without ground plane for reconstruction; Second image (a) assigns 0 to fully metal areas, and (b) shows $|H_x|$ variation.
    \item First image (i) is probe array-measured magnetic field (1.84 GHz); First image (f) uses HFSS data with ground plane for reconstruction; Second image (a) assigns 1 to partially metal areas, and (c) shows $|H_\gamma|$ variation.
    \item First image (g) is sampled field from HFSS without ground plane (2 GHz); First image (j) reconstructs field 4 mm from the coupler; Second image (a) assigns 0.5 to fully dielectric areas, and (b) shows $|H_\gamma|$ variation.
    \item First image (c) is single probe-measured field (2 GHz); First image (h) reconstructs field 0.5 mm from the coupler; Second image (a) assigns 1 to fully metal areas, and (c) shows $|H_x|$ variation.
    \item First image (b) is S-parameters of the filter (1.84 GHz); First image (f) uses single probe-measured data for reconstruction; Second image (a) assigns 0 to partially dielectric areas, and (b) shows $|H_x|$ variation.
    \item First image (d) uses probe array-measured data for reconstruction (2 GHz); First image (i) is HFSS-simulated field with ground plane; Second image (a) assigns 1 to fully dielectric areas, and (c) shows $|H_\gamma|$ variation.
    \item First image (e) is probe array-measured field (1.84 GHz); First image (j) reconstructs field 0.5 mm from the filter; Second image (a) assigns 0 to fully dielectric areas, and (b) shows $|H_\gamma|$ variation.
    \item First image (g) is sampled field from HFSS with ground plane (2 GHz); First image (d) reconstructs field 4 mm from the coupler; Second image (a) assigns 1 to partially metal areas, and (c) shows $|H_x|$ variation.
\end{enumerate}

\textbf{\emph{\textcolor{DeepPurple}{Steps}}}

\textbf{\textcolor{CaseOrange}{Step 1.}}

\begin{center}
    \centering
    \captionsetup{type=figure}
    \includegraphics[width=0.98\linewidth]{imgs/sci-exp-case/information/1.png}
\end{center}

\textbf{\textcolor{CaseOrange}{Step 2.}}

\begin{center}
    \centering
    \captionsetup{type=figure}
    \includegraphics[width=0.98\linewidth]{imgs/sci-exp-case/information/2.png}
\end{center}

\textbf{\textcolor{CaseOrange}{Step 3.}} Extract core features of the first image (structure + frequency + measurement/simulation + reconstruction distance)

\textbf{\textcolor{CaseOrange}{Step 4.}} Structure \& frequency: The first image (a) is an SIW coupler (not filter), and (b) its S-parameters are measured at 2 GHz (not 1.84 GHz, which is the second image's frequency).

\textbf{\textcolor{CaseOrange}{Step 5.}} Measurement/simulation source: (c)/(g) = HFSS-simulated field: (c) = no ground plane, (g) = with finite ground plane; (e)/(i) = measured field: (e) = single probe, (i) = probe array;

\textbf{\textcolor{CaseOrange}{Step 6.}} Reconstruction distance: All reconstructed fields (d)/(f)/(h)/(j) are 0.5 mm from the coupler; measurement plane distance = 4 mm (not reconstruction distance).

\textbf{\textcolor{CaseOrange}{Step 7.}} Eliminate options with first image errors: Option 1 (a=filter, second image (a)=1 for metal, (b)=$|H_\gamma|$): Structure error + material assignment error + field component error. Option 2 (second image (a)=0 for dielectric, (c)=$|H_x|$): Material assignment error + field component error. Option 4 (i=1.84 GHz, f=HFSS with ground plane, (a)=1 for partial metal): Frequency error + reconstruction source error + material assignment error. Option 5 (g=no ground plane, j=4 mm reconstruction, (a)=0.5 for dielectric): Simulation source error + reconstruction distance error + material assignment error. Option 6 (c=single probe-measured, (a)=1 for metal, (c)=$|H_x|$): Field source error + material assignment error + field component error. Option 7 (b=filter S-parameters, 1.84 GHz, (a)=0 for partial dielectric): Structure/frequency error + material assignment error. Option 8 (d=probe array data, i=HFSS-simulated): Reconstruction source error + field source error. Option 9 (e=probe array-measured, 1.84 GHz, a=filter, (a)=0 for dielectric, (b)=$|H_\gamma|$): Measurement method error + frequency/structure error + material assignment/field component error. Option 10 (d=4 mm reconstruction, (a)=1 for partial metal, (c)=$|H_x|$): Reconstruction distance error + material assignment error + field component error.

\textbf{\textcolor{CaseOrange}{Step 8.}} Extract core features of the second image (material assignment + field components).

\textbf{\textcolor{CaseOrange}{Step 9.}} Material assignment rule: (a) 0 = fully metal-covered, 1 = fully dielectric-covered, 0-1 = partially metal-covered (not reverse or arbitrary values).

\textbf{\textcolor{CaseOrange}{Step 10.}} Field components: (b) = $|H_x|$ variation, (c) = $|H_\gamma|$ variation (not mixed).

\textbf{\textcolor{CaseOrange}{Step 11.}} Verify remaining option 3: First image part: "First image (e) is single probe-measured magnetic field (2 GHz)" → matches (e)=single probe, 2 GHz; "First image (d) uses HFSS data without ground plane for reconstruction" → (d) is reconstructed from (c)=HFSS no ground plane, correct. Second image part: "Second image (a) assigns 0 to fully metal areas" → matches material rule; "Second image (b) shows $|H_x|$ variation" → matches (b)=$|H_x|$, correct. Confirm option 3 is completely correct. All parts of option 3 align with the first image's structure/frequency/field source/reconstruction rule and the second image's material assignment/field component definition, with no contradictions.

\textbf{\emph{\textcolor{DeepPurple}{Answer}}}

C

\end{tcolorbox}

\begin{tcolorbox}[
    breakable,
    title=Example of Experimental Reasoning in Life,
    colback=LighterGray,
    colframe=DeepPurple,
    colbacktitle=DeepPurple,
    coltitle=White,
]

\textbf{\emph{\textcolor{DeepPurple}{Images}}}

\begin{center}
    \centering
    \captionsetup{type=figure}
    \includegraphics[width=0.98\linewidth]{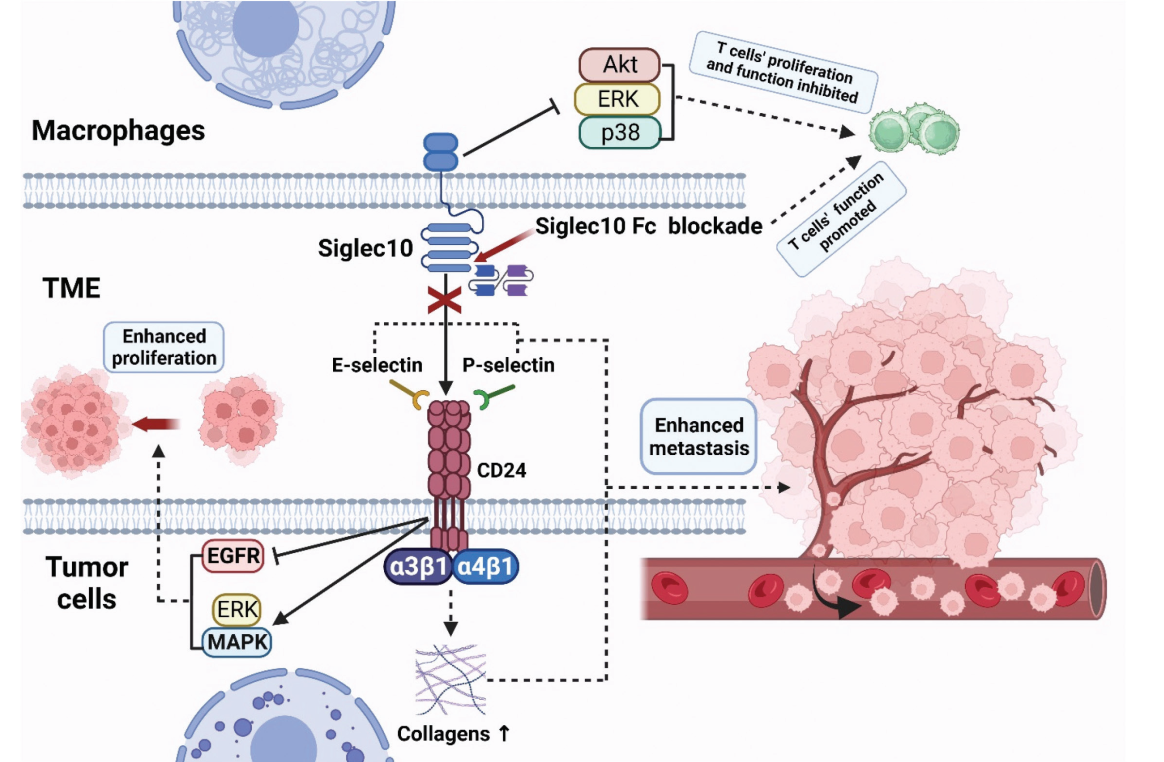}
\end{center}

\begin{center}
    \centering
    \captionsetup{type=figure}
    \includegraphics[width=0.98\linewidth]{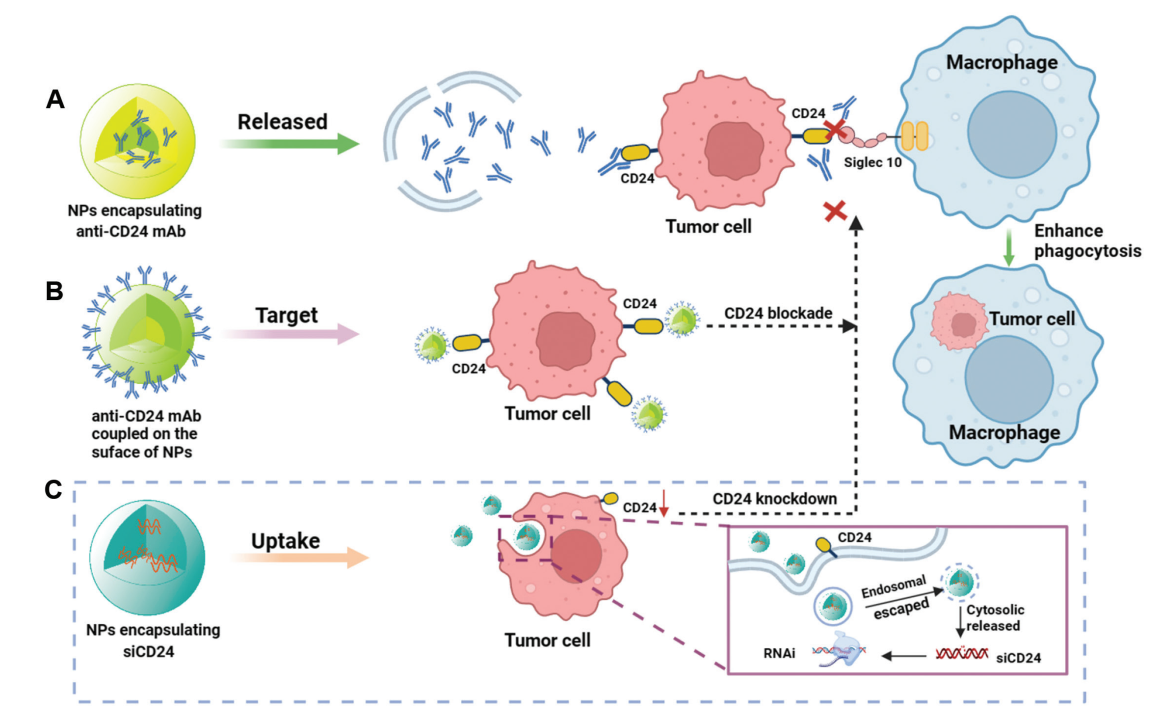}
\end{center}

\begin{center}
    \centering
    \captionsetup{type=figure}
    \includegraphics[width=0.98\linewidth]{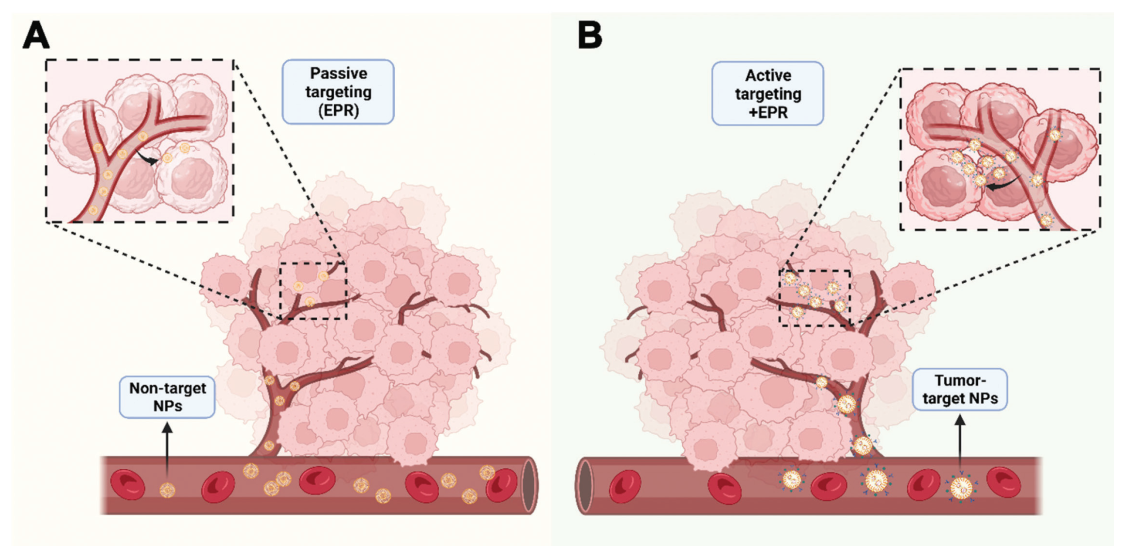}
\end{center}

\textbf{\emph{\textcolor{DeepPurple}{Question}}}

According to the first image, if one wants to inhibit tumor development by targeting non-tumor cells within the body, which cells should the monoclonal antibody be made against? Using which method from the second image to deliver the antibody can achieve a inhibition of tumor development from a deeper level? Which type in the third image does this method belong to?Please choose from the given options:

\textbf{\emph{\textcolor{DeepPurple}{Options}}}

\begin{enumerate}[label=\Alph*.]
    \item Siglec-10;A;A
    \item Siglec-10;A;B
    \item Siglec-10;B;A
    \item Siglec-10;B;B
    \item Siglec-10;C;A
    \item CD24;A;A
    \item CD24;A;B
    \item CD24;B;A
    \item CD24;B;B
    \item CD24;C;A
\end{enumerate}

\textbf{\emph{\textcolor{DeepPurple}{Steps}}}

\textbf{\textcolor{CaseOrange}{Step 1.}}

\begin{center}
    \centering
    \captionsetup{type=figure}
    \includegraphics[width=0.98\linewidth]{imgs/sci-exp-case/life/1.png}
\end{center}

\textbf{\textcolor{CaseOrange}{Step 2.}} The proteins identified in the image that can serve as targets are mainly Siglec-10 and CD24.

\textbf{\textcolor{CaseOrange}{Step 3.}} The topic requires starting from non-tumor cells, so Siglec-10 was chosen.

\textbf{\textcolor{CaseOrange}{Step 4.}} 

\begin{center}
    \centering
    \captionsetup{type=figure}
    \includegraphics[width=0.98\linewidth]{imgs/sci-exp-case/life/2.png}
\end{center}

\textbf{\textcolor{CaseOrange}{Step 5.}} Identify the three main strategies for NP-mediated CD24-Siglec10 axis-targeted therapy shown in the figure.

\textbf{\textcolor{CaseOrange}{Step 6.}} Among them, strategies A and B both use antibodies to directly block signal transduction on the cell surface, whereas strategy C uses siRNA to inhibit the expression of the target protein at the nucleic acid level.

\textbf{\textcolor{CaseOrange}{Step 7.}} Strategy C is a deeper approach to suppress tumor development.

\textbf{\textcolor{CaseOrange}{Step 8.}}

\begin{center}
    \centering
    \captionsetup{type=figure}
    \includegraphics[width=0.98\linewidth]{imgs/sci-exp-case/life/3.png}
\end{center}

\textbf{\textcolor{CaseOrange}{Step 9.}} Identifying two modes of nanoparticle-based drug delivery systems in the image.

\textbf{\textcolor{CaseOrange}{Step 10.}} The surface of the nanomaterials delivering siRNA does not carry antibodies and is passively targeted.

\textbf{\emph{\textcolor{DeepPurple}{Answer}}}

E

\end{tcolorbox}

\begin{tcolorbox}[
    breakable,
    title=Example of Experimental Reasoning in Material,
    colback=LighterGray,
    colframe=DeepPurple,
    colbacktitle=DeepPurple,
    coltitle=White,
]

\textbf{\emph{\textcolor{DeepPurple}{Images}}}

\begin{center}
    \centering
    \captionsetup{type=figure}
    \includegraphics[width=0.98\linewidth]{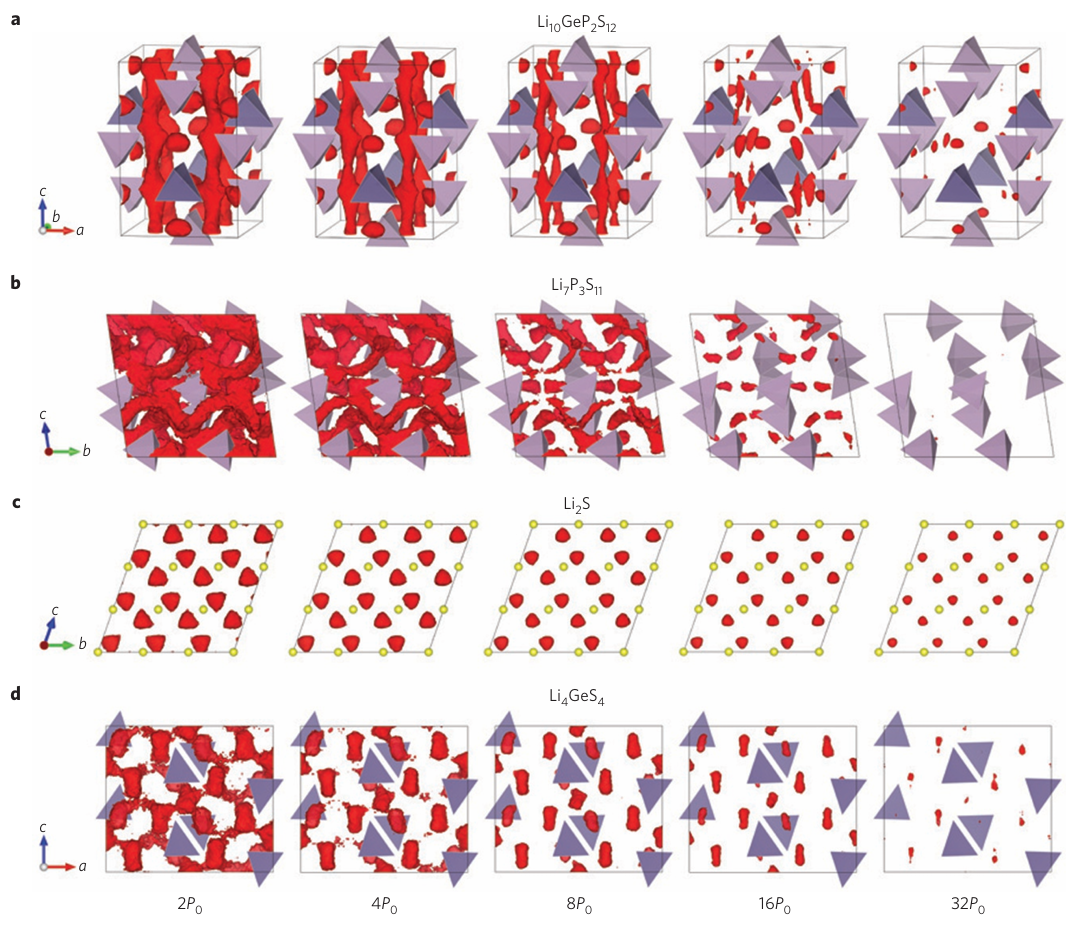}
\end{center}

\textbf{\emph{\textcolor{DeepPurple}{Question}}}

Images are Li-ion probability densities in Li-ion conductors. Li-ion probability densities are colored red. Which material does represent the best Li-ion conductivity?

\textbf{\emph{\textcolor{DeepPurple}{Options}}}

\begin{enumerate}[label=\Alph*.]
    \item Li10GeP2S12
    \item Li7P3S11
    \item Li2S
    \item $\gamma$-Li3PS4
    \item Li4GeS4
    \item Li3.25Ge0.25P0.75S4
    \item Li2S-P2S5
    \item Li10SnP2S12
    \item Li10SiP2S12
    \item Li6PS5Cl
\end{enumerate}

\textbf{\emph{\textcolor{DeepPurple}{Steps}}}

\textbf{\textcolor{CaseOrange}{Step 1.}}

\begin{center}
    \centering
    \captionsetup{type=figure}
    \includegraphics[width=0.98\linewidth]{imgs/sci-exp-case/material/1.png}
\end{center}

\textbf{\textcolor{CaseOrange}{Step 2.}} Find the Li-ion probability densities of materials in the figure.

\textbf{\textcolor{CaseOrange}{Step 3.}} Determine the largest region of the Li-ion probability densities. The answer is Li10GeP2S12.

\textbf{\emph{\textcolor{DeepPurple}{Answer}}}

A

\end{tcolorbox}

\begin{tcolorbox}[
    breakable,
    title=Example of Experimental Reasoning in Neuroscience,
    colback=LighterGray,
    colframe=DeepPurple,
    colbacktitle=DeepPurple,
    coltitle=White,
]

\textbf{\emph{\textcolor{DeepPurple}{Images}}}

\begin{center}
    \centering
    \captionsetup{type=figure}
    \includegraphics[width=0.98\linewidth]{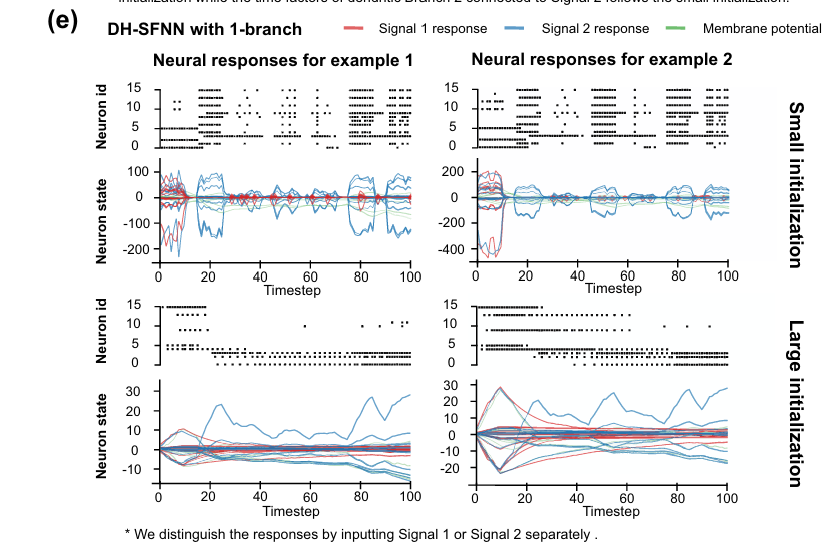}
\end{center}

\begin{center}
    \centering
    \captionsetup{type=figure}
    \includegraphics[width=0.5\linewidth]{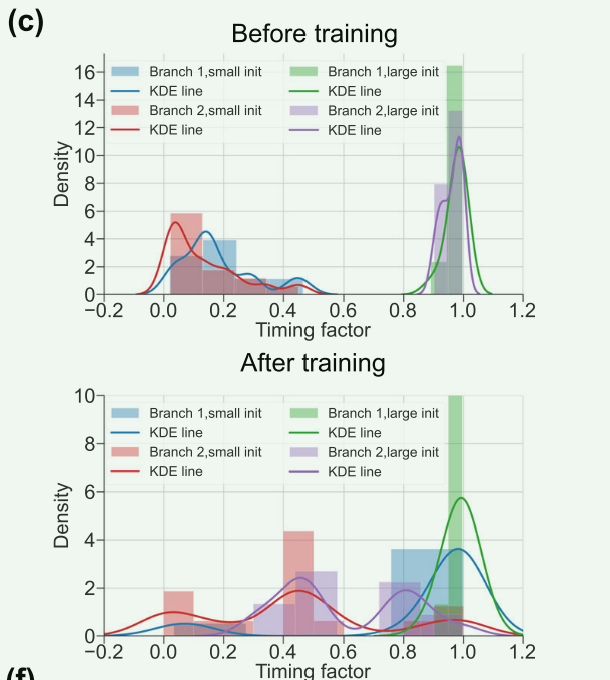}
\end{center}

\textbf{\emph{\textcolor{DeepPurple}{Question}}}

Please answer based on the first image: How many peaks exceeding 20 appeared in the first 60 timesteps of the Large initialization for Signal 2 response in each of the two examples?Based on the second image, after training, does Branch 1 with a small initialization increase (+) or decrease (-), and does Branch 2 with a large initialization increase (+) or decrease (-)?

\textbf{\emph{\textcolor{DeepPurple}{Options}}}

\begin{enumerate}[label=\Alph*.]
    \item 1,3;+-
    \item 0,0;++
    \item 1,2;--
    \item 1,1;++
    \item 1,3;-+
    \item 2,1;--
    \item 2,2;-+
    \item 3,1;-+
    \item 3,2;+-
    \item 0,3;-+
\end{enumerate}

\textbf{\emph{\textcolor{DeepPurple}{Steps}}}

\textbf{\textcolor{CaseOrange}{Step 1.}}

\begin{center}
    \centering
    \captionsetup{type=figure}
    \includegraphics[width=0.98\linewidth]{imgs/sci-exp-case/neuroscience/1.png}
\end{center}

\textbf{\textcolor{CaseOrange}{Step 2.}} Realize: Signal 2 response is blue line.

\textbf{\textcolor{CaseOrange}{Step 3.}} Define the counting range: Large initialization, Neuron state $> 20$,Timestep $< 60$, in each of the two examples.

\textbf{\textcolor{CaseOrange}{Step 4.}} Find out that there is 1 in example1 and 3 in example2. Answer: 1,3.

\textbf{\textcolor{CaseOrange}{Step 5.}} 

\begin{center}
    \centering
    \captionsetup{type=figure}
    \includegraphics[width=0.98\linewidth]{imgs/sci-exp-case/neuroscience/2.png}
\end{center}

\textbf{\textcolor{CaseOrange}{Step 6.}} Branch 1 small init: the KDE line and histogram show a shift. Before training, Branch 1 small init was lower around 0-0.2, after training, it's higher around 0.8-1.0, so increase (+)

\textbf{\textcolor{CaseOrange}{Step 7.}} Branch 2 large init: before training, it was a peak around 1.0, after training, the density decreases there, so decrease (-).

\textbf{\textcolor{CaseOrange}{Step 8.}} Conclude:1,3;+-

\textbf{\emph{\textcolor{DeepPurple}{Answer}}}

A

\end{tcolorbox}

\subsection{Supplementary Evaluation Results}

\begin{figure}[ht]
\centerline
{\includegraphics[width=16cm]{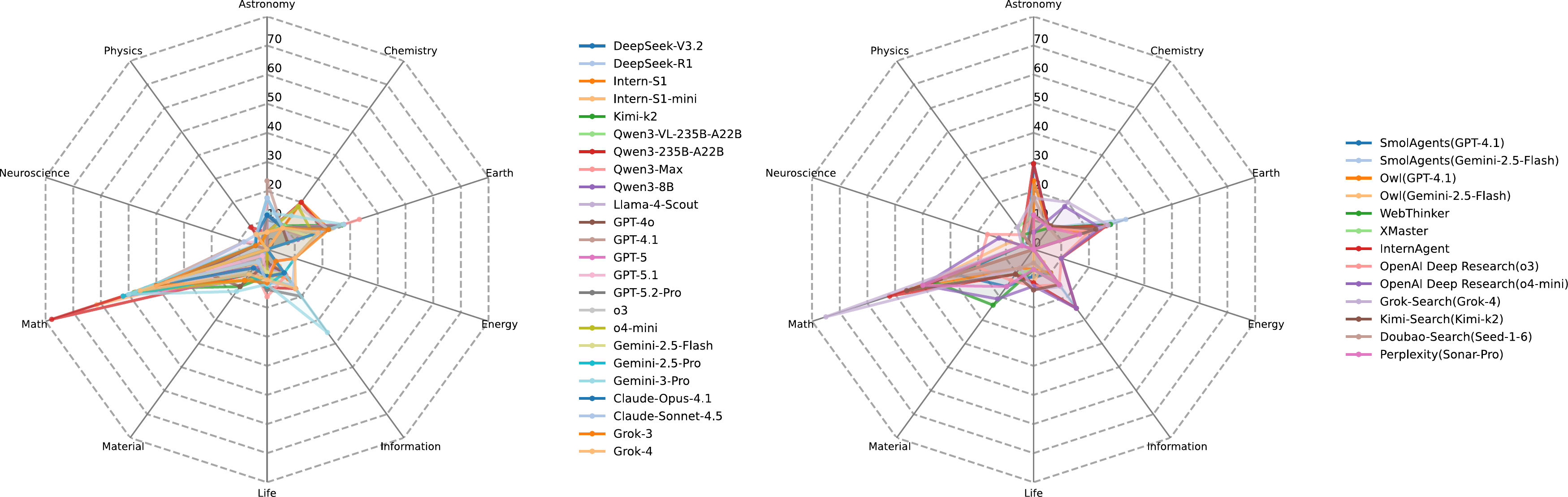}}
\caption{\textbf{Scientific Deep Research Across Subjects}: Combined subject-wise performance of LLMs and agents on deep research tasks.}
\label{fig: deep_research_subject}
\end{figure}

\begin{figure}[ht]
\centerline
{\includegraphics[width=8cm]{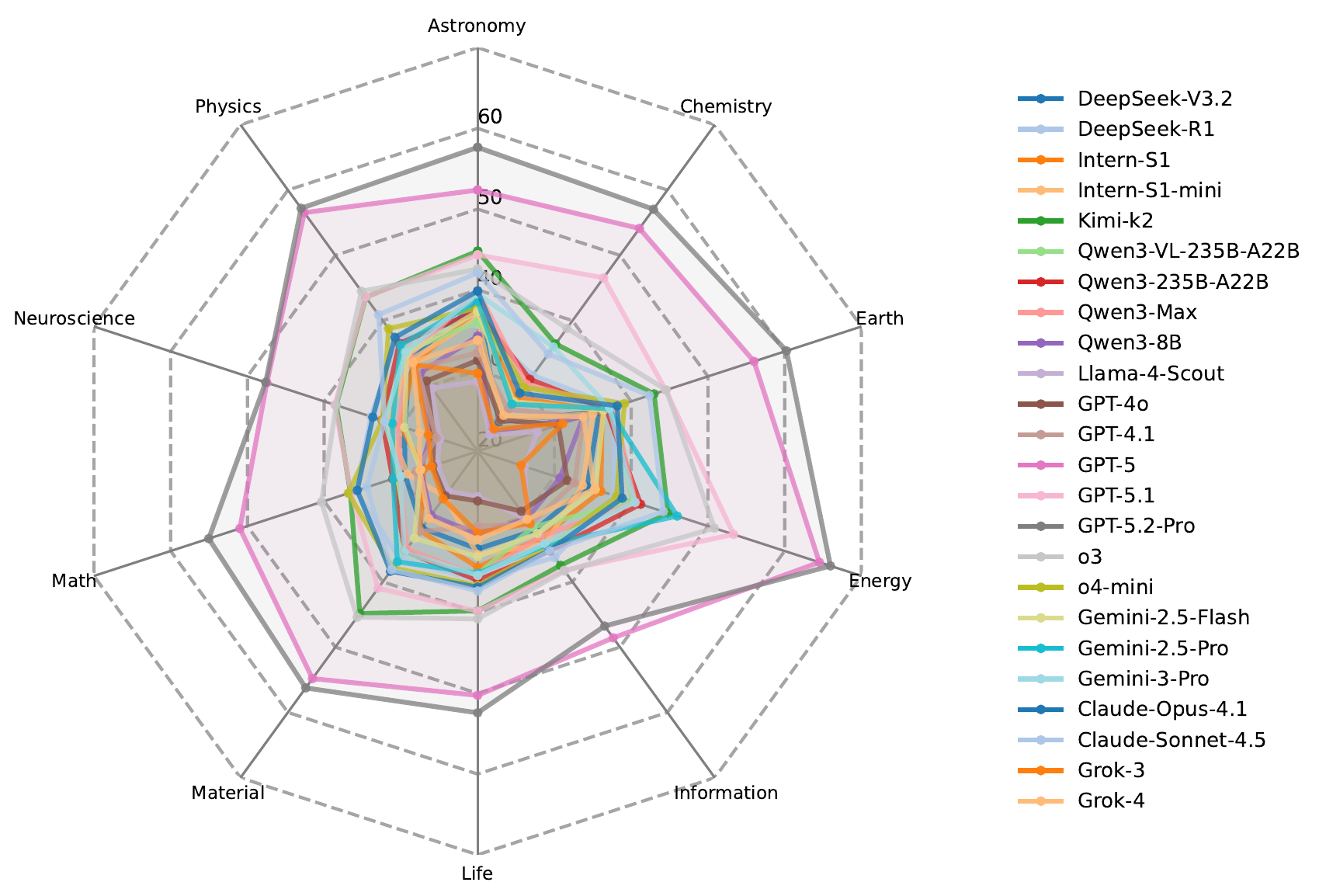}}
\caption{\textbf{Idea Generation Across Subjects}: Subject-wise scores for idea generation.}
\label{fig: idea_subject}
\end{figure}

\begin{figure}[ht]
\centerline
{\includegraphics[width=8cm]{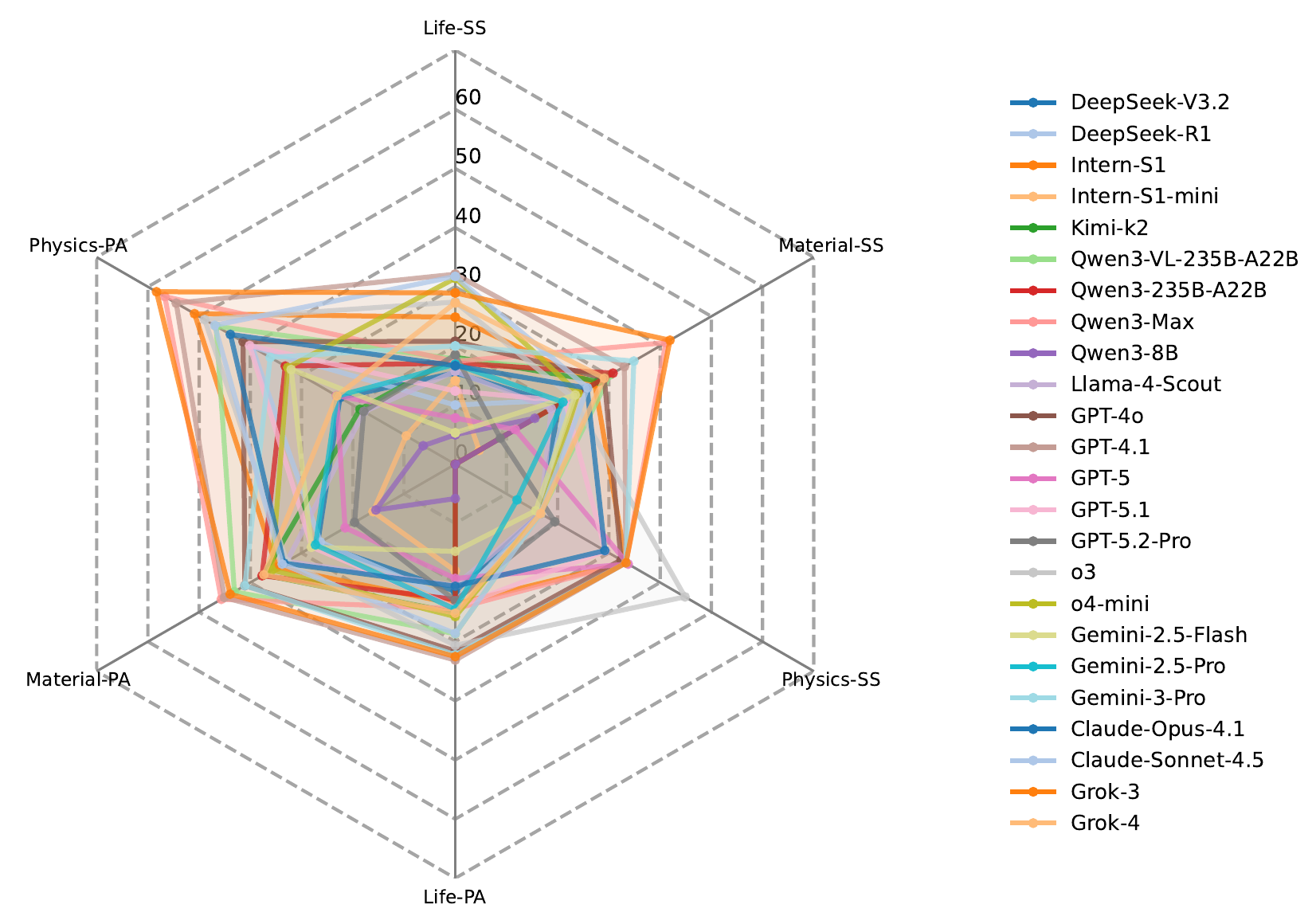}}
\caption{\textbf{Wet Experiment Across Subjects}: Subject-wise Action Sequence Similarity (SS) and Parameter Accuracy (PA) performance in wet experiments.}
\label{fig: wet_exp_subject}
\end{figure}


\begin{table}[t]
\centering
\renewcommand{\arraystretch}{0.9}
\setlength{\tabcolsep}{2pt}
\tiny
\resizebox{14cm}{!}{
\begin{tabular}{lcccc}
\toprule
\textbf{Model} & \textbf{Properties} & \textbf{Micro-experiments} & \textbf{Macro-experiments} & \textbf{Data} \\
\midrule
DeepSeek-V3.2 & 6.62 & 21.57 & 15.38 & 9.80 \\
DeepSeek-R1 & 10.61 & 23.47 & 15.38 & 10.00 \\
Intern-S1 & 7.14 & 24.64 & 20.00 & 12.12 \\
Intern-S1-mini & 5.88 & 19.18 & 17.39 & 5.41 \\
Kimi-k2 & 8.09 & 20.21 & 20.00 & 10.00 \\
Qwen3-VL-235B-A22B & 7.30 & 19.19 & 12.50 & 10.20 \\
Qwen3-235B-A22B & 11.94 & 23.75 & 11.54 & 6.12 \\
Qwen3-Max & 7.00 & 30.00 & 0.00 & 13.79 \\
Qwen3-8B & 5.84 & 14.42 & 3.85 & 3.92 \\
Llama-4-Scout & 5.11 & 14.42 & 3.85 & 3.92 \\
GPT-4o & 5.84 & 12.50 & 7.69 & 3.92 \\
GPT-4.1 & 7.30 & 17.31 & 15.38 & 7.84 \\
GPT-5 & 10.22 & 21.15 & 26.92 & 5.88 \\
GPT-5.1 & 8.03 & 18.27 & 15.38 & 5.88 \\
GPT-5.2-Pro & 10.22 & 23.08 & 23.08 & 11.76 \\
o3 & 10.95 & 17.31 & 19.23 & 5.88 \\
o4-mini & 8.76 & 18.27 & 11.54 & 7.84 \\
Gemini-2.5-Flash & 9.49 & 16.35 & 11.54 & 1.96 \\
Gemini-2.5-Pro & 11.68 & 23.08 & 15.38 & 7.84 \\
Gemini-3-Pro & 15.00 & 26.14 & 22.73 & 10.87 \\
Claude-Opus-4.1 & 8.82 & 20.19 & 15.38 & 7.84 \\
Claude-Sonnet-4.5 & 8.03 & 23.08 & 15.38 & 9.80 \\
Grok-3 & 9.49 & 20.19 & 11.54 & 11.76 \\
Grok-4 & 10.37 & 21.65 & 15.38 & 4.00 \\
\bottomrule
\end{tabular}
}
\caption{\textbf{Deep Research Task Metrics (LLMs)}: Category-wise scores across Properties, Micro/Macro-Experiments, and Data. Note: Because different subjects have different characteristics, the number of questions in each category is not the same (Figure~\ref{fig: data_distribution}). Therefore, the overall performance of the model cannot be obtained by directly averaging the values in the table.}
\label{tab:llms_deep_research_task_metric}
\end{table}

\begin{table}[t]
\centering
\renewcommand{\arraystretch}{0.9}
\setlength{\tabcolsep}{2pt}
\tiny
\resizebox{16cm}{!}{
\begin{tabular}{lcccc}
\toprule
\textbf{Agent} & \textbf{Properties} & \textbf{Micro-experiments} & \textbf{Macro-experiments} & \textbf{Data} \\
\midrule
SmolAgents(GPT-4.1) & 13.87 & 16.35 & 26.92 & 5.88 \\
SmolAgents(Gemini-2.5-Flash) & 12.41 & 24.04 & 26.92 & 11.76 \\
Owl(GPT-4.1) & 6.57 & 18.27 & 19.23 & 9.80 \\
Owl(Gemini-2.5-Flash) & 6.61 & 14.29 & 9.52 & 8.33 \\
WebThinker & 13.87 & 18.27 & 26.92 & 3.92 \\
XMaster & 13.14 & 17.31 & 19.23 & 5.88 \\
InternAgent & 13.24 & 24.04 & 26.92 & 9.80 \\
OpenAI Deep Research(o3) & 16.06 & 14.42 & 11.54 & 9.80 \\
OpenAI Deep Research(o4-mini) & 14.60 & 22.12 & 19.23 & 11.76 \\
Grok-Search(Grok-4) & 14.18 & 22.73 & 19.23 & 11.76 \\
Kimi-Search(Kimi-k2) & 9.49 & 22.92 & 11.54 & 14.00 \\
Doubao-Search(Seed-1-6) & 7.35 & 16.50 & 0.00 & 3.92 \\
Perplexity(Sonar-Pro) & 6.57 & 21.15 & 19.23 & 3.92 \\
\bottomrule
\end{tabular}
}
\caption{\textbf{Deep Research Task Metrics (Agents)}: Category-wise scores across Properties, Micro/Macro-Experiments, and Data. Note: Because different subjects have different characteristics, the number of questions in each category is not the same (Figure~\ref{fig: data_distribution}). Therefore, the overall performance of the model cannot be obtained by directly averaging the values in the table.}
\label{tab:agents_deep_research_task_metric}
\end{table}

\begin{table}[t]
\centering
\renewcommand{\arraystretch}{0.9}
\setlength{\tabcolsep}{2pt}
\tiny
\resizebox{17cm}{!}{
\begin{tabular}{lcccccc}
\toprule
\textbf{Model} & \textbf{Numerical Calculation} & \textbf{Statistical Analysis} & \textbf{Simulation} & \textbf{Metric Calculation} & \textbf{Data Processing} & \textbf{Predictive Modeling} \\
\midrule
DeepSeek-V3.2 & 19.30 & 19.05 & 26.32 & 35.71 & 42.86 & 27.27 \\
DeepSeek-R1 & 31.76 & 23.81 & 26.32 & 39.29 & 47.62 & 45.45 \\
Intern-S1 & 25.61 & 28.57 & 26.32 & 39.29 & 42.86 & 27.27 \\
Intern-S1-mini & 14.62 & 19.05 & 15.79 & 25.00 & 28.57 & 9.09 \\
Kimi-k2 & 26.90 & 23.81 & 31.58 & 35.71 & 52.38 & 18.18 \\
Qwen3-VL-235B-A22B & 25.15 & 23.81 & 26.32 & 39.29 & 47.62 & 27.27 \\
Qwen3-235B-A22B & 25.29 & 28.57 & 31.58 & 35.71 & 47.62 & 27.27 \\
Qwen3-Max & 29.24 & 38.10 & 31.58 & 39.29 & 47.62 & 45.45 \\
Qwen3-8B & 18.71 & 14.29 & 15.79 & 25.00 & 23.81 & 0.00 \\
Llama-4-Scout & 20.59 & 19.05 & 15.79 & 21.43 & 23.81 & 18.18 \\
GPT-4o & 25.15 & 23.81 & 26.32 & 32.14 & 38.10 & 27.27 \\
GPT-4.1 & 32.75 & 38.10 & 26.32 & 39.29 & 47.62 & 27.27 \\
GPT-5 & 25.73 & 28.57 & 31.58 & 39.29 & 52.38 & 27.27 \\
GPT-5.1 & 29.24 & 23.81 & 26.32 & 42.86 & 42.86 & 27.27 \\
GPT-5.2-Pro & 26.90 & 19.05 & 26.32 & 32.14 & 38.10 & 36.36 \\
o3 & 28.65 & 42.86 & 26.32 & 42.86 & 38.10 & 27.27 \\
o4-mini & 35.09 & 28.57 & 26.32 & 39.29 & 52.38 & 36.36 \\
Gemini-2.5-Flash & 16.96 & 23.81 & 21.05 & 32.14 & 38.10 & 18.18 \\
Gemini-2.5-Pro & 19.30 & 23.81 & 21.05 & 21.43 & 42.86 & 36.36 \\
Gemini-3-Pro & 33.53 & 33.33 & 35.29 & 46.43 & 50.00 & 45.45 \\
Claude-Opus-4.1 & 30.99 & 28.57 & 31.58 & 53.57 & 47.62 & 36.36 \\
Claude-Sonnet-4.5 & 33.33 & 38.10 & 26.32 & 42.86 & 47.62 & 45.45 \\
Grok-3 & 22.81 & 33.33 & 31.58 & 35.71 & 47.62 & 18.18 \\
Grok-4 & 32.12 & 19.05 & 31.58 & 40.74 & 42.86 & 54.55 \\
\bottomrule
\end{tabular}
}
\caption{\textbf{Dry Experiment Function Categories}: Completion scores across six function types. Note: Because different subjects have different characteristics, the number of questions in each category is not the same (Figure~\ref{fig: data_distribution}). Therefore, the overall performance of the model cannot be obtained by directly averaging the values in the table.}
\label{tab:dry_task_metric_table}
\end{table}

\begin{table}[t]
\centering
\renewcommand{\arraystretch}{0.9}
\setlength{\tabcolsep}{2pt}
\tiny
\resizebox{17cm}{!}{
\begin{tabular}{lcccc}
\toprule
\textbf{Model} & \textbf{Signal Perception} & \textbf{Attribute Understanding} & \textbf{Comparative Reasoning} & \textbf{Causal Reasoning} \\
\midrule
Intern-S1 & 39.29 & 21.88 & 28.57 & 37.50 \\
Intern-S1-mini & 17.86 & 10.94 & 18.29 & 20.83 \\
Qwen3-VL-235B-A22B & 32.14 & 26.56 & 32.00 & 41.67 \\
Qwen3-VL-Max & 50.00 & 34.38 & 36.57 & 41.67 \\
Qwen3-VL-8B & 21.43 & 21.88 & 23.43 & 29.17 \\
Llama-4-Scout & 28.57 & 17.19 & 28.57 & 25.00 \\
GPT-4o & 39.29 & 26.56 & 33.71 & 29.17 \\
GPT-4.1 & 46.43 & 40.62 & 34.29 & 54.10 \\
GPT-5 & 53.57 & 32.81 & 37.71 & 37.50 \\
GPT-5.1 & 21.43 & 25.00 & 36.57 & 54.17 \\
GPT-5.2-Pro & 53.57 & 39.06 & 38.29 & 29.17 \\
o3 & 35.71 & 26.56 & 33.14 & 41.67 \\
o4-mini & 39.29 & 35.94 & 30.29 & 41.67 \\
Gemini-2.5-Flash & 35.71 & 37.50 & 30.29 & 54.17 \\
Gemini-2.5-Pro & 50.00 & 42.19 & 38.29 & 50.00 \\
Gemini-3-Pro & 50.00 & 40.62 & 42.86 & 29.17 \\
Claude-Opus-4.1 & 53.57 & 35.94 & 34.86 & 58.33 \\
Claude-Sonnet-4.5 & 35.71 & 35.94 & 38.86 & 37.50 \\
Grok-4 & 42.86 & 26.56 & 28.00 & 41.67 \\
\bottomrule
\end{tabular}
}
\caption{\textbf{Experimental Reasoning by Type (Multi-choice Accuracy)}: Scores across signal, attribute, comparative, and causal reasoning. Note: Because different subjects have different characteristics, the number of questions in each category is not the same (Figure~\ref{fig: data_distribution}). Therefore, the overall performance of the model cannot be obtained by directly averaging the values in the table.}
\label{tab:mcp_task_metric_table}
\end{table}

\begin{table}[t]
\centering
\renewcommand{\arraystretch}{0.9}
\setlength{\tabcolsep}{2pt}
\tiny
\resizebox{17cm}{!}{
\begin{tabular}{lcccccccccc}
\toprule
\textbf{Model} & \textbf{Astronomy} & \textbf{Chemistry} & \textbf{Earth} & \textbf{Energy} & \textbf{Information} & \textbf{Life} & \textbf{Material} & \textbf{Math} & \textbf{Neuroscience} & \textbf{Physics} \\
\midrule
DeepSeek-V3.2 & 11.76 & 10.00 & 20.75 & 0.00 & 10.47 & 0.00 & 7.89 & 44.00 & 0.00 & 3.12 \\
DeepSeek-R1 & 6.25 & 9.09 & 24.00 & 0.00 & 16.67 & 0.00 & 7.89 & 52.00 & 4.17 & 6.67 \\
Intern-S1 & 0.00 & 20.00 & 22.45 & 0.00 & 12.50 & 8.00 & 0.00 & 47.62 & 0.00 & 0.00 \\
Intern-S1-mini & 0.00 & 9.09 & 23.26 & 0.00 & 7.14 & 6.25 & 7.14 & 61.54 & 0.00 & 0.00 \\
Kimi-k2 & 5.88 & 10.00 & 27.08 & 0.00 & 5.26 & 9.30 & 15.79 & 43.48 & 0.00 & 0.00 \\
Qwen3-VL-235B-A22B & 5.88 & 10.00 & 19.61 & 0.00 & 16.67 & 9.41 & 5.26 & 40.00 & 0.00 & 6.25 \\
Qwen3-235B-A22B & 5.88 & 20.00 & 20.83 & 0.00 & 16.67 & 13.10 & 10.53 & 77.78 & 0.00 & 9.38 \\
Qwen3-Max & 11.11 & 0.00 & 33.33 & 0.00 & 11.11 & 16.28 & 7.89 & 44.00 & 4.17 & 3.12 \\
Qwen3-8B & 11.76 & 0.00 & 11.11 & 0.00 & 10.00 & 5.75 & 7.89 & 32.00 & 0.00 & 0.00 \\
Llama-4-Scout & 11.76 & 9.09 & 9.26 & 0.00 & 10.00 & 6.90 & 5.26 & 20.00 & 4.17 & 3.12 \\
GPT-4o & 5.88 & 9.09 & 7.41 & 0.00 & 10.00 & 4.60 & 15.79 & 24.00 & 4.17 & 0.00 \\
GPT-4.1 & 23.53 & 9.09 & 12.96 & 0.00 & 5.00 & 9.20 & 5.26 & 44.00 & 8.33 & 0.00 \\
GPT-5 & 5.88 & 9.09 & 27.78 & 0.00 & 10.00 & 9.20 & 13.16 & 52.00 & 0.00 & 3.12 \\
GPT-5.1 & 17.65 & 9.09 & 18.52 & 10.00 & 5.00 & 9.20 & 2.63 & 36.00 & 8.33 & 3.12 \\
GPT-5.2-Pro & 11.76 & 9.09 & 25.93 & 0.00 & 20.00 & 13.79 & 10.53 & 48.00 & 0.00 & 3.12 \\
o3 & 5.88 & 18.18 & 22.22 & 0.00 & 10.00 & 9.20 & 7.89 & 44.00 & 4.17 & 3.12 \\
o4-mini & 5.88 & 18.18 & 16.67 & 0.00 & 0.00 & 9.20 & 13.16 & 48.00 & 0.00 & 3.12 \\
Gemini-2.5-Flash & 5.88 & 9.09 & 14.81 & 0.00 & 10.00 & 8.05 & 5.26 & 40.00 & 4.17 & 6.25 \\
Gemini-2.5-Pro & 17.65 & 9.09 & 18.52 & 10.00 & 10.00 & 12.64 & 10.53 & 52.00 & 4.17 & 6.25 \\
Gemini-3-Pro & 12.50 & 14.29 & 27.66 & 0.00 & 35.29 & 12.00 & 17.86 & 50.00 & 4.76 & 6.25 \\
Claude-Opus-4.1 & 11.76 & 9.09 & 22.22 & 0.00 & 10.00 & 9.30 & 7.89 & 40.00 & 4.17 & 6.25 \\
Claude-Sonnet-4.5 & 17.65 & 9.09 & 20.37 & 10.00 & 15.00 & 11.49 & 5.26 & 36.00 & 8.33 & 6.25 \\
Grok-3 & 5.88 & 9.09 & 22.22 & 10.00 & 5.00 & 11.49 & 13.16 & 40.00 & 4.17 & 3.12 \\
Grok-4 & 5.88 & 9.09 & 18.37 & 10.00 & 16.67 & 10.47 & 10.53 & 45.83 & 0.00 & 6.45 \\
\bottomrule
\end{tabular}
}
\caption{\textbf{Deep Research Across Subjects (LLMs)}: Subject-wise scores across ten scientific domains. Note: Because different subjects have different characteristics, the number of questions in each category is not the same (Figure~\ref{fig: data_distribution}). Therefore, the overall performance of the model cannot be obtained by directly averaging the values in the table.}
\label{tab:llms_deep_research_subject_metric_table}
\end{table}

\begin{table}[t]
\centering
\renewcommand{\arraystretch}{0.9}
\setlength{\tabcolsep}{2pt}
\tiny
\resizebox{17cm}{!}{
\begin{tabular}{lcccccccccc}
\toprule
\textbf{Agent} & \textbf{Astronomy} & \textbf{Chemistry} & \textbf{Earth} & \textbf{Energy} & \textbf{Information} & \textbf{Life} & \textbf{Material} & \textbf{Math} & \textbf{Neuroscience} & \textbf{Physics} \\
\midrule
SmolAgents(GPT-4.1) & 29.41 & 9.09 & 27.78 & 0.00 & 10.00 & 9.20 & 15.79 & 28.00 & 4.17 & 3.12 \\
SmolAgents(Gemini-2.5-Flash) & 23.53 & 9.09 & 33.33 & 0.00 & 25.00 & 11.49 & 10.53 & 44.00 & 4.17 & 3.12 \\
Owl(GPT-4.1) & 23.53 & 9.09 & 18.52 & 0.00 & 10.00 & 6.90 & 7.89 & 44.00 & 4.17 & 0.00 \\
Owl(Gemini-2.5-Flash) & 6.25 & 10.00 & 15.79 & 0.00 & 5.56 & 11.54 & 0.00 & 41.67 & 8.33 & 0.00 \\
WebThinker & 5.88 & 9.09 & 27.78 & 0.00 & 15.00 & 6.90 & 23.68 & 36.00 & 4.17 & 6.25 \\
XMaster & 11.76 & 9.09 & 25.93 & 0.00 & 15.00 & 6.90 & 10.53 & 44.00 & 0.00 & 9.38 \\
InternAgent & 29.41 & 9.09 & 26.42 & 10.00 & 25.00 & 11.49 & 10.53 & 52.00 & 0.00 & 6.25 \\
OpenAI Deep Research(o3) & 11.76 & 9.09 & 20.37 & 10.00 & 15.00 & 12.64 & 13.16 & 20.00 & 16.67 & 6.25 \\
OpenAI Deep Research(o4-mini) & 5.88 & 18.18 & 24.07 & 10.00 & 25.00 & 12.64 & 21.05 & 40.00 & 12.50 & 0.00 \\
Grok-Search(Grok-4) & 17.65 & 20.00 & 26.92 & 0.00 & 15.79 & 13.95 & 7.89 & 75.00 & 4.17 & 9.38 \\
Kimi-Search(Kimi-k2) & 11.76 & 10.00 & 22.45 & 0.00 & 15.00 & 13.95 & 10.53 & 45.83 & 0.00 & 3.12 \\
Doubao-Search(Seed-1-6) & 17.65 & 9.09 & 9.43 & 0.00 & 15.00 & 4.65 & 8.11 & 32.00 & 0.00 & 6.25 \\
Perplexity(Sonar-Pro) & 11.76 & 9.09 & 16.67 & 0.00 & 15.00 & 6.90 & 15.79 & 40.00 & 4.17 & 0.00 \\
\bottomrule
\end{tabular}
}
\caption{\textbf{Deep Research Across Subjects (Agents)}: Subject-wise scores across ten scientific domains. Note: Because different subjects have different characteristics, the number of questions in each category is not the same (Figure~\ref{fig: data_distribution}). Therefore, the overall performance of the model cannot be obtained by directly averaging the values in the table.}
\label{tab:agents_deep_research_subject_metric_table}
\end{table}

\begin{table}[t]
\centering
\renewcommand{\arraystretch}{0.9}
\setlength{\tabcolsep}{2pt}
\tiny
\resizebox{17cm}{!}{
\begin{tabular}{lcccccccccc}
\toprule
\textbf{Model} & \textbf{Astronomy} & \textbf{Chemistry} & \textbf{Earth} & \textbf{Energy} & \textbf{Information} & \textbf{Life} & \textbf{Material} & \textbf{Math} & \textbf{Neuroscience} & \textbf{Physics} \\
\midrule
DeepSeek-V3.2 & 38.38 & 24.49 & 35.85 & 34.24 & 32.09 & 32.11 & 31.77 & 29.46 & 29.53 & 34.20 \\
DeepSeek-R1 & 35.76 & 31.63 & 37.89 & 37.73 & 36.26 & 35.96 & 36.47 & 35.88 & 33.88 & 37.21 \\
Intern-S1 & 37.53 & 28.20 & 36.22 & 36.07 & 33.30 & 34.38 & 32.15 & 27.00 & 30.07 & 33.46 \\
Intern-S1-mini & 36.49 & 24.77 & 35.00 & 33.68 & 34.21 & 32.80 & 26.96 & 29.16 & 31.91 & 34.02 \\
Kimi-k2 & 44.80 & 36.44 & 42.99 & 44.80 & 37.48 & 39.78 & 44.86 & 36.58 & 38.43 & 43.59 \\
Qwen3-VL-235B-A22B & 36.00 & 30.06 & 37.90 & 40.09 & 31.62 & 35.28 & 35.59 & 30.56 & 32.18 & 35.31 \\
Qwen3-235B-A22B & 37.38 & 31.02 & 36.78 & 41.24 & 35.25 & 35.98 & 35.34 & 31.06 & 32.46 & 36.52 \\
Qwen3-Max & 39.80 & 30.28 & 37.74 & 40.56 & 33.12 & 35.42 & 34.98 & 30.12 & 30.31 & 34.54 \\
Qwen3-8B & 34.25 & 22.91 & 33.78 & 30.72 & 30.35 & 30.26 & 29.80 & 27.42 & 26.20 & 32.05 \\
Llama-4-Scout & 28.65 & 22.50 & 27.79 & 26.10 & 30.47 & 25.62 & 26.14 & 25.26 & 24.94 & 29.65 \\
GPT-4o & 31.27 & 24.79 & 30.50 & 31.70 & 29.19 & 26.17 & 26.83 & 25.86 & 25.72 & 30.77 \\
GPT-4.1 & 32.20 & 26.40 & 33.79 & 32.64 & 31.15 & 29.28 & 32.30 & 27.99 & 25.37 & 32.78 \\
GPT-5 & 52.37 & 54.12 & 56.01 & 64.53 & 48.58 & 50.25 & 54.82 & 50.99 & 47.46 & 56.55 \\
GPT-5.1 & 44.34 & 46.56 & 44.50 & 53.35 & 38.24 & 39.80 & 41.00 & 36.49 & 38.61 & 43.61 \\
GPT-5.2-Pro & 57.65 & 57.06 & 60.24 & 65.97 & 46.78 & 52.41 & 56.25 & 55.04 & 47.62 & 57.21 \\
o3 & 42.57 & 38.83 & 44.58 & 50.85 & 38.35 & 40.77 & 45.42 & 40.36 & 38.43 & 44.50 \\
o4-mini & 37.74 & 29.78 & 39.14 & 38.08 & 34.79 & 36.63 & 37.86 & 36.86 & 32.42 & 38.78 \\
Gemini-2.5-Flash & 37.32 & 27.61 & 36.42 & 35.33 & 32.59 & 33.06 & 33.34 & 27.42 & 29.51 & 34.93 \\
Gemini-2.5-Pro & 38.64 & 27.22 & 37.10 & 46.00 & 34.39 & 35.12 & 36.93 & 31.00 & 31.12 & 36.28 \\
Gemini-3-Pro & 39.51 & 35.97 & 37.17 & 40.49 & 34.14 & 35.35 & 35.49 & 30.03 & 32.14 & 35.18 \\
Claude-Opus-4.1 & 39.85 & 28.89 & 38.19 & 38.83 & 35.19 & 36.85 & 38.39 & 35.69 & 33.66 & 37.44 \\
Claude-Sonnet-4.5 & 42.11 & 34.89 & 42.38 & 44.20 & 35.24 & 37.31 & 38.14 & 34.44 & 32.13 & 40.90 \\
Grok-3 & 29.66 & 23.40 & 31.10 & 25.66 & 31.04 & 30.11 & 27.29 & 26.01 & 26.43 & 33.26 \\
Grok-4 & 33.75 & 25.48 & 33.78 & 35.22 & 30.44 & 30.96 & 30.30 & 27.54 & 27.58 & 33.61 \\
\bottomrule
\end{tabular}
}
\caption{\textbf{Idea Generation Across Subjects}: Subject-wise scores. Note: Because different subjects have different characteristics, the number of questions in each category is not the same (Figure~\ref{fig: data_distribution}). Therefore, the overall performance of the model cannot be obtained by directly averaging the values in the table.}
\label{tab:idea_subject_metric_table}
\end{table}

\begin{table}[t]
\centering
\renewcommand{\arraystretch}{0.9}
\setlength{\tabcolsep}{2pt}
\tiny
\resizebox{17cm}{!}{
\begin{tabular}{lcccccccccc}
\toprule
\textbf{Model} & \textbf{Astronomy} & \textbf{Chemistry} & \textbf{Earth} & \textbf{Energy} & \textbf{Information} & \textbf{Life} & \textbf{Material} & \textbf{Math} & \textbf{Neuroscience} & \textbf{Physics} \\
\midrule
DeepSeek-V3.2 & 31.25 & 0.00 & 20.83 & 10.00 & 14.29 & 27.50 & 44.44 & 29.41 & 16.67 & 17.24 \\
DeepSeek-R1 & 37.50 & 20.00 & 33.33 & 10.00 & 35.71 & 33.75 & 55.56 & 29.41 & 29.17 & 24.14 \\
Intern-S1 & 37.50 & 0.00 & 25.00 & 10.00 & 28.57 & 33.75 & 48.15 & 18.18 & 16.67 & 24.14 \\
Intern-S1-mini & 12.50 & 0.00 & 18.75 & 10.00 & 14.29 & 21.25 & 33.33 & 0.00 & 4.17 & 17.24 \\
Kimi-k2 & 43.75 & 0.00 & 22.92 & 20.00 & 21.43 & 33.75 & 44.44 & 16.67 & 20.83 & 34.48 \\
Qwen3-VL-235B-A22B & 37.50 & 0.00 & 29.17 & 10.00 & 14.29 & 35.00 & 40.74 & 16.67 & 20.83 & 24.14 \\
Qwen3-235B-A22B & 31.25 & 0.00 & 25.00 & 30.00 & 14.29 & 35.00 & 44.44 & 17.65 & 20.83 & 27.59 \\
Qwen3-Max & 50.00 & 0.00 & 31.25 & 30.00 & 28.57 & 37.50 & 48.15 & 22.22 & 25.00 & 24.14 \\
Qwen3-8B & 25.00 & 0.00 & 18.75 & 10.00 & 7.14 & 20.00 & 33.33 & 5.56 & 12.50 & 20.69 \\
Llama-4-Scout & 18.75 & 0.00 & 18.75 & 10.00 & 14.29 & 25.00 & 33.33 & 17.65 & 12.50 & 17.24 \\
GPT-4o & 37.50 & 0.00 & 27.08 & 10.00 & 14.29 & 35.00 & 51.85 & 22.22 & 20.83 & 20.69 \\
GPT-4.1 & 43.75 & 20.00 & 33.33 & 40.00 & 28.57 & 33.75 & 48.15 & 27.78 & 29.17 & 34.48 \\
GPT-5 & 37.50 & 0.00 & 27.08 & 40.00 & 35.71 & 31.25 & 40.74 & 22.22 & 20.83 & 27.59 \\
GPT-5.1 & 31.25 & 0.00 & 27.08 & 30.00 & 28.57 & 38.75 & 44.44 & 22.22 & 12.50 & 31.03 \\
GPT-5.2-Pro & 43.75 & 0.00 & 22.92 & 10.00 & 21.43 & 33.75 & 44.44 & 27.78 & 20.83 & 17.24 \\
o3 & 37.50 & 0.00 & 33.33 & 10.00 & 28.57 & 35.00 & 51.85 & 22.22 & 20.83 & 27.59 \\
o4-mini & 37.50 & 0.00 & 33.33 & 20.00 & 28.57 & 40.00 & 51.85 & 22.22 & 37.50 & 34.48 \\
Gemini-2.5-Flash & 18.75 & 0.00 & 18.75 & 10.00 & 14.29 & 23.75 & 37.04 & 27.78 & 16.67 & 13.79 \\
Gemini-2.5-Pro & 25.00 & 0.00 & 18.75 & 0.00 & 21.43 & 25.00 & 33.33 & 22.22 & 16.67 & 27.59 \\
Gemini-3-Pro & 37.50 & 0.00 & 32.61 & 30.00 & 38.46 & 38.46 & 55.56 & 37.50 & 34.78 & 28.57 \\
Claude-Opus-4.1 & 43.75 & 20.00 & 33.33 & 40.00 & 28.57 & 33.75 & 48.15 & 27.78 & 29.17 & 34.48 \\
Claude-Sonnet-4.5 & 43.75 & 20.00 & 35.42 & 30.00 & 21.43 & 41.25 & 51.85 & 27.78 & 25.00 & 27.59 \\
grok-3 & 31.25 & 0.00 & 29.17 & 20.00 & 14.29 & 32.50 & 40.74 & 11.11 & 20.83 & 24.14 \\
Grok-4 & 37.50 & 20.00 & 27.66 & 20.00 & 30.77 & 37.97 & 51.85 & 43.75 & 25.00 & 22.22 \\
\bottomrule
\end{tabular}
}
\caption{\textbf{Dry Experiment Across Subjects}: Subject-wise scores. Note: Because different subjects have different characteristics, the number of questions in each category is not the same (Figure~\ref{fig: data_distribution}). Therefore, the overall performance of the model cannot be obtained by directly averaging the values in the table.}
\label{tab:dry_subject_metric_table2}
\end{table}

\begin{table}[t]
\centering
\renewcommand{\arraystretch}{0.9}
\setlength{\tabcolsep}{2pt}
\tiny
\resizebox{14cm}{!}{
\begin{tabular}{lcccccc}
\toprule
\textbf{Model} & \textbf{Life-SS} & \textbf{Material-SS} & \textbf{Physics-SS} & \textbf{Life-PA} & \textbf{Material-PA} & \textbf{Physics-PA} \\
\midrule
DeepSeek-V3.2 & 15.47 & 20.20 & 16.67 & 21.48 & 26.87 & 22.50 \\
DeepSeek-R1 & 10.00 & 21.83 & 16.67 & 23.14 & 26.32 & 39.67 \\
Intern-S1 & 24.86 & 27.23 & 33.33 & 23.62 & 34.87 & 50.90 \\
Intern-S1-mini & 14.15 & 4.67 & 0.00 & 18.02 & 16.04 & 9.58 \\
Kimi-k2 & 18.10 & 28.04 & 0.00 & 25.13 & 37.42 & 18.61 \\
Qwen3-VL-235B-A22B & 17.78 & 30.25 & 16.67 & 28.66 & 43.11 & 46.68 \\
Qwen3-235B-A22B & 17.11 & 30.80 & 0.00 & 22.87 & 37.71 & 33.18 \\
Qwen3-Max & 17.37 & 41.11 & 33.33 & 24.44 & 45.67 & 56.70 \\
Qwen3-8B & 4.99 & 15.54 & 0.00 & 5.81 & 15.49 & 6.25 \\
Llama-4-Scout & 15.72 & 18.75 & 16.67 & 20.53 & 32.86 & 17.78 \\
GPT-4o & 20.79 & 29.10 & 32.38 & 31.58 & 41.06 & 41.41 \\
GPT-4.1 & 32.13 & 33.02 & 33.33 & 33.11 & 45.06 & 54.47 \\
GPT-5 & 7.81 & 11.72 & 33.76 & 19.31 & 21.50 & 23.18 \\
GPT-5.1 & 12.38 & 21.44 & 29.30 & 24.24 & 28.00 & 40.14 \\
GPT-5.2-Pro & 18.50 & 8.81 & 19.44 & 23.05 & 19.65 & 17.86 \\
o3 & 27.43 & 22.79 & 44.86 & 30.63 & 32.87 & 48.92 \\
o4-mini & 31.46 & 24.01 & 16.67 & 25.76 & 35.78 & 32.70 \\
Gemini-2.5-Flash & 5.31 & 23.44 & 15.71 & 14.73 & 28.09 & 32.03 \\
Gemini-2.5-Pro & 16.90 & 21.02 & 12.06 & 24.52 & 27.28 & 23.03 \\
Gemini-3-Pro & 20.00 & 34.88 & 33.33 & 32.21 & 41.07 & 36.12 \\
Claude-Opus-4.1 & 16.65 & 25.74 & 29.21 & 20.63 & 33.45 & 43.90 \\
Claude-Sonnet-4.5 & 31.75 & 25.83 & 16.67 & 28.62 & 33.78 & 46.97 \\
grok-3 & 28.97 & 41.93 & 33.33 & 32.52 & 43.94 & 58.32 \\
Grok-4 & 27.29 & 29.10 & 16.67 & 25.19 & 37.35 & 23.09 \\
\bottomrule
\end{tabular}
}
\caption{\textbf{Wet Experiment Across Subjects}: Scores across Action Sequence Similarity (SS) and Parameter Accuracy (PA) categories. Note: Because different subjects have different characteristics, the number of questions in each category is not the same (Figure~\ref{fig: data_distribution}). Therefore, the overall performance of the model cannot be obtained by directly averaging the values in the table.}
\label{tab:wet_subject_metric_table}
\end{table}

\begin{table}[t]
\centering
\renewcommand{\arraystretch}{0.9}
\setlength{\tabcolsep}{2pt}
\tiny
\resizebox{17cm}{!}{
\begin{tabular}{lccccccccc}
\toprule
\textbf{Model} & \textbf{Astronomy} & \textbf{Chemistry} & \textbf{Earth} & \textbf{Energy} & \textbf{Information} & \textbf{Life} & \textbf{Material} & \textbf{Neuroscience} & \textbf{Physics} \\
\midrule
Intern-S1 & 47.06 & 27.27 & 27.78 & 40.00 & 25.00 & 29.41 & 26.67 & 33.33 & 16.00 \\
Intern-S1-mini & 23.53 & 27.27 & 18.52 & 30.00 & 10.00 & 18.82 & 8.89 & 12.50 & 16.00 \\
Qwen3-VL-235B-A22B & 58.82 & 36.36 & 31.48 & 50.00 & 15.00 & 29.41 & 31.11 & 33.33 & 24.00 \\
Qwen3-VL-Max & 52.94 & 36.36 & 31.48 & 50.00 & 35.00 & 41.18 & 40.00 & 37.50 & 24.00 \\
Qwen3-VL-8B & 29.41 & 36.36 & 24.07 & 60.00 & 20.00 & 25.88 & 13.33 & 16.67 & 16.00 \\
Llama-4-Scout & 41.18 & 27.27 & 27.78 & 30.00 & 30.00 & 23.53 & 31.11 & 20.83 & 0.80 \\
GPT-4o & 41.18 & 54.55 & 37.04 & 60.00 & 20.00 & 29.41 & 31.11 & 20.83 & 28.00 \\
GPT-4.1 & 35.29 & 36.36 & 37.04 & 60.00 & 45.00 & 42.35 & 37.78 & 33.33 & 24.00 \\
GPT-5 & 70.59 & 36.36 & 37.04 & 30.00 & 50.00 & 37.65 & 33.33 & 41.67 & 20.00 \\
GPT-5.1 & 47.06 & 45.45 & 33.33 & 40.00 & 35.00 & 31.76 & 42.22 & 16.67 & 28.00 \\
GPT-5.2-Pro & 52.94 & 18.18 & 31.48 & 30.00 & 40.00 & 42.35 & 44.74 & 50.00 & 31.25 \\
o3 & 58.82 & 45.45 & 29.63 & 50.00 & 35.00 & 29.41 & 24.44 & 50.00 & 16.00 \\
o4-mini & 64.71 & 45.45 & 31.48 & 30.00 & 25.00 & 34.12 & 26.67 & 33.33 & 28.00 \\
Gemini-2.5-Flash & 52.94 & 27.27 & 33.33 & 40.00 & 45.00 & 36.47 & 24.44 & 37.50 & 24.00 \\
Gemini-2.5-Pro & 52.94 & 36.36 & 38.89 & 30.00 & 50.00 & 38.82 & 37.78 & 58.33 & 36.00 \\
Gemini-3-Pro & 47.06 & 45.45 & 35.19 & 50.00 & 45.00 & 40.00 & 48.89 & 62.50 & 20.00 \\
Claude-Opus-4.1 & 58.82 & 45.45 & 25.93 & 60.00 & 40.00 & 37.65 & 33.33 & 50.00 & 44.00 \\
Claude-Sonnet-4.5 & 52.94 & 36.36 & 37.04 & 20.00 & 40.00 & 35.29 & 42.22 & 29.17 & 44.00 \\
Grok-4 & 52.94 & 18.18 & 33.33 & 40.00 & 35.00 & 30.59 & 26.67 & 20.83 & 20.00 \\
\bottomrule
\end{tabular}
}
\caption{\textbf{Experimental Reasoning Across Subjects (Multi-choice Accuracy)}: Subject-wise scores across 10 scientific disciplines. Note: Because different subjects have different characteristics, the number of questions in each category is not the same (Figure~\ref{fig: data_distribution}). Therefore, the overall performance of the model cannot be obtained by directly averaging the values in the table.}
\label{tab:mcp_subject_metric_table}
\end{table}

%% file: sections/codes/code_00.tex
\begin{tcolorbox}[
    breakable,
    title=Example of Dry Experiment in Astronomy,
    colback=LighterGray,
    colframe=DeepPurple,
    colbacktitle=DeepPurple,
    coltitle=White,
]
\textbf{\emph{\textcolor{DeepPurple}{Background}}}

The Zwicky Transient Facility (ZTF) is an advanced optical time-domain sky survey utilizing the Palomar 48-inch Schmidt telescope equipped with a custom wide-field CCD camera. This camera covers a 47.7 square degree field of view with 16 large-format CCDs, enabling a survey speed over an order of magnitude faster than its predecessor. The system achieves a median image quality of approximately 2.0 arcseconds full-width at half-maximum (FWHM) across g, r, and i bands, with typical 5-sigma limiting magnitudes near 20.8 (g), 20.6 (r), and 19.9 (i) in 30-second exposures, improving under dark-sky conditions.

The optical design addresses the Schmidt telescopes curved focal surface through a combination of a modified Schmidt corrector, a meniscus dewar window, faceted cold plate mounting, and individual field flattener lenses above each CCD. The cameras cryostat and readout electronics are optimized for minimal beam obstruction and rapid 8.2-second readout with low noise (~10 electrons median). A robotic observing system and scheduler maximize volumetric survey speed by selecting fields on a fixed grid with minimal dithering, enabling efficient coverage of the Northern sky and Galactic plane.

ZTFs data system performs near-real-time image processing, including bias subtraction, flat-fielding, astrometric and photometric calibration, and image differencing using the ZOGY algorithm to detect transient and variable sources. Alerts containing rich contextual information and machine-learning-based Real-Bogus scores are distributed via a scalable streaming system to community brokers. The system also supports solar system science by detecting both point-like and streaked moving objects, linking detections into orbits, and reporting to the Minor Planet Center.

Early scientific results demonstrate ZTFs capability to discover and classify supernovae, including young Type II events, and to conduct rapid follow-up of multi-messenger triggers such as neutrinos and gamma-ray bursts. The facility also enables studies of variable stars, exemplified by light curves of Be stars and RR Lyrae, and solar system objects, including near-Earth asteroids, asteroid rotation periods, comet activity, and Centaur outbursts.

ZTFs public surveys include a three-day cadence Northern Sky Survey and a nightly Galactic Plane Survey, with observations typically taken twice per night in g and r bands. The surveys moderate depth and high cadence complement future facilities by providing early discovery and characterization of bright transients accessible to moderate-aperture telescopes. ZTF serves as a pathfinder for next-generation surveys, offering a prototype alert stream and extensive time-domain data products to the astronomical community.

\textbf{\emph{\textcolor{DeepPurple}{Data Code}}}

\begin{lstlisting}
#!/usr/bin/env python3
# -*- coding: utf-8 -*-
"""
Paper: The Zwicky Transient Facility: System Overview, Performance, and First Results
Authors: Eric C. Bellm, Shrinivas R. Kulkarni, Matthew J. Graham, et al.
Year: 2019

This script generates synthetic asteroid light curve data based on the descriptions
in Section 6.4.2 of the paper.

Python Version: 3.10.12
"""

import sys
assert sys.version_info >= (3, 10), "This code requires Python 3.10 or higher"

# Dependencies
# pip install numpy==1.24.3 pandas==2.0.3

import numpy as np
import pandas as pd
from pathlib import Path
from typing import Tuple

# Global constants
DATA_DIR = Path("data")
DEFAULT_FILE_PATH = DATA_DIR / "asteroid_light_curve.csv"
RANDOM_SEED = 42 # Ensure reproducible results

def generate_asteroid_light_curve(
    file_path: str,
    n_points: int = 150,
    period_hr: float = 2.25,
    amplitude: float = 0.2,
    mag_range: Tuple[float, float] = (17.8, 18.2),
    noise_level: float = 0.03
) -> None:
    """
    Generates a synthetic asteroid light curve and saves it to a CSV file.
    The light curve is modeled as a simple cosine function with added Gaussian noise.
    This mimics the data for an asteroid like (11014) Svatopluk in Figure 10(a).

    Tag: [Data download]

    Args:
        file_path (str): The path to save the output CSV file.
        n_points (int): The number of data points to generate.
        period_hr (float): The rotation period of the asteroid in hours.
        amplitude (float): The amplitude of the light curve variation in magnitudes.
        mag_range (Tuple[float, float]): The approximate magnitude range of the asteroid.
        noise_level (float): The standard deviation of the Gaussian noise to add to the magnitudes.

    Returns:
        None

    Examples:
        >>> Path("data").mkdir(exist_ok=True)
        >>> file_path = "data/test_light_curve.csv"
        >>> generate_asteroid_light_curve(file_path, n_points=50)
        >>> import pandas as pd
        >>> df = pd.read_csv(file_path)
        >>> print(df.shape)
        (50, 3)
    """
    # Set random seed to ensure reproducibility
    np.random.seed(0+RANDOM_SEED)

    # Generate unevenly sampled time points to simulate real observations
    # Observation time span is about 4 hours
    observation_span_hr = period_hr * 1.8
    times = np.sort(np.random.rand(n_points) * observation_span_hr)

    # Calculate the baseline magnitude of the light curve
    base_magnitude = np.mean(mag_range)

    # Use cosine function to simulate asteroid brightness variation
    # Multiply by 2 because a full rotation period usually contains two peaks and two troughs
    magnitudes_true = base_magnitude - amplitude * np.cos(2 * np.pi * times / period_hr * 2)

    # Add Gaussian noise to the observed data
    noise = np.random.normal(0, noise_level, n_points)
    magnitudes_obs = magnitudes_true + noise

    # Generate error for each data point, related to noise level
    errors = np.random.normal(noise_level, noise_level / 4, n_points)
    errors = np.maximum(errors, noise_level / 2) # Ensure errors are not too small

    # Create a DataFrame to store the data
    df = pd.DataFrame({
        'time_hr': times,
        'magnitude': magnitudes_obs,
        'error': errors
    })

    # Save to CSV file
    df.to_csv(file_path, index=False)
    print(f"Successfully generated synthetic light curve data and saved to: {file_path}")

if __name__ == "__main__":
    # Ensure data directory exists
    DATA_DIR.mkdir(exist_ok=True)

    # Generate simulated data
    generate_asteroid_light_curve(
        file_path=str(DEFAULT_FILE_PATH),
        n_points=150,
        period_hr=2.25, # Asteroid period corresponding to Figure 10(a)
        amplitude=0.15, # Amplitude
        mag_range=(17.8, 18.1), # Magnitude range
        noise_level=0.02 # Noise level
    )
\end{lstlisting}

\textbf{\emph{\textcolor{DeepPurple}{Main Code with Incomplete Functions}}}

\begin{lstlisting}
#!/usr/bin/env python3
# -*- coding: utf-8 -*-
"""
Paper: The Zwicky Transient Facility: System Overview, Performance, and First Results
Authors: Eric C. Bellm, Shrinivas R. Kulkarni, Matthew J. Graham, et al.
Year: 2019

This script implements the asteroid light curve analysis from Section 6.4.2.
It determines the rotation period of an asteroid from its light curve using
a Lomb-Scargle periodogram and Fourier series fitting.

Python Version: 3.10.12
"""

import sys

assert sys.version_info >= (3, 10), "This code requires Python 3.10 or higher"

# Dependencies
# pip install numpy==1.24.3 pandas==2.0.3 scipy==1.10.1

import numpy as np
import pandas as pd
from scipy.signal import lombscargle
from typing import Tuple
from pathlib import Path

# Global constants
DATA_FILE_PATH = "data/asteroid_light_curve.csv"
# Order of Fourier series, the paper mentions second-order Fourier series
FOURIER_N_TERMS = 2
# Numerical stability constant
EPSILON = 1e-9

def load_light_curve_data(file_path: str) -> pd.DataFrame:
    """
    Loads asteroid light curve data from a CSV file.

    Tag: [Data loading]

    Args:
        file_path (str): The path to the CSV file.

    Returns:
        pd.DataFrame: A DataFrame containing the light curve data with
                      columns 'time_hr', 'magnitude', and 'error'.

    Examples:
        >>> Path("data").mkdir(exist_ok=True)
        >>> data = {'time_hr': [0, 1], 'magnitude': [18.0, 18.1], 'error': [0.01, 0.01]}
        >>> df = pd.DataFrame(data)
        >>> df.to_csv("data/dummy.csv", index=False)
        >>> loaded_df = load_light_curve_data("data/dummy.csv")
        >>> print(loaded_df.shape)
        (2, 3)
    """
    try:
        return pd.read_csv(file_path)
    except FileNotFoundError:
        print(f"Error: Data file not found at '{file_path}'")
        print("Please run 'data.py' first to generate the data file.")
        sys.exit(1)

def calculate_lomb_scargle_periodogram(
        times: np.ndarray,
        magnitudes: np.ndarray,
        min_period: float = 0.5,
        max_period: float = 5.0,
        num_periods: int = 10000
) -> Tuple[np.ndarray, np.ndarray]:
    """
    Calculates the Lomb-Scargle periodogram for unevenly sampled data.

    Tag: [Numerical calculation]

    Args:
        times (np.ndarray): Array of time points.
        magnitudes (np.ndarray): Array of magnitude measurements.
        min_period (float): The minimum period to test.
        max_period (float): The maximum period to test.
        num_periods (int): The number of period points to evaluate.

    Returns:
        Tuple[np.ndarray, np.ndarray]: A tuple containing the periods tested
                                       and the corresponding periodogram power.

    Examples:
        >>> times = np.linspace(0, 4, 50)
        >>> magnitudes = 18 + 0.1 * np.sin(2 * np.pi * times / 2.0)
        >>> periods, power = calculate_lomb_scargle_periodogram(times, magnitudes)
        >>> print(periods.shape, power.shape)
        (10000,) (10000,)
    """
    pass # [Please complete the code]

def find_best_period_from_periodogram(
        periods: np.ndarray,
        power: np.ndarray
) -> float:
    """
    Finds the period corresponding to the highest power in the periodogram.

    Tag: [Numerical calculation]

    Args:
        periods (np.ndarray): Array of periods.
        power (np.ndarray): Array of periodogram powers.

    Returns:
        float: The period with the highest power.

    Examples:
        >>> periods = np.array([1.0, 2.0, 3.0])
        >>> power = np.array([0.1, 0.8, 0.2])
        >>> best_period = find_best_period_from_periodogram(periods, power)
        >>> print(best_period)
        2.0
    """
    pass # [Please complete the code]

def build_fourier_design_matrix(
        times: np.ndarray,
        period: float,
        n_terms: int
) -> np.ndarray:
    """
    Builds the design matrix for a Fourier series linear least-squares fit.

    Tag: [Predictive modeling]

    Args:
        times (np.ndarray): Array of time points.
        period (float): The fundamental period of the Fourier series.
        n_terms (int): The number of Fourier terms (harmonics) to include.

    Returns:
        np.ndarray: The design matrix for the fit.

    Examples:
        >>> times = np.array([0, 1, 2])
        >>> period = 4.0
        >>> n_terms = 1
        >>> matrix = build_fourier_design_matrix(times, period, n_terms)
        >>> print(matrix.shape)
        (3, 3)
    """
    # Fundamental frequency
    omega = 2 * np.pi / (period + EPSILON)
    # Initialize a column vector for the constant term
    design_matrix = [np.ones_like(times)]
    # Loop to add sin and cos terms for each order
    for i in range(1, n_terms + 1):
        design_matrix.append(np.sin(i * omega * times))
        design_matrix.append(np.cos(i * omega * times))
    return np.vstack(design_matrix).T

def fit_fourier_model(
        design_matrix: np.ndarray,
        magnitudes: np.ndarray,
        errors: np.ndarray
) -> np.ndarray:
    """
    Fits a Fourier model using weighted linear least squares.

    Tag: [Predictive modeling]

    Args:
        design_matrix (np.ndarray): The design matrix from build_fourier_design_matrix.
        magnitudes (np.ndarray): Array of magnitude measurements.
        errors (np.ndarray): Array of measurement errors for weighting.

    Returns:
        np.ndarray: The array of fitted Fourier coefficients.

    Examples:
        >>> times = np.linspace(0, 4, 10)
        >>> period = 2.0
        >>> magnitudes = np.sin(2 * np.pi * times / period)
        >>> errors = np.full_like(magnitudes, 0.1)
        >>> matrix = build_fourier_design_matrix(times, period, 1)
        >>> coeffs = fit_fourier_model(matrix, magnitudes, errors)
        >>> print(len(coeffs))
        3
    """
    # Use errors for weighting
    weights = 1.0 / (errors + EPSILON)
    weighted_matrix = design_matrix * weights[:, np.newaxis]
    weighted_magnitudes = magnitudes * weights

    # Solve using least squares
    coeffs, _, _, _ = np.linalg.lstsq(weighted_matrix, weighted_magnitudes, rcond=None)
    return coeffs

def evaluate_fourier_model(
        design_matrix: np.ndarray,
        coeffs: np.ndarray
) -> np.ndarray:
    """
    Evaluates the Fourier model at given time points.

    Tag: [Numerical calculation]

    Args:
        design_matrix (np.ndarray): The design matrix.
        coeffs (np.ndarray): The fitted Fourier coefficients.

    Returns:
        np.ndarray: The predicted magnitudes from the model.

    Examples:
        >>> times = np.array([0, 1, 2])
        >>> period = 4.0
        >>> n_terms = 1
        >>> matrix = build_fourier_design_matrix(times, period, n_terms)
        >>> coeffs = np.array([18.0, 0.1, 0.0])
        >>> model_mags = evaluate_fourier_model(matrix, coeffs)
        >>> print(model_mags.shape)
        (3,)
    """
    return np.dot(design_matrix, coeffs)

def calculate_reduced_chi_squared(
        observed: np.ndarray,
        expected: np.ndarray,
        errors: np.ndarray,
        num_fit_params: int
) -> float:
    """
    Calculates the reduced chi-squared statistic for a fit.

    Tag: [Metric calculation]

    Args:
        observed (np.ndarray): The observed data values.
        expected (np.ndarray): The model's expected values.
        errors (np.ndarray): The errors on the observed values.
        num_fit_params (int): The number of free parameters in the model.

    Returns:
        float: The reduced chi-squared value.

    Examples:
        >>> obs = np.array([1, 2, 3])
        >>> exp = np.array([1.1, 2.2, 2.9])
        >>> err = np.array([0.2, 0.2, 0.2])
        >>> r_chi2 = calculate_reduced_chi_squared(obs, exp, err, 1)
        >>> print(f"{r_chi2:.2f}")
        1.25
    """
    # Calculate chi-squared value
    chi_squared = np.sum(((observed - expected) / (errors + EPSILON)) ** 2)
    # Calculate degrees of freedom
    degrees_of_freedom = len(observed) - num_fit_params
    # Avoid division by zero
    if degrees_of_freedom <= 0:
        return np.inf
    return chi_squared / degrees_of_freedom

if __name__ == "__main__":
    print("--- ZTF Asteroid Light Curve Analysis ---")

    # 1. Load data
    print(f"\n[1/4] Loading light curve data from '{DATA_FILE_PATH}'...")
    light_curve_df = load_light_curve_data(DATA_FILE_PATH)
    times = light_curve_df['time_hr'].values
    magnitudes = light_curve_df['magnitude'].values
    errors = light_curve_df['error'].values
    print(f"Successfully loaded {len(times)} data points.")

    # 2. Calculate Lomb-Scargle periodogram
    print("\n[2/4] Calculating periodogram using Lomb-Scargle method...")
    # Set a reasonable period search range
    min_p = 0.5  # hours
    max_p = (times.max() - times.min())  # observation span as max period
    periods, power = calculate_lomb_scargle_periodogram(times, magnitudes, min_period=min_p, max_period=max_p)
    best_period = find_best_period_from_periodogram(periods, power)
    print(f"Periodogram analysis complete. Most likely light curve period is: {best_period:.4f} hours.")

    # 3. Fit Fourier series model
    print(f"\n[3/4] Fitting {FOURIER_N_TERMS}-order Fourier series using found period {best_period:.4f} hours...")
    design_matrix = build_fourier_design_matrix(times, best_period, FOURIER_N_TERMS)
    fourier_coeffs = fit_fourier_model(design_matrix, magnitudes, errors)
    model_magnitudes = evaluate_fourier_model(design_matrix, fourier_coeffs)
    print("Fourier model fitting complete.")
    print(f"Fitted coefficients: {np.round(fourier_coeffs, 4)}")

    # 4. Evaluate goodness of fit
    print("\n[4/4] Evaluating model goodness of fit...")
    num_params = 1 + 2 * FOURIER_N_TERMS  # 1 constant term + n*2 sin/cos terms
    r_chi2 = calculate_reduced_chi_squared(magnitudes, model_magnitudes, errors, num_params)
    print(f"Reduced chi-squared of the model: {r_chi2:.4f}")
    if 0.5 < r_chi2 < 2.0:
        print("This is a reasonable fit.")
    else:
        print("Warning: The fit may be poor or error estimates inaccurate.")

    # Final output
    # The core goal in the paper for finding asteroid rotation period is to obtain the period value
    print("\n--- Analysis Complete ---")
    print("[Final Output]")
    # Since the asteroid rotation causes two peaks in the light curve per rotation,
    # the physical rotation period is twice the light curve period found by Lomb-Scargle
    # This is a common convention in astronomy
    rotation_period = best_period * 2.0
    print(f"{rotation_period:.4f}")
\end{lstlisting}

\textbf{\emph{\textcolor{DeepPurple}{Answer}}}

\begin{lstlisting}
def calculate_lomb_scargle_periodogram(
        times: np.ndarray,
        magnitudes: np.ndarray,
        min_period: float = 0.5,
        max_period: float = 5.0,
        num_periods: int = 10000
) -> Tuple[np.ndarray, np.ndarray]:
    """
    Calculates the Lomb-Scargle periodogram for unevenly sampled data.

    Tag: [Numerical calculation]

    Args:
        times (np.ndarray): Array of time points.
        magnitudes (np.ndarray): Array of magnitude measurements.
        min_period (float): The minimum period to test.
        max_period (float): The maximum period to test.
        num_periods (int): The number of period points to evaluate.

    Returns:
        Tuple[np.ndarray, np.ndarray]: A tuple containing the periods tested
                                       and the corresponding periodogram power.

    Examples:
        >>> times = np.linspace(0, 4, 50)
        >>> magnitudes = 18 + 0.1 * np.sin(2 * np.pi * times / 2.0)
        >>> periods, power = calculate_lomb_scargle_periodogram(times, magnitudes)
        >>> print(periods.shape, power.shape)
        (10000,) (10000,)
    """
    # 1. Define and generate the search range for ordinary frequencies (f = 1/P)
    safe_min_period = max(min_period, EPSILON)
    min_freq = 1.0 / max_period
    max_freq = 1.0 / safe_min_period
    ordinary_frequencies = np.linspace(min_freq, max_freq, num_periods)

    # 2. [Key modification] Convert ordinary frequencies to angular frequencies (omega = 2*pi*f)
    # scipy.signal.lombscargle requires angular frequencies as input
    angular_frequencies = 2 * np.pi * ordinary_frequencies

    # 3. Center magnitude data to remove DC component
    magnitudes_centered = magnitudes - np.mean(magnitudes)

    # 4. Calculate periodogram power using angular frequencies
    power = lombscargle(times, magnitudes_centered, angular_frequencies, normalize=True)

    # 5. Return periods (P = 1/f) corresponding to ordinary frequencies and power
    periods = 1.0 / (ordinary_frequencies + EPSILON)
    return periods, power

def find_best_period_from_periodogram(
        periods: np.ndarray,
        power: np.ndarray
) -> float:
    """
    Finds the period corresponding to the highest power in the periodogram.

    Tag: [Numerical calculation]

    Args:
        periods (np.ndarray): Array of periods.
        power (np.ndarray): Array of periodogram powers.

    Returns:
        float: The period with the highest power.

    Examples:
        >>> periods = np.array([1.0, 2.0, 3.0])
        >>> power = np.array([0.1, 0.8, 0.2])
        >>> best_period = find_best_period_from_periodogram(periods, power)
        >>> print(best_period)
        2.0
    """
    best_period_index = np.argmax(power)
    return periods[best_period_index]
\end{lstlisting}

\end{tcolorbox}

%% file: sections/codes/code_01.tex
\begin{tcolorbox}[
    breakable,
    title=Example of Dry Experiment in Earth,
    colback=LighterGray,
    colframe=DeepPurple,
    colbacktitle=DeepPurple,
    coltitle=White,
]
\textbf{\emph{\textcolor{DeepPurple}{Background}}}

Surface ozone is a secondary air pollutant formed by photochemical reactions involving carbon monoxide (CO), volatile organic compounds (VOCs), nitrogen oxides (NOx = NO + NO2), and sunlight. It poses significant risks to human health, including respiratory and cardiovascular effects, and damages vegetation by reducing crop yields and ecosystem productivity. While stringent emission controls since the 1990s have reduced ozone pollution in many Western regions, rapid industrialization and urbanization in East Asia, particularly China, have led to increasing ozone precursor emissions and elevated surface ozone levels.

Recent nationwide monitoring in China, initiated around 2013, reveals that although median ozone concentrations during the warm season (AprilSeptember) are comparable to those in industrialized regions such as Japan, South Korea, Europe, and the United States, the frequency and magnitude of high-ozone events are substantially greater in China. Key metrics include the fourth highest daily maximum 8-hour average ozone (4MDA8), the number of days exceeding 70 ppb (NDGT70), and cumulative exposure indices like SOMO35 (sum of ozone means over 35 ppb). Chinas warm-season 4MDA8 averages around 86 ppb, exceeding other regions by 630\%, while NDGT70 values are 93575\% higher, indicating more frequent episodes of elevated ozone. Vegetation exposure metrics such as AOT40 and W126, which correlate with ozone-induced plant damage, are also significantly elevated in China, suggesting greater risks to agricultural productivity and ecosystem health.

Spatially, ozone pollution hotspots in China are concentrated in densely populated and industrialized regions including the North China Plain, Yangtze River Delta, and Pearl River Delta, with some western areas affected due to topography and local emissions. Seasonal patterns show ozone peaks in late spring and early summer, influenced by regional meteorology such as the Asian summer monsoon, which modulates photochemical activity and pollutant transport.

Temporal analysis from 2013 to 2017 indicates a rising trend in ozone levels across Chinese cities, with annual increases in exposure metrics ranging from approximately 3.7\% to over 15\% per year. This contrasts with stable or declining ozone trends in Europe and the United States over recent decades. The increase in ozone occurs despite reductions in primary pollutants like SO2, NO2, CO, and fine particulate matter (PM2.5), reflecting complex photochemical interactions. In particular, reductions in NOx or PM2.5 can paradoxically enhance ozone formation in VOC-sensitive regimes prevalent in eastern China. Rising VOC emissions and meteorological factors such as hotter, drier summers also contribute to elevated ozone.

These findings highlight China as a current global hotspot for surface ozone pollution, with greater human and vegetation exposure than other industrialized regions with extensive monitoring. The severity and increasing trend of ozone pollution pose challenges for air quality management, indicating a need for targeted control strategies focusing on VOC emissions and comprehensive understanding of chemical and meteorological influences on ozone formation.

\textbf{\emph{\textcolor{DeepPurple}{Data Code}}}

\begin{lstlisting}
#!/usr/bin/env python3
# -*- coding: utf-8 -*-
"""
Paper: Severe Surface Ozone Pollution in China: A Global Perspective
Authors: Xiao Lu, Jiayun Hong, Lin Zhang, et al.
Year: 2018

Data generation script for simulating hourly surface ozone data.
Python Version: 3.10.12
"""

import sys

assert sys.version_info >= (3, 10), "This code requires Python 3.10 or higher"

# Dependencies
# pip install numpy==1.24.3 pandas==2.0.3

import numpy as np
import pandas as pd
from pathlib import Path
import os

def generate_hourly_ozone_data(
        n_sites: int,
        start_date: str,
        end_date: str,
        region_params: dict
) -> pd.DataFrame:
    """
    Generate synthetic hourly ozone data for multiple sites.
    Tag: [Simulation]

    Args:
        n_sites (int): The number of monitoring sites to simulate.
        start_date (str): The start date for the data series (e.g., '2013-01-01').
        end_date (str): The end date for the data series (e.g., '2017-12-31').
        region_params (dict): A dictionary containing parameters for the region.
            Keys should include 'base_mean', 'seasonal_amp', 'daily_amp',
            'noise_level', 'event_prob', 'event_strength'.

    Returns:
        pd.DataFrame: A DataFrame with columns ['site_id', 'timestamp', 'ozone_ppb'].

    Examples:
        >>> params = {
        ...     'base_mean': 40, 'seasonal_amp': 15, 'daily_amp': 20,
        ...     'noise_level': 5, 'event_prob': 0.02, 'event_strength': 40
        ... }
        >>> df = generate_hourly_ozone_data(2, '2017-01-01', '2017-01-31', params)
        >>> print(df.shape)
        (1488, 3)
    """
    np.random.seed(0+42)
    timestamps = pd.to_datetime(np.arange(
        np.datetime64(start_date),
        np.datetime64(end_date) + np.timedelta64(1, 'D'),
        np.timedelta64(1, 'h')
    ))
    n_hours = len(timestamps)

    # Prepare time feature vectors
    day_of_year = timestamps.dayofyear
    hour_of_day = timestamps.hour

    all_sites_data = []
    for site_id in range(n_sites):
        # Base signal = seasonal cycle + daily cycle
        seasonal_cycle = region_params['seasonal_amp'] * np.sin(2 * np.pi * (day_of_year - 90) / 365.25)
        daily_cycle = region_params['daily_amp'] * np.sin(2 * np.pi * (hour_of_day - 8) / 24)

        base_signal = region_params['base_mean'] + seasonal_cycle + daily_cycle

        # Add random noise
        noise = np.random.randn(n_hours) * region_params['noise_level']

        # Simulate high pollution events
        events = np.zeros(n_hours)
        for i in range(n_hours):
            if np.random.rand() < region_params['event_prob']:
                # Pollution event lasts 24-72 hours
                duration = np.random.randint(24, 73)
                event_end = min(i + duration, n_hours)
                event_shape = np.sin(np.linspace(0, np.pi, event_end - i))
                events[i:event_end] += region_params['event_strength'] * event_shape

        # Compose final signal
        ozone_concentration = base_signal + noise + events
        # Ensure concentration values are non-negative
        ozone_concentration = np.maximum(ozone_concentration, 0)

        site_df = pd.DataFrame({
            'site_id': f'site_{site_id}',
            'timestamp': timestamps,
            'ozone_ppb': ozone_concentration
        })
        all_sites_data.append(site_df)

    return pd.concat(all_sites_data, ignore_index=True)

def save_data_to_csv(df: pd.DataFrame, file_path: str):
    """
    Save a DataFrame to a CSV file.
    Tag: [Data saving]

    Args:
        df (pd.DataFrame): The DataFrame to save.
        file_path (str): The path to the output CSV file.

    Returns:
        None

    Examples:
        >>> data = pd.DataFrame({'col1': [1, 2], 'col2': [3, 4]})
        >>> save_data_to_csv(data, 'data/test.csv')
    """
    # Ensure directory exists
    output_dir = os.path.dirname(file_path)
    if not os.path.exists(output_dir):
        os.makedirs(output_dir)
    df.to_csv(file_path, index=False)
    print(f"Data has been saved to: {file_path}")

if __name__ == "__main__":
    # Set different parameters for China and JKEU regions according to the paper description
    # China region parameters: higher base value, stronger seasonal and daily variations, more frequent and stronger high pollution events
    china_params = {
        'base_mean': 45,
        'seasonal_amp': 20,
        'daily_amp': 25,
        'noise_level': 8,
        'event_prob': 0.015,  # Higher event occurrence probability
        'event_strength': 50  # Stronger event intensity
    }

    # JKEU region parameters: relatively moderate pollution levels
    jkeu_params = {
        'base_mean': 35,
        'seasonal_amp': 15,
        'daily_amp': 20,
        'noise_level': 5,
        'event_prob': 0.005,  # Lower event occurrence probability
        'event_strength': 30  # Weaker event intensity
    }

    # To reduce runtime, we only simulate one year of data
    START_DATE = '2017-01-01'
    END_DATE = '2017-12-31'
    NUM_SITES = 10  # Simulate 10 sites

    print("Generating simulated ozone data for China region...")
    china_ozone_data = generate_hourly_ozone_data(NUM_SITES, START_DATE, END_DATE, china_params)

    print("Generating simulated ozone data for JKEU region...")
    jkeu_ozone_data = generate_hourly_ozone_data(NUM_SITES, START_DATE, END_DATE, jkeu_params)

    # Create data directory and save files
    data_dir = Path("data")
    data_dir.mkdir(exist_ok=True)

    save_data_to_csv(china_ozone_data, str(data_dir / "china_ozone_data.csv"))
    save_data_to_csv(jkeu_ozone_data, str(data_dir / "jkeu_ozone_data.csv"))
\end{lstlisting}

\textbf{\emph{\textcolor{DeepPurple}{Main Code with Incomplete Functions}}}

\begin{lstlisting}
#!/usr/bin/env python3
# -*- coding: utf-8 -*-
"""
Paper: Severe Surface Ozone Pollution in China: A Global Perspective
Authors: Xiao Lu, Jiayun Hong, Lin Zhang, et al.
Year: 2018

Implementation of ozone pollution metrics calculation.
Python Version: 3.10.12
"""

import sys

assert sys.version_info >= (3, 10), "This code requires Python 3.10 or higher"

# Dependencies
# pip install numpy==1.24.3 pandas==2.0.3

import numpy as np
import pandas as pd

def load_ozone_data(file_path: str) -> pd.DataFrame:
    """
    Load hourly ozone data from a CSV file.
    Tag: [Data loading]

    Args:
        file_path (str): The path to the CSV data file.

    Returns:
        pd.DataFrame: A DataFrame with a datetime index and ozone data.

    Examples:
        >>> df = load_ozone_data('data/china_ozone_data.csv')
        >>> print(df.columns)
        Index(['site_id', 'ozone_ppb'], dtype='object')
    """
    df = pd.read_csv(file_path, parse_dates=['timestamp'])
    df = df.set_index('timestamp')
    return df

def calculate_mda8(daily_hourly_data: pd.Series) -> float:
    """
    Calculate the Daily Maximum 8-hour Average (MDA8) ozone concentration.
    Tag: [Numerical calculation]

    Args:
        daily_hourly_data (pd.Series): A Series of 24 hourly ozone values for a single day.

    Returns:
        float: The MDA8 value in ppb. Returns np.nan if data is insufficient.

    Examples:
        >>> hours = pd.to_datetime(pd.date_range('2023-07-01', periods=24, freq='h'))
        >>> data = pd.Series(np.sin(np.linspace(0, 2*np.pi, 24)) * 20 + 50, index=hours)
        >>> mda8 = calculate_mda8(data)
        >>> print(round(mda8, 2))
        67.68
    """
    pass # [Please complete the code]

def calculate_4mda8(mda8_series: pd.Series) -> float:
    """
    Calculate the 4th highest MDA8 value for a given period.
    Tag: [Numerical calculation]

    Args:
        mda8_series (pd.Series): A Series of daily MDA8 values.

    Returns:
        float: The 4th highest MDA8 value. Returns np.nan if data is insufficient.

    Examples:
        >>> data = pd.Series([80, 90, 70, 100, 110, 60])
        >>> val = calculate_4mda8(data)
        >>> print(val)
        80.0
    """
    valid_mda8 = mda8_series.dropna()
    if len(valid_mda8) < 4:
        return np.nan
    return valid_mda8.sort_values(ascending=False).iloc[3]

def calculate_ndgt70(mda8_series: pd.Series) -> int:
    """
    Calculate the total number of days with MDA8 values > 70 ppb.
    Tag: [Numerical calculation]

    Args:
        mda8_series (pd.Series): A Series of daily MDA8 values.

    Returns:
        int: The count of days where MDA8 > 70 ppb.

    Examples:
        >>> data = pd.Series([65, 71, 85, 70, 70.1])
        >>> count = calculate_ndgt70(data)
        >>> print(count)
        3
    """
    return (mda8_series > 70).sum()

def calculate_aot40(hourly_data: pd.Series) -> float:
    """
    Calculate the AOT40 (Accumulated Ozone over a Threshold of 40 ppb).
    Tag: [Numerical calculation]

    Args:
        hourly_data (pd.Series): A Series of hourly ozone data for the entire period.
                                 The function will filter for daytime hours (08:00-19:59).

    Returns:
        float: The total AOT40 value in ppb-hours.

    Examples:
        >>> hours = pd.to_datetime(pd.date_range('2023-07-01', periods=24, freq='h'))
        >>> data = pd.Series(np.arange(30, 54), index=hours)
        >>> aot40 = calculate_aot40(data)
        >>> print(aot40)
        91.0
    """
    daytime_data = hourly_data[(hourly_data.index.hour >= 8) & (hourly_data.index.hour <= 19)]
    # Calculate the portion exceeding 40 ppb each hour
    exceedances = np.maximum(0, daytime_data - 40)
    return exceedances.sum()

def calculate_w126(hourly_data: pd.Series) -> float:
    """
    Calculate the W126 metric, a weighted cumulative exposure index.
    Tag: [Numerical calculation]

    Args:
        hourly_data (pd.Series): A Series of hourly ozone data for the entire period.
                                 The function will filter for daytime hours (08:00-19:59).

    Returns:
        float: The total W126 value in ppb-hours.

    Examples:
        >>> hours = pd.to_datetime(pd.date_range('2023-07-01', periods=24, freq='h'))
        >>> data = pd.Series(np.full(24, 80), index=hours) # Constant 80 ppb
        >>> w126 = calculate_w126(data)
        >>> print(round(w126, 2))
        954.16
    """
    pass # [Please complete the code]

def analyze_regional_metrics(df: pd.DataFrame) -> dict:
    """
    Analyze and compute all key ozone metrics for a given region's data.
    Tag: [Statistical analysis]

    Args:
        df (pd.DataFrame): The DataFrame containing hourly ozone data for a region.

    Returns:
        dict: A dictionary of regionally-averaged ozone metrics.

    Examples:
        >>> df = load_ozone_data('data/china_ozone_data.csv')
        >>> metrics = analyze_regional_metrics(df)
        >>> print(metrics.keys())
        dict_keys(['avg_4mda8', 'avg_ndgt70', 'avg_aot40', 'avg_w126'])
    """
    site_metrics = []
    # Group by site for calculation
    for site_id, site_data in df.groupby('site_id'):
        # Filter warm season data (April-September)
        warm_season_data = site_data[(site_data.index.month >= 4) & (site_data.index.month <= 9)]

        # Calculate daily MDA8
        daily_mda8 = warm_season_data['ozone_ppb'].resample('D').apply(calculate_mda8).dropna()

        if daily_mda8.empty:
            continue

        # Calculate various metrics
        m4da8 = calculate_4mda8(daily_mda8)
        ndgt70 = calculate_ndgt70(daily_mda8)
        aot40 = calculate_aot40(warm_season_data['ozone_ppb'])
        w126 = calculate_w126(warm_season_data['ozone_ppb'])

        site_metrics.append({
            'site_id': site_id,
            '4mda8': m4da8,
            'ndgt70': ndgt70,
            'aot40': aot40,
            'w126': w126
        })

    # Calculate regional averages
    metrics_df = pd.DataFrame(site_metrics).dropna()
    if metrics_df.empty:
        return {
            'avg_4mda8': 0, 'avg_ndgt70': 0, 'avg_aot40': 0, 'avg_w126': 0
        }

    regional_avg = {
        'avg_4mda8': metrics_df['4mda8'].mean(),
        'avg_ndgt70': metrics_df['ndgt70'].mean(),
        'avg_aot40': metrics_df['aot40'].mean(),
        'avg_w126': metrics_df['w126'].mean()
    }
    return regional_avg

if __name__ == "__main__":
    # Load data
    try:
        china_df = load_ozone_data('data/china_ozone_data.csv')
        jkeu_df = load_ozone_data('data/jkeu_ozone_data.csv')
    except FileNotFoundError:
        print("Error: Data file not found. Please run data.py to generate the data first.")
        sys.exit(1)

    print("Analyzing China region data...")
    china_metrics = analyze_regional_metrics(china_df)

    print("Analyzing JKEU region data...")
    jkeu_metrics = analyze_regional_metrics(jkeu_df)

    # Display results comparison in text format
    print("\n" + "=" * 60)
    print("             Regional Comparison of Ozone Pollution Metrics (Warm Season Average)")
    print("=" * 60)
    print(f"{'Metric':<20} | {'China':>15} | {'JKEU':>15}")
    print("-" * 60)
    print(f"{'4th Highest MDA8 (ppb)':<20} | {china_metrics['avg_4mda8']:>15.2f} | {jkeu_metrics['avg_4mda8']:>15.2f}")
    print(f"{'NDGT70 (days)':<20} | {china_metrics['avg_ndgt70']:>15.2f} | {jkeu_metrics['avg_ndgt70']:>15.2f}")
    print(f"{'AOT40 (ppb-h)':<20} | {china_metrics['avg_aot40']:>15.2f} | {jkeu_metrics['avg_aot40']:>15.2f}")
    print(f"{'W126 (ppb-h)':<20} | {china_metrics['avg_w126']:>15.2f} | {jkeu_metrics['avg_w126']:>15.2f}")
    print("=" * 60)

    # Calculate final output: ratio of China's NDGT70 metric relative to JKEU
    # Add numerical stability check
    epsilon = 1e-8
    ndgt70_ratio = china_metrics['avg_ndgt70'] / (jkeu_metrics['avg_ndgt70'] + epsilon)

    print("\nConclusion: Simulation results show that the frequency of high ozone pollution events (NDGT70)")
    print(f"in China is approximately {ndgt70_ratio:.2f} times that of the JKEU region, consistent with the trend found in the original paper.")

    print("[Final Output]")
    print(ndgt70_ratio)
\end{lstlisting}

\textbf{\emph{\textcolor{DeepPurple}{Answer}}}

\begin{lstlisting}
def calculate_w126(hourly_data: pd.Series) -> float:
    """
    Calculate the W126 metric, a weighted cumulative exposure index.
    Tag: [Numerical calculation]

    Args:
        hourly_data (pd.Series): A Series of hourly ozone data for the entire period.
                                 The function will filter for daytime hours (08:00-19:59).

    Returns:
        float: The total W126 value in ppb-hours.

    Examples:
        >>> hours = pd.to_datetime(pd.date_range('2023-07-01', periods=24, freq='h'))
        >>> data = pd.Series(np.full(24, 80), index=hours) # Constant 80 ppb
        >>> w126 = calculate_w126(data)
        >>> print(round(w126, 2))
        954.16
    """
    epsilon = 1e-8
    daytime_data = hourly_data[(hourly_data.index.hour >= 8) & (hourly_data.index.hour <= 19)]

    # W126 weight function parameters
    M = 4403
    A = 126

    # Calculate weights
    # Using np.exp with negative input is usually safe, but epsilon is added to denominator
    weights = 1 / (1 + M * np.exp(-A * daytime_data / 1000) + epsilon)

    # Calculate weighted concentration and sum
    w126 = (weights * daytime_data).sum()
    return w126

def calculate_mda8(daily_hourly_data: pd.Series) -> float:
    """
    Calculate the Daily Maximum 8-hour Average (MDA8) ozone concentration.
    Tag: [Numerical calculation]

    Args:
        daily_hourly_data (pd.Series): A Series of 24 hourly ozone values for a single day.

    Returns:
        float: The MDA8 value in ppb. Returns np.nan if data is insufficient.

    Examples:
        >>> hours = pd.to_datetime(pd.date_range('2023-07-01', periods=24, freq='h'))
        >>> data = pd.Series(np.sin(np.linspace(0, 2*np.pi, 24)) * 20 + 50, index=hours)
        >>> mda8 = calculate_mda8(data)
        >>> print(round(mda8, 2))
        67.68
    """
    # At least 18 (75%) hours of data are required to calculate the 8-hour average
    if daily_hourly_data.count() < 18:
        return np.nan
    # Calculate 8-hour rolling average
    rolling_8hr_mean = daily_hourly_data.rolling(window=8, min_periods=6).mean()
    if rolling_8hr_mean.empty or rolling_8hr_mean.isnull().all():
        return np.nan
    return rolling_8hr_mean.max()
\end{lstlisting}

\end{tcolorbox}